# Research on the Spatial Data Intelligent Foundation Model

Wang Shaohua[1,*], Xie Xing[2], Li Yong[3], Guo Danhuai[4], Cai Zhi[5], Liu Yu[6], Yue Yang[7], Pan Xiao[8], Lu Feng[9], Wu Huayi[10], Gui Zhipeng[10], Ding Zhiming[11], Zheng Bolong[12], Zhang Fuzheng[13], Wang Jingyuan[14], Chen Zhengchao[1], Lu Hao[15], Li Jiayi[10], Yue Peng[10], Yu Wenhao[16], Yao Yao[16], Sun Leilei[14], Zhang Yong[5], Chen Longbiao[17], Du Xiaoping[18], Li Xiang[19], Zhang Xueying[20], Qin Kun[10], Gong Zhaoya[6], Dong Weihua[21], Meng Xiaofeng[22,*]

[1]State Key Laboratory of Remote Sensing Science, Aerospace Information Research Institute, Chinese Academy of Sciences; [2]Microsoft Research Asia; [3]Tsinghua University; [4]Beijing University of Chemical Technology; [5]Beijing University of Technology; [6]Peking University; [7]Shenzhen University; [8]Shijiazhuang Railway University; [9]Institute of Geographic Sciences and Natural Resources Research, Chinese Academy of Sciences; [10]Wuhan University; [11]Research Institute of Software, Chinese Academy of Sciences; [12]Huazhong University of Science and Technology; [13]Fast Natural Language Processing Center and Audio Center; [14]Beihang University; [15]SuperMap Software Co. Ltd; [16]China University of Geosciences (Wuhan); [17]Xiamen University; [18]Key Laboratory of Digital Geography, Chinese Academy of Sciences; [19]East China Normal University; [20]Nanjing Normal University; [21]Beijing Normal University; [22]Renmin University of China

E-mails for corresponding authors: Meng Xiaofeng (xfmeng@ruc.edu.cn),
Wang Shaohua (wangshaohua@aircas.ac.cn)

**Abstract:** This report, organized and authored by the ACM SIGSPATIAL China Chapter, builds upon the *National Spatial Data Intelligence Annual Development Report (2022)* and the *Spatial Data Intelligence and Urban Metaverse White Paper (2023)*. Officially released at the 5th Spatial Data Intelligence Academic Conference (SpatialDI 2024), the report centers around the spatial data intelligent foundation model, discussing its principles, methodologies, and cutting-edge applications. It provides a comprehensive overview of the model's definition, evolution, current state, emerging trends, and challenges. Additionally, the report elaborates on key technologies relevant to large-scale spatial data intelligence models and their applications across various domains, including urban environments, air and space remote sensing, geography, and transportation. The document also explores current use cases of large spatial data intelligence models in fields such as smart cities, multimodal data processing, remote sensing, intelligent transportation systems, and resource management. It highlights recent examples in areas like resource and environmental management, offering insights into future developments for spatial data intelligence models. The report outlines the foundational concept of spatial data intelligent models, delves into their three-stage development process, and examines their research status and future trends. It identifies three primary challenges facing these models today and reviews progress across four key thematic areas: urban planning, remote sensing from air and space, geographical analysis, and transportation networks. Furthermore, the report systematically presents the key technologies, benefits, research developments, and future directions of spatial data intelligent models, covering spatiotemporal big data platforms, distributed computing, 3D virtual reality, foundational model performance, spatial analysis and visualization, geospatial intelligent computing, deep learning, high-performance data processing, geographic knowledge graphs, and intelligent multi-scenario simulations. It analyzes the application of these technologies in spatial data modeling and their role in advancing smart foundation models. In summarizing recent applications of large-scale spatial data intelligence models across five major fields—urban environments, multimodal data, remote sensing, intelligent transportation, and resource and environmental management—the report forecasts future developments and shifts in spatial data analysis. It anticipates three main development trends for spatial data intelligent models, offering a roadmap for future progress in industry, academia, and research. The report aims to foster the growth of large-scale spatial data intelligence models in the AGI era, promoting their application in diverse areas such as urban planning, air and space remote sensing, geography, and transportation. It also seeks to enhance academic exchanges in theory, technology, and applications to address the key challenges and bottlenecks facing the spatial data intelligent model industry.

**Key words:** Spatial data intelligent foundation model; Intelligent computation; AGI; GeoAI; Multi-model

# CONTENTS

# 1 Background of The Spatial Data Intelligent Foundation Model

The development of artificial intelligence technology has driven continuous innovation, with large language model, ChatGPT, and other foundational AI models becoming increasingly mature. The combination of geography and artificial intelligence has given rise to GeoAI (Geospatial Artificial Intelligence) and foundation models for spatial data intelligence. This encompasses a wide range of research related to geography and artificial intelligence, such as developing intelligent computer programs that simulate human perception of environmental and spatial reasoning, discovering new knowledge about geographical phenomena, and enhancing the understanding of human-environment interactions with Earth systems. These studies share a spatial perspective, focusing on solving complex geographical problems and addressing major societal challenges to achieve sustainable development goals. Currently, the related applications are not limited to geography and earth sciences; they have been successfully applied to downstream tasks, such as humanitarian relief, precision agriculture, urban planning, transportation, supply chain management, and climate change mitigation (Gao et al., 2023).

## 1.1 Foundation model for Spatial Data Intelligence and its Development History

Spatial data intelligence is an interdisciplinary research field that utilizes advanced communication technology, artificial intelligence methods, big data analysis, advanced computing technology, and other techniques to better perceive, collect, share, manage, analyze, and apply spatial data. The development of a series of foundation models, such as ChatGPT, signifies that the information society has entered a new stage dominated by these models. Spatial data analysis experiences a revolutionary transformation—the era of foundation models for spatial data intelligence. In this era, the integration of various advanced technologies, particularly generative artificial intelligence, reinforcement learning, and natural language processing, drives the development of foundation model for spatial data intelligence.

A foundation model for spatial data intelligence refers to a comprehensive model built using advanced communication technology, artificial intelligence methods, massive big data analysis, and advanced computing technology. This model can perform comprehensive and deep analysis and process vast and heterogeneous spatial data. It can efficiently integrate various spatial data resources, achieving the fusion and cross-application of multi-source data, and intelligently extracting the potential value and patterns of spatial data. This provides precise spatial information services and decision support for various industries. The foundation model for spatial data intelligence encompasses key development directions, such as data perception, data management, data analysis, and data security. By achieving comprehensive data perception, meticulous management, in-depth analysis, and security assurance, it enables the whole intelligent process and application of spatial data. This model not only focuses on data acquisition and perception but also emphasizes data storage and management, processing and deep analysis, as well as data privacy and security, ensuring the integrity, accuracy, and reliability of spatial data.

Spatial intelligence is a basic feature of the spatial data intelligence foundation model. Fei-Fei Li, a professor of computer science at Stanford University, systematically explained the concept of "spatial intelligence", that is, the perception, understanding and interaction capabilities of the foundation model in three-dimensional space. This concept goes beyond the limitations of traditional two-dimensional vision and gives the foundation model a deep understanding of space (especially three-dimensional space), enabling it to navigate, operate and make decisions in a complex three-dimensional world like humans (Krishna et al., 2022). The spatial intelligence feature will enable the spatial data intelligence foundation model to understand and interact with the 3D world. Based on the summary of the development trend of AI-generated images and videos from the ImageNet project to the current stage, Professor Fei-Fei Li pointed out that the core value of spatial intelligence is to transform visual information (or multimodal three-dimensional spatial input information) into actionable wisdom (Gupta et al., 2021). Therefore, spatial intelligence empowering artificial general intelligence (AGI) is an important goal of the current spatial data intelligence foundation model. The spatial intelligence technology foundation of the spatial data intelligence foundation model will include the following four aspects:

(1) 3D spatial sensing: 3D visual sensing technology is the eyes of spatial intelligence. It enables the foundation model to perceive and understand 3D space like humans. This technology involves stereoscopic vision, depth perception, object recognition, and scene reconstruction. Through the input of high-precision 3D spatial data such as LiDAR, stereo cameras, and depth cameras, the foundation model can capture detailed 3D information about the surrounding environment. After being processed by complex algorithms, this information is converted into a model that the foundation model can understand, providing a basis for subsequent decisions and actions.

(2) Spatiotemporal data processing and analysis: Spatiotemporal data processing and analysis is the brain of spatial intelligence. Spatial data contains not only location information, but also the time dimension, which makes data processing more complicated. Effective spatiotemporal data processing requires powerful computing power and advanced algorithms. This includes data cleaning, feature extraction, pattern recognition, and predictive analysis. Through in-depth analysis of spatiotemporal data, foundation models can understand the movement of objects and the changing trends of the environment, so as to make more intelligent decisions.

(3) Application of deep learning in spatial intelligence: Deep learning is the learning mechanism of spatial intelligence. It enables foundation models to learn and reason by imitating the neural network structure of the human brain. In the field of spatial intelligence, deep learning is used in many aspects such as image recognition, speech processing, and natural language understanding. Especially in three-dimensional visual perception, deep learning can help foundation models recognize objects and understand scenes more accurately. With the continuous advancement of deep learning technology, the learning ability of foundation models will become stronger and stronger, and the performance of spatial intelligence will become more and more outstanding.

(4) Comprehensive solutions: including large-scale processing of 3D data, real-time perception in complex environments, and decision-making under uncertain conditions. In addition, the autonomy and adaptability of foundation models are also the key areas that current technologies need to tackle. Only by solving these problems can spatial intelligence be truly integrated into general artificial intelligence, and large spatial data intelligence models can enter people's daily lives.

Compared to traditional artificial intelligence model, foundation models for spatial data intelligence have the following notable features: First, they enable multi-source data fusion, integrating spatial data from various sources such as geographic information systems, remote sensing technology, and sensor networks. This allows for comprehensive, multidimensional spatial information acquisition and analysis. Second, they have the capability for cross-disciplinary applications. These models are not limited to the field of computer science but can also integrate data and knowledge from other fields such as mathematics, remote sensing, meteorology, and geology, to achieve comprehensive cross-disciplinary analysis and intelligent decision-making. Furthermore, they possess the ability to efficiently process massive amounts of data. They can handle large-scale, high-dimensional spatial data, by utilizing distributed computing and high-performance computing platforms to achieve rapid processing and analysis of vast datasets. Lastly, they feature intelligent reasoning and prediction functions. By learning the patterns and trends in spatial data, these models can perform intelligent reasoning and forecasting, to provide users with precise spatial information services and decision support.

The development process of foundation models for spatial data intelligence can be divided into the following three stages:

The first stage is spatial data mining. During this stage, spatial data analysis relies predominantly on conventional data mining methods. In hopes of better understanding and utilizing this data, researchers focus on uncovering hidden patterns and trends from massive spatial data. Data mining methods include clustering, classification, and association rule mining, with strive to discover potential patterns and correlations through data analysis and mining. However, during this stage, the data mining process mainly depends on manually formulated rules and logic, failing to fully utilize the characteristics and inherent the structure of the data itself. Therefore, although data mining has made crucial advancements in certain specific scenarios, it often proves inadequate when dealing with large-scale, high-dimensional spatial data.

During the data mining stage, the primary goal of spatial data analysis is to discover potential patterns and trends within the data, which provides support for subsequent decision-making and applications. However, due to the limitations of data mining methods, spatial data analysis often struggles to handle complex spatial data and require high data quality and integrity. Therefore, despite achieving some success in simple scenarios, the effectiveness of spatial data analysis in practical applications is often unsatisfactory.

The second stage is the application stage of traditional machine learning and deep learning. With the rapid development of machine learning and deep learning technologies, spatial data analysis has gradually introduced these advanced methods. Traditional machine learning methods, such as Support Vector Machine (SVM), Decision Trees, as well as deep learning methods, such as Convolutional

Neural Networks (CNN), Recurrent Neural Networks (RNN), have infused new vitality to spatial data analysis. These methods, through feature engineering and data preprocessing, enable the extraction of features and the classification of spatial data, leading to significant progress in areas, such as remote sensing image recognition and geographic information extraction.

The introduction of traditional machine learning and deep learning methods has significantly improved the effectiveness and accuracy of spatial data analysis. These methods can not only handle large-scale, high-dimensional spatial data but also fully exploit the potential patterns and trends within the data. In particular, the application of deep learning methods has brought spatial data analysis to unprecedented heights, providing new insights and approaches.

The third stage is the stage of foundation model for spatial data intelligence. With the continuous development of big data technology and artificial intelligence algorithms, generative artificial intelligence provides a new perspective for the development of foundation model for spatial data intelligence. Through technologies such as deep learning, these models can delve into the inherent patterns and features of spatial data, producing more precise and diverse data. This not only compensates for the shortcomings of missing data but also enriches the hierarchy and dimensions of the data, which makes spatial data analysis more comprehensive and in-depth. The integration of generative artificial intelligence not only enhances the intelligence level of the model but also broadens their application scenarios and depth. Through interdisciplinary data learning, model can integrate a more diverse range of knowledge, providing richer and deeper insights for spatial data analysis. This interdisciplinary integration not only improves the accuracy and efficiency of analysis but also promotes communication and integration between different fields, injecting new momentum into the development and innovation of spatial data intelligence applications.

By dividing these stages, we can clearly see the trajectory of the development of foundation models for spatial data intelligence from its inception to its growth, as well as the technological innovation and application transformation they bring. With the continuous advancement of technology and the expansion of application scenarios, foundation model for spatial data intelligence will continue to play an important role, driving the development and innovation of the field of spatial data analysis. Looking ahead, we anticipate more innovative applications based on foundation models for spatial data intelligence, providing more effective support and guarantees for the sustainable development and intelligent process of human society.

## 1.2 Definition and Development History of Foundation Model

Foundation models refer to large-scale deep learning models in the field of machine learning, which include vast numbers of parameters and architectures, typically consisting of tens of thousands of neurons and millions to billions of parameters. Hence, these models can handle various complex and detailed tasks. The emergence of foundation model has significantly advanced the development of artificial intelligence, enabling machines to better understand and process human language and image information.

As foundation model technology continues to progress, various industries are incorporating the latest advancements to customize specialized foundation model tailored to their needs. Foundation models for spatial data intelligence are the result of the intersection of geography, spatial sciences, and artificial intelligence. They have already found extensive applications in fields such as transportation, smart cities, defense, healthcare, and commercial marketing.

### 1.2.1 Foundational Foundation Model

In 2006, Hinton published a paper on deep learning, sparking a wave of interest in the field. In 2012, Hinton and his student designed the first modern convolutional neural network model, AlexNet, which emerged victorious in the ImageNet competition. In 2015, Kaiming He proposed the residual network structure, making it a standard for deep networks and significantly increasing the number of layers in neural networks. In 2017, Google's research team introduced the core "self-attention mechanism" of the Transformer architecture, abandoning the sequential structure of recurrent neural networks (RNNs) and incorporating the idea of human focus into the network. In 2021, Google presented the Vision Transformer at ICLR, extending the Transformer model architecture to the field of computer vision and replacing convolutional neural networks (CNNs) as the mainstream algorithm. In 2022, a review by Fei-Fei Li and others on foundation model comprehensively introduced the capabilities and technical principles of foundation model, along with their applications in law, healthcare, education, and their social impacts concerning inequality, misuse, economic environment effects, and legal and ethical considerations. Following the release of ChatGPT, major Chinese companies have also released their own foundational foundation model.

Deep learning is an advanced machine learning technology that has become the mainstream model in the field. Its core principle involves constructing multi-layer nonlinear transformations, capturing complex inputs by continuously increasing the number of layers and nodes to achieve more accurate outputs. The main deep learning model include Deep Neural Networks (DNN), Convolutional Neural Networks (CNN), Recurrent Neural Networks (RNN), and Reinforcement Learning (RL). AI technologies represented by deep neural networks are driving the successful implementation of intelligent applications such as computer vision, autonomous driving, natural language processing, and speech recognition. With the rapid increase in the scale of model parameters and architectures, foundation models have emerged as a revolutionary breakthrough in AI technology.

Current general foundational model can be roughly divided into four categories (Mai et al., 2023):

(1) Large language model, such as PaLM, LLAMA, GPT-3, InstrucGPT, and ChatGPT.

(2) Large foundational vision model, such as Imagen, Stable Diffusion, DALL·E2, and SAM.

(3) Large foundational multimodal model, such as CLIP, OpenCLIP, BLIP, OpenFlamingo, KOSMOS-1, and GPT-4.

(4) Large foundational reinforcement learning model, such as Gato.

The following section will introduce several typical foundation models:

(1) GPT

Since 2018, OpenAI has released the GPT series of foundation model, using the Transformer architecture and pre-training on vast amounts of internet text data, achieving excellent performance in various language tasks. In 2019, OpenAI released the GPT-2 model, featuring a larger model size and more pre-training parameters, enabling it to produce smoother and more coherent language generation results. In 2020, OpenAI introduced the GPT-3 model with 175 billion parameters, demonstrating remarkable performance across various natural language processing tasks. Based on given prompt texts, it can generate coherent and creative articles, dialogues, etc. However, the high computational resources and costs make the use of GPT-3 still quite limited. On March 15, 2023, OpenAI officially launched GPT-4 with even larger training data, supporting multimodal input and output forms including images and text, which possessing powerful image recognition capabilities. Currently, with foundation model such as ChatGPT and its counterparts generating significant interest, GPT has achieved the ability to create knowledge. These models have improved machines' understanding of natural language, grasp of world knowledge, and logical reasoning abilities. In the future, even more powerful and intelligent GPT versions will continue to emerge.

GPT-3 is a large language model developed by OpenAI with 175 billion parameters, which can generate high-quality text, answering questions, performing text classification, summarization, and other tasks. Based on the Transformer architecture, GPT-3 utilizes pre-training and fine-tuning methods. Through self-supervised learning on vast amounts of text data, it acquires extensive language knowledge. After pre-training, the model can be fine-tuned to adapt to various specific natural language processing tasks, such as text generation, question answering, and text classification. Additionally, GPT-3 achieves the best results in some natural language processing benchmark tests, demonstrating its outstanding performance across various natural language processing tasks. However, due to the complexity of GPT-3 and its computational resource requirements, its usage and development also face challenges. Moreover, training on large amounts of pre-training data has raised concerns about data privacy and fairness. Therefore, researchers and society at large need to collaborate and address these issues to ensure that large language model like GPT-3 better serve humanity.

On November 30, 2022, OpenAI released ChatGPT, a conversational language model based on the GPT-3.5 architecture. This model excels in natural language generation, allowing users to interact in a natural conversational format. It can generate high-quality, coherent, and logical text, enabling tasks such as automated question answering, text classification, automatic summarization, machine translation, and chat dialogues. It can even perform creative tasks like writing, composing, and code generation, as well as understanding and executing multi-step instructions and learning new tasks from examples. ChatGPT shows exceptional performance in open-domain natural language understanding. without requiring parameter adjustments, it often surpasses model specifically designed and trained with supervised data for certain tasks with just a few examples. When faced with various text generation tasks proposed by users, ChatGPT can generate fluent, logical, and diverse long texts in most cases. Shortly after the release of ChatGPT, OpenAI subsequently released GPT-4, increasing the window length from GPT-3.5's 4096 tokens to 32768 tokens. In addition to recognizing and extracting information from images and providing textual feedback, GPT-4 can quickly generate website code based on hand-drawn sketches. Moreover, GPT-4 can answer questions in numerous fields such as mathematics, programming, vision, medicine, law, and psychology without needing specially designed instructions, performing far better than ChatGPT and almost reaching human-level proficiency. Although, it is not perfect yet, it can reasonably be considered as an early general artificial intelligence system (Che et al., 2023).

(2) SAM

Segment Anything Model (SAM) is a new image segmentation model and task recently open-sourced by Facebook Research. SAM can generate high-quality object masks from input prompts (such as points or boxes) and can be used to generate masks for all objects in an image. It has been trained on a dataset containing 11 million images and 1.1 billion masks, enabling zero-shot transfer to new image distributions and tasks. Its segmentation performance is relatively impressive, so it is currently the best algorithm in terms of segmentation quality. The model consists of three components: an image encoder, a prompt encoder, and a mask decoder. It leverages the concept of prompts from NLP tasks, providing prompts for image segmentation tasks to achieve rapid segmentation of any target. Prompts can be sets of foreground/background points, rough boxes, masks, any form of text, or any information indicating the parts of the image that need to be segmented. The task's input is the original image and some prompts. The task’s output is the mask information for different objects in the image (Kirillov et al., 2023).

(3) CLIP and BLIP

Vision-Language Pre-training (VLP) has improved the performance of many vision-language tasks. Contrastive Language-Image Pre-training (CLIP), a groundbreaking work in the VLP field, is one of the earliest and most widely adopted frameworks for joint training of visual and language modalities. Released by OpenAI in 2021, it is a classic in multimodal research. The model collects a large amount of paired internet data and pre-trains on a big dataset of 400 million data points, using self-supervised contrastive learning to learn joint embeddings of visual and textual features (Radford et al., 2021). Bootstrapping Language-Image Pre-training (BLIP) is a new VLP framework that improves CLIP by training on captions generated from images collected from the internet. It employs three vision-language objectives—image-text contrastive learning, image-text matching, and image-conditioned language modeling—for joint pre-training (Li et al., 2022).

(4) Gemini

Gemini is an AI model released by Google DeepMind on December 6, 2023, capable of simultaneously recognizing five types of information: text, images, audio, video, and code. It can also understand and generate high-quality code in mainstream programming languages, such as Python, Java, and C++, and has comprehensive security assessment capabilities. The first version, Gemini 1.0, includes three different model sizes: Gemini Ultra, for handling "highly complex tasks"; Gemini Nano, for handling multiple tasks; and Gemini Pro, for handling "specific tasks on end devices."

(5) Sora

Sora is a new AI text-to-video foundation model launched by OpenAI on February 15, 2024. OpenAI considers it as a "world simulator." This model can create real and imagined scenes based on textual descriptions, featuring capabilities such as text-to-video generation, complex scene and character generation, language comprehension, multi-camera generation, video generation from static images, and physical world simulation. As a general visual data model, its training relies on a vast amount of video data with text captions. Its excellence lies in its ability to generate videos and images spanning different durations, aspect ratios, and resolutions, including high-definition videos up to one minute long. Sora aims to help people solve problems requiring interaction with the real world.

### 1.2.2 Geographic Foundation Model

The intersection of artificial intelligence and geographic spatial science research has historical roots. Hence, the application of AI technology in the fields of geography and earth sciences is not new. Smith (1984) and Couclelis (1986) discussed the potential role of AI in solving geographic problems as early as the 1980s; Openshaw (1997) also published a monograph on geographic artificial intelligence. New methods and technologies, including artificial intelligence, are needed to solve many of the scientific challenges arising from natural geospatial and socio-cultural geospatial issues; spatial and temporal data, such as remote sensing satellite data, population mobility data, and vehicle operation trajectory data, which are constantly being generated, can also support the training of artificial intelligence model and the development of new algorithms (Gao Song, 2020; Wu et al., 2019).

The development of machine learning (ML) and artificial intelligence (AI) has brought significant success to foundational general foundation model, but exploration of dedicated foundation model related to geographic spatial artificial intelligence (GeoAI) is relatively sparse. The key technical challenge lies in overcoming the inherent multimodal characteristics of GeoAI. The core data modalities of GeoAI include text, images (remote sensing images and street view images), trajectory data, knowledge graphs, and geospatial vector data (such as map layers from OpenStreetMap), all contain critical geospatial information (geometric and semantic information). Each modality's data has unique structures and requires data model for spatial representation. Therefore, effectively integrating these representations with appropriate inductive biases in a single model requires careful design. The multimodal nature of GeoAI hinders the direct application of existing pre-trained foundational model to GeoAI tasks.

Geography encompasses various subfields and is an extensive discipline, including geospatial semantics, health geography, urban geography, remote sensing science, and more. Existing large language model perform well on certain geospatial tasks, such as place name recognition, location description recognition, and time series prediction of dementia, which often surpasses fully supervised, task-specific ML/DL model. However, in tasks involving diverse data modalities such as point data, street view images, and remote sensing images, existing foundational model still lag specialized model from a spatial thinking perspective. Given the increasing availability and importance of spatial data, GeoAI research will also contribute to broader problem-solving and intelligent digital assistants. As a subfield of spatial data science, GeoAI leverages advancements in technology and data services to support the creation of more intelligent geographic information, methods, systems, and services for various downstream tasks. These include image classification, object detection, scene segmentation, simulation and interpolation, link prediction, retrieval and question answering, real-time data integration, and geographic enrichment.

After 2015, geospatial science research integrated with deep learning (DL) technologies, such as convolutional neural networks, generative adversarial networks, and graph neural networks, has continued to emerge. Today, ML has become a core component of spatial analysis in geographic information, used for classification, clustering, and prediction. DL and AI algorithms have been successfully developed and applied to numerous geographic information applications. DL is integrated with geospatial data, employing different AI methods for classification, semantic segmentation, or object detection, depending on the data type. Useful information is automatically extracted from satellite, aerial, or drone images through image classification, object detection, and semantic and instanced segmentation (Pierdicca and Paolanti, 2022).

Geospatial location serves as a link associating multiple thematic layers (weather, hydrology, soil, urban buildings, etc.), multiple elements (people, events, geographic objects), and heterogeneous data (images, text, videos, etc.). The application of AI technology to geospatial research primarily involves two types of modeling methods: spatial implicit model and spatial explicit model. Spatial implicit model refers to AI model that treat geographic locations as ordinary dimensions within a multidimensional feature vector, without giving special consideration to spatial locations or incorporating spatial relationships and constraints into the model. For instance, incorporating geographic coordinates into a simple K-means clustering model is an example of a spatial implicit machine learning model. However, constructing a clustering model with spatial constraints using a Delaunay triangulation network falls under spatial explicit modeling. For example, a dataset containing urban geographic locations and population data, where the task is to rank cities solely based on population size without considering geographic location, is not a spatial explicit model. Conversely, if the task is to determine whether cities with high population densities are spatially clustered, it requires an explicit spatial analysis perspective to constitute a spatial explicit model.

Research has demonstrated that spatial explicit AI model outperform classical machine learning model that do not consider spatial factors in visual tasks such as computer image classification and intelligent inductive reasoning tasks based on geographic knowledge graphs. Therefore, when developing new machine learning model to support geospatial knowledge discovery and intelligent decision-making, it is essential to consider how to combine the characteristics of geospatial data with AI model features to design reasonable model. Successful GeoAI research must address important geospatial issues through the development of spatial explicit model and demonstrate how to integrate graphical data and new methods developed at both the symbolic and sub-symbolic levels into today's GIS workflows (Janowicz et al., 2020).

Research on spatial data intelligent foundation model primarily focuses on spatial representation learning, spatiotemporal prediction and spatial interpolation, Earth resource and environmental monitoring, cartography, and geographic text semantic analysis (Gao Song, 2020).

(1) Spatial Representation Learning

The success of many ML algorithms often hinges on the quality of data representation and feature engineering. Therefore, spatial feature learning or representation learning is particularly crucial for developing spatial explicit AI model and advancing GeoAI innovation. By enhancing the predictive accuracy of ML model, researchers leverage representation learning techniques to extract latent geospatial features.

For example, Yan et al. proposed the Place2Vec model, which adopts natural language processing approaches for feature representation learning of points of interest (POI) data, built environments and contextual semantics of surrounding areas. This model improves place information retrieval and intelligent recommendation capabilities and is used as machine learning features for urban land use classification. Similarly, Liu et al. introduced the Road2Vec model, which quantifies implicit traffic interactions between roads based on large-scale taxi trajectory data. This model captures latent spatial heterogeneity and nonlinear interaction characteristics, enhancing the accuracy of traffic volume predictions for road segments. Crivellari and Beinat proposed the Mot2Vec model, generating feature vectors for activity locations by training on large-scale crowd mobility data. This model characterizes the associative properties and similarities of places. Jean et al. presented the Tile2Vec model for remote sensing data, an unsupervised representation learning algorithm that extends the distributional hypothesis in natural language processing to spatial data distribution. This spatial representation learning significantly improves the performance of predictive tasks, such as land cover type classification and identifying poverty areas in developing countries. Lastly, Mai et al. innovatively proposed the multi-scale spatial location encoding method Space2Vec, which encodes absolute locations and spatial relationships through a representation learning model, outperforming established ML methods in location modeling and image classification tasks.

(2) Spatiotemporal Prediction and Spatial Interpolation

The basic idea of spatiotemporal prediction is to estimate the value of a target object or geographic variable at an unknown time or location based on multidimensional attribute variables. Spatial interpolation is a common spatial analysis function in GIS, which uses attribute values from known locations to estimate the same attribute values at unknown points. Traditional spatial interpolation methods include Inverse Distance Weighting (IDW), Triangular Irregular Network (TIN), and Kriging. The application of ML and DL methods to explore new approaches for spatiotemporal prediction and spatial interpolation has made significant advances in fields such as surveying and mapping, social sensing, and intelligent transportation.

Zhu et al. designed a new deep learning architecture named Conditional Encoder-Decoder Generative Adversarial Networks (CEDGANs) for spatial interpolation, appling to elevation spatial interpolation in DEM. Li et al. extracted crowd activity locations and movement patterns from sparsely sampled mobile phone location data and proposed a new fuzzy Long Short-Term Memory (LSTM) network trajectory prediction model (TrjPre-FLSTM). Bao et al. built a BiLSTM-CNN model based on spatial clustering and deep neural networks using geotagged social media data to enhance the prediction accuracy of user area locations. Liang et al. improved the classic commercial geography Huff model by incorporating temporal dynamic attributes and combined location big data to intelligently estimate the spatiotemporal probability of customers visiting stores. Xing et al. proposed a general spatial data-driven end-to-end intelligent prediction framework, Neighbor-ResNet, to estimate human activity levels based on the multi-layer feature perception of regional landscape physical characteristics from remote sensing images. Pourebrahim et al. compared the performance of spatial interaction gravity model and Convolutional Neural Networks (CNNs) in predicting travel spatial distribution. Yao et al. compared the performance of several classic spatial interaction model and Graph Neural Network (GNN) model in predicting spatial point interaction flows.

Considering that human travel activities mainly follow road traffic networks, related research based on traffic networks is also abundant. Murphy et al. used CNNs to classify the distance error of GPS trajectory data on given travel routes to conditionally select the best estimate for driving paths between raw GPS trajectory data and map-matched routes. Zhang et al. trained deep convolutional model on large-scale street view image data in cities to predict the spatiotemporal types of traffic flow along streets. Zhang and Cheng proposed the GLDNet model, a sparse network spatiotemporal point process prediction model based on deep graph learning, suitable for analyzing data with apparent spatial clustering characteristics but relatively random temporal distribution, such as traffic accidents and street crimes. For dense spatiotemporal data, Ren et al. proposed a model using Residual LSTM networks for city-scale traffic flow prediction. Zhao et al. introduced a novel spatiotemporal graph convolutional network (T-GCN) for traffic prediction tasks, employing GCN to learn complex road network topologies to capture spatial dependencies and using Gated Recurrent Units (GRU) to learn the temporal dynamic changes of traffic states to capture temporal dependencies.

With the emergence of multi-source geographic big data, research integrating remote sensing data and social sensing data that is also continuously emerging. Professor Liu Yu's team from Peking University proposed a theoretical and technical framework for perceiving urban spatial differentiation patterns by integrating multi-source geographic big data and machine learning methods from the four dimensions of "people-land-static-dynamic." Zhang et al. proposed an intelligent analysis framework that quantitatively characterizes places from multiple dimensions, such as place type, visit volume, crowd information, and surrounding environment, by integrating social media user check-in data and urban street view images using machine learning methods. Helbich and Yao combined urban street view data with data sources such as urban resident activities and survey questionnaires for multi-dimensional perception and urban dynamic modeling, discovering the impact of environmental visual variables on people's mental states. Cao et al. used ResNet, Spatial Pyramid Pooling (SPP-Net), and stacked bidirectional LSTM (LSTM-Net) to learn multidimensional features from social sensing data and remote sensing data for intelligent classification of urban functional areas, comparing three different fusion methods: connection, element-wise addition, and element-wise max pooling. Ye et al. integrated social media and street view data for precise identification of urban functions. Law et al. developed a CNN-based model named Street-Frontage-Net by integrating OpenStreetMap and street view image data for intelligent assessment of urban street quality.

(3) Earth Resource and Environmental Monitoring

In recent years, the number of Earth observation satellites has surged globally, leading to a significant increase in the volume of observational data from satellite remote sensing and aerial imagery. This provides a rich data source for surveying and dynamically monitoring land resources, forest cover, environmental changes to analyze urban expansion and land use change trends. However, the characteristics of multi-source, multi-temporal, multi-spectral, and multi-resolution remote sensing data also pose certain challenges for practical application analysis. Various methods leveraging deep learning model combined with multi-source remote sensing data to extract spatiotemporal features are being explored. Reichstein et al. suggested coupling physical process model with data-driven machine learning to form a hybrid modeling approach. Scott et al. used transfer learning and network tuning techniques, data augmentation techniques, and deep convolutional network model to improve land cover classification accuracy. Huang et al. proposed a semi-transfer deep convolutional neural network model (STDCNN), which generated high-precision urban land use maps from WorldView high-resolution imagery. Peng et al. designed the Patch Similarity-based Convolutional Neural Network (PSNet), which uses spectral reflectance values rather than raw image values for model training, reducing numerical errors caused by inconsistent lighting. Yuan et al. discussed various fusion methods of multi-source remote sensing big data, spatiotemporal information, and deep learning model.

Simultaneously, geographic foundation model has garnered wide international attention. In August 2023, IBM and NASA jointly open-sourced the geospatial foundation model Prithvi, addressing key factors in geospatial foundation model, effective pre-training of foundation model, and promoting their application in the field of geographic sciences using the distinct characteristics of training data. The model is based on NASA's Harmonized Landsat Sentinel-2 (HLS) satellite imagery, achieving multi-temporal image reconstruction and high-resolution map applications for flood, fire, and other geographic scene changes, revealing the processes of environmental development and change. The model employs a ViT architecture and a Masked AutoEncoder (MAE) learning strategy to develop a self-supervised encoder, training continuous HLS imagery. It includes spatial attention across multiple patches and temporal attention within each patch, considering both the spatial positional relationships of different regions and the temporal evolution patterns of the same region (Jakubik et al., 2023).

(4) Cartography

The era of big data and artificial intelligence has driven the innovative development of cartographic science in several key areas. First, deep convolutional model can automatically extract multi-category geographic features, map symbols, and textual annotation information from maps and images. Second, advanced deep learning methods can accurately annotate the spatial positions of contemporary geographic features on historical scanned maps. Third, generative adversarial network model can be used for map style transfer learning, automatic rendering of topographic map shadows, and the use of synthetic information to improve cartographic style design or achieve electronic deception of map locations in national security applications. Fourth, the integration of artificial intelligence with map design could partially automate cartographic workflows, such as simplifying and aggregating building polygons, simplifying road network lines, and merging them based on connectivity. Additionally, the development of emerging technologies such as neuromorphic computing and brain-computer interfaces (BCI) has paved the way for a deep interdisciplinary combination of cartography and neuroscience. Utilizing methods and results from cognitive neuroscience to analyze maps has also promoted the integration and deep application of AI in the fields of cartography and geographic information (Zhong Ershun, 2022).

(5) Geographic Text Semantic Analysis

Digital gazetteers based on geographic texts and unstructured geographic text data play a crucial role in geographic information retrieval, spatiotemporal knowledge organization, and location data-driven intelligent decision-making. Most gazetteer databases are compiled by authoritative institutions, characterized by large data volumes, high production costs, and slow update cycles. Thus, the ability to automatically collect and extract geographic text information from vast amounts of natural language texts and social media big data is becoming increasingly important. The main steps in geographic text data semantic analysis include toponym recognition, toponym disambiguation and matching, and spatial coordinate extraction. Hu summarized various analytical methods for processing geographic text data, such as topic modeling, rule-based matching, and deep learning model. Intelligent applications of geographic text semantic analysis include extracting opinions and sentiments about places and living environments from user text comments, automatically recognizing and understanding user spatial query statements, and intelligently recommending GIS spatial analysis functions and matching operational tools. Moreover, using improved deep learning model to analyze geotagged social media text data can more accurately extract user locations during natural disasters, aiding emergency response decision-making and rescue efforts.

Foundation models, as cutting-edge achievements in artificial intelligence technology, are widely used in natural language processing fields, such as text classification, sentiment analysis, summarization, and translation. These models can be applied in areas like automated writing, chatbots, virtual assistants, voice assistants, and automated translation. They also have broad applications in text processing, image recognition, and multimodal data processing. Currently, foundation models are revolutionizing the state of natural language processing tasks, fostering more powerful and intelligent language technologies, and gradually establishing themselves the core force driving technological and societal development.

## 1.3 Development Trends of Foundation Model

The research on foundation models and the development of other disciplines, including geospatial science, are mutually reinforcing rather than a unidirectional knowledge production process driven by technological input. Existing general-purpose foundation models have been widely applied in various fields and interdisciplinary explorations, including natural language and audio processing, drug discovery, and even psychometrics. However, these models still have shortcomings, particularly in understanding spatial relationships within vertical domains. The future development of spatial data intelligent foundation model faces several significant challenges: First, improving the sharing mechanisms for large-scale geospatial annotated datasets; second, enhancing model transferability and interpretability; third, improving geospatial semantic analysis and reasoning capabilities. For instance, future users might ask about vacation spots their parents visited, an audiobook about the area they are driving through, or a quiet hotel located downtown, rather than inquiring about the construction date of the Eiffel Tower or the time it takes to drive to the airport. These and related questions require identifying the user's location, determining distances to other features, reasoning about topological relationships, and understanding vague cognitive regions, which current models still struggle to achieve (Janowicz et al., 2020).

With the increasing urbanization worldwide, the emergence of global issues, and the growing challenges of transportation, combining the rapid development of spatial big data with foundation model has become a pressing concern. Leveraging spatiotemporal big data from transportation and urban environments through recent technologies such as artificial intelligence, 5G, and digital twins, we aim to create competitive spatial data intelligent foundation model with enhanced spatial perception and analysis capabilities. This represents a new opportunity to maximize the utilization efficiency of spatial data and empower various industries, presenting a significant challenge for governments, companies, and scientists. In recent years, under the guidance of national digitalization planning and the construction of Digital China, Chinese research institutions and related enterprises have increased their research investments in this direction, maintaining relatively strong growth in the global frontier research and development of intelligent spatial data.

As AI technology continues to innovate and upgrade, spatial data intelligent foundation models are gradually entering a comprehensive commercialization phase, significantly impacting economic development, industrial transformation, national governance, and people's lives. In urban transportation applications, these models accelerate the transition from digitalization to intelligent urban transportation, making travel for passengers, and the management services of transportation agencies and government departments, more intelligent. In urban disaster prevention and emergency response, they can predict natural disasters in advance, enable emergency dispatch during disasters, and manage post-disaster recovery intelligently, reducing economic losses and effectively preventing secondary disasters. In epidemic prevention, they assist in early warning, spread prediction, epidemic investigation, and resource allocation. In the energy sector, they significantly enhance digitalization, automation, and intelligence, accelerating the achievement of dual carbon goals. In national spatial planning, they integrate multi-source geographic data from various departments, realizing a "one-map" approach to land planning, providing government departments with accurate and standardized data support (Song Xuan et al., 2022). With the advent and full development of the 5G era, 5G's critical features of high speed and low latency have significantly improved the real-time processing capabilities of the entire process of spatial data perception, collection, processing, and analysis. The widespread adoption of 5G base stations and 5G smartphones offers new opportunities for latency-sensitive applications, such as intelligent transportation and emergency dispatch management, enhancing the timeliness of spatial data intelligent foundation model.

The future of spatial data intelligent foundation model requires joint support from academia, industry, and government. It involves integrating systems thinking, spatial thinking, and computational thinking, and harnessing the wisdom of scholars and practitioners from fields such as earth system science, geography, computer science, mathematics, and physics. Together, they will explore major scientific challenges in geospatial science and the development, deployment, and deep application of spatial data intelligent foundation model (Yue et al., 2020).

## 1.4 Challenges Of Spatial Data Intelligent Foundation Model

The spatial data intelligent foundation model is a new type of artificial intelligence model that can learn from vast amounts of spatial data. It can generate new spatial data, perform spatial analysis, and create spatial content. The development of spatial data intelligent foundation model has brought significant transformations to the field of spatial information but also faces several challenges. Considering the current hot topics in spatial data intelligent foundation model, this section discusses the challenges they face from three aspects: scaling laws, effectiveness, and generative intelligence.

### 1.4.1 Scaling Laws of Foundation Model

With the development of deep learning technology, the application of foundation models in various fields is becoming increasingly common. However, effectively designing and training these foundation models has become a challenge. To address this, the scaling law of foundation models is an important theoretical tool. It helps us understand and predict the performance of foundation models and guides us in making more informed decisions in model design and training. In the field of spatial data intelligence, which often involves large-scale data and complex spatial relationships, models' performance typically changes with the increase in data scale. Therefore, understanding the scaling law of foundation models can help us better design and optimize models to meet the data processing needs of different scales.

The concept of the scaling law for foundation models were proposed by OpenAI in 2020. It is briefly defined as follows: as the model size, dataset size, and the number of floating-point operations (FLOPs) used for training increase, the models' performance improvements. To achieve optimal performance, all three factors must be scaled up simultaneously. When not constrained by the other two factors, the model's performance has a power-law relationship with each individual factor. This power-law relationship allows us to predict the model's performance in advance. Specifically, the scaling law of foundation model includes the following points:

(1) For decoder-only model, the computation C (FLOPs), model parameters $N$, and data size D (number of tokens) satisfy the relation: $C \approx 6ND$.

(2) The final performance of the model is primarily related to the computation $C$, the number of model parameters $N$, and the data size $D$, and is largely independent of the specific structure of the model (layers/depth/width). That is, by fixing the total number of model parameters and adjusting the layers/depth/width, the performance differences among different models are minimal, usually within 2%.

(3) When the model is not constrained by the other two factors, the model performance has a power-law relationship with each factor. As shown in the figure below, representing model performance with foundation model loss rate, the three variables computation $C$, the number of model parameters $N$, and data size $D$ show a high-fitting power-law model with the model loss rate, indicating a linear relationship between the logarithms of the three variables and the model loss rate.

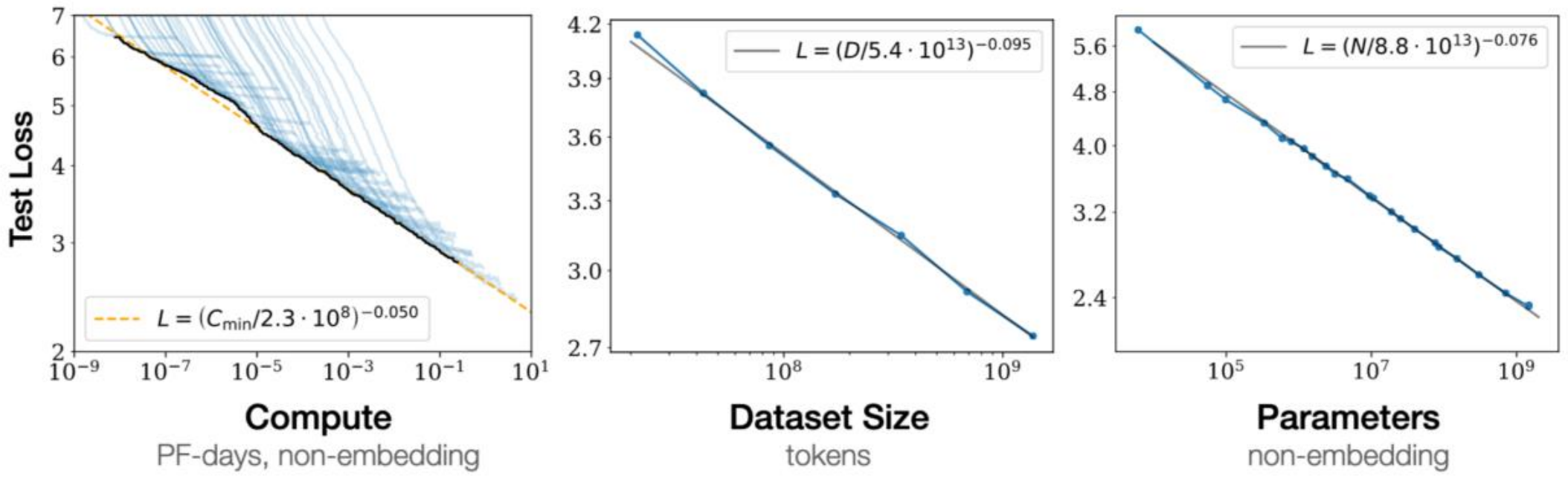

Fig. 1-1 The scaling law of the foundation model

(4) To improve model performance, the number of model parameters $N$ and data size $D$ need to be scaled up simultaneously, but there is still a debate on the exact ratio for scaling up the model and data.

(5) The scaling law for foundation model applies not only to large language models but also to other modalities and cross-modal tasks.

Summarizing the contents of the scaling law for foundation models, we can derive the core formula of the scaling law:

$$L(x) = L_\infty + \left(\frac{x_0}{x}\right)^\alpha$$

where $L_\infty$ represents the irreducible loss that cannot be reduced by increasing the model scale and can be considered the entropy of the data itself (e.g., noise in the data); $\left(\frac{x_0}{x}\right)^\alpha$ represents the reducible loss that can be reduced by increasing the computation and can be considered the difference between the model's fitted distribution and the actual distribution. According to the formula, increasing $x$ (e.g., computation $D$) decreases the overall loss rate and improves model performance; as $x$ approaches infinity, the model can fit the true distribution of the data, making $\left(\frac{x_0}{x}\right)^\alpha$ approach zero; the overall loss tends to $L_\infty$.

Currently, OpenAI has published the relationship curve between the computation and performance of its latest language model framework, GPT-4. The horizontal axis represents the normalized computation, assuming GPT-4's computation is 1. By using a computational scale 10,000 times smaller, we can predict the final performance of GPT-4. The vertical axis is "Bits per word," which is a unit of cross-entropy. When calculating cross-entropy using the base-2 logarithm, the unit of cross-entropy is "bits per word," consistent with the concept of bits in information theory. Therefore, the lower this value, the better the model's performance. The results show that GPT-4's computation and model performance also exhibit a clear power-law relationship.

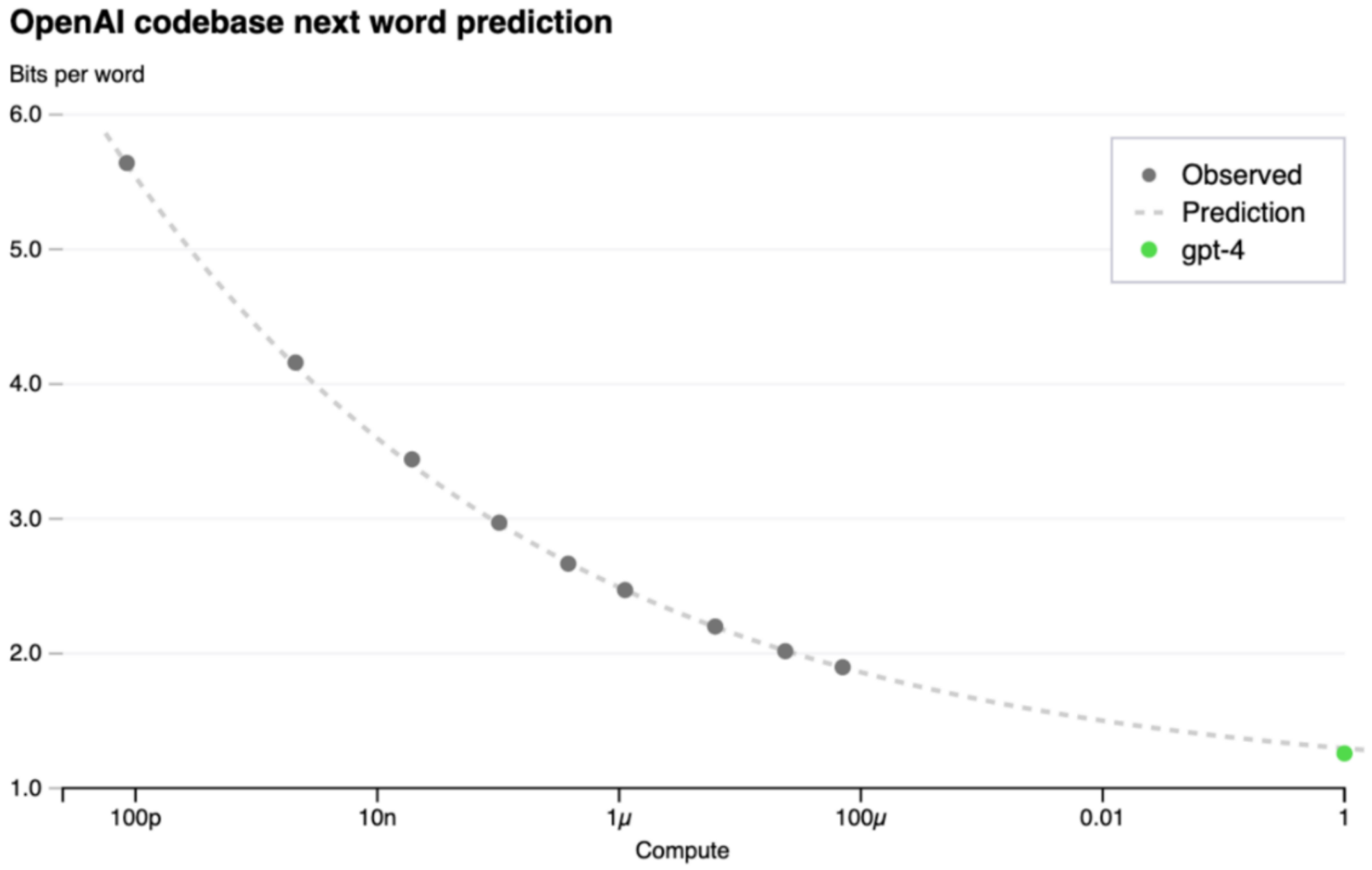

Fig. 1-2 The power law between computation and performance of GPT-4

In addition to the power-law relationship between a single variable and model performance $L$, we can also establish a joint power-law relationship between $D$, $N$, and $L$. According to the following formula, Kaplan et al. (2020) derived that for a model with a parameter count N, the dataset size $D$ needs to be greater than $(5\times10^3)\ N^{0.74}$ to ensure the model does not overfit.

$$L(N,D)=\left[\left(\frac{N_C}{N}\right)^{\frac{\alpha_N}{\alpha_D}}+\frac{D_C}{D}\right]^{\alpha_D}$$

The power-law relationship between $C$ and $L$ indicates that with every tenfold increase in computation, the model's performance improves to a certain extent. When the computation budget is limited, the question of how to allocate the dataset size $D$ and the model parameter count $N$ to achieve optimal model performance arises. OpenAI suggests that with every tenfold increase in computation, the dataset size should increase by about 1.8 times and the model parameter count by about 5.5 times, indicating that the model parameter count is more critical. DeepMind, on the other hand, suggests that with every tenfold increase in computation, both the dataset size and the model parameter count should increase by about 3.16 times.

According to Kaplan et al.'s research, the power-law relationships and joint power-law relationships revealed by the scaling law of foundation model can lead to some contradictions. These contradictions may help us understand the limits of the scaling law and explore the challenges it brings to the future development of foundation models. Specifically, these include the following three aspects:

(1) If we continually increase the model parameter count $N$ and the dataset size $D$ in a 5.5:1.8 ratio, there will inevitably be a point $N_*$ and $D_*$ such that $D_* \ll (5\times10^3)N_*^{0.74}$. In other words, after reaching $N_*$ and $D_*$, further increasing the model parameter count and dataset size will continue to lower the loss $L$. However, according to the joint power-law relationship, the model will begin to overfit, causing $L$ to rise instead of fall. Therefore, Kaplan et al. believe that the scaling law will fail before $N$ and $D$ grow to $N_*$ and $D_*$; the loss at the $N_*$ and $D_*$ point represents the irreducible error inherent in natural language data. In practical foundation model applications, there is still a distance from $N_*$ and $D_*$ as there are other modalities of data beyond natural language, such as image and speech data, which also follow similar scaling laws. However, reaching the scaling law limits in multimodal datasets is more challenging.

(2) As the loss $L$ decreases, the performance of some downstream tasks may exhibit abrupt changes, known as emergent phenomena, which cannot be accurately predicted by the scaling law. It remains unknown whether further increases in $N$ and $D$ and further reductions in $L$ will lead to more emergent phenomena. If the scaling law's limits have not been reached, there is still an opportunity to significantly enhance the "intelligence" of large language model, even if the loss $L$ does not decrease substantially.

(3) In a network with multiple interacting agents, there might be an empirical rule similar to Metcalfe's Law. This rule suggests that as the number of interactive agents in the network increases, the overall "intelligence" of the network also continuously improves.

### 1.4.2 Effectiveness of Foundation Model

The effectiveness of foundation models is a key metric for evaluating their performance and value. By discussing the effectiveness of foundation model, we can understand their strengths and weaknesses, which in turn guides their development and application. For instance, if a foundation model is found to be ineffective for a specific task, targeted improvements can be made to enhance its performance. The effectiveness of foundation model can be evaluated from the following aspects:

(1) Task Accuracy: The effectiveness of a foundation model is primarily reflected in its accuracy in performing specific tasks. For example, in remote sensing image classification tasks, a foundation model can accurately identify different land cover types in images, such as roads, buildings, and vegetation.

(2) Analytical Insights: Foundation models can go beyond simple spatial data processing to extract valuable information and insights. For example, a foundation model can analyze urban population density and traffic flow, predicting future urban development trends.

(3) Creative Content: Foundation models can use spatial data to create content that often reflects the characteristics and underlying patterns of the spatial data.

(4) Application Scenarios: The effectiveness of a foundation model is also evident in its wide range of application scenarios. Foundation models can be applied in natural resource management, urban planning, environmental monitoring, emergency management, and more.

(5) User Satisfaction: The effectiveness of a foundation model is also measured by user satisfaction. If users recognize the model's performance and features and apply it in practical work, it indicates that the model is effective.

Spatial data intelligent foundation models are designed to generate data content that meets specific user needs based on input information such as text, language, charts, and data, which makes them a multimodal foundation model in essence. By integrating multimodal big data, these models obtain richer and more comprehensive information, thereby enhancing their performance in understanding, analyzing, and generating spatial information. When trained on multimodal datasets, foundation model can learn the common features among different types of data, improving their generalization ability to new data types, which makes them widely applicable in fields like remote sensing image analysis, spatial planning, and virtual reality. In this context, spatial data intelligent foundation model, as multimodal foundation model, undoubtedly demonstrate high effectiveness, manifested in the following ways:

(1) Integration of Multi-Source Spatial Information to Enhance Spatial Understanding: Spatial data intelligent foundation model can integrate spatial data from various sources such as remote sensing images, spatial maps, and spatial text descriptions to perform multimodal information fusion, leading to a more comprehensive and accurate understanding of spatial information. For example, in land use classification tasks, these models can utilize remote sensing images, spatial maps, and spatial text descriptions simultaneously, considering the visual features, spatial structure, and semantic information of land cover to improve classification accuracy.

(2) Mining Complex Spatial Relationships to Aid Spatial Analysis: These models can uncover complex relationships from multimodal spatial data, such as the relationship between land use and transportation or vegetation and climate, providing new insights and methods for spatial analysis. For example, they can analyze urban spatial data to identify the relationship between urban layout and traffic congestion and provide suggestions for urban planning.

(3) Generating Creative Spatial Content to Enrich Spatial Expression: Beyond processing and analyzing spatial data, these models can create creative spatial content such as poetry, novels, and paintings, reflecting the characteristics and underlying patterns of spatial data, possessing high artistic and cultural value. For instance, a model can generate a poem about landscapes based on remote sensing images or a painting of urban scenery based on spatial maps.

(4) Enhancing Model Generalization Ability to Adapt to New Applications: The multimodal data processing capability of these models allow them to learn common features among different data modalities, improving their generalization ability to new data types. For example, a model trained on remote sensing image datasets can directly apply its multimodal data processing capabilities to new spatial maps or spatial text description datasets without additional training.

(5) Unlocking New Application Scenarios to Drive Spatial Intelligence Development: The multimodal data processing capabilities of these model can unlock new application scenarios in spatial intelligence, such as video summarization, sentiment analysis, and virtual assistants, bringing new transformations and development to the spatial information field. For example, a spatial data intelligent foundation model can be used to build an intelligent video analysis system that automatically recognizes and understands spatial information in videos and provides relevant services to users.

(6) Enhancing User Experience for Natural Interaction: These models can provide a more natural and smooth user experience, such as constructing intelligent customer service systems to offer more personalized spatial information services. Users can interact with the model through natural language. For example, if the user asks about traffic conditions at a specific location or nearby restaurant information, the model can provide accurate and personalized information services based on user needs.

As multimodal foundation model, spatial data intelligent foundation model demonstrates effectiveness in integrating multi-source spatial information, mining complex spatial relationships, generating creative spatial content, enhancing model generalization ability, unlocking new application scenarios, and improving user experience. They show great application potential and development prospects. With technological progress and the accumulation of data, spatial data intelligent foundation model will play an increasingly important role in the spatial information field, driving the flourishing development of spatial intelligence.

### 1.4.3 Generative Intelligence of Foundation Model

Generative intelligence refers to artificial intelligence systems which can generate new content, such as text, images, audio, etc. Generative intelligence plays an important role in foundation models, and they can learn a large amount of data to generate content with a certain structure and semantics, which is highly creative and expressive. The spatial data intelligent foundation model integrates spatial data, artificial intelligence, and natural language processing technology, which can understand, analyze, and generate spatial data. The generative intelligence of the spatial data intelligent foundation model refers to its ability to generate new and original spatial data. This includes new remote sensing images (simulating remote sensing images under different time and weather conditions, or higher resolution remote sensing images), new spatial maps (higher precision spatial maps, thematic maps containing more information), and new spatial text descriptions (generating new spatial text descriptions based on existing spatial data, such as automatically generating remote sensing image explanations or generating spatial map explanations). These capabilities meet people's diverse needs for spatial data, thereby reducing the cost of obtaining real spatial data and helping people better understand and analyze spatial data.

The generative intelligence of spatial data intelligent foundation model is still in the early stages of development, but it has achieved some remarkable results. For example, the DALL-E 3 model of OpenAI can generate realistic images, including landscapes, people, objects, etc. The Earth Engine platform of Google AI can generate various types of spatial data, such as remote sensing images, land use data, population data, etc. Overall, compared to traditional analytical model, spatial data intelligent foundation model based on generative intelligence can learn from a large amount of multimodal data and generate new samples that are similar to the original training data based on the learned and mined data patterns. At the same time, the distribution and attributes of generated samples can be controlled by adjusting model parameters to generate thematic data information that meets the needs of remote sensing and geospatial analysis, which is data-driven, creative, and controllable. Therefore, generative intelligence brings obvious challenges to the construction and design of spatial data intelligent foundation model, including huge data demands, high complexity model, as well as security and ethical issues. We propose several issues that need to be considered in the development direction and trend of generative intelligence for comprehensive spatial data foundation model:

(1) Discriminant AI or Generative AI

A fundamental issue in the construction and design of generative intelligence for spatial data intelligent foundation model is to distinguish between discriminative AI and generative AI, to clarify the design direction of "generative" intelligence.

The main objective of discriminative model is to establish the relationship between input data and related outputs, which is to learn the conditional probability distribution $p(y|x)$, where Y represents the output label or category; X represents the input feature. This model focuses on how to classify or predict based on input data and directly model decision boundaries. Discriminative model focus on how to predict output under given input conditions, so they typically focus more on category boundaries and decision surfaces. Discriminative model typically performs well on specific tasks because they focus on category boundaries, making classification more accurate. Common discriminative model includes logistic regression, support vector machines, decision trees, classifiers in neural networks, etc. However, discriminative AI model that only modeling with $p(y|x)$ , which is not sufficient to understand semantic information. It is also difficult to make correct and stable decisions.

The basic idea of generative AI can be represented by the following diagram. Two piles of points can represent two distributions, a binary classification problem. Parameterized conditional distribution $p(y|x)$ neural network model will identify the boundary between the two and use it as a basis to classify new data. For the new black ✖, the model will analyze ✖ is in the blue area and away from the dividing line, so black will be produced ✖ confirmatory conclusion belonging to blue color. This is obviously unreasonable. Although black ✖ Located in the blue area, but it is also far from the blue data gathering point. Therefore, categorizing it as blue is reckless. If we introduce $p(x)$ in the above model, we can obtain that although $p(y|x)$ is very high; $p(x)$is very low, so the final score $p(x,y)$ will not be high, leading to an uncertain conclusion or blue data is low credibility. Thus, the model not only enables decision-making, but also has an extent of belief in making decisions.

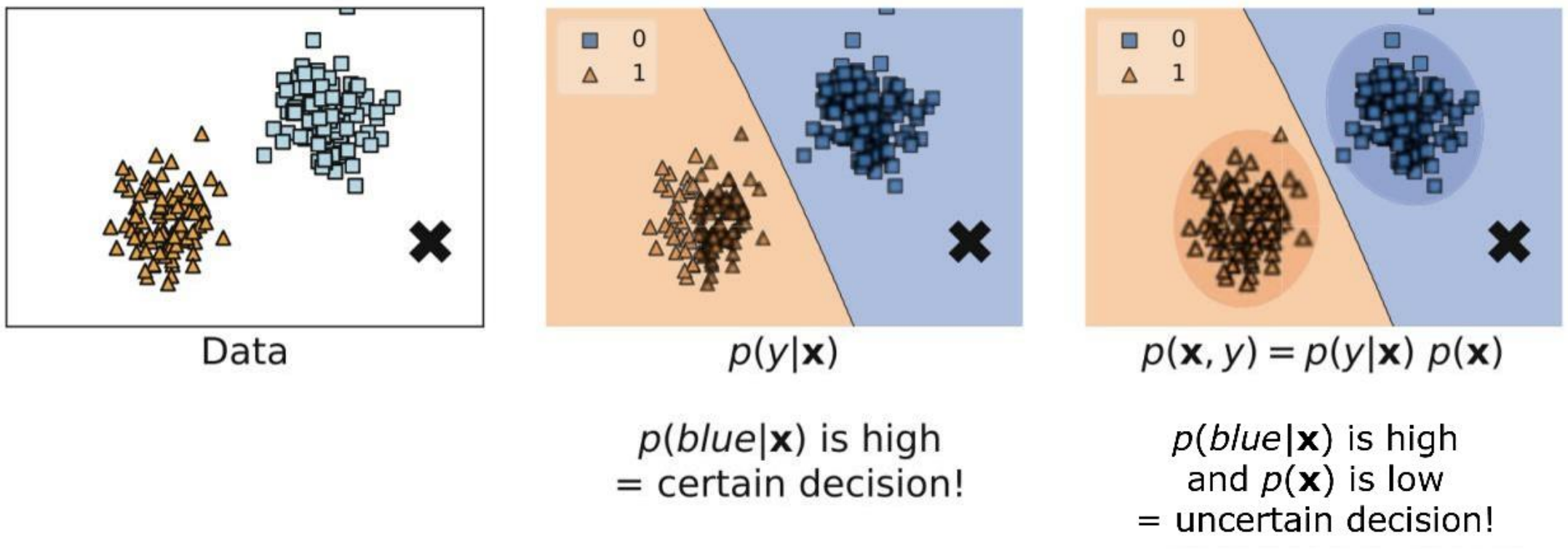

Fig. 1-3 The basic thought of generative AI

In summary, the construction of $p(x)$ is utmost importance. One of the core tasks of generative intelligence is to solve the modeling problem of $p(x)$. By designing a generative intelligence model for spatial data intelligence, we can create solutions for interacting with the environment. For example, if the model expresses doubt about some data, we can consider whether to manually intervene in checking data marks or creating a new data category. In addition, we can also use this technology to evaluate the uncertainty of the environment and provide reference for the construction of machine learning systems in the future.

(2) Chat or Agent

Dialogue based and agent based are the two development directions of generative intelligence in today's spatial data intelligence foundation model. From a characteristic perspective, dialogue type generative intelligence has interactivity, language model attributes, and personalized characteristics, which is similar to chat robots, dialogue type generative intelligence relies on powerful natural language processing capabilities and emphasizes interaction and communication with users. It is necessary to continuously optimize language model to improve the understanding and the quality of generated language; Agent type generative intelligence has the characteristics of multiple executing tasks, strong decision-making ability, multi-modal interaction, etc. It focuses on executing specific tasks, possessing certain decision-making ability, and can to process various types of input and output, including speech, text, ordinary digital images and remote sensing images, sensor data, geographic vector data, maps, etc., and making responses based on this.

At present stage, conversational generative intelligence (large language model) has been quite mature and developed. There are many mature application cases and model products, such as ChatGPT, Gemini, Claude, ERNIE Bot, etc., as well as conversational generative intelligence that accepts text input to generate other modal data information, such as text to video model SORA, text to picture model DALL-E 3, etc. However, in terms of generative intelligence for spatial data foundation model, it is clearly insufficient to only can process and output text information and complete user interaction in the form of chat. It is necessary to develop task-oriented proxy generative intelligence based on dialogue functions. Agent based generative intelligence is an artificial intelligence system that goes beyond simple text generation. It uses Large Language model (LLMs) as its core computing engine, enabling it to engage in dialogue, perform tasks, deduction, and exhibit a certain degree of autonomy. In short, an Agent is a system with complex reasoning abilities, memory, and means of executing tasks. Agent based generative intelligence mainly has four key components: ①Planning: Sub objective decomposition can break down large tasks into smaller manageable sub objectives, enabling effective processing of complex tasks; Through historical actions can be self-criticized and self-reflected, planning learned from mistakes and improved in subsequent steps to optimize the quality of the final result. ②Memory: The memory includes short-term memory for contextual learning and long-term memory achieved through external vector storage and retrieval. ③ Tool use: For the information of model weights lost, the agent learns to call external APIs to obtain additional information, including current information, code execution capabilities, access to proprietary information sources, etc. ④Action: The action module is the part of an intelligent agent which actually executes decisions or responses. Faced with different tasks, intelligent agent system has a complete set of action strategies, which can select the actions that need to be executed during decision-making, such as well-known memory retrieval, reasoning, learning, programming, etc.

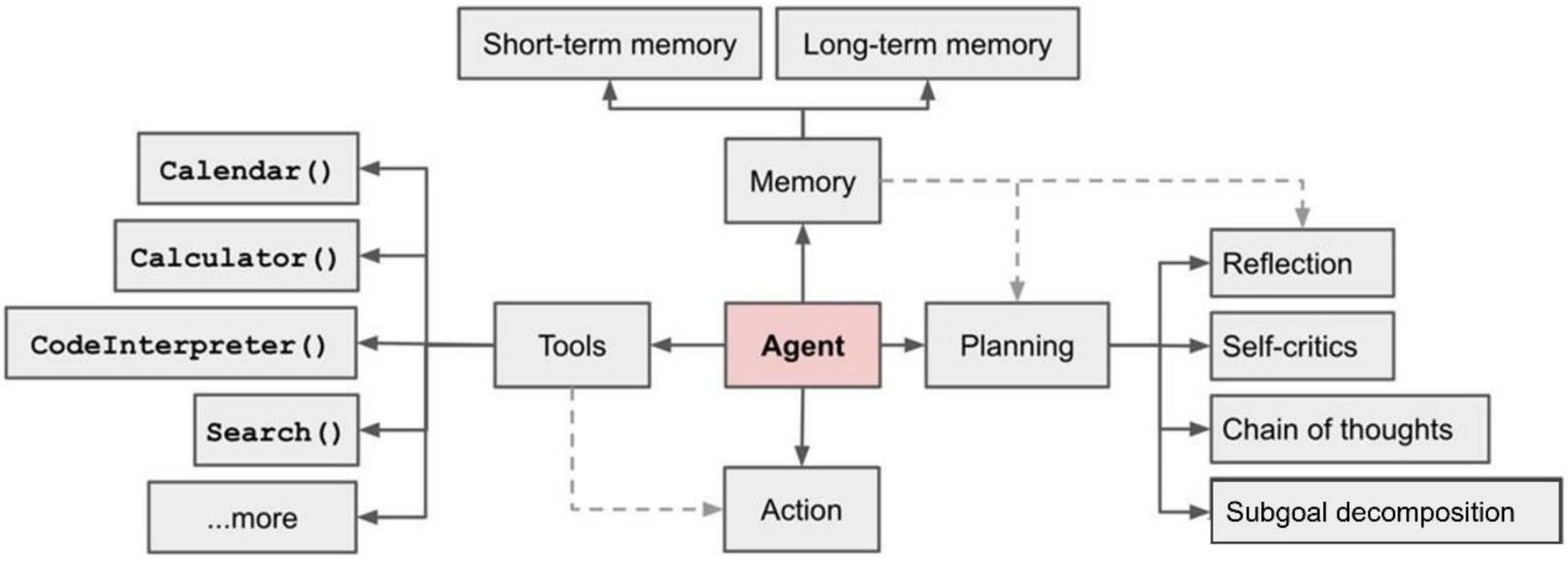

Fig. 1-4 The 4 components of agent generative AI

Based on foundation models, agents can not only provide everyone with exclusive intelligent assistants to enhance their abilities, but also change the mode of human-machine collaboration, bringing about a wider range of human-machine integration. The revolutionary evolution of generative intelligence has presented three modes of human-machine collaboration to this day:

① Embedding model: Users communicate with generative intelligence through language, use prompt words to set goals. Generative intelligence assists users in achieving these goals. For example, ordinary users inputting prompt words into generative intelligence to create novels, music works, 3D content, etc. In this mode, the role of generative intelligence is equivalent to execute commands, while humans play the roles of decision-makers and commanders.

② Copilot model: In this mode, humans and generative intelligence are more like partners, participating together in the workflow and playing their respective roles. Generative intelligence intervenes in the workflow, from providing suggestions to assisting in completing various stages of the process. For example, in software development, generative intelligence can help programmers write code, detect errors, or optimize performance. Human beings and generative intelligence work together in this process, complementing each other's abilities. Generative intelligence is more like a knowledgeable partner than a mere tool. For example, the Copilot foundation model developed by Microsoft has evolved into foundation model products such as Dynamics 365 Copilot, Microsoft 365 Copilot, and Power Platform Copilot. This has proposed the concept that "Copilot is a completely new way of working".

③ Agent model: Human set goals and provide necessary resources, such as computing power. Generative intelligence independently undertakes most of the work. Finally, humans monitor the process and evaluate the results. In this mode, generative intelligence fully embodies the interactive, autonomous, and adaptive characteristics of intelligent agents, approaching independent actors, while humans are more likely to serve as supervisors and evaluators.

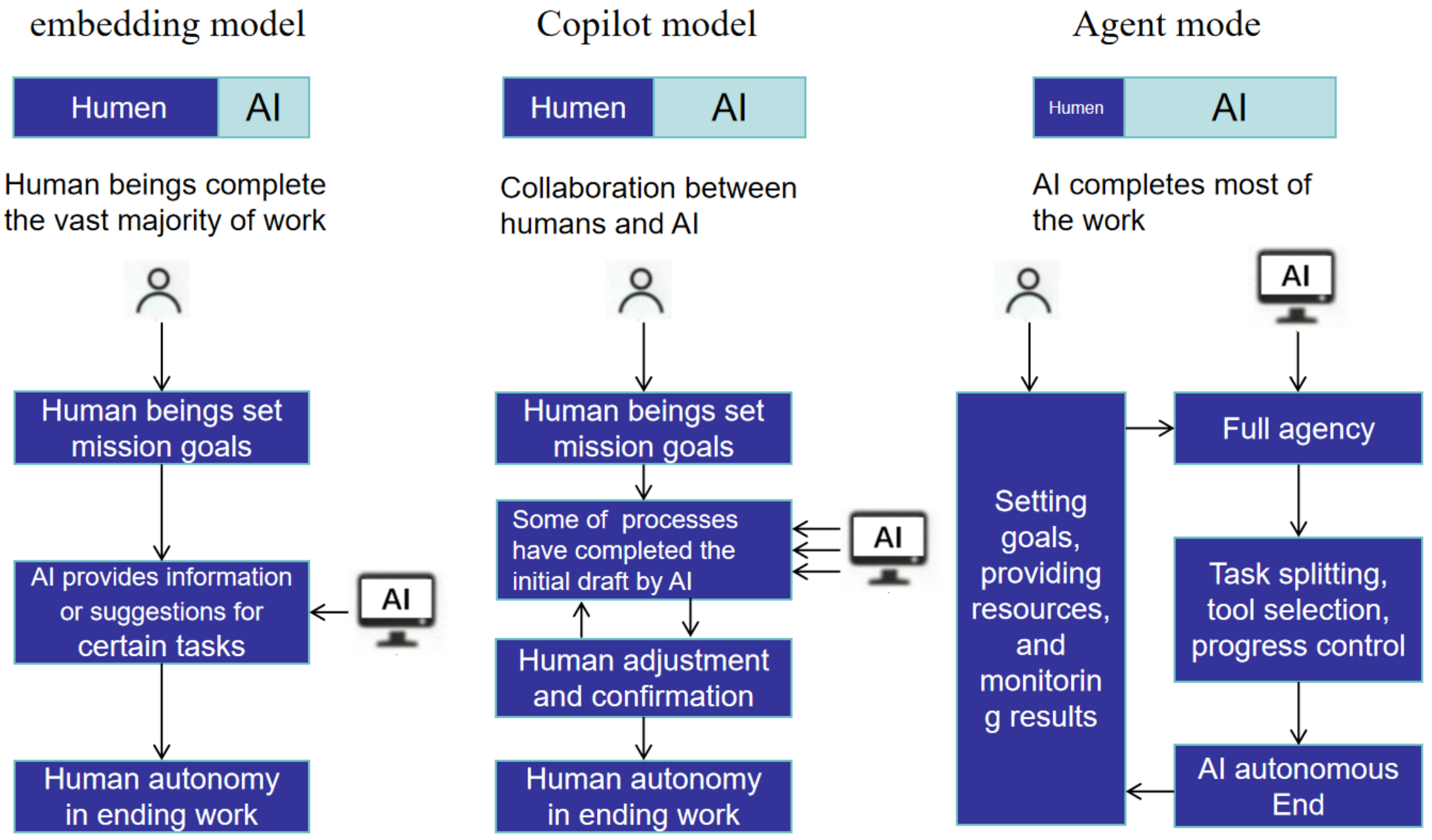

Fig. 1-5 Three modes of collaboration between humans and generative AI

From the perspective of technological optimization iteration and implementation, the development of generative intelligence also faces some bottlenecks:

① Limited context length: The limited context capacity limits the inclusion of historical information, detailed explanations, API call context, and responses. The design of the system must adapt to this limited communication bandwidth and mechanisms, such as self-reflection learned from past mistakes, will benefit from long or infinite contextual windows. Although vector storage and retrieval can provide access to larger knowledge bases, their representation ability is not as powerful as full attention.

② The challenges of long-term planning and task decomposition: Long term planning and effective exploration of solution space still pose challenges. Foundation models find it difficult to adjust their plans when encountering unexpected errors, which makes them less robust compared to humans learning from trial and error.

③ Reliability of natural language interfaces: Current Agent systems rely on natural language as the interface between foundation model and external components, such as memory and tools. However, the reliability of the model output is questionable as foundation models may exhibit formatting errors and occasionally exhibit rebellious behavior, such as refusing to follow instructions. Therefore, most of the Agent demonstration code focuses on parsing the model output.

(3) The challenge of data complexity

The training of generative intelligence for spatial data intelligent foundation model requires a large amount of spatial data, including remote sensing images, spatial maps, spatial text descriptions, etc. These data are generally large in size, diverse in format. They require a lot of storage and computing resources, such as a high-resolution remote sensing image, whose data volume can reach tens of GB. Therefore, a large amount of data processing also poses challenges to generative intelligence in terms of data complexity, including geographic accuracy, geographic bias, temporal bias, spatial scale, universality, and spatial heterogeneity.

① Geographic accuracy: in a geographic environment, generating geographically accurate results is particularly important for almost all generative intelligent tasks. For example, based on the picture, the expected answer should be ‘Washington, North Carolina’. However, ChatGPT only shows that North Carolina does not include Washington. In fact, the largest city in Washington state is supposed to be Seattle, which does not contain a city called Washington. The following figure shows four remote sensing images generated by Stable Diffusion. Although these images seem to be similar to satellite images, it is easy to see that they are artificial remote sensing images, because the geographical feature layout in these images clearly does not come from any city in the world. In fact, generating geographically accurate remote sensing images is an important remote sensing task, where geometric accuracy is crucial for downstream tasks.

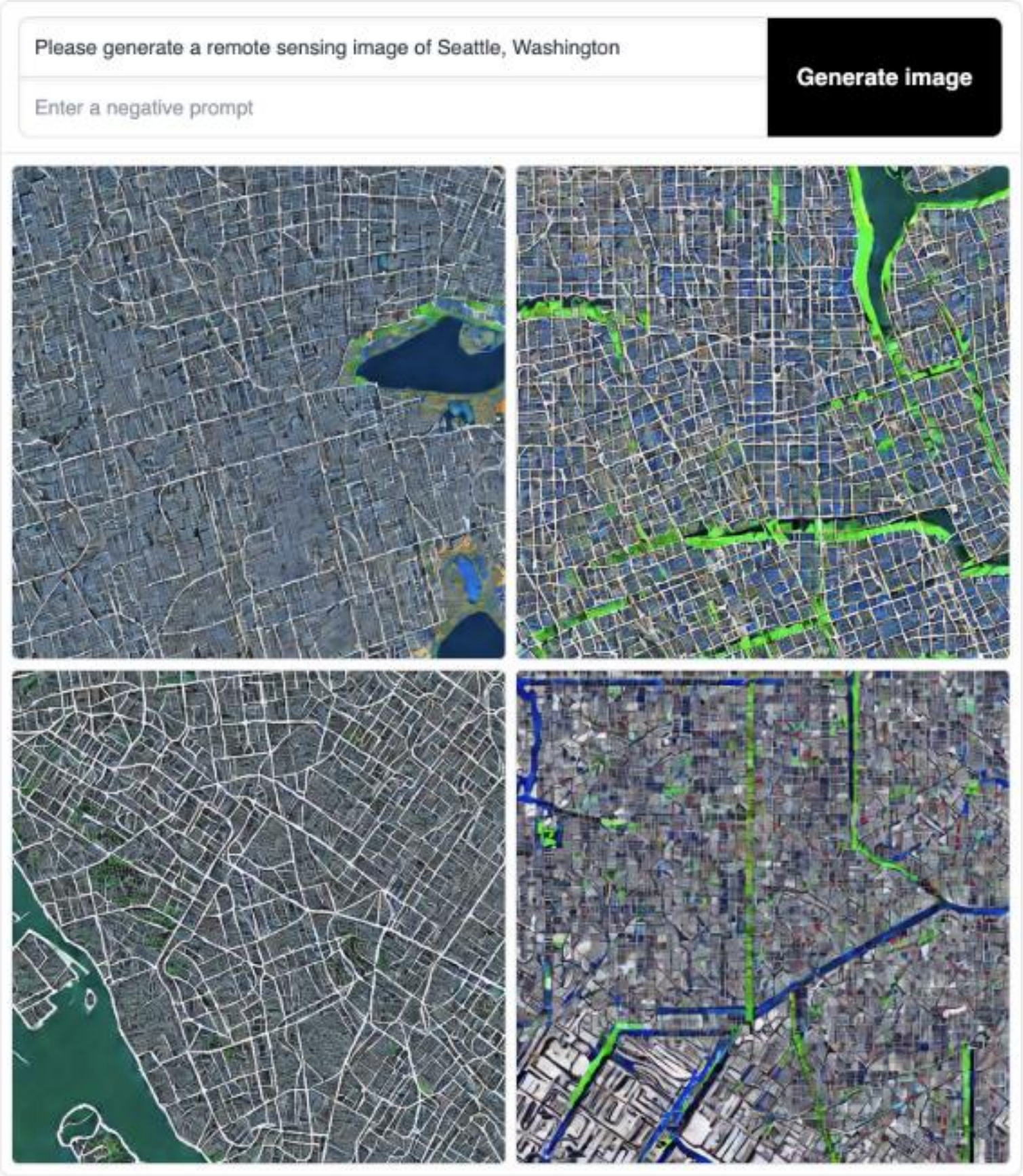

Fig. 1-6 Inaccurate results generated by Stable Diffusion

② Geographic bias: Generative intelligence may overlook existing social inequalities and biases in data. Almost all current geographic parsers are highly biased towards data rich regions, such as GPT-4, which generates inaccurate results due to inherited geographic biases in these models. Compared to San Jose in California, USA, San Jose in the Philippines is a less popular place name in many text corpora; Similarly, compared to the state of Washington and the capital city of Washington D.C. in the United States, Washington in New York is also an unpopular place name, which is why ChatGPT and GPT-4 have misinterpreted these place names. Compared to specific task model, generative intelligence is more susceptible to geographical biases. As training data is collected on a large scale, the result may be dominated by overrepresented communities or regions; Secondly, the large number of learnable parameters and complex model structures make model interpretation and unbinding more difficult; At the same time, the geographical bias of foundation model is easily inherited by all downstream adaptive model, resulting in greater negative impacts. Therefore, there is an urgent need to design appropriate geographic unbiased frameworks in generative intelligence.

③ Time deviation: Similar to geographical bias, generative intelligence is also affected by time bias in terms of redundancy, as the current training data for geographic entities is much more abundant than historical data. Time bias can also lead to inaccurate results. Two models, ChatGPT and GPT-4, were asked for their locations in Newport and Oceania in 1878 and 1923, respectively. However, GPT-4 was unable to answer these two questions as it heavily relied on pre trained data biased towards current geographic knowledge. Time and geographical biases are currently the key challenges that generative intelligent development needs to address.

④ Spatial scale: Geographic information can be represented at different spatial scales, which means that the same geographic phenomenon/object can have completely different spatial representations (points and polygons) in generative intelligent tasks. For example, urban traffic prediction model must represent San Francisco as a complex polygon, while geographic parsers typically represent it as a single point. As foundation models are developed for various downstream tasks, they need to be able to process geospatial information at different spatial scales and infer the correct spatial scale for a given downstream task. Developing such modules is a key component of effective generative intelligence.

⑤ Universality and spatial heterogeneity: An open issue of generative intelligence in spatial data modeling is how to achieve model generalization or replicability in space while still allowing model to capture spatial heterogeneity. Given geospatial data with different spatial scales, the generative intelligence of spatial data model needs to be able to learn general spatial trends while still remembering specific location details. However, the following question still needs to be considered: Will this universality introduce inevitable inherent model bias in downstream generative intelligent tasks? Will the local information of this memory lead to overly complex prediction surfaces for global prediction problems?

(4) Security and Ethics of Generative Intelligence

Due to the generative intelligence of large spatial data intelligent model, which require the processing of large-scale and complex spatial data and the generation of new thematic data information, it is inevitable that there will be deviations in accuracy and content. Meanwhile, generative intelligence may be used to generate false or misleading information, which may have a negative impact on society and bring about ethical issues such as bias, discrimination, privacy, etc.

① Security issues: Generative intelligence can be used to generate realistic false information, such as fake news reports, social media posts, images, videos; It can be used to generate malicious code, network attack tools, etc., to launch network attacks. These attacks may lead to serious consequences such as data leakage and system paralysis; It can be used to create deep fake videos, such as grafting someone's voice or image onto another person.

② Ethical issues: Generative intelligent model may learn biases in training data and reflect them in the generated samples; It could be used to create discriminatory content and generate content that violates personal privacy.

Specifically, the ethical issues of generative intelligence are mainly reflected in aspects such as fairness, transparency, and accountability. The generative intelligent model should be fair and impartial, avoiding bias and discrimination; At the same time, it should be transparent and explainable, allowing people to understand its working principle. The development and application of generative intelligent models should be strictly regulated to ensure that they are used for legitimate purposes. Therefore, the government should formulate relevant regulations and policies to regulate the development and application of generative intelligence. For example, it can be stipulated that the training data, training process, and model parameters of generative intelligent model need to be made public; It can be stipulated that generative intelligent applications require ethical review, etc. Researchers should develop technologies that can detect and prevent generative intelligence from being used for malicious purposes. For example, technology can be developed to detect false information and malicious content; Technologies that can defend against deep forgery attacks can be developed. At the same time, it is important to raise public awareness of the security and ethical issues of generative intelligence, helping people identify and resist false information and malicious content. For example, the media literacy of the public can be improved through education, publicity, and other means; The public can be encouraged to actively report false information and malicious content.

## 2 Foundation Model for Spatial Data Intelligence

The 2nd "Strategic Symposium on Spatial Data Intelligence" was successfully held at the Beijing Friendship Hotel, organized by ACM SIGSPATIAL China Chapter. The conference, themed " Foundation Models and Spatial Data Intelligence", focused on the assistance of foundation models in spatial data intelligence and the design and training challenges in vertical fields. In his opening remarks, Professor Xiaofeng Meng emphasized the importance of the GIS and CS academic communities and explored the role of foundation models in spatial data intelligence. The conference covered general topics and four vertical domain topics, and included panels, poster presentations, and forums, providing attendees with a comprehensive platform for communication. Experts and scholars engaged in in-depth discussions on the fundamental issues of foundation models in various fields such as urban, aerospace remote sensing, geography, and transportation, sharing their research achievements and cutting-edge views. The roundtable forum, chaired by Professor Yu Liu, saw invited guests discussing the characteristics and significant issues of spatiotemporal foundation models, calling for collective efforts to advance the progress in foundation model-related work.

Urban foundation models, aerospace remote sensing foundation models, geographic foundation models, and transportation foundation models all share the characteristics of being driven by big data, empowered by artificial intelligence, and application oriented. They utilize AI technology to extract information and construct a cognitive system for complex systems and operational rules in specific fields. These foundation models play a crucial role in urban planning, disaster management, resource exploration, and traffic management, providing strong support for solving real-world problems. Different foundation models have distinct differences in data types, focuses, and application scenarios. Urban foundation models use data on urban planning, population, and transportation to focus on urban structure and operational rules for urban planning, traffic management, and emergency management. Aerospace remote sensing foundation models utilize remote sensing images and satellite data, focusing on surface features and changes, which can be used for disaster monitoring, resource exploration, and environmental monitoring. Geographic foundation models use geographic information data, focusing on the geographical environment and resource distribution, which can be used for land use, resource management, and ecological protection. Transportation foundation models use traffic flow and road network data, emphasizing the dynamic changes in traffic flow for traffic planning, control, and safety.

### 2.1 Fundamental Issues of Foundation Models

Foundation models are becoming a core force driving technological and social development. They have broad commercial application prospects in various fields of generative AI. These include improving enterprise operational efficiency, optimizing decision-making, advancing intelligent assistants, and content creation. At the societal level, they can promote educational equity and smart city planning. Applying foundation models to scientific research, especially in biomedicine, can extract knowledge from vast literature, predict the impact of protein mutations on diseases, even generate candidate drug molecules for specific diseases, significantly improving drug discovery efficiency and accelerating technological progress (Thirunavukarasu et al., 2023). However, it is essential to use social science methods to emphasize algorithm fairness, privacy protection, technology transparency, and public education, ensuring these powerful tools enhance human welfare while maintaining ethical and cultural values. This approach will ensure AI truly benefits humanity and promotes sustainable social development.

#### 2.1.1 Commercial and Societal Application Potential

Foundation models have vast application prospects in multiple fields of generative AI, such as natural language processing, computer vision, and speech recognition. They can help enterprises improve operational efficiency and optimize decision-making, and advance areas like intelligent assistants, content creation, and customer service. At the societal level, they can promote educational equity and provide accessible services for people with disabilities. Foundation model technology can also be used for climate simulation and smart city planning, promoting sustainable development.

Foundation model technology has demonstrated capabilities exceeding human average and even top-level performance in certain fields, potentially bringing significant social and commercial value. The KwaiYi foundation model developed by Kuaishou AI team includes large-scale language models and multimodal foundation models. It has achieved state-of-the-art results in most authoritative Chinese/English benchmarks such as MMLU, C-Eval, CMMLU, and HumanEval with the same model size. Additionally, the KwaiYi foundation model has excellent language understanding and multimodal generation capabilities, supporting content creation, information consultation, mathematical logic, code writing, and multi-round dialogue tasks. Manual evaluation results indicate that the KwaiYi foundation model has reached industry-leading levels. Besides superior general technology capabilities, the KwaiYi foundation model also has significant business value and is widely used in various Kuaishou business scenarios.

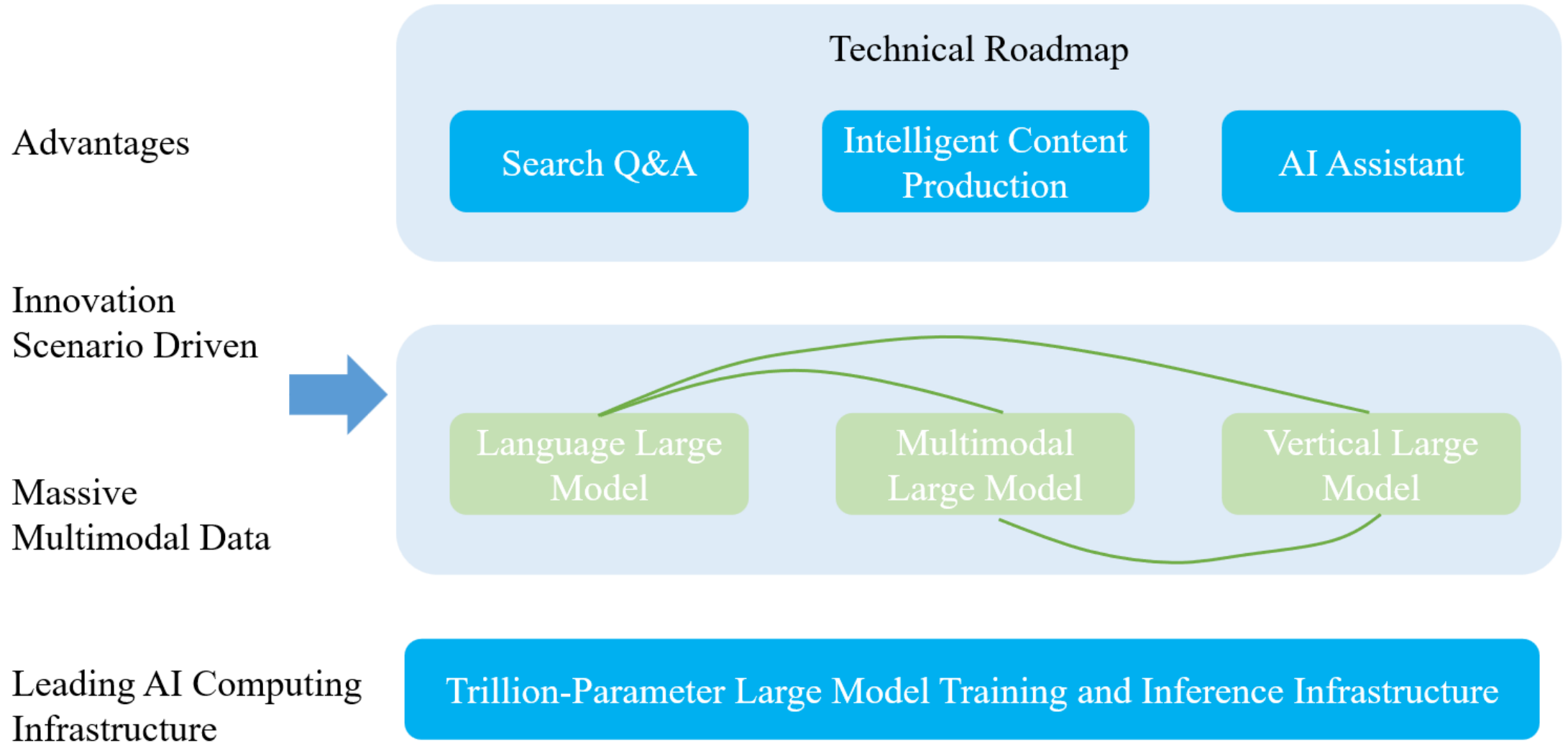

Fig. 2-1 Research process of KwaiYi foundation model

As a technology company driven by AI, Kuaishou has timely recognized the important value and development trend of foundation models. In early 2023, Kuaishou launched a significant investment in the KwaiYi foundation model research project, aiming to create an independently controllable, industry-leading large-scale language model and cross-modal foundation model. Kuaishou has several advantages in foundation model research and development: innovative scene-driven approaches that better combine real needs, vast multimodal data, including video, images, and text. It provides valuable data resources for foundation model training, and leads AI computing infrastructure, laying a solid foundation for large-scale model training. The Kuaishou foundation model will focus on core scenarios such as search and question-answering, intelligent content production, and AI assistants, including language foundation models, multimodal foundation models, and vertical foundation models. The company is also building training and inference infrastructure to support foundation models with hundreds of billions of parameters. During the pre-training data preparation phase, Kuaishou has accumulated trillions of tokens from PB+ raw data, covering multiple fields such as encyclopedias, news, books, reviews, recipes, papers, Q&A communities, and blogs. To ensure data quality, the team adopted measures such as filtering out sensitive and inappropriate content, privacy data filtering, quality model evaluation, deduplication, and anomaly detection and removal. During training, advanced techniques like mixed-precision training and Spike automatic recovery were utilized to improve training efficiency and model performance. This drives foundation model technology innovation and industrial application, creating more value for enterprises and society.

### 2.1.2 Scientific Research Applications, including Drug Discovery

Applying foundation models in scientific research has broad prospects. In biomedicine, foundation models can extract knowledge from vast literature, predict the impact of protein mutations on diseases, and even generate candidate drug molecules for specific diseases. This can significantly improve drug discovery efficiency and reduce costs, helping to solve many intractable diseases. Foundation models can also be applied in other scientific fields such as materials design and new energy, accelerating technological progress.

Foundation models in various fields are showing profound influence, especially in scientific discovery, and hold tremendous potential for application. Drug discovery is undoubtedly a focal point. The Microsoft Research AI Center team has made a series of breakthrough achievements in building foundation models for scientific fields, focusing on drug discovery. They proposed BioGPT, a large language model specifically trained in biomedicine, playing a key role in data mining and knowledge extraction in drug discovery. BioGPT has shown excellent results in target discovery with companies like Insilico Medicine. It uses biomedical-trained large language models to predict and discover nine potential anti-aging targets, opening new avenues for disease treatment.

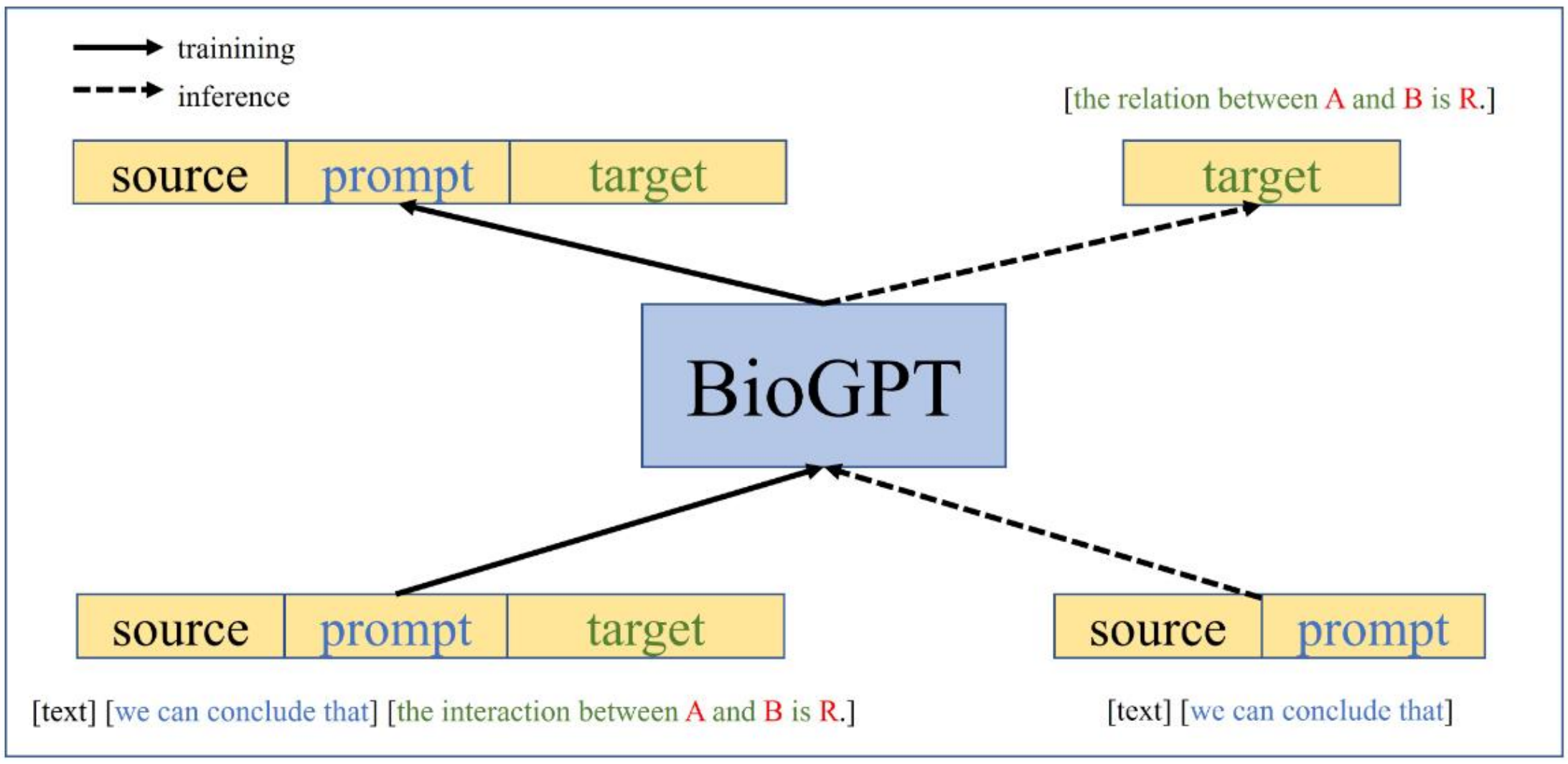

Fig. 2-2 BioGPT framework for downstream tasks

The goal of the Microsoft Research AI Center is to establish a unified foundation model for scientific fields to support broader natural science applications. This model will be based on scientific priori knowledge, described by fundamental physical laws. It needs to have multimodal input and output capabilities, accepting text, one-dimensional sequence data (molecular conformations), two-dimensional images (protein structure diagrams), and three-dimensional data (molecular dynamics simulation trajectories). It must handle molecular systems of varying scales and complexities, including proteins, DNA, and RNA. To improve the model's intelligence and knowledge accumulation, it needs to integrate large language model controllers, generalization tools, and large-scale knowledge bases, combining scientific priori knowledge with big data knowledge. This foundation model will become a general technological core for natural sciences, generating highly intelligent foundation models that can predict and solve scientific problems, providing powerful AI tools for academic research and industrial applications. Hence, it can accelerate scientific discovery processes in various fields. The model can ultimately offer intelligent services to fields such as chemistry, biology, materials, and energy through API interfaces, contributing significantly to scientific progress.

In the life sciences, the application prospects of scientific foundation models are vast. They can extract key knowledge from vast biomedical literature, accurately predict protein mutations and molecule-target affinities, and even directly generate new candidate molecule structures. These capabilities significantly enhancing drug discovery efficiency and success rates, potentially accelerating the advent of new drugs for various challenging diseases.

### 2.1.3 Ethics and Value Maintenance

The rapid development of foundation model technology brings unprecedented opportunities and challenges to human society. It can greatly enhance productivity and quality of life. However, without necessary value guidance and ethical constraints, it may lead to negative impacts and risks. Therefore, it is crucial to use social science methodologies to ensure these powerful AI tools enhance human welfare while maintaining ethical and cultural values. For example, attention to algorithmic fairness is essential to avoid unfair treatment and discriminatory outcomes due to data or model biases. Strengthening privacy protection is necessary to prevent misuse of personal sensitive data. Increasing AI system transparency is also vital, enabling public supervision and accountability and fostering trust (Jobin, et al., 2019). To address these issues, it is urgent to build an interdisciplinary research system that closely monitors the integration of AI and social sciences, ensuring AI advancements align with our core values. Social science research methods can help tackle the diverse challenges AI development brings. For instance, applying psychometrics to AI system evaluation can provide a more comprehensive and objective assessment of cognitive capability structures, shaping AI to align with human values.

Applying psychometric frameworks provides scientifically rigorous methodologies for AI system evaluation. This interdisciplinary integration can deepen our understanding of AI and human intelligence. Initially, construct identification is needed to clarify the latent cognitive structures and capability factors in AI systems, which may include logical reasoning, pattern recognition, language comprehension, and creativity. This requires theoretical analysis and exploratory research. Next, construct measurement involves designing test scenarios and items for each capability factor, establishing quantitative scoring standards. These tests should cover various difficulty levels and knowledge domains, ensuring measurement reliability. The third step is to perform test validation, involving long-term testing of multiple subjects, collecting extensive data, and verifying the measurement scheme's reliability and validity. For AI systems, large-scale computer-simulated experiments can be conducted; for humans, lab and online/offline assessments are needed. This validation step is critical for refining and optimizing the evaluation system. A well-constructed psychometric evaluation framework can provide a comprehensive and objective assessment of AI systems' capabilities, fostering deeper integration of AI with psychology and cognitive science, revealing commonalities and differences between human and machine intelligence. This interdisciplinary integration will significantly advance AI technology, aligning it with human cognitive patterns and enhancing its capabilities, making AI more than just a computational power game but a true reflection of human wisdom.

AI technology is being widely applied globally with profound impacts on economic, political, and cultural domains. Developing value alignment systems consistent with human ethics will bring AI decision-making closer to human moral judgments. This interdisciplinary perspective is essential for fostering responsible, transparent, and human-aligned AI development. The future of AI will deeply influence human civilization's trajectory, so managing this significant transformation is crucial. While guiding technology development with rigorous social science methodologies, embracing innovation with an open and inclusive mindset will ensure AI and human wisdom merge and coexist harmoniously.

## 2.2 Urban Foundation Model

With the rapid development of foundation model technology, the machine's ability to understand natural language, the level of world common sense, and logical reasoning ability have all achieved unprecedented improvement. While the general foundation model performs well in many areas, there is still considerable room for improvement in supporting and understanding urban issues that involve the concept of time and space. The so-called "urban foundation model" refers to the intelligent urban management and service system built based on massive urban data and advanced artificial intelligence technology. It makes comprehensive use of multi-source heterogeneous data in various urban fields such as transportation, energy, environment, medical and health care, and analyzes and predicts the urban operation status by establishing a mathematical model, providing strong support for scientific decision-making and intelligent management. Urban foundation models have the basic characteristics such as strong complexity, high computational performance requirements, strong generalization ability, good adaptability, end-to-end learning support, transfer learning ability, and highly interpretability. It aims to integrate multi-source big data of urban spatial and temporal dimension. This includes deeply integrate key factors such as urban geographic information, structural layout, and functional zoning. The goal is to form a more comprehensive and in-depth understanding of the dynamic evolution and development trend of the city. This understanding provides strong support for urban planning, intelligent operation and management, and sustainable development. This section will focus on the related topics of urban models, focusing on the current situation, challenges, and future development trends of urban models in the data processing, analysis and application. In-depth discussion on how to optimize and customize the foundation model architecture to better meet the needs of urban planning, architectural design, transportation and other specific fields related to spatial relations. It also looks forward to the new trend of technology development in the future.

### 2.2.1 Urban Foundation Model Roadmap and Data Activation Technology System

Modern urban management does face many major challenges, including urban planning management, public safety management, public health management and other fields. To effectively respond to these challenges, we need to use advanced big data and artificial intelligence technology to realize intelligent urban management.

(1) Urban Information

Based on the concept of intelligent management, urban informatization is the primary task, which aims to transform the data in the physical space into the information space. The transformation involves urban information infrastructure, such as GPS, RFID, smart phones, LBS, wearable devices, etc., and the data generated by these facilities includes huge urban data such as mobile phone signaling data and microblog check-in data. By combining the method of artificial intelligence, data mining, machine learning and other cutting-edge technologies, in-depth analysis and processing of these massive urban data will help to provide strong support for smart city management, intelligent urban life, and business intelligence services (Ismagilova, et al., 2019). The core methods of concern involve cutting-edge technologies such as artificial intelligence, data mining and machine learning, which constitute the basic prototype of urban foundation models.

(2) Classification of Foundation Models

According to the different modes of input data, foundation models can be divided into two categories: one is the question-answer foundation language model (foundation Language Models) of "input text". The other is the "input picture speaking image-language model" (Visual-Language foundation models). At present, although foundation models can recognize words and images, their ability to identify spatio-temporal data is relatively limited. They cannot accurately understand spatiotemporal information (Birhane et al., 2023). To enable the foundation model to comprehensively identify text, pictures and spatiotemporal data, it is necessary to adopt the uniform vector representation method of multimodal heterogeneous data; uniformly transform the data of different modes into vector forms, and then input it into the base foundation model for processing. In this way, the foundation model can output the results of text, pictures and spatiotemporal data, such as POI points, road network lines, region surface, trajectory sequence and other forms of output results.

(3) Characterization Learning Method for Urban Road Network

Urban road network is an important part of urban spatial structure, which is of great significance for urban planning and traffic management. In order to fully model road network related information, representation learning is needed to represent urban road network nodes as vectors in European space to capture the topological structure and functional characteristics of road network. In this field, the hierarchical graph neural network model HRNR (Hierarchical Road Network Representation) provides an effective solution. The model organizes the road network into a three-level hierarchy, including functional area layer, such as commercial district, residential area, etc., structural area layer, such as block layer, and road section layer. Two probability distribution matrices introduced are respectively responsible for the allocation of road sections to structural areas and structural areas to functional areas, nodes of different levels can be associated to reflect the hierarchical characteristics of the road network. At the same time, the HRNR model uses the adjacency matrix reconstructed based on the network topology structure and the connectivity matrix reconstructed based on the actual trajectory data, to effectively capture the structure and functional characteristics of the road network. Within the model, the node embedding representation is learned on the whole network through the hierarchical update mechanism, finally realizing the discovery and characterization of the urban spatial pattern structure. This method maps urban road network nodes into vector representation, which not only retains the topological structure information of the road network, but also integrates semantic information such as functional areas and actual travel trajectory, providing strong support for urban traffic analysis and planning decision making.

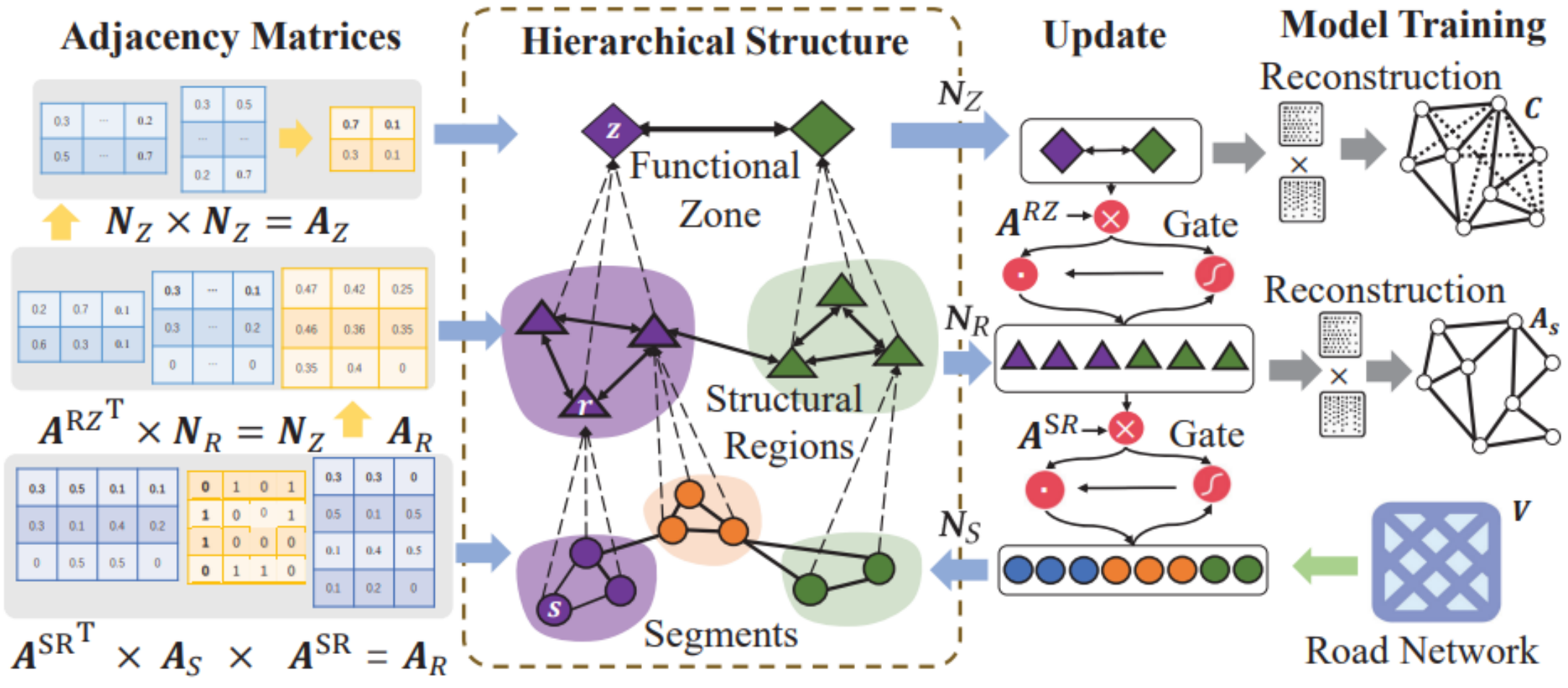

Fig. 2-3 Overall structure of the HRNR model

(4) Characterization Learning Methods for Individual Trajectories

Besides the representational learning of urban road networks, the characterization of individual travel trajectories is equally important. By characterizing individual travel tracks as vectors in European space, the information about individual travel can be fully reflected, laying a foundation for applications such as personalized travel services and traffic flow prediction. Trajectory representation learning (Trajectory Representation Learning, TRL) is a powerful tool to achieve this goal. The goal of TRL is to transform complex raw trajectory data into low-dimensional representation vectors that are not only small and computationally efficient but can also be applied to downstream tasks such as trajectory classification, clustering, and similarity calculation. Jiang et al. (2023) proposed an innovative self-supervised trajectory representation learning framework, START (Self-supervised trajectory representation learning framework with TemporAl Regularities and Travel semantics), to solve this problem very well. The framework integrates temporal rules and travel semantic information and consists of two stages: The first stage is the trajectory pattern enhancement graph attention network, which converts the road network features, such as road topology relationship, and travel semantics, such as travel destination, into representation vectors of road segments. The second stage is the time-aware trajectory encoder, which encodes the road segment representation vector of the same trajectory as the trajectory representation vector, and the time-sensitive trajectory encoder simultaneously integrates time rule information, such as peak hours, weekdays / holidays, etc., so that the trajectory representation can better reflect the spatial and temporal characteristics. This framework method can not only adapt to heterogeneous trajectory data sets across different cities, but also integrate road network information, semantic information, and time information into trajectory representation, providing a new perspective and tool for individual travel behavior analysis, which is of great significance to improving the analysis ability in related fields.

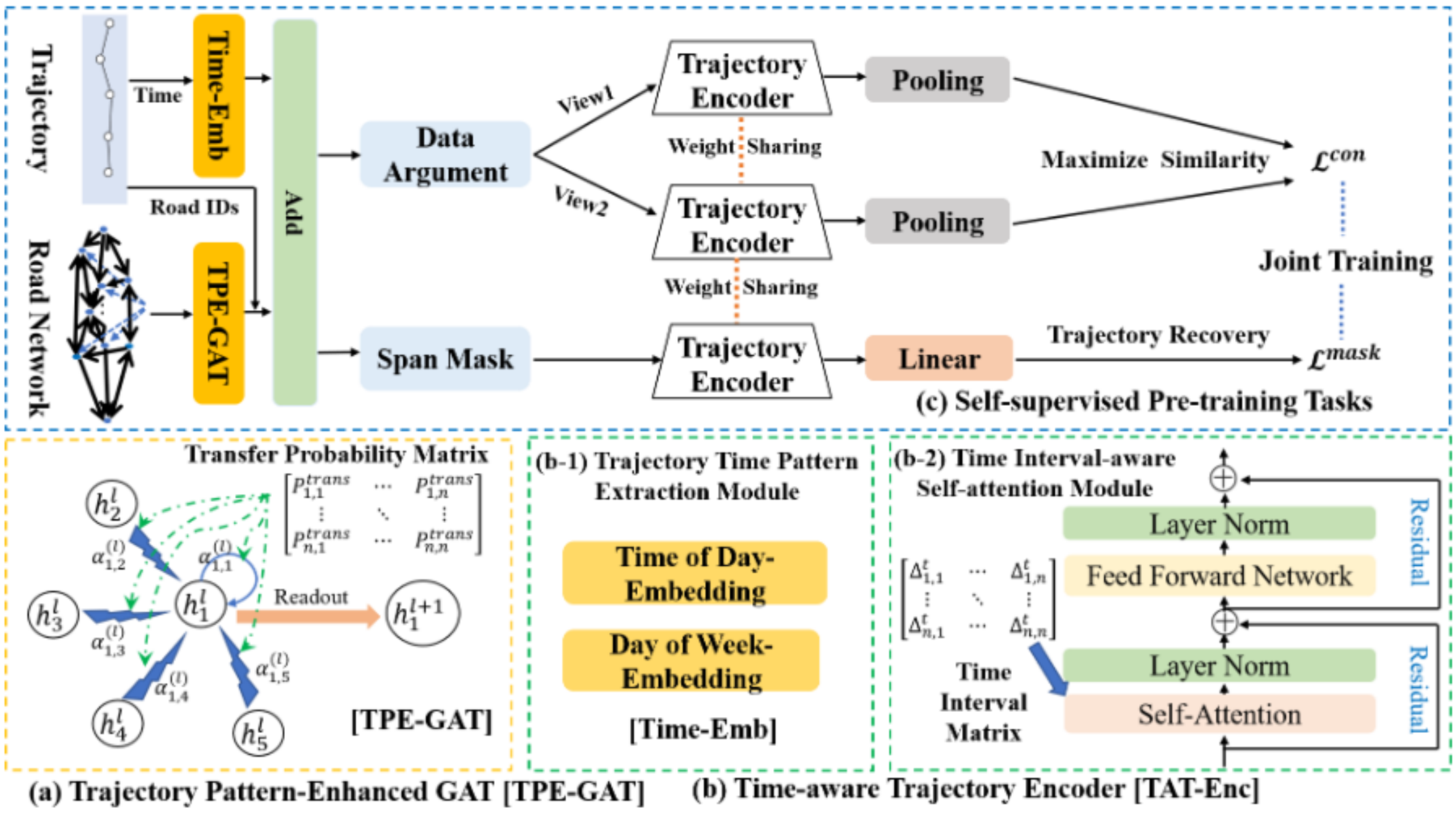

Fig. 2-4 Overall structure of START

(5) Urban Foundation Model and Urban Data Activation

Urban data activation refers to the use of the rich data resources generated in the city. Through the analysis, processing, and application of these data, it aims to improve the operation efficiency of the city. This process also enhances the living quality of life of residents and enhance the ability of urban planning and management. This concept encompasses the entire process, from data collection, collation, to analysis, and application. Based on vector map, the infrastructure of urban spatiotemporal foundation model based on general map representation integrates multi-modal general foundation model and urban digital twin platform. It integrates the basic model of map element representation algorithm and individual trajectory representation algorithm. This architecture provides important support and contribution to major areas of challenges such as planning management, safety management, and public health. By integrating multi-source data and models, the architecture can provide city decision makers with comprehensive urban operation status and trend analysis, helping them to better formulate policies, plan urban development and improve emergency response capacity. Hence, it promotes the sustainable development of cities and the improvement of residents' quality of life.

### 2.2.2 Simulation and Planning Decision of Urban Agent Based on Foundation-scale Model

City is the carrier of human activities and resources highly concentrated in space-time dimension. The orderly operation and sustainable development of city depend on the complicated interaction mode between human and urban environment. In recent years, foundation language model technology has made rapid development. With its excellent reasoning and planning ability in the field of agent simulation, foundation language model technology shows unprecedented application potential. The system will give full play to the advantages of foundation language models and effectively solve existing urban agents, such as vehicles, pedestrians, drones. The simulation model has pain points, such as poor environmental perception ability, vague behavior motivation, and poor behavior coherence, so as to achieve high-fidelity simulation of urban agent behavior patterns. The simulation system will build interaction bridges between agents and urban POI and infrastructure, making the simulation scene closer to the real situation. This advanced simulation system generates realistic traffic flow patterns, crowd trajectories, and emergency response, providing city managers with more accurate decision-making references. Foundation language models also have great development potential in the field of multi-agent city simulation, which will promote the intelligent, optimized and sustainable development of future cities.

(1) Construction and Application of Urban Knowledge Atlas

A generative artificial intelligence (AIGC)-driven urban mobility simulation system is established, which can simulate the physical elements of cities with tens of millions of people at different scales, such as transportation, energy, water resources, etc., and social factors, such as population migration, economic activities, etc., to build a full-factor cross-scale urban simulation model. Building such a foundation urban simulation system requires a combination of advanced technologies (Xu et al.,2021). In terms of interactive systems, Web2D visualization, 3D virtual reality visualization, and decision optimization SDK are needed to provide decision makers with efficient visual interaction and decision support, intuitive data presentation, and decision-making assistance tools. The simulation system requires parallel computing, heterogeneous computing, and other technologies to achieve high-performance computing. It also uses cloud native technologies, such as containers and microservices, to achieve scalable distributed computing architecture, including data standardization tool chain, distributed coordination tool chain, database, and other infrastructure. The data system is responsible for generating high-precision maps from multi-source data. It identifies areas of interest and mines network topology knowledge. Additionally, it discovers movement rules and generates personal trajectories, etc. This system provides data support for knowledge map construction and simulation system.

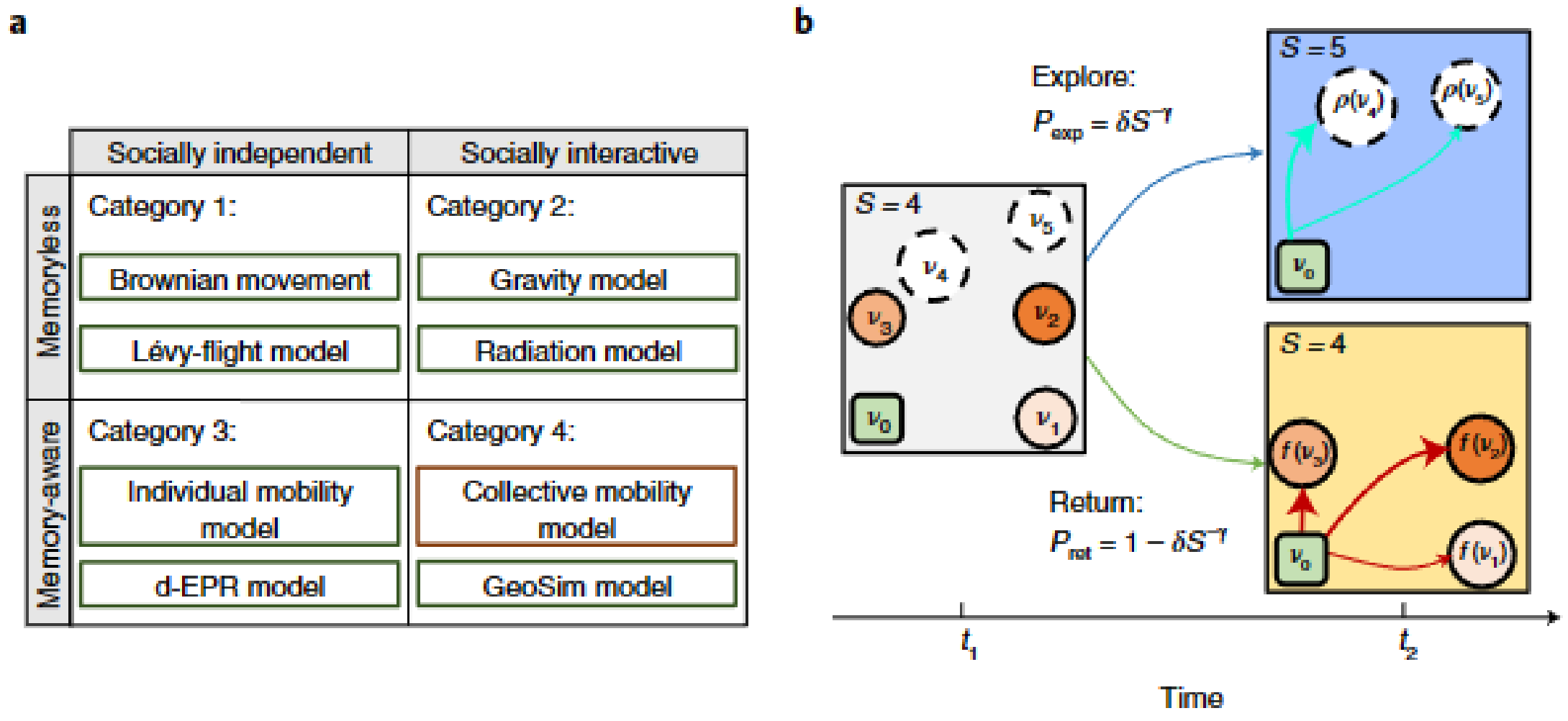

Fig. 2-5 The paradigm of human motion model and the proposed collective flow model

(2) High Performance Simulation Framework

For foundation-scale urban simulation system, high-performance distributed computing framework is the key technical support. Traditional single-machine computing architecture can no longer meet the computing needs of massive data and complex models, so distributed computing mode is needed to disperse computing tasks to multiple nodes for parallel execution, so as to obtain foundation-scale computing power. Spatial region segmentation is a common method to realize distributed computing. Because urban simulation involves geospatial data, the whole urban area can be divided into several sub-areas according to certain rules; the data and computing tasks of each sub-area are assigned to a computing node; and the parallel computing framework is used to maximize the use of computing resources in distributed systems. At the same time, we need to consider the data interaction in the boundary area. Different nodes need to exchange necessary synchronization information to ensure the consistency of the overall data. This requires establishing an efficient communication mechanism between nodes, such as using Redis, a high-performance distributed memory database for information exchange. In addition to spatial segmentation, other task division strategies can also be adopted for different model characteristics and computational characteristics, such as assigning different models to different nodes based on model decomposition, dispersing calculations at different time steps to different nodes based on temporal segmentation. In a distributed computing framework, a central coordination module is required to be responsible for task scheduling and resource allocation, dynamically monitor the computing state of each node, dynamically adjust task allocation according to load conditions, and realize Load Balancer and failover. At the same time, log monitoring and operation and maintenance management mechanisms are also required to ensure high availability of the system (Zheng et al., 2023). In addition to parallel computing, heterogeneous accelerated computing is also an important means to improve computing performance. GPU, FPGA, TPU and other heterogeneous computing accelerators can be combined to deploy computing modules suitable for accelerators to accelerators for higher efficiency.

(3) Support the Urban Science Theory Research and Intelligent Decision-Making Technology Research

This research involves real-time communication optimization, short-term resource allocation, and long-term urban planning. By revealing the micro-behavioral mechanisms behind the macro-development laws of cities, it optimizes urban functions and improves the quality of life of residents. The combination of dynamic models and artificial intelligence simulation technologies is the core of this research field. By establishing a dynamic model based on the microscopic migration behavior of urban residents, combining the key factors of human migration behavior, such as long-term memory and dynamic social interaction, are combined. Using efficient artificial intelligence simulation technology, the laws of urban macro-evolution can be revealed. These laws include the law of urban size distribution, the super-linear relationship between population and urban area, and the distribution of urban population density. This approach establishes a theoretical bridge for understanding the relationship between individual migration behavior and urban evolution law. This method of simulation combined with decision-making has multiple advantages and can be applied to solve problems, such as mobile network energy consumption optimization, mining analysis based on foundation-scale real network traffic and energy consumption data, to build wireless network twin simulation systems to find energy-saving strategies for on-the-spot deployment. In practice, compared with traditional mathematical modeling and operational optimization methods, simulation combined with decision making can significantly improve the carbon efficiency of the network by more than 40%. It can also help 71% of provinces avoid falling into carbon efficiency traps (Li et al., 2023). In addition, this work also achieved precise prevention and control of genetic diseases in the context of limited medical resources, as well as fine-grained, rapid-response epidemic simulation and policy formulation. Compared with the baseline model, the prediction accuracy of daily epidemic increased by more than 31%. The heterogeneity of infectious disease risk within the population was successfully characterized. In terms of vaccine strategy, the overall utility and multidimensional fairness of the designed vaccine strategy can be guaranteed regardless of the number of vaccines or the timing of vaccination (Chen et al., 2022).

(4) All-Element Cross-Scale Urban Simulation Model and System Realization

In the all-element cross-scale urban simulation model and system realization, through simulating the spatial planning of urban communities, attention is paid to the basic elements of urban community spatial planning, and reasonable spatial layout of land. Roads and other elements are carried out to realize urban development. The main idea is community planning based on urban simulation and reinforcement learning decision, which simulates the process of urban development through plot cutting, road construction and other operations, combines action selection, representation extraction, urban spatial topology modeling, and other technologies to make reinforcement learning decision, so as to generate planning schemes that meet the needs of urban development. Through simulation and decision interactive training, the urban planning decision model is continuously optimized, and at the same time, the huge solution space is efficiently searched to provide decision support for urban planning. The simulation environment itself can also feedback the decision model through the life circle community simulation, forming an interactive learning cycle. As a complex system, the demand of digitalization is increasing day by day. Urban system is a multi-level, multi-element, and highly interactive system, covering all aspects of city. It forms a dynamic network, in which the complex interaction between human activities and urban environment is constantly evolving. To cope with this complexity, it is necessary to establish a city generative intelligent infrastructure platform. The architecture of this platform should be built on an open digital basis. It should realize the interaction between the city simulator and the city knowledge graph through data flow transmission. Generative pre-trained model enables interaction with users. Such a platform can provide practical application scenarios, such as travel planning, location optimization, travel investigation, etc., for urban planning, so it improves the efficiency and accuracy of urban planning.

By combining with agent, city GPT can generate individual and collective behaviors in city simulator and provide decision support and assistance for decision makers. All-element cross-scale urban simulation model and system realization is a challenging and promising project, which combines simulation technology, artificial intelligence, and urban planning. It provides new ideas and methods for urban sustainable development.

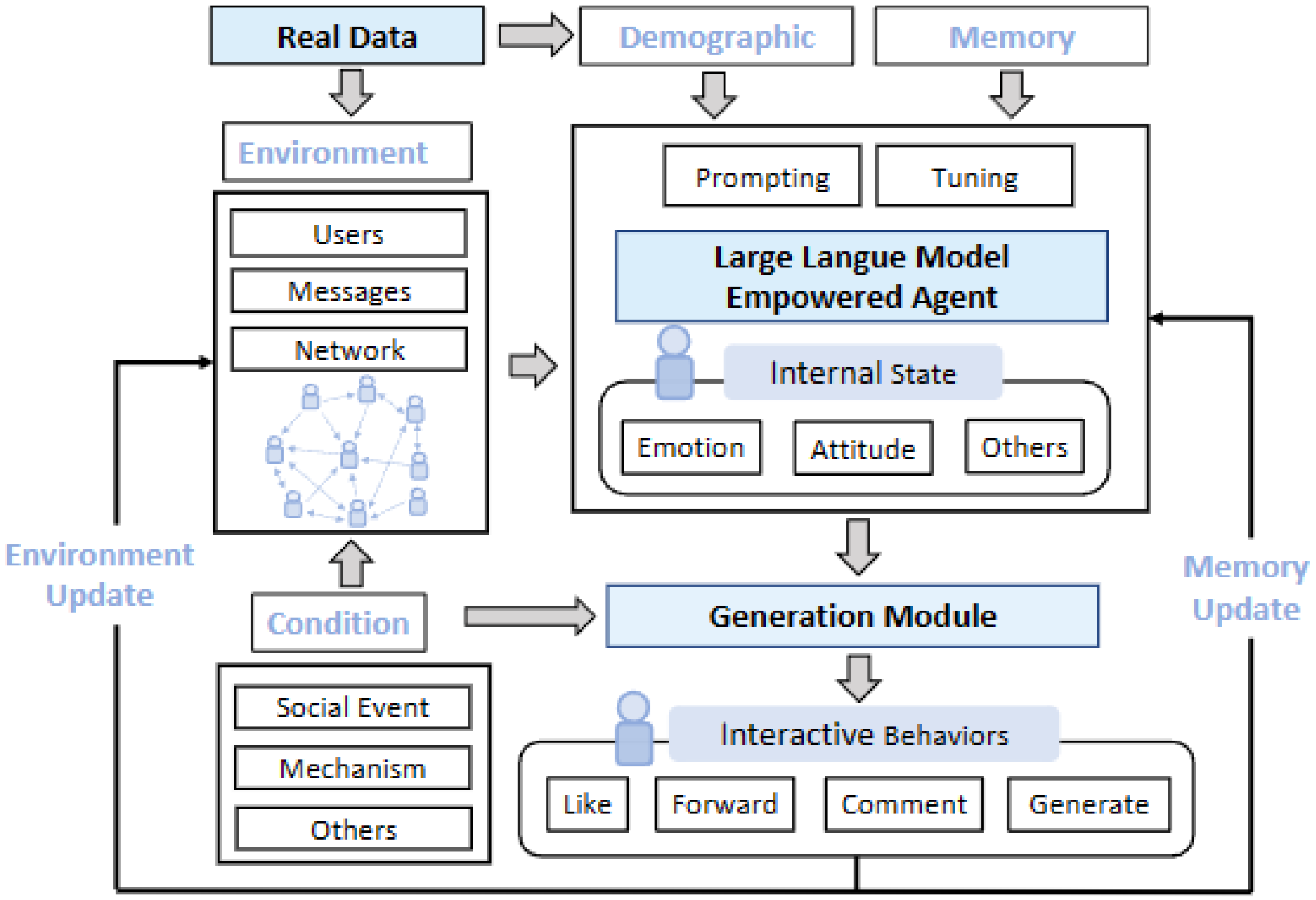

Fig. 2-6 Social network simulation system

### 2.2.3 Construction and Practice of Urban Space-Time Foundation Model

The rise of pre-trained models and foundation language models has undoubtedly revolutionized the field of artificial intelligence. These advanced models have greatly enhanced the processing and understanding of natural language by machines, opening new possibilities for AI applications in various industries, and ushering a new era of integration with AI in various fields. However, in the face of such complex urban systems, to take full advantage of new technologies requires deep thinking about how to apply generic language models to urban computing. While these models perform well in natural language processing, they cannot be directly applied to urban data and physical scenes. City data usually contains complex space-time dimensions and heterogeneous information from multiple sources. It requires professional adjustment and optimization of the model to adapt to the special needs of urban computing. At the same time, how to build a foundation model system based on city database is another problem to consider. As a huge organic system, the city contains rich data resources, such as transportation, environment, population, architecture, and other multi-dimensional information. Through deep mining and modeling of this data, we can construct foundation models specifically for urban problems to more effectively support the fields of urban planning, management, and decision-making (Sassite et al., 2022).

Smart space is the trend of future city development. With the help of artificial intelligence (AI) and digital technology, the boundary between real world and virtual city boundary is gradually blurred, thereby realizing deep fusion. This trend makes the interaction of the world evolve from traditional physical space to a new realm combining virtual and physical. In urban space, multiple activities such as life, transportation, commerce, and governance jointly construct a complex network of space-time relations. To effectively manage and utilize these spatiotemporal relationships, the spatiotemporal IA technology system has emerged. The spatiotemporal IA technology system is a comprehensive technology framework that aims to develop around data modeling, AI computing and XR interaction. It takes digital twins as the cornerstone and meta-universe as the ultimate goal. In the spatiotemporal AI technology system, it includes spatiotemporal perception technology, spatiotemporal data engine, and spatiotemporal atlas engine. Spatio-temporal sensing technology is responsible for collecting and sensing spatio-temporal data from all over the world. Spatio-temporal data engine is responsible for processing and managing these data, while spatio-temporal graph engine is responsible for building and maintaining the relationships and connections of these data. Foundation models are coupled with spatio-temporal AI technology, which is a further embodiment of intelligent space and a key to the development of intelligent space. Foundation models are divided into general foundation models and domain foundation models. General foundation models use a unified model architecture and adopt the same learning mode to build a general vocabulary that can be applied to multiple modalities, unifying all tasks into sequence tasks. Domain foundation models combine general pre-training with specialized domain pre-training to form professional business scenario applications. Domain foundation model is the key to realize intelligent space. There are many ways to construct domain foundation model. We can integrate general foundation language model with vertical domain knowledge to create vertical domain foundation model on top of general foundation language model. Alternatively, we can directly construct domain foundation model (vertical domain small model) through vertical domain data. The combination of ChatGPT and domain foundation model. The coupling application scenarios of spatio-temporal AI technology and foundation model are rich and diverse, which can provide intelligent decision support for urban sustainable development. These application scenarios include but are not limited to site selection recommendation, network planning, regional research and judgment, market demand assessment, traffic convenience assessment, etc., providing a brand-new perspective and solution for urban management and planning.

As an important combination of modern technology and smart city construction, foundation urban models have broad development prospects and huge application potential. With the rapid development of artificial intelligence technology, the application scope of foundation urban models is expanding. Traffic management and planning, smart city construction, environmental monitoring, governance, as well as urban planning and land use, etc. have begun to use this technology to realize intelligent management and operation. This has not only increased the efficiency of city operations, but also greatly improved the quality of life of the people. Although in the fields of natural language processing and image recognition, foundation models have made significant progress, making it possible to handle foundation amounts of data and achieve fine management. However, foundation urban models face many challenges and difficulties in their application, including model interpretability, data privacy protection, and cost control, which are still important tasks to be solved. If these challenges can be solved, it will help foundation urban models maximize their functions while fully respecting individual rights and interests, thereby promoting intelligent urban progress. It can be predicted that in the future, the importance of foundation-scale urban models in urban intelligent management, sustainable development, and improvement of residents 'quality of life will gradually increase.

### 2.3 Space and Space Remote Sensing Foundation Model

At present, the world has entered the era of hourly fast response and submeter remote sensing observation. Remote sensing technology uses electromagnetic waves as an information carrier, greatly expanding the range of human perception ability. With the development of interdisciplinary and cross-border integration, the application field of remote sensing has been further expanded, bringing a broader application prospect. This section will introduce the key technologies, methods, and applications of the foundation space remote sensing model. Including the preliminary cognition and practical application of remote sensing AI foundation models, these foundation-scale models use deep learning and other technologies to process massive remote sensing data, so as to provide accurate information support for geological exploration, environmental monitoring, and other fields hold. The exploration and practice of integrated intelligent technology of remote sensing and geographic information system (GIS) combine remote sensing data and geographic information. This combination can realize more accurate spatial data analysis and application. Additionally, it provides more comprehensive decision support for urban planning and resource management. The self-supervised deep learning method for foundation-scale hyperspectral image interpretation automatically extracts the feature information in the hyperspectral image. This is achieved through the deep learning algorithm to realize the accurate classification and identification of ground object types. Based on the understanding of the above technical methods, the application practice mode of the remote sensing model of air and space information. The transition of the remote sensing model is further explored driving force. Through in-depth research and application of space remote sensing model, remote sensing data can be better understood and analyzed. Then, this data can be applied to environmental monitoring, resource management, urban planning and other fields.

2.3.1 Preliminary Cognition and Practical Application of Remote Sensing AI Foundation Model

The emergence of ChatGPT marks the entry of artificial intelligence into the era of foundation models. The foundation-scale neural network model brings more general application capabilities to artificial intelligence, which also provides new opportunities and challenges for remote sensing data analysis. At the same time, with the advent of the era of remote sensing big data, earth observation and remote sensing technology has experienced rapid development. Satellite constellations are constantly emerging, providing us have more types and more data amount of remote sensing data than ever before. This means that human beings have entered the era of remote sensing big data. Remote sensing big data analysis system is the basis of foundation model promotion in the field of remote sensing. For instance, Cangling system is an intelligent information extraction system based on deep learning. This kind of remote sensing information extraction system provides data samples for remote sensing AI foundation model. A foundation number of data samples and relatively convenient access methods have promoted the breakthrough progress of foundation language model and foundation visual model, and then led the whole society into the era of foundation model. Based on the current development background of artificial intelligence and intelligent interpretation of remote sensing information, this section summarizes and analyzes the development status of remote sensing model. It also expounds the preliminary trend of the development of remote sensing model combined with the existing research work. The development of foundation remote sensing models should not only focus on foundation-scale. The training and processing technology of data should also combine with the characteristics of intelligent interpretation of remote sensing information. This approach fully excavates the valuable information in remote sensing data and provides more accurate and efficient interpretation ability (Hong et al., 2021). Through the development and application of remote sensing model, the analysis and decision-making ability of remote sensing data can be further improved, so as to promote the application of remote sensing technology in various fields.

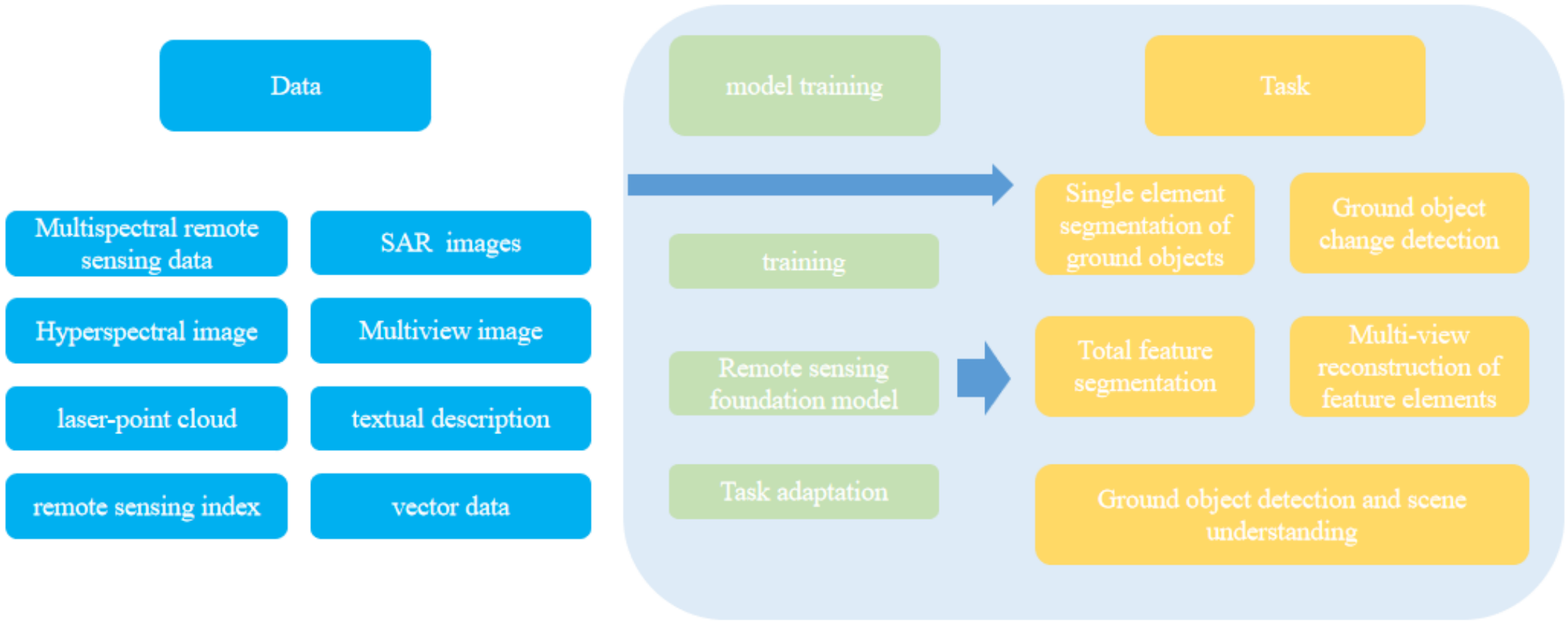

Fig. 2-7 The flowchart of remote sensing foundation model

(1) Segment Anything Models, SAM

Segment Anything Models (SAM) refers to a kind of neural network model, used for image segmentation tasks, so there are still some problems in the application of remote sensing image. SAM lacks understanding because of the current training dataset. SAM performs better in dividing high-resolution remote sensing images, which can accurately segment all kinds of ground objects, but performs poorly in processing low-resolution global land use data. Moreover, because remote sensing images need to have semantic information, the Mask generated by SAM is missing Lack of labels, which makes the prompt semantic information difficult. Designed for segmentation and detection tasks, SAM cannot perform some remote sensing specific tasks, such as change detection and vector output. When the boundary of ground objects in remote sensing images is poorly defined due to complex scenes, it is difficult for SAM to comprehensively segment the remote sensing image target, so the results will depend heavily on the type, location, and number of prompts. The diversity of remote sensing data is also a problem. SAM is mainly concentrated in Prompt; while outside Prompt, SAM concentrated only in natural images. SAM has the problem of network structure limitation as a common image encoding. It is difficult to meet the remote sensing fine-grained task in terms of efficiency and precision.

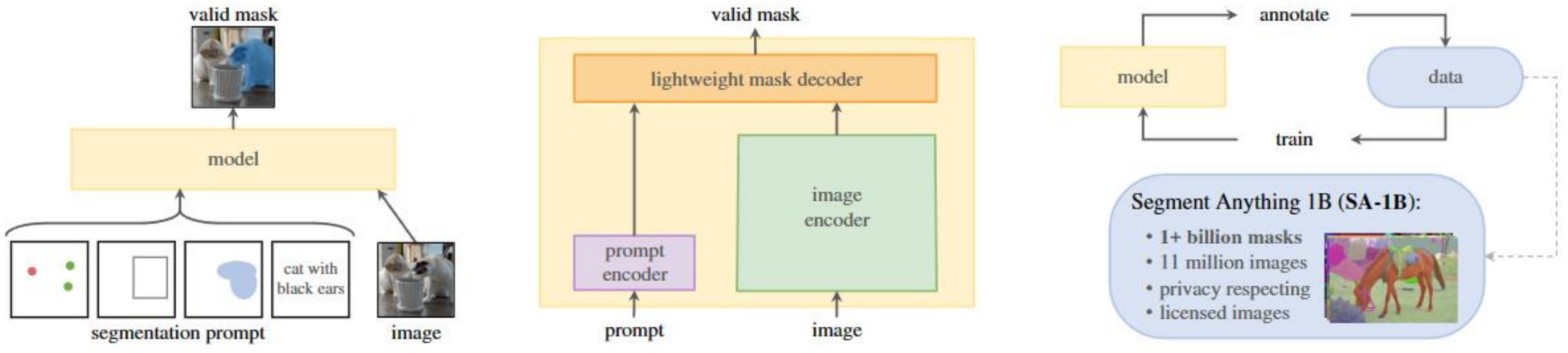

Fig. 2-8 Segmentation fundamental model

However, SAM also has some advantages in remote sensing applications. Although SAM is only trained for natural images, it can identify and analyze high-resolution remote sensing images, showing a strong ability to generalize and understand images. This provides a basis for experimental proof for the study of visual multimodal foundation models, which proves that visual foundation models are feasible. The data engine designed by SAM provides an effective way for the construction of foundation data sets. It also provides a guarantee for the training data of the foundation visual model. The network structure design of SAM conforms to two rules: multimodal data embedding and fusion. The model structure completely breaks the information barriers between images, text, rectangular boxes, and other prior knowledge; the moderate number of SAM model parameters provides flexibility for model training and deployment. SAM application scenarios exist in 3D applications, video tracking, image generation, interactive annotation tools, image segmentation, target detection, and image repair (Osco et al., 2023). Since SAM was proposed and opened to training weights, remarkable results have been achieved in the field of image and vision. Its foundation model shows excellent image understanding ability and wisdom in Each downstream task can play a role. It will also have a positive impact on the remote sensing vision tasks.

Although SAM has some problems in the application of remote sensing image, it also provides the visual multimodal foundation model of the basis of experimental proof. In the multiple visual tasks, it showed powerful image understanding ability. Based on its secondary work, each downstream task can assign. On some basic remote sensing visual tasks have the shadow of the SAM. Further research and development can further improve the performance of SAM in remote sensing image applications and provide better support for research and decision-making in the field of remote sensing.

(2) Current Status of Foundation Remote Sensing Model

Existing remote sensing foundation models include ViTAEv2, RingMO, and RS5M. The advantage of ViTAEv2 lies in the use of ViTAE network for pre-training and optimization through millions of remote sensing image classification tasks, covering a variety of remote sensing image data sources with different sensors, image sizes and resolutions. It can reduce the bias of the CV foundation model in training data and improve the efficiency and precision with the improved transformer module. However, the disadvantage of ViTAEv2 is that the pre-training task is scene classification. With relatively coarse feature granularity, it requires foundation-scale supervised training data, training data set acquisition cost to only support as pre-training, which is lack of direct application of deep feeling task. RingMO Two million remote sensing images are used for MAE pre-training. Image data sources are diverse, covering a foundation number of domestic satellite images with different phases, different resolutions, and different regional remote sensing images. Its remote sensing downstream task reduces the bias of CV foundation models in training data. Also, it uses the Swin series of advanced transformer structures to perform better in image reconstruction tasks. However, the disadvantage of RingMO is that its input mode is relatively single and lacks many the embedding of spectral, vector, text, and other modes. RingMo only supported as pre-trained models, which lacks the ability of direct generalization to remote sensing tasks. RS5M realizes the migration of general foundation models to the field of remote sensing by building five million-scale image-text matching datasets. It shows a great ability in the image classification task. However, the disadvantage of RS5M is that the quality of text description in the dataset remains to be improved. Additionally, RS5M’s ability to build more powerful remote sensing foundation models based on foundation datasets is lacking. It still lacks in fine-grained processing details. The existing remote sensing foundation model deals with the relationship between natural scene images and remote sensing images. There are also some deficiencies in the domain differences, leading to its poor performance on remote sensing tasks. High-quality remote sensing image data sets are still lacking in terms of foundation range and similar application scenarios, which also limits the application development of existing foundation models. The subsequent research orientation needs to be improved for the above problems to improve the performance and wide application of remote sensing foundation models.

(3) Research Idea of the Remote Sensing Foundation Model

At present, there are many problems in the field of foundation model processing remote sensing information, such as model based on scattered small data set training, remote sensing information mining, and expression, such as prior knowledge, model accuracy, and generalization ability, is poor, which demonstrates single remote sensing data information limitations, segmentation task training cost a foundation number of researchers low level repeated problem. At the same time, the foundation model itself has higher dimension of training data information, which is beneficial to learning essential features; the model applies self-supervised learning algorithm to reduce training development cost, learning task-independent general knowledge, supporting low-cost segmentation task generalization, and further breaking the existing model Type structure accuracy limited potential and other advantages. Therefore, based on the existing problems and potentials of the above foundation model, the research idea of remote sensing foundation model tends to two directions, which is the remote sensing adaptation based on other existing foundation models and the reconstruction of pre-training with remote sensing data.

| Foundation data | Model characteristics | Application potential of remote sensing | Instance |
|---|---|---|---|
| MAE | Self-supervised learning | Large-scale RS image pre-training | SatMAE |
| SAM | zero-shot Instance segmentation | RS image semantic segmentation and sample labeling | SAM-CD |
| Grounding-DINO | Open set object detection | Object detection in RS images based on text prompts | Text2Seg |
| CLIP | Graphic matching | RS image classification, RS image-text data set construction | RS5M |
| BLIP | image description | RS image classification, RS image-text data set construction | RS5M |
| DELL.E | Image generation based on text prompts | Assisted remote sensing image generation | - |

Fig. 2-9 Summary of existing foundation models

The first direction is the remote sensing adaptation based on other existing foundation models. The existing CV / NLP model has a strong ability of general knowledge learning and expression, which can be well adapted to remote sensing tasks with a small amount of remote sensing knowledge guidance or prompt. For example, MAE has the characteristics of self-supervised learning and has the potential of foundation-scale remote sensing image pre-training. SAM has zero-shot instance segmentation characteristics, which can be used for remote sensing image semantic segmentation and remote sensing sample annotation. Grounding-DINO has great potential in remote sensing image object detection through open set object detection. CLIP and BLIP are based on the characteristics of graphic matching and image description for remote sensing image classification and remote sensing image-text data set construction. DELL.E, with the ability of text prompt image generation, can be used to assist the remote sensing image generation. The above model was used to assist in the generation of remote sensing pre-training datasets for subsequent processing. By dividing foundation model (SAM, FastSAM) and remote sensing prompt, the model can conduct semi-automatic annotation of remote sensing images. It uses graph-text matching foundation model (BLIP, CoCa) to realize the construction of remote sensing image-text matching datasets or remote sensing scene classification dataset. The map image generation model foundation model (DELL.E) realizes the automatic simulation and generation of remote sensing images. The foundation visual model is used to extract the features of remote sensing images. Then it embeds adapter or fine-tuning classifier to reduce the number of remote sensing samples and to improve the generalization ability. By extracting features in Fast SAM and completing the adaptation of remote sensing image feature extraction through adapter, the change detection task is completed. Based on the remote sensing model, we generate prompts for the foundation model and complete the remote sensing task in conjunction with the foundation model. Based on the remote sensing change monitoring network, changing monitoring points are independently generated to generate point tips. Additionally, SAM is used to feature draw. The semantic segmentation of remote sensing is completed by using the general performance of foundation models such as vision and text (Yang et al., 2021). Adapting remote sensing with existing models to construct remote sensing models.

Foundation model construction based on pre-training of remote sensing data is another important direction. It requires a complete process from data to model training to task implementation. Starting from the remote sensing sample library construction, large-scale parameters are used to mine remote sensing data information. This includes sample images, phenology information, vector files, ground observation information, POI information. The remote sensing multimodal knowledge of remote sensing sample library, build a multimodal remote sensing mapping knowledge base. This base includes high quality, complete scene, modal diversity, and foundation-scale training data It meets the need of the model for multiple sensors, multiple phases, more meteorological conditions, across regions, and cross resolution applications for the demand of the scene. On this basis, in view of the remote sensing model training, establish model remote sensing knowledge understanding and rules. Put forward "object-data-scene-task" integration of knowledge understanding and expression method. Use multimodal remote sensing samples involving knowledge and rules, the diversity of data sources, and modal diversification of remote sensing data model training and optimization. In addition, a remote sensing foundation model network based on deformable convolution can be designed. This network model addresses the problem of insufficient acquisition features at the image encoder level and insufficient fusion of remote sensing feature features of existing foundation models. It constructs a synesthesia foundation model based on formable convolution image encoder to learn common features of complex structures at multiple scales, learning the multi-level features of multi-scale remote sensing features. The pre-training strategy of foundation model based on self-supervised learning effectively utilizes large-scale data. Therefore, most of the current foundation model training strategies use self-supervised learning to complete the training of foundation data amount. The common training methods include MAE based on image mask-reconstruction and CLIP based on image-text matching. MAE and CLIP have the advantages and disadvantages of each other; the two training modes can be integrated. Finally, the remote sensing tasks are realized through the task migration optimization generalization breakthrough.

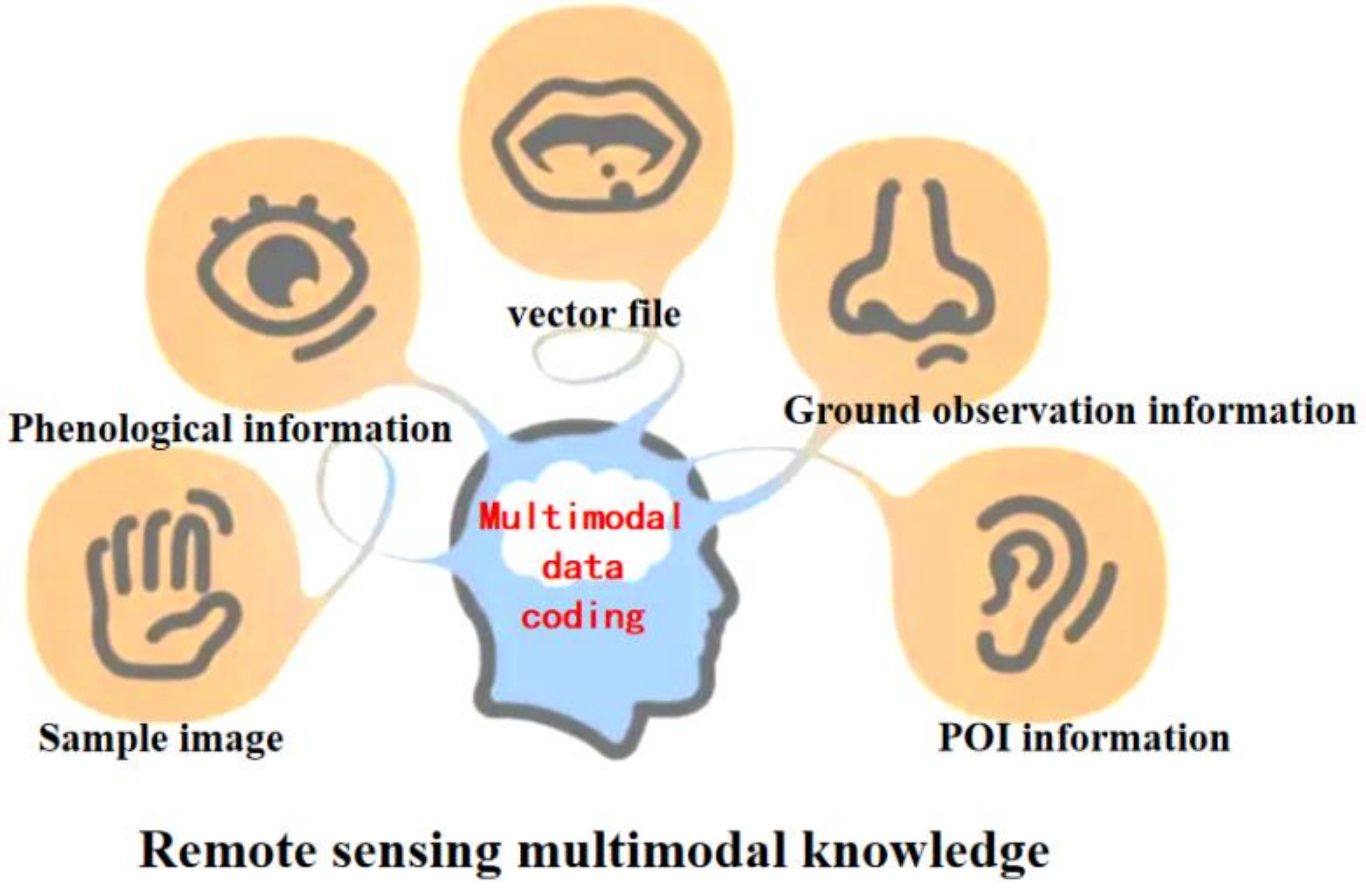

Fig. 2-10 Remote sensing multi-model knowledge

Foundation remote sensing model is an effective and inevitable way to solve the rapid and intelligent extraction of remote sensing information. At present, there are still some defects in the existing visual foundation model applied to remote sensing data, including the lack of training data, the limitations of network structure, and the differences in application scenarios. Therefore, to further improve the processing efficiency and accuracy of remote sensing data, a series of strategies are needed to develop remote sensing AI foundation models. On the one hand, the existing foundation models such as vision, natural language, and text can be used to improve their performance in the field of remote sensing by adjusting and adapting the remote sensing data and application scenarios. For example, knowledge can be known by introducing remote sensing data Knowledge and prior information, transferring the learning or fine-tuning of existing foundation models to make them more suitable for processing remote sensing data. This approach can save the training cost while maintaining the generalization ability and precision of foundation models. On the other hand, we can start from using remote sensing data to design more network structures based on the characteristics of remote sensing data and application scenarios, so as to improve the accuracy and efficiency of foundation models. Through pre-training on large-scale remote sensing data and combining domain knowledge and prior information, foundation models are able to better understand and analyze remote sensing data and achieve rapid intelligent extraction. With the development of remote sensing foundation models, it can be oriented from a single task foundation model with good generalization ability and precision gradually turn to pre-trained base models with generality ability in a wide range of tasks and domains. This pre-trained basic model can provide basic support for different remote sensing tasks, reducing the cost of repeated training, while having high accuracy and generalization ability. By adapting the existing foundation model and using remote sensing data for pre-training, a more intelligent and efficient remote sensing foundation model can be developed, which can bring better effect and efficiency for the extraction of remote sensing information and application.

2.3.2 Exploration and Practice of Intelligent Technology Integrating Remote Sensing and GIS

Advances in big data and cloud computing have brought new opportunities for the development of remote sensing and GIS integrated platform software. The application of these new technologies makes remote sensing and GIS data processing more efficient and flexible. At the same time, the development of foundation model technology has also made the integration of artificial intelligence, remote sensing, and GIS technology more closely. The deep integration of artificial intelligence in the field of GIS and remote sensing provides a new vision for the development of spatial intelligence technology, optimizing the design of algorithms and models. By combining deep learning and remote sensing image processing algorithms, the automatic extraction and analysis of ground objects, landscape, and spatial information in remote sensing images can be realized further application to the automatic interpretation of remote sensing images, ground object classification, object detection, and change monitoring. The integration of artificial intelligence and GIS technology can also strengthen the intelligent management and analysis of remote sensing and GIS data. Through the intelligent storage, retrieval, processing, and analysis of remote sensing and GIS data, the efficient management and utilization of foundation-scale and high-dimensional geographic data can be realized (Song et al., 2020). Based on practical application, intelligent decision-making and planning provide scientific support for the fields of urban planning, environmental management, and resource utilization. In the future, along with the development of artificial intelligence technology with continuous progress, space intelligence technologies and products will be further developed to provide more accurate and efficient solutions for earth observation, resource management and environmental protection.

(1) GIS Integrated Intelligent Technology

GIS integrated intelligence technology combines geospatial intelligence, Artificial intelligence, and business intelligence to extract information and knowledge from spatial data. It uses artificial intelligence technology for data analysis and decision support to improve business and management decisions. In this area, the GI Pyramid provides a framework for the development of spatial intelligence, including Geo-control, Geo-design, Geo-decision, Geo-visualization, and Geo-perception. The integration of AI technologies enables spatial intelligence to better perform Geo-control, Geo-design, Geo-decision, and Geo-visualization, and other tasks. At the same time, some GIS technology companies such as SuperMap also provide technology systems including big data GIS, artificial intelligence GIS, new-generation three-dimensional GIS, distributed GIS, and cross-platform GIS to meet the needs of GIS integrated intelligent technology (Song et al., 2021). By combining GIS and artificial intelligence, more intelligent, efficient, accurate geospatial data analysis, and management can be achieved.

(2) Exploration and Practice of the Remote Sensing Foundation Model

SuperMap, the technical features of artificial intelligence GIS, include advanced model algorithm, complete process tools, rich AI functions, and rich pre-training model, which can improve the application effect of spatial intelligence. Advanced model algorithms, such as Segformer, EffcientNet, Cascade R-CNN, Siam-Segformer, SFNet, and RTMDet, are used to process remote sensing data with superior performance. Sample management, training data generation, model training, model reasoning, post-processing of reasoning results, model evaluation, and so on are used the complete process tool makes the whole remote sensing data processing process more efficient and complete. Using rich AI functions, such as binary classification, object detection, object classification, object extraction, scene classification, change detection, to meet the needs of different remote sensing application scenarios. In addition, SuperMap also provides a wealth of pre-training models, including urban water extraction model, domestic urban building extraction model, aircraft and ship target detection model, domestic cultivated land extraction model, domestic greenhouse extraction model, etc. These models are trained with over 1 billion labels and have high accuracy and generalization ability. From the SuperMap AI foundation model integration exploration, its built-in SAM image segmentation foundation model contains more than 1 billion label training, providing batch and interactive remote sensing image segmentation capability support (Song et al., 2019). For SAM image segmentation foundation model, its built-in model supports batch and interactive remote sensing image segmentation tasks. Combined with the pre-training model of image target detection, semantic information and target prompt box can be output to improve the extraction effect of small objects in remote sensing images. The model has a flexible structure and supports custom substitutions. Input is considered as the original image (for batch segmentation) or prompt information (for interactive segmentation). Output is considered as the ground object split the results. In addition, SuperMap has also made rich progress in the spatial empowerment of AI technology, including the development of 3D GIS visualization, simulation from sun to night. In addition, SuperMap also uses retrieval augmented generation (RAG) technology to achieve professional field capabilities beyond traditional foundation models by combining foundation language models with external knowledge bases, such as traditional web queries. These explorations and practices provide a new vision and direction for the software research and development of the remote sensing and GIS integrated platform, continuing to promote the integration of spatial intelligence technology and artificial intelligence.

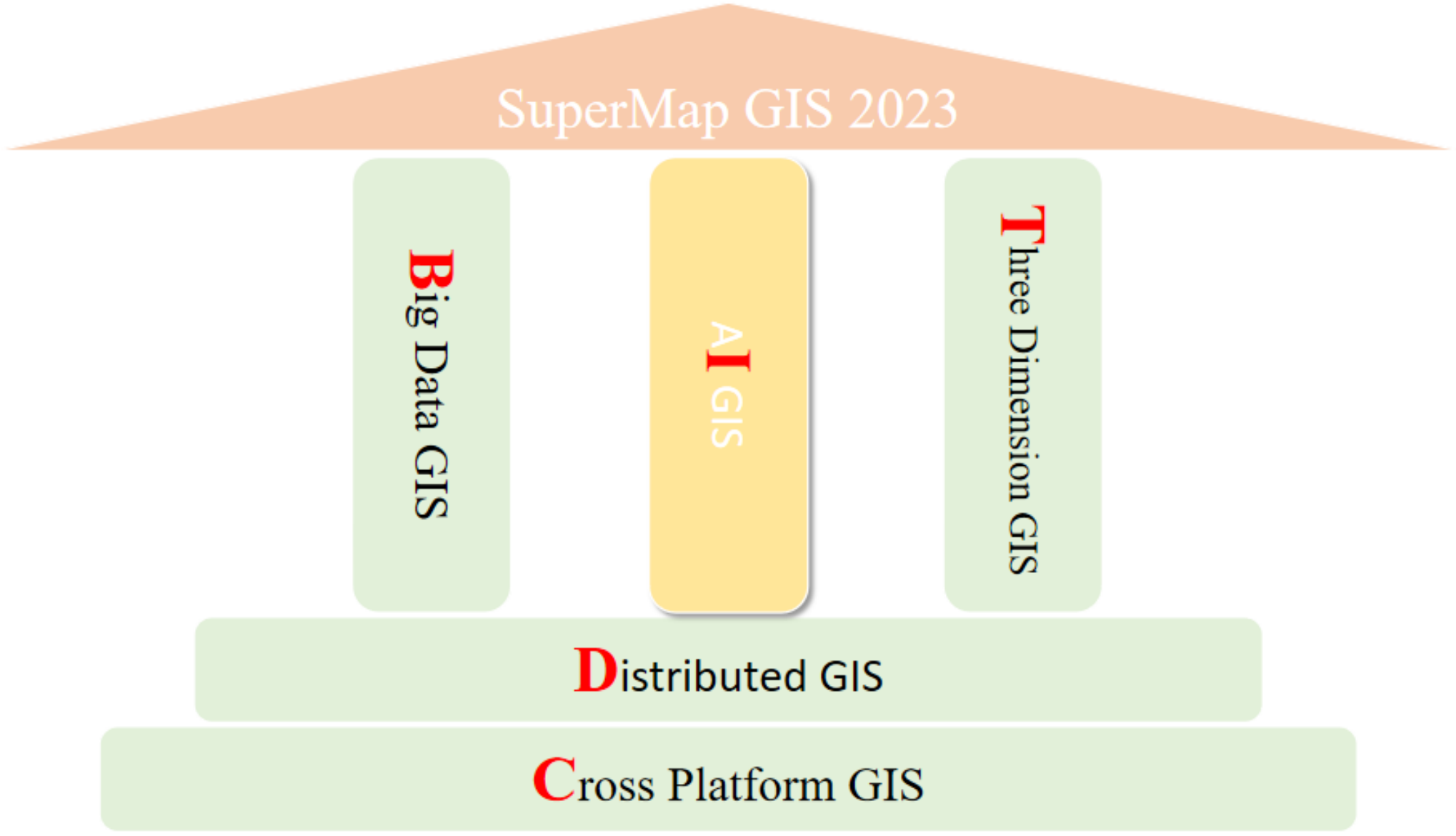

Fig. 2-11 Technology systems of SuperMap GIS (BitDC)

### 2.3.3 Data Benchmark and Learning Paradigm for Hyperspectral Remote Sensing Foundation Model

The data benchmark and learning paradigm for hyperspectral remote sensing foundation model is a solution proposed to overcome the current challenges of hyperspectral interpretation. Hyperspectral remote sensing data contains rich spectral information and can provide more detailed and accurate classification and identification results of surface targets. Hyperspectral imaging is one of the most important remote sensing imaging techniques by combining imaging techniques and spectroscopy to detect the spatial and spectral information of ground targets. However, current hyperspectral interpretation efforts are limited by the small size of the datasets and insufficient network generalization ability, leading to the inability to obtain accurate classification results on large-scale data. For solving this problem, a data benchmark for the hyperspectral remote sensing foundation model can be established. The data benchmark contains a large-scale hyperspectral imaging dataset, which contains rich ground target categories and spectral information. By using this data benchmark, more samples and diverse scenarios can be provided to fully train and test hyperspectral foundation models, improving the generalization ability and migration ability of networks. To better use HMS data for deep learning, self-supervised learning methods can be integrated into HRS data interpretation. Self-supervised learning is an unsupervised learning method by using the data itself. The sign is trained without relying on manually annotated labels. By establishing a self-supervised deep learning network suitable for hyperspectral remote sensing data, self-learning can be conducted through the intrinsic structure and information of the data to improve the expression and interpretation ability of the model for hyperspectral data. The data benchmark and learning paradigm for hyperspectral remote sensing foundation model can provide strong support for the research and development of hyperspectral interpretation. By training and testing on large-scale data, as well as using self-supervised learning, the performance and application capability of foundation models of hyperspectral remote sensing data can be further improved, so as to provide more accurate and application for the interpretation and application of hyperspectral remote sensing data reliable support (Huang et al., 2022).

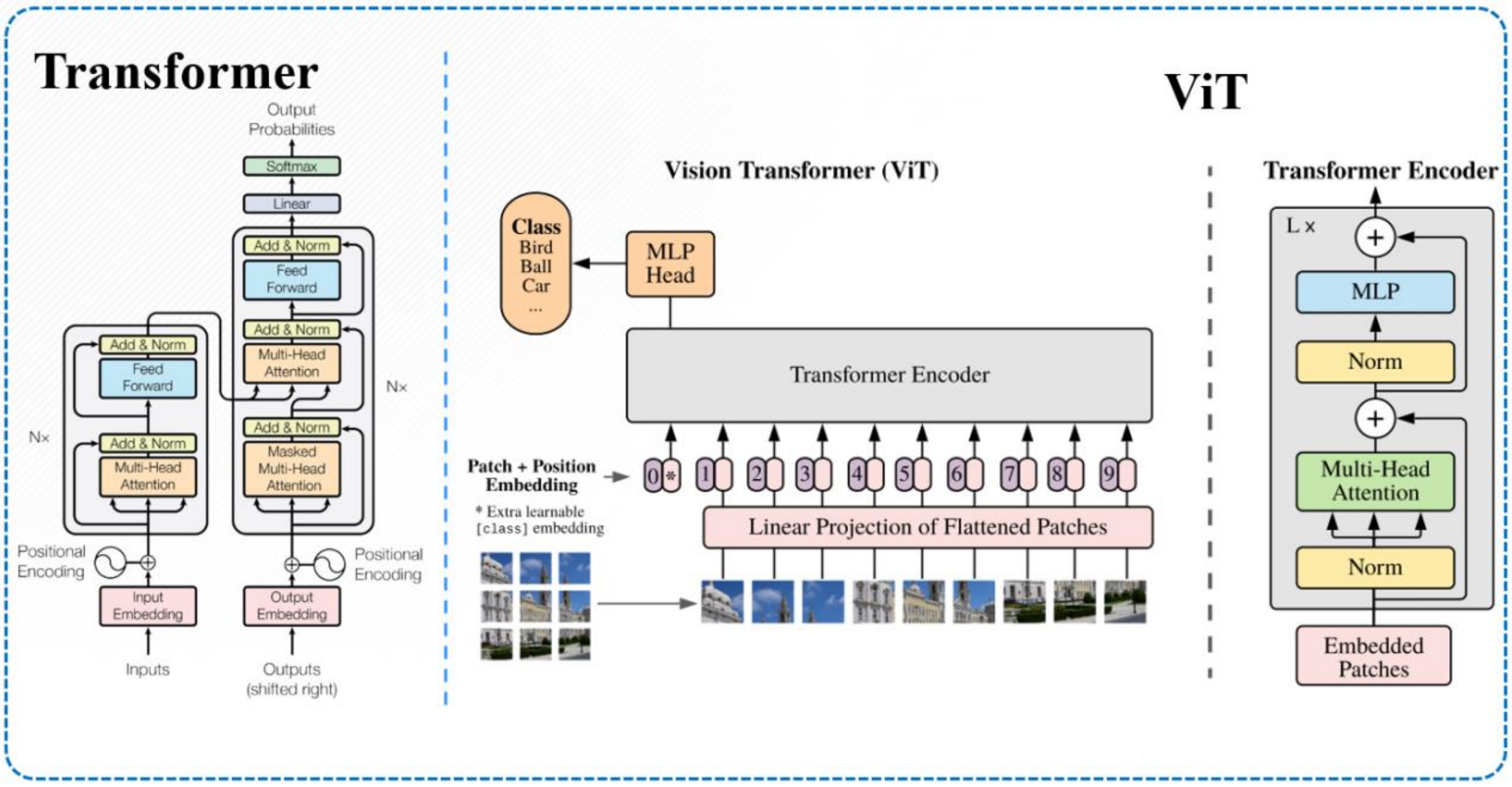

Fig. 2-12 Transformer model

The space remote sensing model has brought a wide range of opportunities and challenges to human society. It provides more accurate and comprehensive information support for scientific research and decision-making. The application of these foundation models in remote sensing image processing not only provides more refined data analysis capabilities for earth observation and environmental monitoring, but also provides important decision support for decision-makers in the fields of natural disaster warning, agricultural development, urban planning, environmental protection, and other fields. Through the remote sensing model, we can better understand the complex changes and interrelationships in the earth's surface and atmosphere, optimize resource management and utilization, reduce energy consumption, improve environmental quality, and protect the ecology system. In addition, remote sensing models can help identify and solve global problems, such as climate change, natural disaster management and population migration. However, the research and development of foundation remote sensing models also faces great challenges. The training of foundation remote sensing models requires a foundation amount of high-quality ground observation data and annotation information, as well as powerful computational resources and algorithm support. The particularity and complexity of remote sensing data make foundation models still face difficulties in multimodal fusion of remote sensing images and accurate analysis of low-resolution data, when processing remote sensing images. To further promote the development and application of the space remote sensing model, it is necessary to continuously improve the algorithm and model structure. The quality and diversity of high remote sensing data can strengthen the understanding and adaptation of remote sensing tasks and improve the stability and reliability of the system. Only by constantly promoting the innovation and development of remote sensing models, can it truly realize its effective application in human society.

## 2.4 Geographic Foundation Model

### 2.4.1 Basic Concepts of the Geographic Foundation Model

A "Geographic Foundation Model" is a model that comprehensively utilizes geographic information and artificial intelligence technologies. By leveraging the advantages of artificial intelligence in large-scale processing and continuous learning, the geographic foundation model better understands and processes various types of geographic data, so it provides more effective and comprehensive solutions for various tasks in the geographic information field (Janowicz et al., 2020).

The composition of the geographic foundation model is shown in Figure 2-13, with the main components including:

(1) Geographic Data Generation

Geographic data generation involves acquiring and integrating various geospatial elements, including POIs (points of interest), trajectory points, imagery, and related datasets. POIs represent specific locations of interest, such as landmark buildings, businesses, shopping centers, or geographic features, while trajectory points represent paths or movements recorded over time. Additionally, imagery includes satellite remote sensing images, aerial photographs, street view images, and other visual representations, providing important contextual information for understanding geographic phenomena. In the geographic foundation model, these components serve as the foundation for geographic analysis and modeling, revealing spatial patterns, trends, and relationships. They provide fundamental information for subsequent spatial reasoning and geographic Q&A.

(2) Geographic Knowledge Database

The geographic knowledge database mainly comprises a geographic text corpus and a spatial database. The geographic text corpus includes a vast amount of geography-related textual materials, such as papers in the field of geography, textual descriptions on maps, place name explanations, etc. Text is one of the primary bases for foundation model learning. The rich textual resources provide strong support for subsequent model-based geographic Q&A and recommendations. The spatial database stores spatial information related to geographic entities and their interrelationships, including the geometric shapes of geographic features, topological relationships, attribute data, etc. it provides fundamental data support for the model to perform geographic analysis and spatial reasoning. The storage and management of this geographic knowledge provide the necessary information foundation for the establishment and application of the geographic foundation model.

(3) Spatial Reasoning

Spatial reasoning is the core component of the geographic foundation model, which is primarily responsible for identifying different types of geographic data and deeply understanding the spatial relationships between geographic data. Through spatial reasoning, the model can accurately analyze and interpret geographic data. It deeply explores the associations between geographic phenomena, providing important support for solving many practical problems, such as urban planning, natural resource management, and environmental protection. Spatial reasoning also plays a significant role in optimizing resource allocation, planning urban development, designing transportation networks. By analyzing geographic data and spatial relationships, it can provide optimization solutions and strategies, maximizing resource utilization efficiency.

(4) Geography Q&A and Recommendations

Geography Q&A and recommendations are key parts of the geographic foundation model, primarily answering various geography-related spatial questions posed by users based on knowledge and spatial reasoning capabilities, such as route recommendations, travel advice, etc. Through this interactive function, users can obtain practical information and suggestions regarding geographic locations, landmarks, traffic routes, travel destinations, etc. Furthermore, geography Q&A and recommendations can learn users' needs and preferences from interactions, providing personalized and customized suggestions, such as helping users better plan trips, explore unknown areas, and solve practical travel problems. This interactive geographic information service provides users with a convenient and efficient way to access geographic spatial information and related advice, representing a core feature of the geographic foundation model.

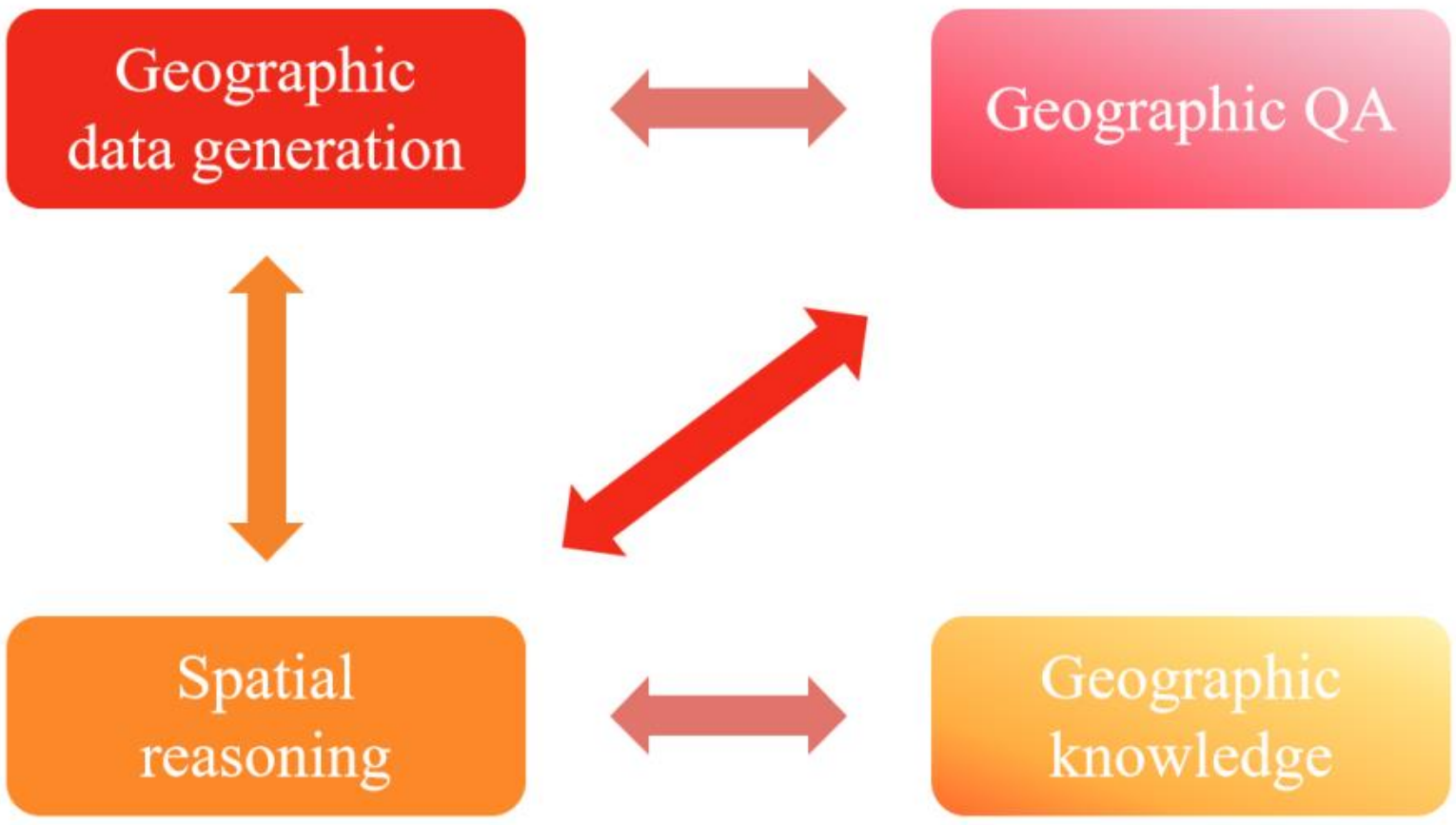

Fig. 2-13 Composition of geographic foundation models

### 2.4.2 Key Technologies of Geographic Foundation Models

In recent years, extremely foundation models trained on internet-scale datasets have achieved advanced performance in various learning tasks, leading to a paradigm shift in the training of modern machine learning (ML) models. Unlike models that learn specific tasks from scratch, pre-trained models, also known as foundation models (FMs), are adjusted through fine-tuning or few/zero-shot learning and then deployed in various domains (Brown et al., 2020). These foundation models enable cross-domain knowledge transfer and sharing while reducing the need for task-specific training data. Foundation models include large language models (LLMs), large visual foundation models, large multimodal foundation models, and large reinforcement learning foundation models. Despite the success of foundation models like ChatGPT, there has been relatively little exploration of foundation models in geospatial artificial intelligence (GeoAI).

The key technological challenge for geographic foundation models lies in the inherent multi-modal nature of geospatial AI. In a geographic foundation model, core data modalities include text, images (e.g., remote sensing or street view images), trajectory data, knowledge graphs, and geospatial vector data (e.g., map layers from OpenStreetMap). All these data contain significant geographic information, such as geometric and semantic details (Hu et al., 2023). Each modality has unique structures and requires specific representations, necessitating the effective integration of all these representations in the geographic foundation model (Hu et al., 2018). This requirement hinders the direct application of existing pre-trained foundation models to all GeoAI tasks. Given these diverse data modalities, the current goal is to develop a geographic foundation model that best integrates all these multimodal data.

Existing multimodal foundation models, such as CLIP (Contrastive Language-Image Pre-Training), generally have the following architecture:

(1) Separate embedding modules are used to encode different modalities of data, such as using a Transformer to process text.

(2) Mixing different modality representations through concatenation (optional).

(3) Further Transformer layers for reasoning across different modalities, enabling some level of semantic alignment, such as associating the text "school" with images of schools (optional).

(4) Generating prediction modules to achieve self-supervised training across different modalities.

However, these architectures lack integration with vector data, which is fundamental for spatial reasoning and key for aligning multimodal data in geographic foundation models. Therefore, vector data can enhance positional encoding for aligning different modalities. For instance, geotagged textual data and remote sensing (street view) images can be easily aligned through their geographic footprints (vector data). This model technology's advantage lies in enabling cross-modal spatial reasoning and knowledge transfer.

In addition to key technological breakthroughs for multimodal characteristics, geographic foundation models need to consider the following critical technologies (Mai et al., 2022):

(1) Geographic Debiasing Framework: Foundation models may amplify social inequalities and biases present in the data. For example, many geographic parsers are heavily biased towards data-rich regions. A key issue for geographic foundation models is geographic bias, which is often overlooked in foundation model research. Foundation models, compared to task-specific models, are more susceptible to geographic bias because:

①The geographic data used for training is often collected on a large scale and may be dominated by overrepresented regions.

②The large number of learnable parameters and complex model structures make it difficult to interpret and debias geographic foundation models.

③Geographic bias in foundation models can easily be inherited by all downstream adapted models.

These factors highlight the importance of an appropriate geographic debiasing framework in geographic foundation models.

(2) Spatial Scale Transformation: Geographic information can be represented at different spatial scales, meaning that the same geographic phenomenon or object can have entirely different spatial representations (point and polygon). For example, an urban traffic prediction model might represent Beijing as a complex polygon with various information, whereas a geographic parser might represent Beijing as a single point. Given that different downstream tasks require models to handle geospatial information at different spatial scales and quickly and accurately infer the correct spatial scale for the task, a module for spatial scale transformation is a crucial component in geographic foundation models.

(3) Generalization and Spatial Heterogeneity: Another key issue for geographic foundation models is how to achieve model generalization, also known as replicability, across different spatial regions while still allowing the model to capture spatial heterogeneity. Given geospatial data at different spatial scales, this requires models to learn general spatial trends while retaining specific location details. However, this key technology has not been effectively resolved with issues, such as whether such generalization could introduce inherent model biases in downstream tasks. Additionally, with the increase of large-scale training data, there are more considerations needed.

As current mainstream geospatial intelligence algorithms are primarily data-driven, training data, or sample data directly affects the accuracy and usability of the trained AI/ML models. High-quality sample data requires complete metadata, traceability, and quality evaluation information, making the training, validation, and testing processes of AI/ML models more accurate. Based on the characteristics and needs of geospatial AI sample data, five core aspects are considered: labeling, traceability, quality, updating, and consistency. Figure 2-14 summarizes the basic conceptual entities necessary for describing sample data (Le et al., 2023).

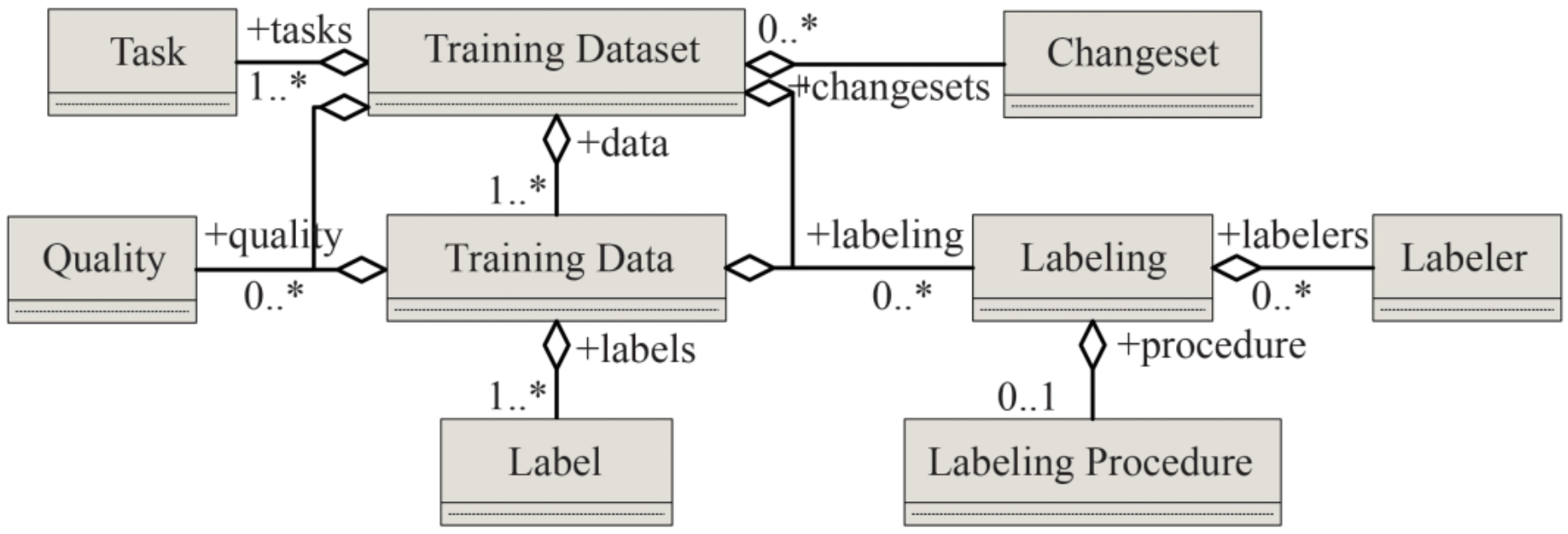

Fig. 2-14 Geographic artificial intelligence sample conceptual model

The training dataset is a collection of multiple sample data units, serving as the unified input for AI/ML models. The training data instance, as a core component of the sample information model, represents a single sample entity within the training dataset. It includes the basic attributes and data content of a single training, validation, or testing sample, providing the necessary input for AI/ML models. The sample label indicates the classification or categorization of each sample, ensuring accuracy during the training process and improving model precision. The sample task describes the goals and tasks involved in the entire training dataset. Sample quality pertains to the quality information of the entire training dataset and individual sample data units, helping data users assess the usability and reliability of the sample dataset. The sample labeling activity describes the information of an artificial labeling activity for producing the sample dataset. The sample labeler describes information about an individual participating in the labeling activity for producing the sample dataset. The sample changeset describes all updates to the sample data between two versions of the dataset.

This geospatial AI sample information model considers labeling, traceability, quality, updating, and consistency, facilitating standardized expression of heterogeneous geospatial intelligence sample data from multiple sources. It provides an informational model foundation for organizing sample databases and supports the sharing of geospatial AI sample data in a networked environment.

Geospatial AI samples can be prepared in advance according to the purposes of AI/ML tasks and organized based on standardized sample information models, directly meeting model input requirements. The complete workflow of the AI-ready process, shown in Figure 2-15, includes six steps: production, mapping, organization, sharing, integration, and training. This process enables a service model that couples sample data with models.

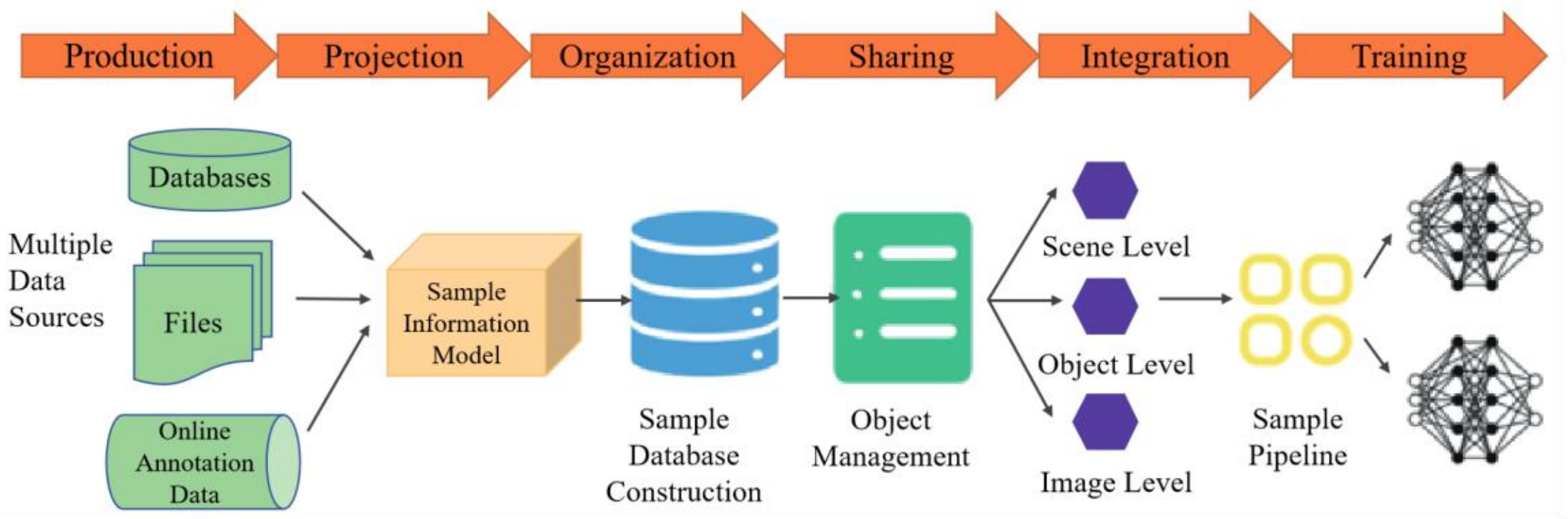

Fig. 2-15 AI-ready workflow

Different geospatial intelligence applications have diverse sample data contents and organizational forms. Building a unified data sample information model is the premise for sharing and interacting with geospatial intelligence sample data and the foundation for constructing geographic foundation models. Addressing core issues such as multimodal characteristics, geographic bias, and spatial scale is fundamental to the design of geographic foundation models, determining the types and quality of geographic information and spatial relationships the model can capture, as well as its interactive performance quality. Currently, research on key technologies for geographic foundation models is thriving. For instance, high-intelligence spatial computing teams are using advanced computational methods and AI technologies to analyze and process spatiotemporal big data for intelligent decision-making and optimization in spatial environments. Models like ReCovNet and SpoNet, based on deep reinforcement learning, solve urban-oriented spatial optimization problems, support decision-making in spatial optimization aspects of geographic foundation models and advancing related technologies.

### 2.5 Intelligent Transportation Foundation Model

The report of the 20th National Congress of the Communist Party of China emphasizes moving faster to boost China's strength in transportation, cyberspace, and digital development. China's intelligent transportation development has entered a period of rapid growth. With the deep integration of emerging technologies such as foundation models, big data, and cloud computing in the transportation sector, the training and learning from massive urban behavioral data enable more precise and efficient processing of spatial-temporal traffic data. This provides new driving intelligence for applications such as autonomous driving, traffic analysis and management, traffic behavior perception, traffic accident handling, and intelligent traffic decision-making (Fig. 2-17).

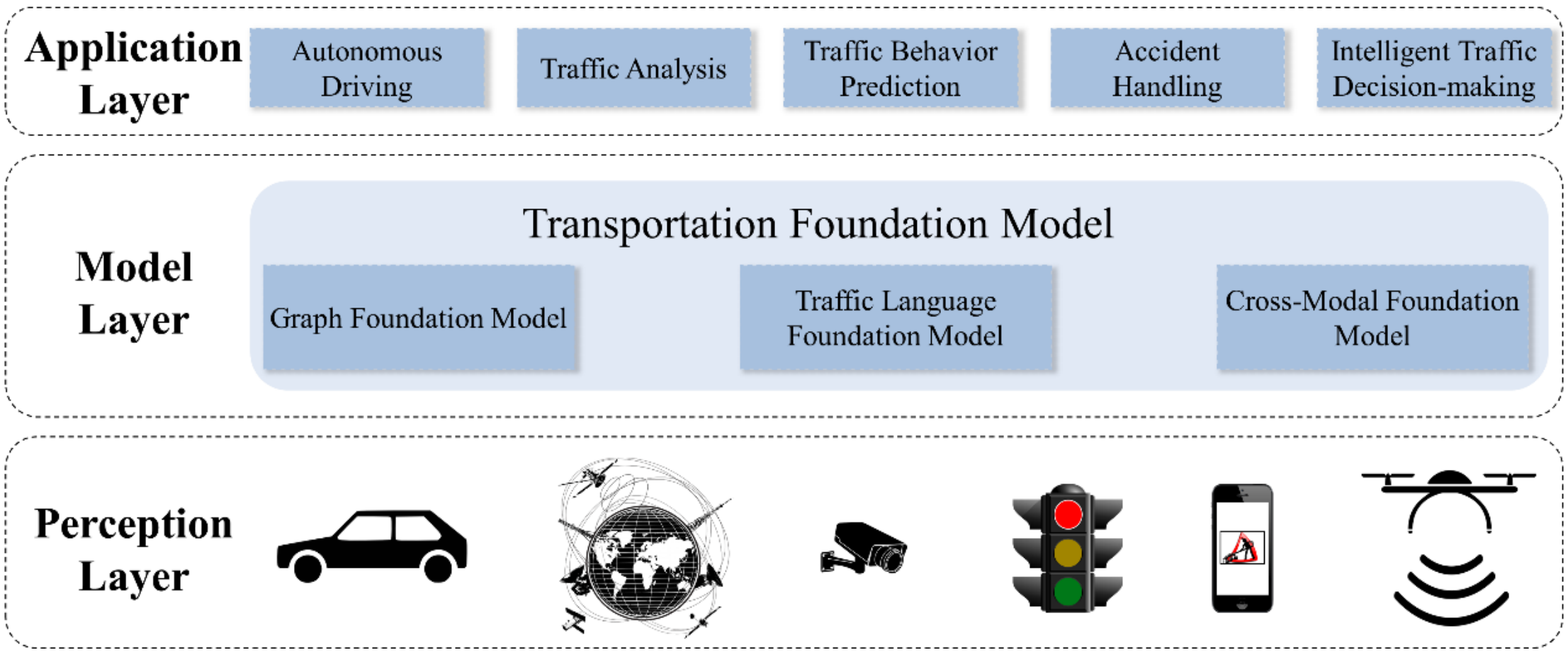

Fig. 2-17 Intelligent transportation foundation model solution

Urban transportation is an exceedingly complex network system influenced by spatiotemporal characteristics, dynamic human mobility, and various environmental factors (Du et al., 2021). It exemplifies a system process that encompasses perception, cognition, and prediction. Consequently, AI models focused on learning and training in specific aspects often find limited and fragmented applications in transportation. The emergence of foundation models offers the potential for achieving comprehensive traffic management.

2.5.1 Graph Foundation Models

Graph foundation model technology utilizes the vast amounts of spatiotemporal image data generated by sensors distributed throughout urban road networks to accomplish traffic management and traffic behavior prediction applications. Initially, these models collect spatiotemporal data from the road network and perform data coordination and fusion processing. This data not only directly reflects traffic conditions but also provides vehicle location information, trajectory data, and traffic flow information. Spatial analysis modules are then used to extract spatial topological relationships and complete data pre-training. The training framework encompasses multi-task learning and transfer learning, enabling the trained model to perform predictive functions. Key technologies of graph foundation models include graph neural network integration, multimodal data fusion, spatiotemporal sequence deep learning models, dynamic traffic condition adaptation algorithms, automatic traffic flow data feature extraction, long-term trend analysis, and multi-task learning.

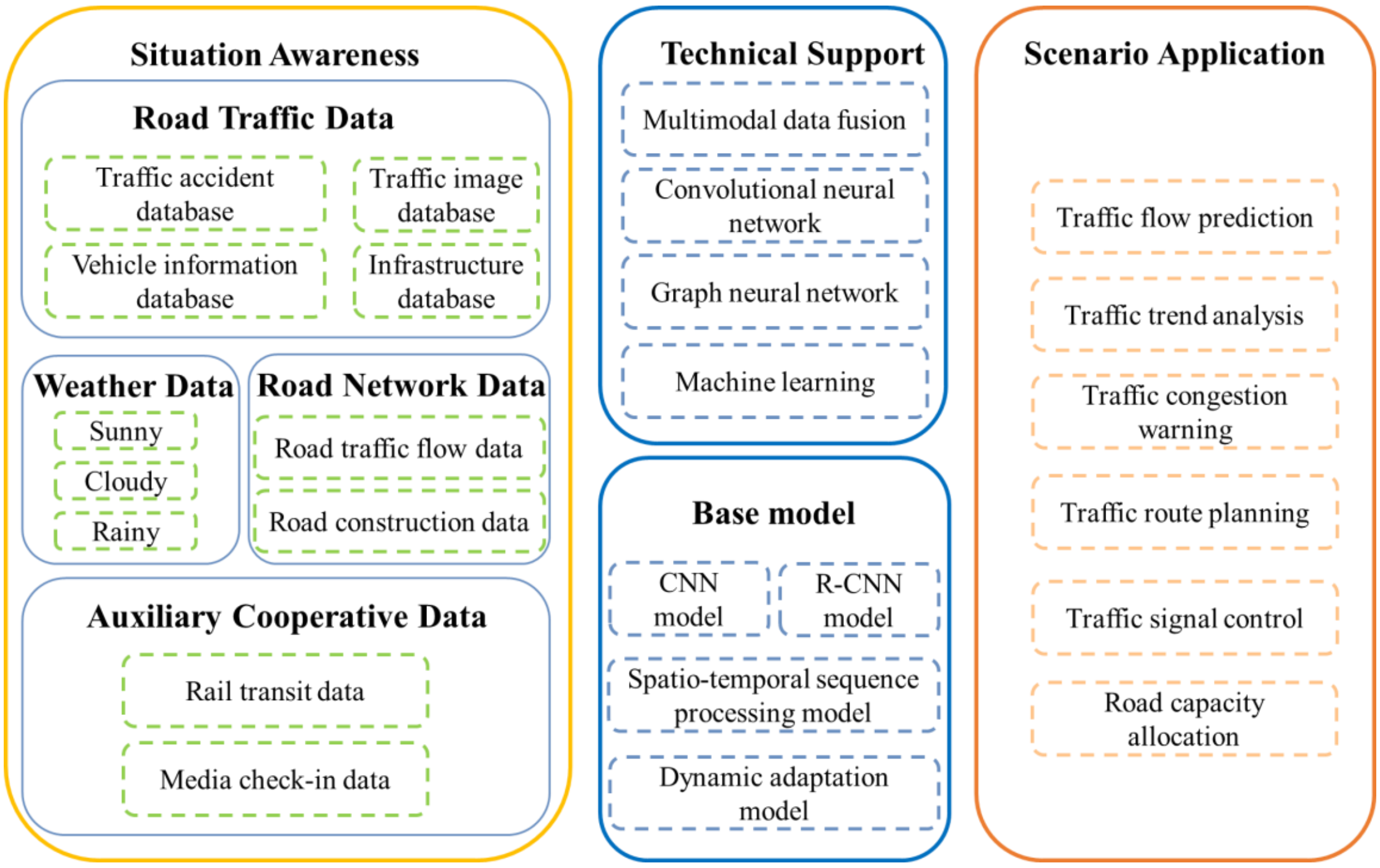

Fig. 2-18 Application of foundation model in the field of transportation

Graph neural network integration: The challenge of traffic management lies in the complexity of road networks. graph convolutional recurrent neural networks (GCRNN) are an effective method for addressing traffic prediction issues in smart cities (Liang et al., 2023). By integrating graph neural networks into traffic foundation models, it becomes possible to accurately capture the complex spatial dependencies of road networks accurately, thereby improving traffic predictions' accuracy. The application of graph neural network integration structures plays a crucial role in learning the intricate spatial structure of traffic systems, ensuring the reliability of decision support in traffic management and planning.

Multimodal data fusion: The data composition in the field of transportation is extremely complex, characterized by multi-source and heterogeneous data. Traffic foundation models need to process various types of data, including video surveillance images, global navigation satellite system (GNSS) tracking data, social media data, and traffic flow images. By integrating big data analysis, advanced machine learning techniques, and reliable traffic knowledge, traffic foundation models can fuse different modal data to obtain more comprehensive traffic flow information and establish lane-level road networks. Foundation models capable of handling multimodal data can provide high-precision decision support for urban traffic planning, network design, transportation infrastructure construction, and traffic management across multiple spatial and temporal scales, enhancing digitalization, informatization, and intelligence in traffic systems.

Spatiotemporal sequence deep learning model: Spatiotemporal sequence data is an important component of data in the transportation field. Analyzing multi-temporal spatial data allows for trajectory inference and traffic flow analysis, providing more comprehensive services for intelligent travel. To better process spatiotemporal data, foundation models integrate spatiotemporal graph convolutional networks (ST-GCN) and recurrent neural networks (RNN). This integration enables simultaneous learning of temporal and spatial features within traffic networks, combining location and spatiotemporal information to predict mobility patterns (Yao et al., 2023). The use of such models significantly improves the accuracy and efficiency of traffic network predictions by precisely capturing the relationships between time and space.

Dynamic traffic condition adaptation algorithm: One of the critical characteristics of transportation networks is the continuous change in traffic conditions. The dynamic traffic condition adaptation algorithm enables foundation models to adjust in real-time. By modifying model parameters based on real-time information from the road network, the model becomes dynamic, adapting its predictive strategies according to current traffic conditions and forecasted changes. This flexibility makes the foundation model an intelligent system, which is capable of better handling the ever-changing traffic environment through its robust data processing and integrated learning capabilities, ultimately providing more accurate predictive services.

Automatic extraction of traffic flow features: Traffic flow characteristics are crucial for effective traffic management. In foundation models, deep learning algorithms are employed to automatically extract key features from vast amounts of traffic data, eliminating the need for manual feature engineering. This method of automatic feature extraction allows for a more comprehensive capture of traffic flow characteristics, enabling the model to understand the traffic flow from the perspective of the overall network structure. As a result, the model can generate solutions for multimodal transportation tasks, enhancing the efficiency and accuracy of traffic management and planning.

Integration of real-time prediction and long-term trend analysis: In traffic management, it is essential to respond to current situations promptly while also predicting future trends. Foundation models integrate real-time data analysis modules and historical data trend learning modules to provide dual support for traffic management and planning. By accumulating real-time prediction data over extended periods and repeatedly learning from it, the model directly transfers this information to the trend learning module. This process enables the foundation model to perform long-term and scenario-specific trend predictions, thereby enhancing its ability to offer accurate and timely decision support for traffic management and planning.

Graph foundation model technology, when integrated with the transportation sector, enables real-time traffic flow prediction, road anomaly detection (Yu et al., 2023), traffic congestion warning, and route recommendations. This information is crucial for traffic managers as it aids in better planning of traffic routes, optimizing traffic signal control, adjusting road capacity distribution, and ultimately improving the efficiency and fluidity of urban traffic. Additionally, it helps reduce congestion and the occurrence of traffic accidents.

### 2.5.2 Traffic Language Foundation Model

The traffic language foundation model is built upon deep learning models and natural language processing (NLP) technologies, learning from vast amounts of data to develop a generalized ability to understand and generate natural language. These general-purpose large language models (LLM) can perform tasks across multiple domains without requiring specialized training. The advent of LLM has significantly enhanced the capability of foundational traffic models in text processing and analysis. By integrating LLM into traffic domain models, it is possible to undertake more complex traffic-related tasks. These tasks include the automatic generation of traffic accident reports, traffic condition summaries, accident scene detection, and analysis, understanding of accident scenes, and the creation of intelligent traffic assistants. This integration allows traffic models to provide more detailed and personalized services to ordinary users, effectively becoming a smart personal traffic assistant for them.

The traffic language foundation model is an integrated model based on multiple sub-domain models and multi-source foundational traffic information data. It interprets user semantic information with user input as text data continuously used for model training. Within the LLM, semantic tasks are autonomously planned and evaluated, invoking foundational traffic models to solve problems. This involves analyzing traffic problem patterns and task types, completing task outputs, and feeding them back to the user (Fig. 2-19). Key technologies in the traffic language foundation model include language-visual cross-modal encoding, integration of language understanding and traffic prediction, transfer learning and domain adaptation, and model fusion and ensemble learning.

Language-visual cross-modal encoding: The transportation domain generates vast amounts of text and visual information. Text information is input into LLM (such as BERT or GPT), while visual information is processed through image models. The resulting text features and visual features are integrated and encoded into a unified feature vector to represent the semantic information of traffic scenarios. These fused multimodal feature representations can be used as input to train deep learning models, such as neural network models for understanding traffic events and predicting traffic flow. Additionally, during actual predictions, new text descriptions and visual information of traffic scenarios can be input into the trained model to obtain predictions of traffic conditions.

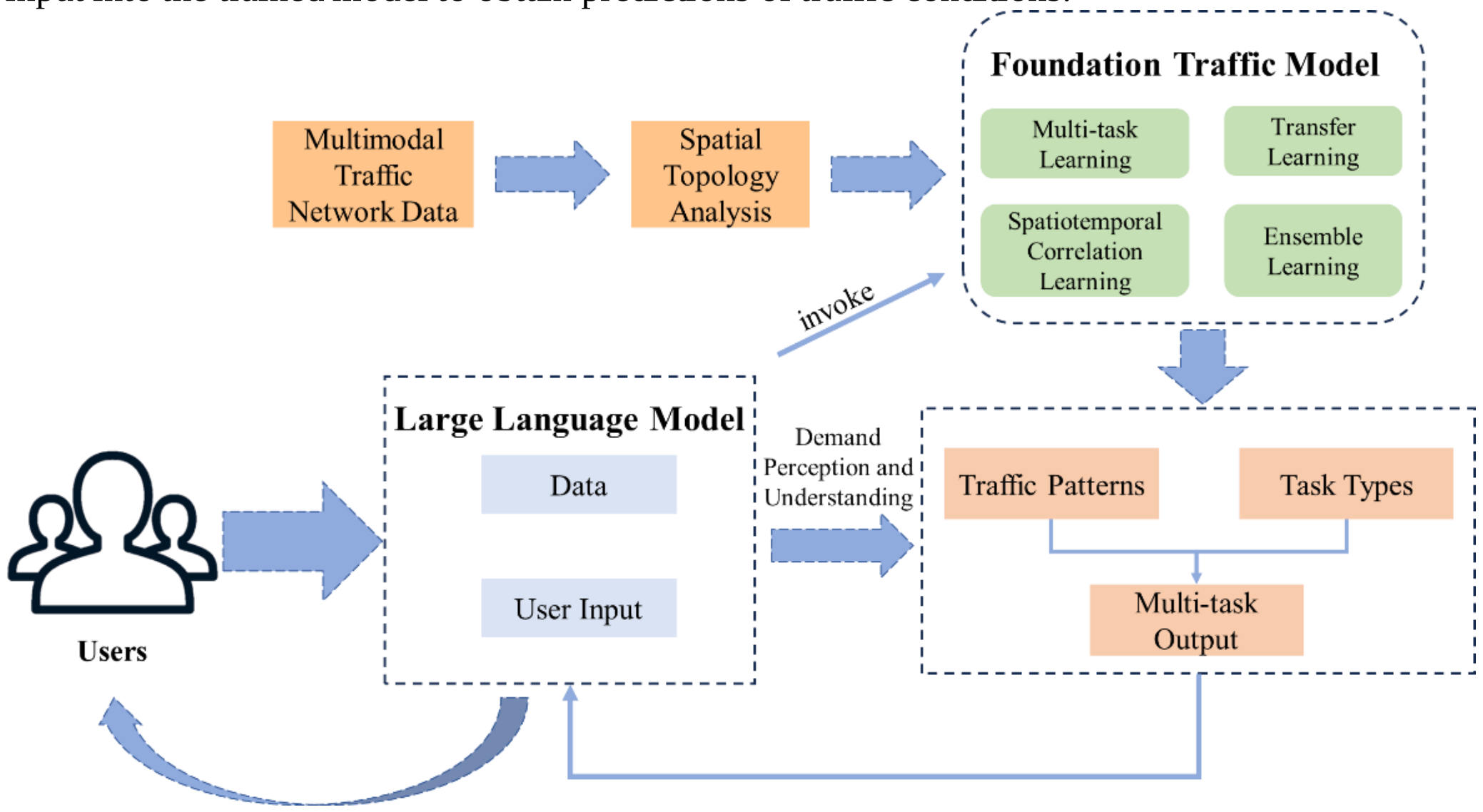

Fig. 2-19 The interaction process of traffic language foundation models

Integration of language understanding and traffic prediction: LLM is used to understand traffic-related natural language texts, such as traffic news reports and social media comments. These models extract key information, event descriptions, and sentiment trends from the text. The traffic events and their development trends described in the text are used as additional features, integrated with other traffic data for comprehensive training. This approach yields a more thorough understanding of traffic events and trends. The results of this understanding can be combined with traffic prediction models to provide more comprehensive decision support for traffic management and planning.

Transfer learning and domain adaptation: Leveraging the pre-training capabilities of LLM in the NLP domain, transfer learning is applied to adapt these models to traffic-related tasks. This involves fine-tuning the language model on traffic-specific datasets to align with the unique characteristics and requirements of traffic data. During fine-tuning, special attention is given to the specific language expressions and terminology used in the traffic domain, enabling the model to better understand traffic-related semantics and context. Additionally, the training strategy of the model is adjusted to match the distribution of traffic data and the objectives of traffic-related tasks.

The traffic language foundation model can address numerical processing and interaction simulation challenges inherent in LLM (Zhang et al., 2024), significantly enhancing data analysis efficiency and more comprehensively unlocking the potential of various stakeholders in the traffic domain. Through the traffic foundation model, managers and operators can access objective and accurate data and report more directly, enabling them to provide objective analyses from a holistic perspective. Basic users, on the other hand, can obtain more real-time and comprehensive traffic prediction information through the traffic foundation model. The advent of the traffic foundation model will not only change the way traffic systems operate but also profoundly impact travel experiences and urban development patterns.

### 2.5.3 Cross-Modal Intelligent Traffic Foundation Model

The cross-modal intelligent traffic foundation model is a comprehensive system built upon multi-source data and various deep learning models. It can integrate multiple data sources, including sensor data (radar data, camera data), video surveillance data, GNSS positioning data, and social media data, performing data fusion in a multimodal manner to convert these diverse data into unified feature vectors. Leveraging intelligent decision support capabilities, the model can predict traffic flow, identify traffic congestion, optimize traffic signal timing, and plan traffic routes, thereby achieving intelligent management and optimization of traffic systems. Additionally, the model dynamically adjusts its parameters and learning strategies based on the constantly changing traffic environment and needs, enabling real-time responses and decisions. This adaptability helps in managing traffic accidents and unexpected events, enhancing the emergency response capability of the traffic system. The traffic foundation model provides traffic managers, planners, and ordinary users with intuitive insights into traffic conditions and trends, offering personalized traffic advice and services. Key technologies include the integration of autonomous driving technologies, reinforcement learning and adaptive decision-making, edge computing and internet of things (IoT) technology applications, and blockchain technology applications.

Integration of autonomous driving technology: As autonomous driving technology advances, the traffic foundation model will incorporate data and technologies related to autonomous driving. This includes data collected from various sources such as autonomous vehicles, traffic signals, and intelligent transportation facilities. The integrated data is fed into the intelligent traffic foundation model to facilitate applications like autonomous driving decision-making, route planning, and emergency incident avoidance. By deeply integrating with autonomous driving technology, the traffic foundation model can more accurately understand and predict traffic behavior, supporting intelligent traffic management and vehicle control.

Reinforcement learning and adaptive decision-making: By leveraging reinforcement learning technology, the traffic foundation model becomes more intelligent, enabling it to autonomously learn and optimize decision-making strategies through interaction with the environment. This adaptive decision-making capability allows the traffic foundation model to better adjust to various traffic scenarios and changing road conditions, thus achieving smarter and more flexible traffic management and services.

Application of edge computing and IoT technologies: With the widespread adoption and development of edge computing and IoT technologies, the traffic foundation model increasingly utilizes data collected by edge devices and sensors. These data include real-time traffic information gathered from vehicle sensors, traffic cameras, and smart traffic signals. By collaborating with cloud-based models, the system achieves faster and more real-time traffic analysis and predictions.

Application of blockchain technology: The decentralized, secure, and transparent nature of blockchain technology holds significant potential for the transportation sector. The traffic foundation model leverages blockchain technology to ensure the security and credibility of traffic data, as well as to facilitate the sharing and exchange of traffic information. By applying blockchain technology, a more reliable and secure traffic data platform can be established, supporting the intelligent and optimized management and service of transportation systems.

The cross-modal intelligent traffic foundation model seamlessly integrates specialized models from multiple domains with the expertise of foundational traffic models. This approach not only aids the advancement of traffic management but also offers a novel perspective on harnessing AI capabilities in this field. The adaptability and flexibility of the intelligent traffic foundation model allow for the incorporation of foundational traffic models according to specific business needs. Additionally, the model can autonomously select and execute foundational traffic models based on task requirements, making it a crucial application for solving complex problems in transportation and urban planning.

## 2.6 New Perspectives on Intelligent Foundation Model for Spatial Data

With the advent of the Big Data Era, spatial data intelligent models, as a core research direction in the field of geographic information science, are gradually demonstrating their significant application potential. Spatial data not only encompass basic information about the location and shape of geographic entities but also contain rich spatial relationships and semantic information, providing strong support for an in-depth understanding of physical world and social spatial phenomena. Maps, serving as a critical representation of the real world, furnish extensive spatial information while offering fundamental perspectives for comprehending physical and social spatial phenomena. Deep reinforcement learning, as an emerging machine learning method, is also showing great potential in spatial optimization problems.

This section will explore new technical points of spatial data intelligent models based on foundation model in urban and remote sensing domains, focusing on the application of maps as modal data in geographic information processing, the exploration of deep reinforcement learning in spatial optimization problems, and the practical application of knowledge graphs coupled with artificial intelligence. By deeply analyzing the characteristics of map data and their integration methods in foundation language model, we discuss how to effectively process geographic text information and adapt it to foundation model frameworks. We will elaborate on the modeling methods and application examples of deep reinforcement learning in spatial optimization problems and explain the mechanisms of combining new-era artificial intelligence with knowledge graphs, thereby making positive contributions to the development and application of geographic information science.

2.6.1 Map as A Modal Data

Maps play a crucial role in understanding complex phenomena in the physical world and social spaces. They serve as a visual representation for observing, interpreting, and comprehending the real world. In practical applications, maps can be created using multi-source data to display specific information, such as point1 of interest (POI), taxi trajectory data, and normalized difference vegetation index (NDVI) data. The next generation of GIS will integrate traditional spatial data processing, spatial analysis techniques, and spatial reasoning capabilities to realize the GPT of geospatial data. In foundation language model architectures, maps can be regarded as a type of modal data similar to audio, images, and text, which can be effectively integrated. Research in deep learning and natural language processing has demonstrated the interrelation between maps and other forms of modal data such as text and images. For example, remote sensing images can be described and analyzed by natural language processing techniques, while the extraction and organization of geographic knowledge graph can rely on the processing of text data. However, previous research on geographic understanding has been limited to latitudes and longitudes and their relationships, while geographic information encompasses much more. There are various complex relationships between geographic entities, such as containment and intersection. Therefore, the current application of geographic information in geographic text processing remains underutilized.

Geographic text, due to its rich expressions and multimodal properties linked with maps, has always been a challenge for automated processing. Thus, how to handle map information and adapt it to foundation model frameworks is a key focus of current research. Current studies have rarely addressed the processing of geographic text information and have primarily concentrated on the geographic text itself. The existing applications of geographic information in geographic text processing mainly include PALM model and GeoBERT model proposed by Didi, STDGAT model and Ernie-GeoL model proposed by Baidu. PALM and STDGAT belong to the model before BERT (Bidirectional Encoder Representations from Transformers). The PALM model discretized the latitude and longitude through CNN, thus enabling the model to learn the distance relationship between the query and the POI. STDGAT builds on PALM by adding the user's time series behavior. Erney-geol in the BERT era focused on integrating various user behaviors on the map,such as hailing, tapping to adopt, multiple typing, into the pre-training task, which added the prediction of latitude and longitude (using GeoHash to express latitude and longitude) to the pre-training task. GeoBERT uses the entity text in the geographic database for graph modeling and graph learning according to the relationship between distance and administrative inclusion, fusing the learned entity text vector with the input geographic text.

Based on the map-text multimodal architecture, we can utilize multi-task pre-training techniques, combining methods such as adversarial attention pre-training, sentence pair pre-training, and multimodal pre-training to train a foundational pre-training model suitable for various geographic text tasks. This enhances performance on a wide range of downstream geographic text processing tasks. In this process, maps as data sources need to be symbolized. Input text data should be tokenized and transformed into vector representations through embedding layers. Depending on the map type (raster or vector), an appropriate data structure should be chosen. Through representation learning, the model can learn joint representations of map and text data and integrate geographic knowledge to improve the understanding of geospatial data. The learned representations can be applied to downstream AI tasks such as route planning or location recommendation. Additionally, corresponding processing rules must be formulated. The relationships between different data representations need to be analyzed to optimize the model’s performance. When handling high-dimensional data, sparse representations can be employed to reduce computational complexity. In this way, map data can be effectively integrated into foundation language model, enhancing their performance and accuracy in processing geospatial information.

2.6.2 Exploration of Deep Reinforcement Learning for Spatial Optimization Applications

Markov Decision Processes (MDP) and deep reinforcement learning (DRL) exhibit strong adaptability in addressing spatial optimization problems. MDP is a state-based decision-making process, where decisions are only dependent on the current state and not on previous states. This property makes MDP highly compatible with DRL. By modeling spatial optimization problems as MDP, we can leverage DRL to solve these problems. DRL combines the strengths of reinforcement learning and deep learning; deep learning is used to perceive the environment and provide current state information, while reinforcement learning is used for decision-making and evaluating the value of actions based on expected returns. Through interaction with the environment and feedback from reward signals, DRL can autonomously learn and optimize the behavior of agents in complex environments. In DRL, decision rules are established through self-learning; performance is continuously improved via trial and error. The core idea of DRL is to use reinforcement learning for decision-making by interacting with the environment to obtain reward signals and adjusting decision strategies based on these signals. This enables the agent to autonomously learn and optimize its behavior in complex environments.

In spatial optimization, DRL can be used to solve problems such as path planning, resource allocation, and layout design. By modeling spatial problems as MDP, in which states represent certain conditions in space, actions represent movements or operations in space, and rewards represent evaluation metrics of optimization goals. DRL algorithms can select appropriate actions based on the current state and evaluate their value through feedback from the environment. By continuously trying different actions, the agent can gradually learn how to make optimal decisions in various states, thus achieving spatial optimization. When applying DRL to spatial optimization, the complexity of the problem and the computational cost must be considered. DRL requires many training samples and computational resources, especially in large-scale spatial problems. The design of optimization algorithms and the tuning of the training process are crucial.

DRL has powerful application potential in micro-scale pedestrian simulation and location selection problems, such as emergency evacuation, logistics distribution, and billboard placement. It can be used to validate the effectiveness of plans and discover optimized actions. Compared to traditional micro-scale pedestrian simulation methods, such as the Social Force model and Pathfinder software, DRL can provide more accurate simulation results, particularly excelling in detailed aspects.

### 2.6.3 Geographic Knowledge Graph Opportunities and Challenges in the AGI Era

In the evolution of artificial intelligence (AI) towards artificial general intelligence (AGI), AI technology has produced a "qualitative leap," yet knowledge graphs based on semantic networks remain indispensable in geosciences. Geographic knowledge, as the product of geographical thinking and reasoning about natural and human phenomena, plays a crucial role in addressing geography-related questions. Geographic knowledge is characterized by its multilevel, diversity, multidimensional, and multi-granular nature, encompassing specialized technical knowledge in technical methods, geographical common sense, foundational knowledge of geography as a discipline, and specialized application knowledge within geographic data. The geographic knowledge graph constructs a geographic knowledge base through semantic network-based representation methods, enabling humans and machines to understand, compute, and interact with geographic knowledge. It creates a human- and machine-understandable, computable, and interactive geographic knowledge system.

As a geo-linguistic system integrating human and machine intelligence, the geographic knowledge graph is founded on disciplinary knowledge. The mechanism of constructing the cognitive system of geographic knowledge involves the digitization and incorporation of human language and disciplinary knowledge generated by human brain intelligence into digital knowledge carriers, using machine language for information comprehension. Building intelligent systems with geographic knowledge as the core and bridge, geo-knowledge bases are collaboratively constructed, and knowledge systems are interactively mapped through human-machine collaboration. Coupling the geographic knowledge base with geospatial big data enables collaborative computing and inferential knowledge discovery, thus addressing geoscientific problems. To achieve efficient geographic knowledge management, a cloud-native architecture-based geo-knowledge base engine can be utilized with the core being a knowledge-centric GIS platform centered on GeoKE. This technological framework employs a unified cloud-native architecture system (OneSls) with the development environment using Java, SpringBoot and Maven project object management model. Data storage employs NebulaGraph (knowledge graph), PostgreSQL (relational data), PostGIS (spatial data), and Alibaba OSS (file data). Core modules include multimodal geographic knowledge storage, management, and query.

The future development of geographic knowledge graph faces multiple challenges. In the realm of interdisciplinary integration, constructing a geographic knowledge system oriented towards artificial intelligence requires the fusion of geographical thinking with AI thinking. This entails combining specialized knowledge in geography with AI technologies to advance the development and application of geographic knowledge graph. Geographic knowledge graph, foundation language model, and geographic models will become critical directions for future development, complementing each other, and jointly advancing GeoAI. This necessitates concurrent research and development in both geography and AI fields. Sustainable construction of geographic knowledge engineering requires adequate funding and long-term mechanism guarantees. This includes establishing open geographic knowledge-sharing platforms to foster cooperation and communication among different research institutions and individuals, promoting the co-construction and sharing of geographic knowledge graph. The practical application of geographic knowledge graph is crucial. Resources need to be opened, and sample projects developed to establish GeoAI's indispensable role in both academic and applied domains. This implies validating and refining geographic knowledge graphs in practical applications and gradually extending them to a broader range of fields and industries.

### 2.6.4 Complementarity of Foundation Models and Knowledge Graphs in GeoAI

In Geographic Artificial Intelligence (GeoAI), foundation model and knowledge graphs exhibit complementary characteristics. Foundation model express knowledge in a parametric form, whereas knowledge graphs represent knowledge in a structured format. Foundation models are adept at handling implicit, non-deterministic expressions, while knowledge graphs provide explicit, deterministic expressions. These two approaches are complementary in knowledge representation and modeling. In GeoAI, the explosive growth of data on various natural elements across the full spatial spectrum poses significant challenges for data acquisition and analysis capabilities. Traditional data mining methods have certain limitations in handling high-dimensional spatial data and have relatively limited exploration in applications. To overcome these limitations, GeoAI introduces vectors as the cornerstone of the artificial intelligence and foundation model era, thereby achieving intelligent geospatial processing. Geographic vectors have distinct spatial structural characteristics, significant spatial topological relationships, complex semantic connections, and specific representations of natural resource elements, characterized by multi-granularity and multi-spatiotemporal scales.

The research trends in GeoAI can be categorized into three areas: spatial management, spatial intelligence, and spatial decision-making. In spatial management, it is necessary to extend the theories and methods of two-dimensional GIS to three-dimensional space, to develop a set of three-dimensional spatial computing methods for full space, all elements, and all content. This aims to achieve the organization, management, and analytical expression of geographic spatial elements, patterns, processes, modes, and rules. In spatial intelligence, the focus is on researching interpretable knowledge graphs and geographic knowledge embedding methods to enhance the intelligence of geospatial data. In spatial decision-making, support is needed for major national strategies and infrastructure construction. These research directions can provide accurate space-time empowerment for various fields.

The complementarity of foundation model and knowledge graphs allows for the incorporation of geographic experience constraints into geographic knowledge graph and the embedding of the structure of knowledge graphs into foundation model. The GeoVector database forms the basis for achieving general geospatial intelligence, embedding Physics Informed Neural Networks (PINN) for constructing multi-spatiotemporal scale frameworks for geographic representation, analysis, prediction, and interpretation. GeoAI not only delves into big data research in the spatiotemporal domains of water, soil, air, life, mountains, forests, fields, lakes, and grass but also reshapes the theories and technologies of surveying and mapping geographic information, supporting the major development strategies of the nation.

# 3 Key Technologies of Spatial Data Intelligent Foundation models

## 3.1 Spatio-Temporal Big Data Storing and Processing Technologies

With the advance of information technology, the need for processing, analysing, and visualising spatio-temporal big data has increased dramatically. GIS is facing new challenges in the era of big data. To overcome the difficulties posed by big data, GIS must develop its technology to cope with big data. Some of the challenges faced by GIS include spatio-temporal big data analysis and processing, spatial big data clustering and distribution, big data indexing and management, and implementing big data computation and visualization in the system while maintaining high performance. Currently, popular big data platforms (e.g. Hadoop and Spark) do not have the capability of executing spatial analyses, spatial computation, or spatial data mining. To achieve distributed storage and management of large-scale spatial data, breakthrough and innovation in distributed spatial data, computation, real-time big data processing, and visualization, it is necessary for GIS to integrate common big data technologies. On the other hand, facing the increasing data volume and types, traditional relational databases are prone to bottlenecks such as low storage efficiency, weak parallel access capability, and difficulties in horizontal expansion. Thus, the development of new spatial data storage technology is imperative. Container technology (e.g. Docker) is conducive to rapid, large-scale deployment of GIS. The optimal synchronization and discovery mechanism in load balancing provide support for dynamic scaling and disaster recovery of GIS services.

If a GIS system attempts to use the data to perform a query or generate a map, the output data from Spark must be converted and transferred to the GIS platform, which is typically a very time consuming and storage intensive process. Additionally, traditional GIS systems only execute computational tasks in the job queue and cannot handle streaming data. Additionally, traditional GIS systems only execute computational tasks in the job queue and cannot handle streaming data. Traditional GIS software and standalone processing architectures are unable to analyze spatial and temporal big data in large quantities (e.g., more than a billion records). Furthermore, these integration processes require high-specification computer hardware and rewriting of most algorithms for big data in GIS. Therefore, the spatio-temporal big data storage and processing platform architecture for spatial data intelligence foundation model should include the performance of mass space virtual storage, distributed computing framework, cloud computing integration, streaming data processing (big data streaming high-performance processing is described in detail in section 3.3.3), 3D virtual reality, fast multi-terminal application, container technology, and continuous delivery.

### 3.1.1 Mass Space Virtual Storage

Data storage is a key issue in spatial data Intelligent foundation model. With the high diversity of data types and low value density of data generated, traditional file systems and databases are no longer able to maintain high performance while continuing to satisfy the demands of big data storage. Recently, technologies and solutions have emerged in the field of virtual storage, many of them have been widely used by Internet platforms for geospatial data, but there is also a need to evolve traditional file system and relational database storage solutions into distributed, virtualized, and software-defined storage systems, so that the storage scalability and processing capacity can handle the coming challenges.

Virtual storage systems can be classified into three categories: distributed file systems, distributed relational databases, and NoSQL/NewSQL storage systems. Distributed file systems are mainly used to overcome the limited storage space and high-cost problems of standalone systems. Running concurrent I/O through multiple replica copies can not only increase the computing bandwidth, but also enhance the system's load balancing, fault tolerance, and dynamic scalability. The system can be deployed in a cloud computing environment to support large file sizes, in-memory caching, space sharing, and REST web services. A popular database of this type is Hadoop; other similar systems include Ceph and IPFS. Distributed relational databases are mainly implemented by adding distributed clustering and distributed transaction processing features to traditional databases (implementation examples include PostgreSQL clustering, MySQL clustering, and CrateDB based on Docker technology). These systems can better support SQL and transaction processing due to the high compatibility with original databases. Data migration and system expansion become easier as the original management methods and software can still be applied. When the system needs to be deployed in a multi-node cluster environment, it is especially important that the system is open source and relatively low cost. NoSQL/NewSQL storage systems focus on reducing the number of ACID transactions, which results in a significant improvement in the performance of data processing.

Nowadays, many different virtual storage systems exist in a variety of environments and are used in different ways. How to fully utilize the advantages of each system while enabling the sharing and transfer of resources between systems? And how to provide a unified data access, read-write method, as well as the ability to store data on multiple platforms, allowing the data to become more valuable? Aiming at solving these problems, based on the seamless integration of multi-source spatial data in SDX+ with the interfaces in GDB-CLI, the model integrates the virtual spatio-temporal integrated service system DaaS (Data as a Service) to realize a unified REST service framework, which can easily connect to many types of data storage systems and work with existing connected database systems. By using a unified data interface, the system can connect with the Hadoop storage ecosystem, MongoDB storage system, PostgreSQL cluster, MySQL cluster, and other existing databases (Figure 3-1).

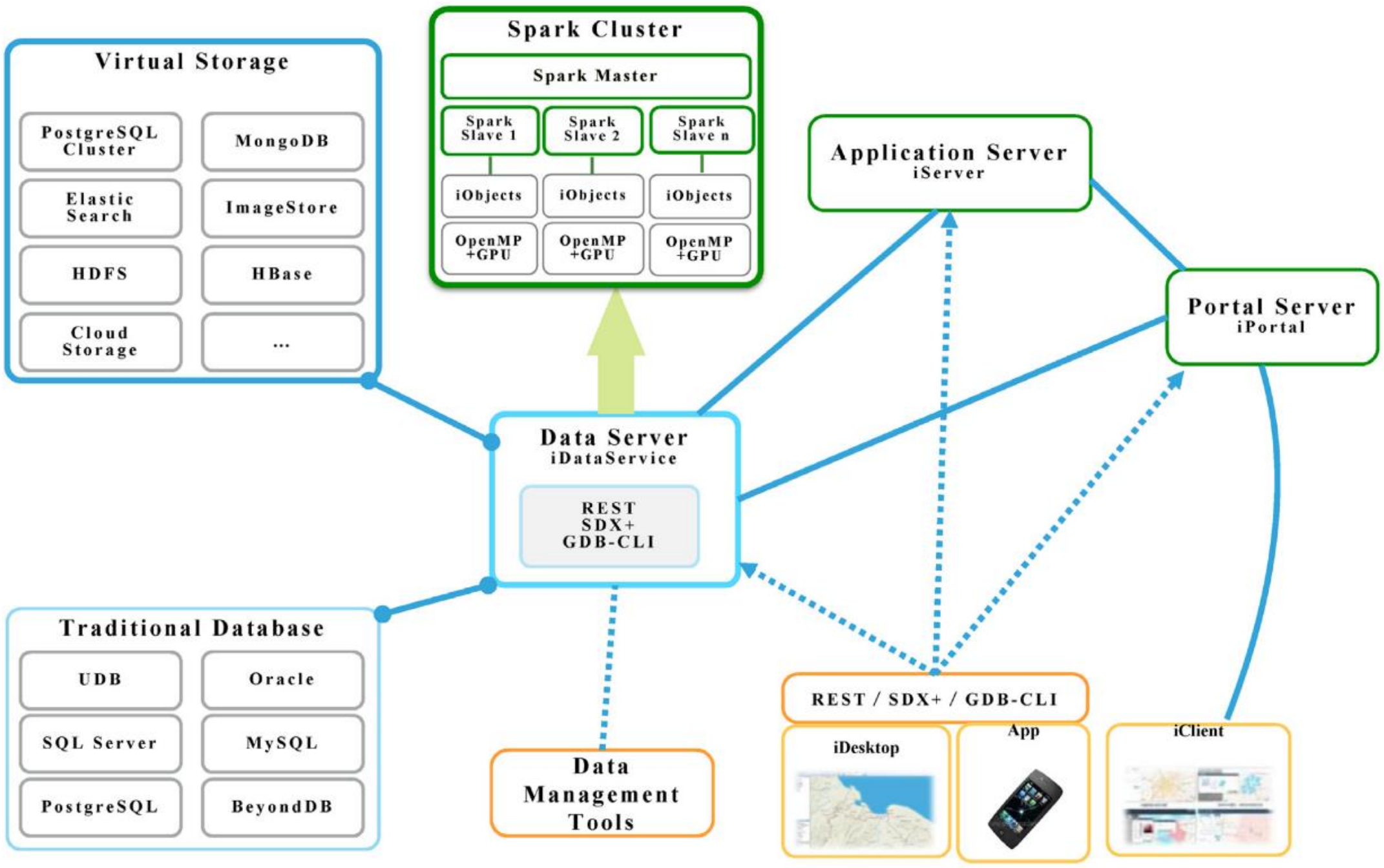

Fig. 3-1 From SDX+ to DaaS

The value of data decreases as storage space requirements increases and maintenance costs rise. If data can be consumed in a sensible period, it will become a more valuable asset. Conversely, if data is not used properly, it can become a burden to the organization. For example, an organization will be exposed to the risk of a sensitive data spillage if there is no adequate investment in data security, which could be detrimental to the company. Just having data does not benefit a business. In fact, the efficiency of data usage determines its value. Therefore, it is critical to establish a continuous data processing infrastructure to address the needs of applications. In addition, maintaining and applying the value of data is a key aspect of developing the foundation model.

### 3.1.2 Distributed Computing Framework

It is difficult to pursue further processor speeds by increasing the clock frequency of the CPU when Moore's Law comes to an end, while the multi-core CPU is becoming the new normal. By using multi-threading and process techniques to manage and parallelize tasks or by using the CUDA and OpenCL parallel computing mechanisms of graphics cards, systems can break through the limitations of computing power in a single CPU. In the model, CUDA's multi-threading support, multi-process services, and OpenMP-based spatial analysis algorithms significantly improve the efficiency of spatial data processing and model analysis, which enables real time running of object visualization functions.

The MapReduce module in Hadoop is specifically designed for batch processing and is considered as the forerunner of the next generation of distributed computing. However, it has many weaknesses such as slow startup, complex deployment, and incapable of executing regression calculations. Modules based on the distributed in-memory computing model and better support for streaming computation built on Flink have begun to be replaced by Spark. The Hadoop/Spark open-source ecosystem, led by the Apache Software Foundation, has become the standard in big data field, and many business solutions are built on this framework, including Databricks, Amazon, IBM, and Oracle's Big Data Service Cloud.

With the improvement of GPS systems, satellite images, UAV photography, and intelligent measurement devices, the requirements for spatial data storage and processing are rapidly increasing. Therefore, it is particularly important to import GIS functions into Spark framework to build a distributed spatio-temporal data processing platform for spatial data intelligent foundation model integration. For example, the latest SuperMap GIS platform is fully supported by Spark computing framework. The model provides a complete big data solution, including three main components of GIS core engine, client SDK, and application system. The GIS core engine can either be imported into the Spark environment as Scala or can be implemented in different front-end big data analytics software by supporting Python. By integrating iObjects for Spark services into the iServer product series, distributed spatial analytics model calculation services can be exposed via REST. The returned results can be easily used and visualized in applications with iObjects, iDesktop, iDesktop Cross, iMobile, iClient, and other 2D/3D linked clients (Figure 3-2).

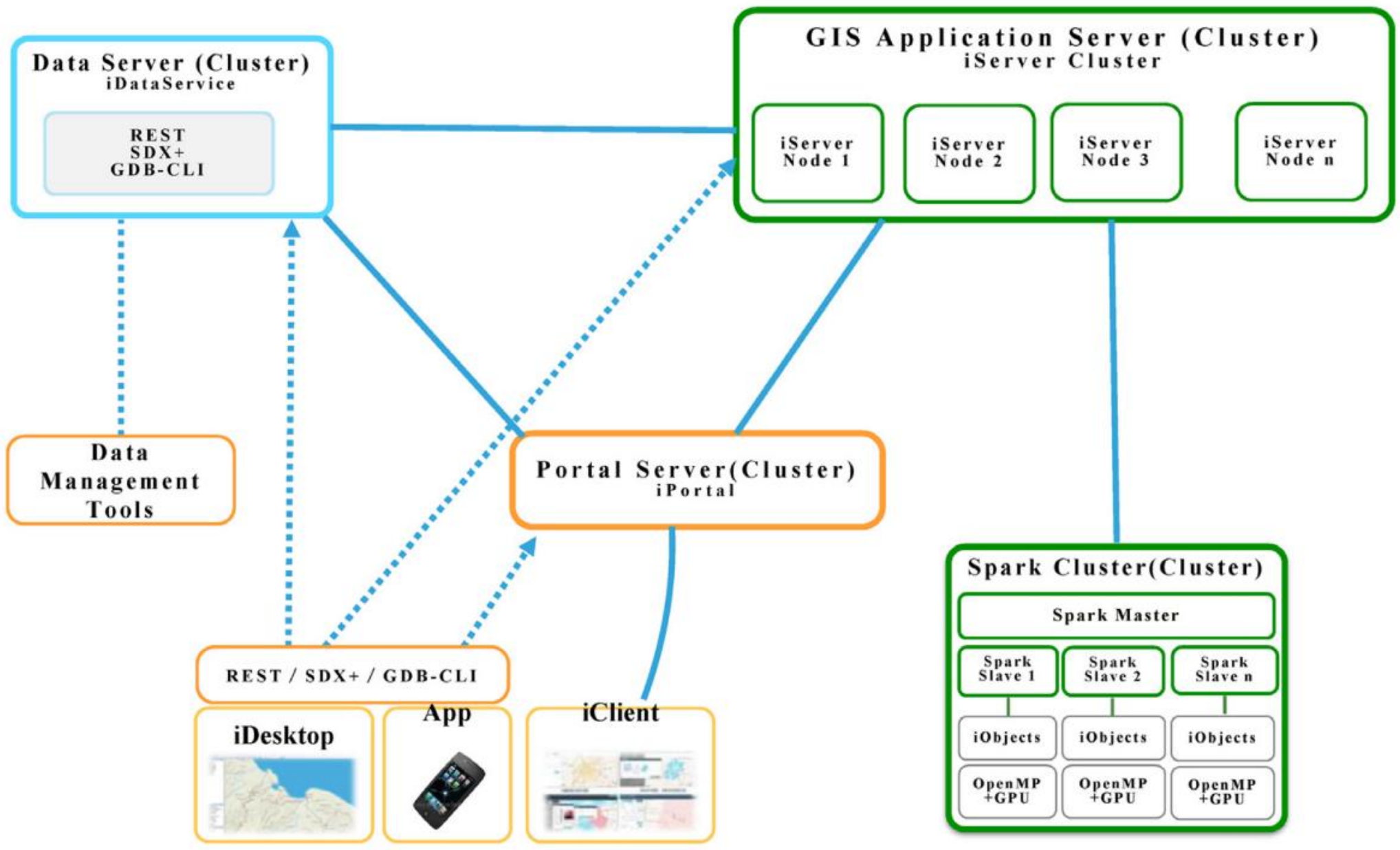

Fig. 3-2 Massive GIS cluster structure

The spatial data intelligent foundation model will be able to make full use of the large-scale storage, distributed memory, and cluster management. Its deployment capabilities brought by modern computing hardware and data centers through this initiative, which can further improve the efficient management of spatial-temporal flow based on spatial-temporal correlation on distributed scheduling and storage (Yueyi Li, Feng Zhang, Zhenhong Du, et al. ,2023). This initiative will also solve the insufficient storage space and computational capacity problems in traditional GIS technology. The model makes it possible to build large-scale application systems or conduct high-accuracy spatial relationship studies, promoting numerous types of applications and breakthroughs in geospatial modelling or algorithms, which will not only elevate geographic information science and geoscience to a new level, but also improve the efficiency of environmental and disaster management, urban planning, and other aspects.

### 3.1.3 Cloud Computing Integration

Cloud computing provides a set of models and methods for sharing computing resources. The dynamic allocation of computational resources using the edge-cloud-hybrid computing paradigm not only improves the system utilization efficiency and computational data collection efficiency (Chen et al., 2022), but also makes it possible to aggregate large-scale computational power in a short period of time. Recent advances in cloud-based remote sensing platforms disrupt the normality of big data processing methods, especially in terms of remote sensing big data (RSBD) analysis (Xu et al., 2022). Amazon, Google, Microsoft, and IBM all provide cloud data center services on a large scale. In China, Aliyun, Baidu Cloud, and Tencent Cloud also provide diverse cloud computing services. In recent years, numbers of startup companies have begun to provide services based on Docker technology, such as QiNiu and QingCloud. All these cloud computing platforms allow users to manage computing resources, rent resources on demand, and quickly build large-scale cloud computing clusters. Traditional server rental services were the major focus in the past. Today, distributed cloud computing clusters based on Hadoop/Spark have become a standard service in large data centers. With the rapid development of Docker container technology, the cloud computing services on which it is based can further reduce maintenance costs and provide a more flexible and agile solution to allocate and deploy resources. Service migration between different data centers or between public and private cloud centers has also become easier using Docker technology. In summary, cloud computing services have shifted from virtual machine-based server rental services to distributed cluster services and microservices based on new technologies such as Docker and Hadoop/Spark.

In Docker, cloud services can be packaged as microservices by business components and can be assembled upon demand at deployment time. Docker instances can be developed, tested, run, and deployed as needed in public, dedicated, industry, and private clouds. This will greatly reduce the cost of maintaining and developing cloud computing services. GIS cloud computing integration infrastructure must be fully combined with Docker technology to design, develop, and deploy systems based on the microservice conceptual model. For example, SuperMap iServer, iExpress, iPortal, iManager already support Docker; services based on its technical standards and microservice structure can be deployed to different cloud computing data centers. For the other features, integration between different types of computing infrastructures and automated management systems are also included. In addition, business users and individual users can access these services directly through maps or online portals (Figure 3-3).

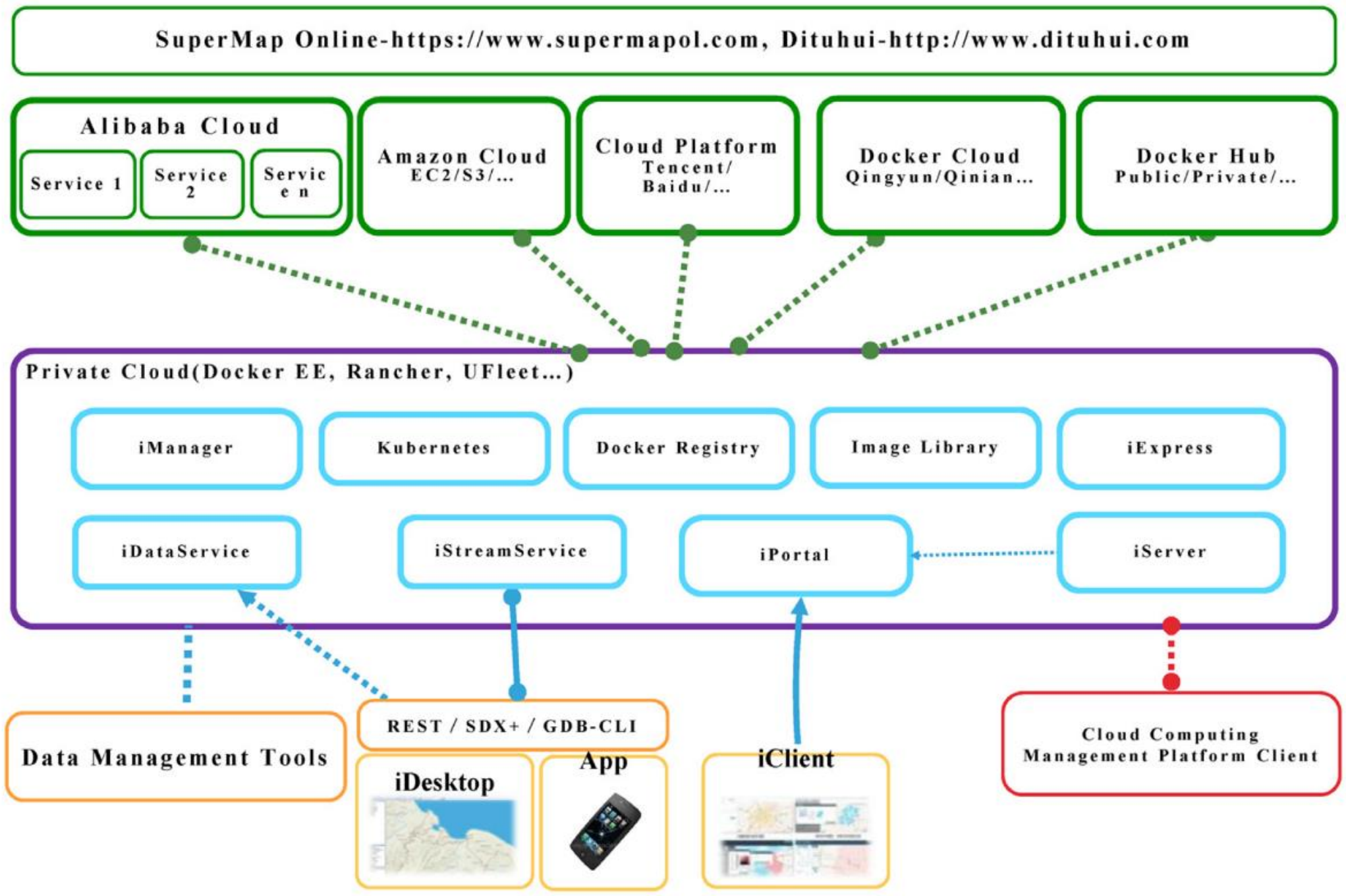

Fig. 3-3 Microservice structure based on cloud computing and Docker

By realizing Docker-based microservices infrastructure, GIS systems can be deployed as cloud computing modules for unified integration and management of multi-clouds. We can also fully integrate geospatial big data into cloud computing infrastructures. These features have become the core capabilities of modern data centers and have even become indispensable system components for many industries such as smart cities and environmental resources. It also provides the core functions of geospatial data management, spatial pattern analysis, geospatial data visualization, API sharing, and other application services.

### 3.1.4 3D and Virtual Reality

In recent years, 3D-related information technology has made significant progress. With the advancement in processing capabilities of graphics cards, the software standards and technologies supported by them, such as OpenGL, OpenCL, WebGL, etc., have also evolved rapidly. The breakthrough of VR/AR headsets and glasses has brought digital 3D applications into a new era. For the sake of the IT revolution, GIS has achieved two key improvements: integration of oblique photography and 2D-3D linkage function. The process of using the data at the end application has been simplified from retrieving complete geospatial data to constructing models. 3D GIS technology for spatial data intelligent foundation models provides a comprehensive solution for importing data, publishing services, analyzing applications, obtaining web access, and improving mobile applications. It is compatible with various types of servers, components, mobile platforms, web, desktop software, existing databases, cloud computing services, and other IT infrastructure.

By combining real 3D and BIM technology with spatial data intelligent foundation models, we can further apply them to multiple micro-management fields, such as building parts, managing component objects, as well as developing support systems for intelligent buildings or IoT networks. By combining VR/AR with spatial data intelligent foundation models, urban planning and management can provide a more enriched user experience, thereby improving the quality of public services in land management, municipal management, urban planning, etc. The 3D virtual reality of spatial data intelligent foundation models not only simplifies the data collection process, but also provides powerful on-site management functions. In addition, it creates a public IT platform that allows users to undertake further spatial planning, application, and optimization (Fig.3-4).

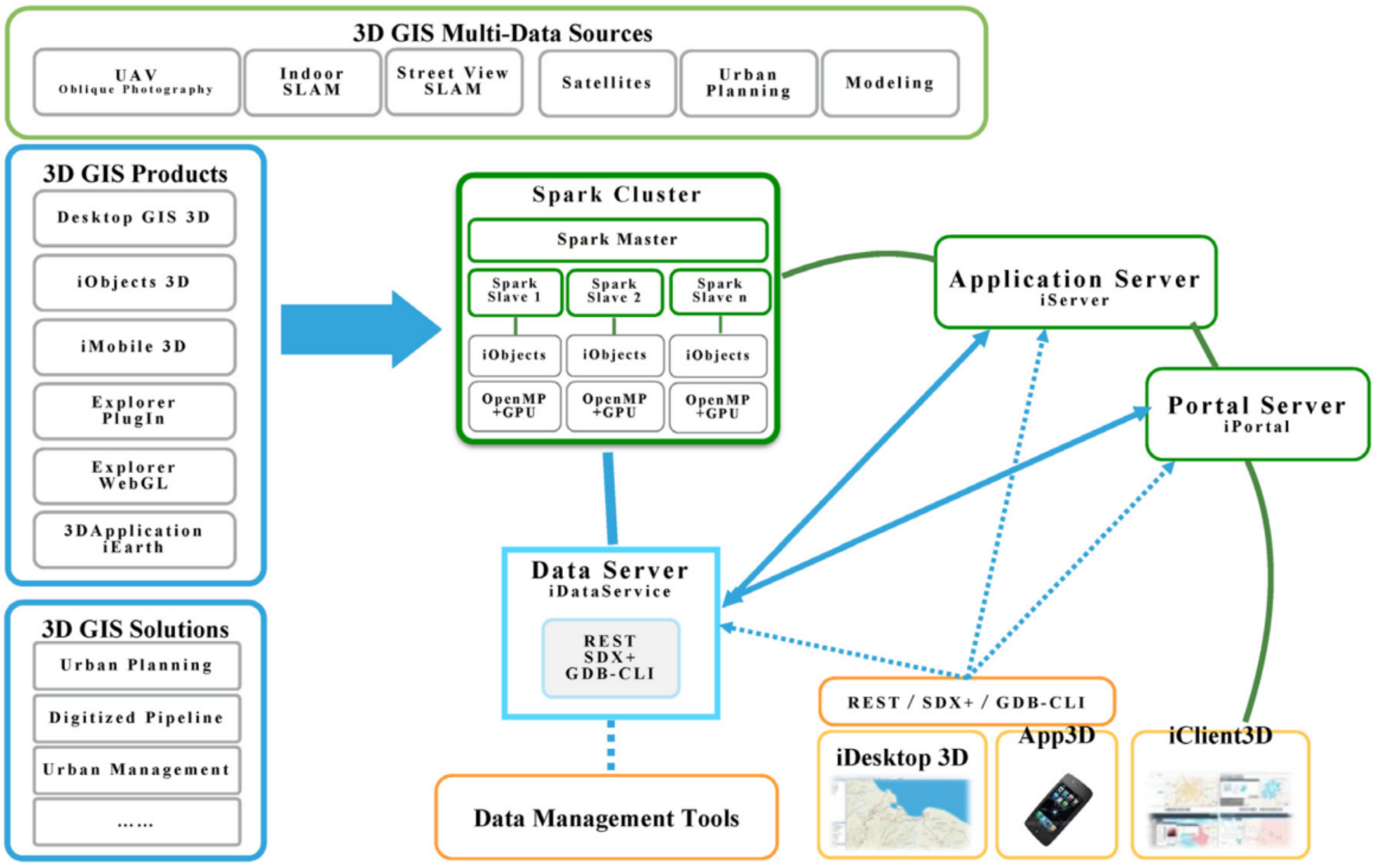

Fig. 3-4 Spatial data intelligent foundation model 3D virtual reality technology and solutions

3D GIS has become a key component of spatial data intelligent foundation models. However, the future 3D GIS will go beyond the current 3D GIS by simulating the real world. It will also support Boolean operations of actual instance models. In addition, introducing physical engines and collision detection algorithms into GIS will make the simulations of models and spatial-temporal environments more realistic. It will drive business applications in planning, design, pipelines, transport, construction, etc. The 3D virtual reality performance of future spatial data intelligent foundation models will also stimulate new advances in high-precision navigation, self-driving cars, airport management, etc.

### 3.1.5 Rapid Multi-Terminal Application

Software is like a magnifier of data value. The more the data is used, the greater the value generated; the higher the software compatibility required. Therefore, spatial data intelligent foundation models need to be equipped with not only powerful data capabilities but also versatile application compatibility, while the foundation model and its software platform should be applicable in different environments and all mobile devices. Clients can be divided into devices, operating systems, hardware infrastructure, and programming languages. The more client types supported, the greater the compatibility. This also means that more users can create more value for data.

In the practice of rapid multi-terminal applications of spatial data intelligent foundation models and their software products, the SuperMap GIS product family offers a wealth of client support. NET-based iDesktop and Java-based iDesktop can directly access cloud computing resources and massive storage. It has the function of professional GIS users processing data, generating maps, and analyzing spatial patterns. iClient provides WebGIS functions and is compatible with different browsers. Its functions include accessing server shared data, performing online analysis, visualizing scenes, etc. And it can be used on various operating systems without the need to install plugin software. iMobile not only provides SDKs developed for iOS and Android but also supports embedded operating systems such as Yuanxin

OS. Since GIS functions are easy to access and portable, SuperMap partners and other vehicle measurement devices have developed many applications on handheld platforms to meet their professional needs.

SuperMap is the GIS platform that supports the most terminal numbers, providing desktop, Web, and mobile SDKs and supporting accessing cloud services via API (Figure 3-5). Users can use the SDKs and plugin framework provided by SuperMap to develop general-purpose applications. It supports many Chinese local CPU brands such as Feiteng and Longxin, as well as the Chinese operating system Kiron OS. In conclusion, the progress made by the GIS system in terms of compatibility will release extra potential from big data, generating more data value.

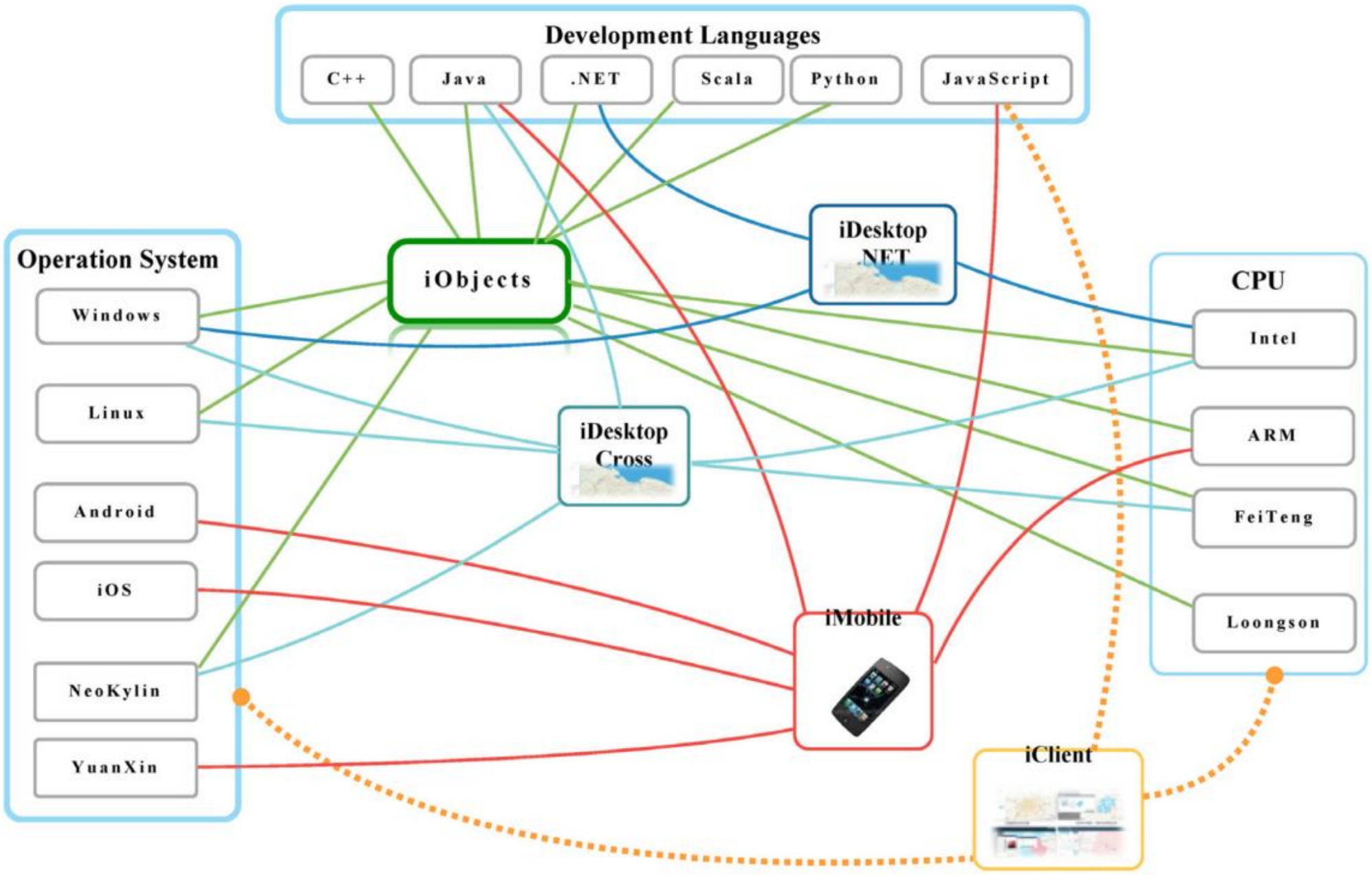

Fig. 3-5 Spatial data intelligent foundation model multi-client solution

### 3.1.6 Container Technology and Continuous Delivery

Under the promotion of Internet technology, software development methodology has undergone thorough development. With the emergence of Git/Gitlab/Github, distributed version control has replaced traditional centralized software development methods. Community development, public code review, automatic test, and continuous integration are now standard development methodologies. Compared with virtualization, Docker containers can be deployed at the system level and run directly on top of the Linux kernel. Docker allows users to compile software into packages and isolate run-time environments. By implementing Docker, it's easy to establish a customizable microservices system framework. Docker also reduces system deployment time and simplifies the migration process between data centers. To meet the latest needs of online platforms, the concept of continuous delivery and DevOps methods have made significant progress. Integration with cluster management systems, such as Mesos, Kubernetes, etc., has also been developed. Automated processes and continuous delivery methods on container frameworks greatly shorten the time for software updates and bug fixes and speed up the response speed of software development. This rapid iteration can speed up software innovation and minimize system risks.

To enable fast iteration, real-time testing, runtime verification, and controlled delivery capabilities, the spatial data intelligent foundation model is divided into three deployment zones: the development zone, validation zone, and production zone. In addition to test data and source code, the development zone also includes development tools, case libraries, and test systems. The validation zone includes validation data, validation systems, and evaluation systems. The production zone includes the production system, which contains the currently running system and the latest updated system. This makes gray publishing possible, which is behavior migrated to the new version of the system through AB testing methods. The SuperMap research team has developed a continuous delivery system that covers the entire solution to verify the reliability of the spatial data intelligent foundation model and solve the complexity problems brought by multiple versions. It has established an automated workflow for software development, integration, and testing. To move towards online platforms, Map Hui and online service portals have gradually built a framework that supports continuous delivery and DevOps. Currently, SuperMap iServer, iExpress, iPortal, iManager, and related products have integrated Docker and microservice framework, which are able to evolve and update themselves. Containers and continuous delivery methods smooth the software migration process and ensure that systems and data can be deployed on demand. This can improve efficiency and increase system availability. Development, testing, validation, deployment, and production maintenance/management/update in traditional software development will all be integrated. The system will enable fast response times, runtime error fixes, and no downtime for updates. In conclusion, building systems for continuous delivery and DevOps workflows using cloud computing frameworks, virtualization technology, and container technology will be the mainstream trend of future software development. This will become a necessary step to adapt to the challenges of big data.

## 3.2 Spatial Analysis and Visualization

### 3.2.1 Spatial Causal Inference

Causal discovery helps in understanding natural or physical mechanisms. Causal relationships play a fundamental role in Earth system science and are garnering increasing attention. However, for spatial scale research, designing and conducting controlled experiments to reveal causal relationships is impractical. Therefore, under the assumption of "cause preceding effect", methods for causal inference from time series data are frequently employed. While temporal reasoning can effectively determine most causal relationships between variables, there are limitations. If the time series is not long enough to capture significant changes, some important causal relationships may be overlooked. This limitation is especially prominent in Earth system science, where the evolution of global changes may take a long time to manifest noticeable changes.

Given the large-scale spatial distribution characteristics of research objects in Earth system science and the often-incomplete time series data, causal inference can be approached from another perspective by fully utilizing spatial differences. Specifically, although changes in a variable might not be detectable over time, the widespread distribution of the variable can make its changes identifiable spatially. The general principle of inferring causal relationships from time series data is based on the time change-response mechanism. Similarly, spatial changes (variations in variables at different spatial locations) and corresponding responses can also be used for causal inference. Causal associations are a crucial component of internal mechanisms that can be identified by observing and analyzing the phenomena they present. Spatial distribution is an important phenomenon for extracting causal associations and complements temporal changes. Corresponding spatial cross-sectional data record spatial processes and their interactions, providing valuable references for understanding causal associations. By formalizing mathematical methods, the reasoning framework can become easy to understand and transferable, allowing researchers from different disciplines or artificial intelligence (AI) to infer causal relationships from big data. From this perspective, spatial causal inference involves the processing, analysis, and induction of large-scale spatiotemporal panel data. For this purpose, integrating a spatial causal inference algorithm module within intelligent spatial data models and coupling it with deep learning algorithms helps in high-performance causal pattern reasoning based on panel data. The derived causal patterns can be integrated into deep learning algorithms to provide feature information from the dimension of geographical causal relationships.

Considering the need for causal inference from spatial cross-sectional data in Earth system science and the limitations of existing spatiotemporal causal models, a novel algorithm, denominated as Geographical Convergent Cross Mapping (GCCM), can be formulated through the utilization of dynamical systems theory and the generalized embedding theory, thereby incorporating them into intelligent spatial data models. This approach enables the swift identification and extraction of causal relationships from spatial data.

GCCM can identify causal relationships between spatial cross-sectional variables and estimate the corresponding causal effects. For two spatial variables X and Y on the same set of spatial units, organized either as a regular grid (raster data) or irregular polygons (vector data), their values and spatial lags can be viewed as observational functions reading values from each spatial unit. According to the generalized embedding theorem, their shadow manifolds $M_x$ and $M_y$ can be constructed, with s being the focal unit under study. For a given x, its corresponding y value can be predicted based on its nearest neighbors identified from $M_x$. This nearest-neighbor-based prediction is defined as cross-mapping prediction:

$$\hat{Y}_s \mid M_x = \sum_{i=1}^{L+1} (w_{si} Y_{si} \mid M_x)$$

Here, s represents the spatial unit where the value of Y needs to be predicted, $\hat{Y}_s$ is the prediction result, L is the embedding dimension, $si$ is the spatial unit used in the prediction, $Y_{si}$ is the observed value at location $si$, and is also the first component of the state in $M_y$, denoted as $\psi(y, s_i)$.

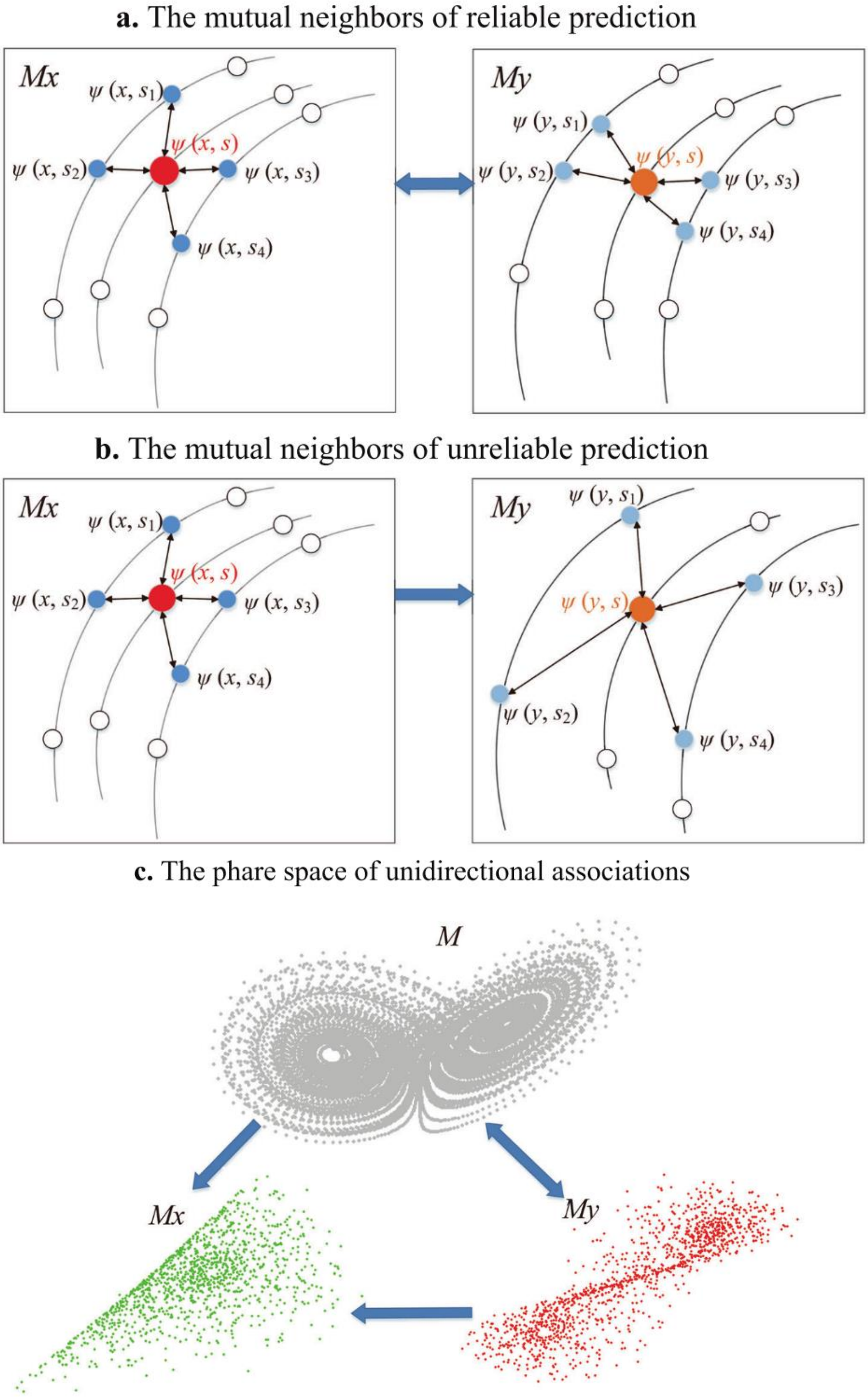

Fig. 3-6 Mutual neighborhood for cross-mapping prediction

Fig. 3-6 illustrates the basic idea of GCCM. In Figure 3-6 (a), the inter-neighbor points in the reconstructed manifold are reliable for cross-mapping prediction. The orange points labeled as $\psi(y,s)$ represent the state of the focal unit to be predicted. The four blue points $\psi(y,s_1),\psi(y,s_2),\psi(y,s_3)$ and $\psi(y,s_4)$ are the nearest neighbors included in the prediction. They are found through a one-to-one mapping between $M_x$ and $M_y$. $\psi(x,s)$ is the corresponding state of $\psi(y,s)$ in $M_x$. The nearest neighbor to $\psi(x,s)$ found in Mx is $\psi(x,s_1),\psi(x,s_2),\psi(x,s_3)$ and $\psi(x,s_4)$, and it can be used to identify $\psi(y,s_1),\psi(y,s_2),\psi(y,s_3)$and$\psi(y,s_4)$ in My using mutual spatial positions.

### 3.2.2 Spatial Data Clustering

Clustering is used to discover similar patterns based on the proximity of elements in the feature space. It is widely applied in computer science, biology, earth science, and economics. While partition-based and connectivity-based clustering methods have been developed, their effectiveness is hindered by the weak connectivity and heterogeneous density of the data. For spatial data intelligent foundation model, a built-in spatial clustering algorithm makes the neural network structure of foundation model more sensitive to the connectivity and heterogeneity of spatial data in the training and output process, which will play an important role in improving the clustering efficiency of foundation model spatial data and the stability and accuracy of the result output. The boundary-seeking clustering algorithm uses Clustering by Direction Centrality (CDC) and employs a density-independent metric based on the K-Nearest Neighbor (KNN) distribution to distinguish between internal and boundary points, addressing the limitations of spatial data clustering in terms of connectivity and heterogeneous density. Boundary points generate closed cages to constrain the connections of internal points, thus preventing cross-cluster connections and separating weakly connected clusters.

The core idea of CDC is to distinguish between boundary points and internal points of a cluster based on the distribution of KNN. Boundary points outline the shape of clusters and generate cages to bind the connections of internal points. Internal points of a cluster are surrounded by their neighboring points in all directions, while boundary points only include neighboring points within a certain directional range. To measure this difference in directional distribution, the algorithm defines the variance of angles formed by KNN in 2D space as the Direction Centrality Metric (DCM).

$$DCM = \frac{1}{k}\sum_{i=1}^{k}\left(\alpha_i - \frac{2\pi}{k}\right)^2$$

The KNN of a central point can form k angles $\alpha_1,\alpha_2,\ldots,\alpha_k$ (Figure 3-7a). For 2D angles, the condition $\sum_{i=1}^{k}\alpha_i = 2$ holds if and only if all angles are equal. The DCM reaches its minimum value of 0. This condition means that the KNN of the central point is evenly distributed in all directions. It can be maximized to $\frac{4(k-1)\pi^2}{k^2}$ when one of these angles is $2\pi$ and the others are 0. This extreme case occurs when the KNN is distributed along the same direction. According to the extremum, the DCM can be normalized to the range [0, 1], as shown in the following equation:

$$DCM = \frac{k}{4(k-1)\pi^2}\sum_{i=1}^{k}\left(\alpha_i - \frac{2\pi}{k}\right)^2$$

A sample result of DCM calculation shows that internal points of the cluster have relatively low DCM values, while boundary points have higher values (Figure 3-7b). Therefore, internal points and boundary points can be divided by a threshold TDCM. The partition results of two synthetic datasets, DS5 and DS7, validate the effectiveness (Figure 3-7c, d).

After calculating DCM and connecting internal points, we complete the process by assigning each boundary point to the cluster of its nearest internal point. CDC includes two controllable parameters, k and TDCM. k adjusts the number of nearest neighbors. TDCM determines the partition of internal and boundary points. In practice, considering that TDCM varies with data distribution, we use percentile ratio of internal points to determine TDCM as the DCM at the$[n(1-ratio)]$-th percentile of the sorted DCMs. The parameter ratio has intuitive physical significance and better stability, making it easier to specify than TDCM. According to the experimental results, a recommended ratio default parameter range of 70% to 99% of internal points is suggested to achieve better clustering results. However, when clusters are mixed, more boundary points (lower ratio) are needed to separate closely related clusters.

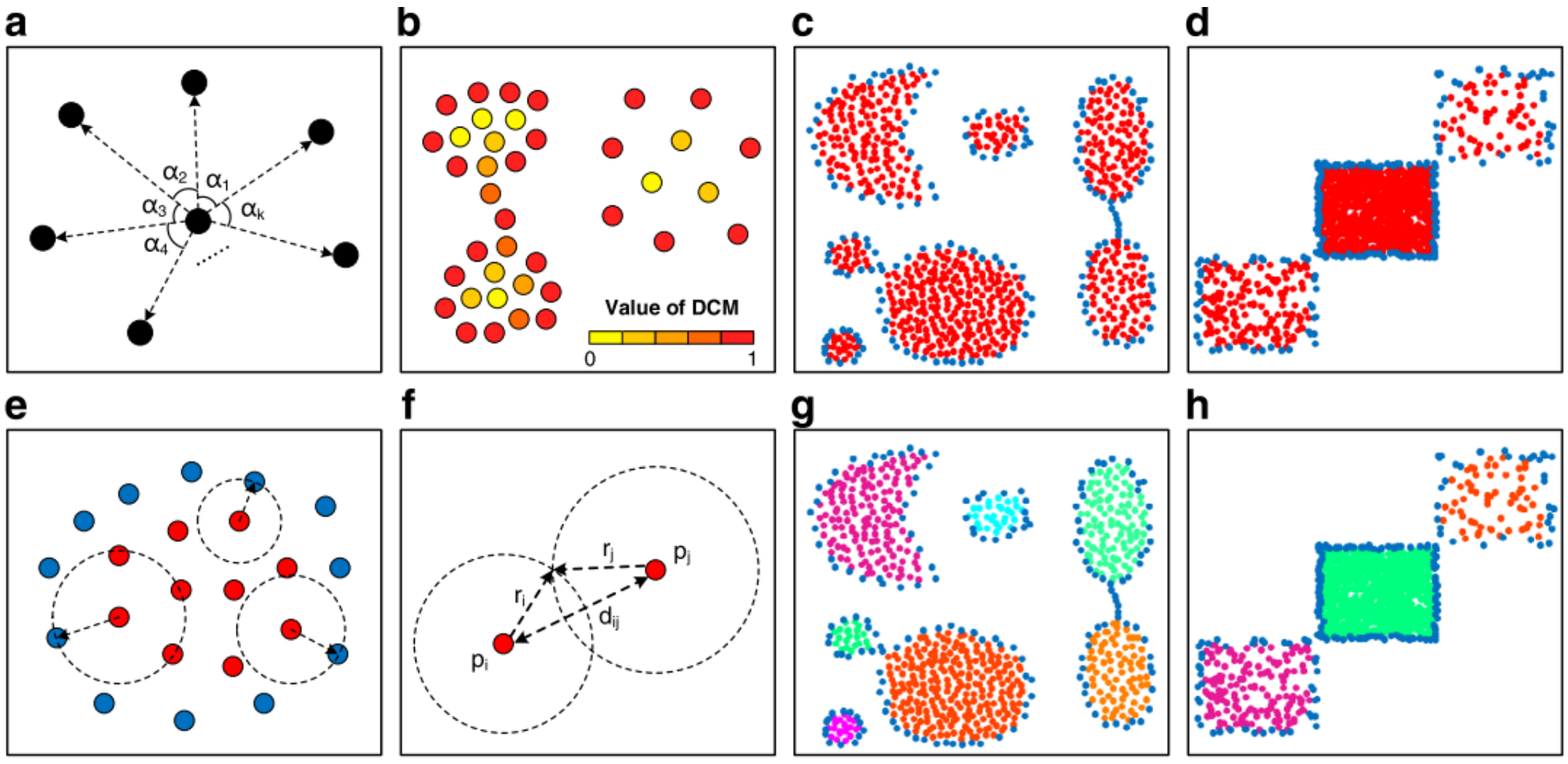

Fig. 3-7 CDC algorithm and intermediate results in 2D space

Figure 3-7 illustrates the CDC algorithm and intermediate results in 2D space. Figure 3-7(a) represents the central angle formed by the KNN of the central point; Figure 3-7(b) represents the DCM calculation result of the sample data; Figures 3-7(c) and 3-7(d) represent the partition results of internal and boundary points on two synthetic datasets, DS5 and DS7, where for DS5, k=10 and TDCM=0.1, and for DS7, k=30 and TDCM=0.1. Red points indicate internal points. Blue points indicate boundary points; Figure 3-7(e) represents the reachability distance of internal points; Figure 3-7(f) represents the association rules connecting internal points; Figures 3-7(g) and 3-7(h) represent the connection results of internal points on DS5 and DS7.

### 3.2.3 Spatial Data Map Visualization

Maps are an ancient but commonly used product that needs to be both "accurate" and "beautiful." Balancing these two aspects requires solid professional skills in map elements, layout design, and other aspects, for making cartography a field with a high threshold. Maps visualize various information to efficiently utilize geographic data, such as displaying the spatial patterns of geographic features, conducting warning analysis for natural disasters, and assessing the differences in population movement (Gao Q L, Yue Y, Tu W, et al., 2021). Currently, there are many attempts to combine mapping with AI, with much discussion on style transfer, image generation, and other AI models. However, such methods often treat maps as a whole and generate them all at once, which may overlook the procedural management of mapmaking and the design of individual map elements, making it difficult to achieve "accuracy." Based on the foundation model intelligent agent framework, the automatic rendering of spatial data is achieved by combining and calling basic mapping tools for "accuracy". Additionally, the DALLE-3 model is embedded to generate creative symbols based on user intent to achieve "beauty", creating the intelligent spatial data map visualization model, MapGPT. This framework is scalable and interactive. If users are dissatisfied with the resulting map elements or decorations, they can interact with the foundation model intelligent agent to adjust and update the content.

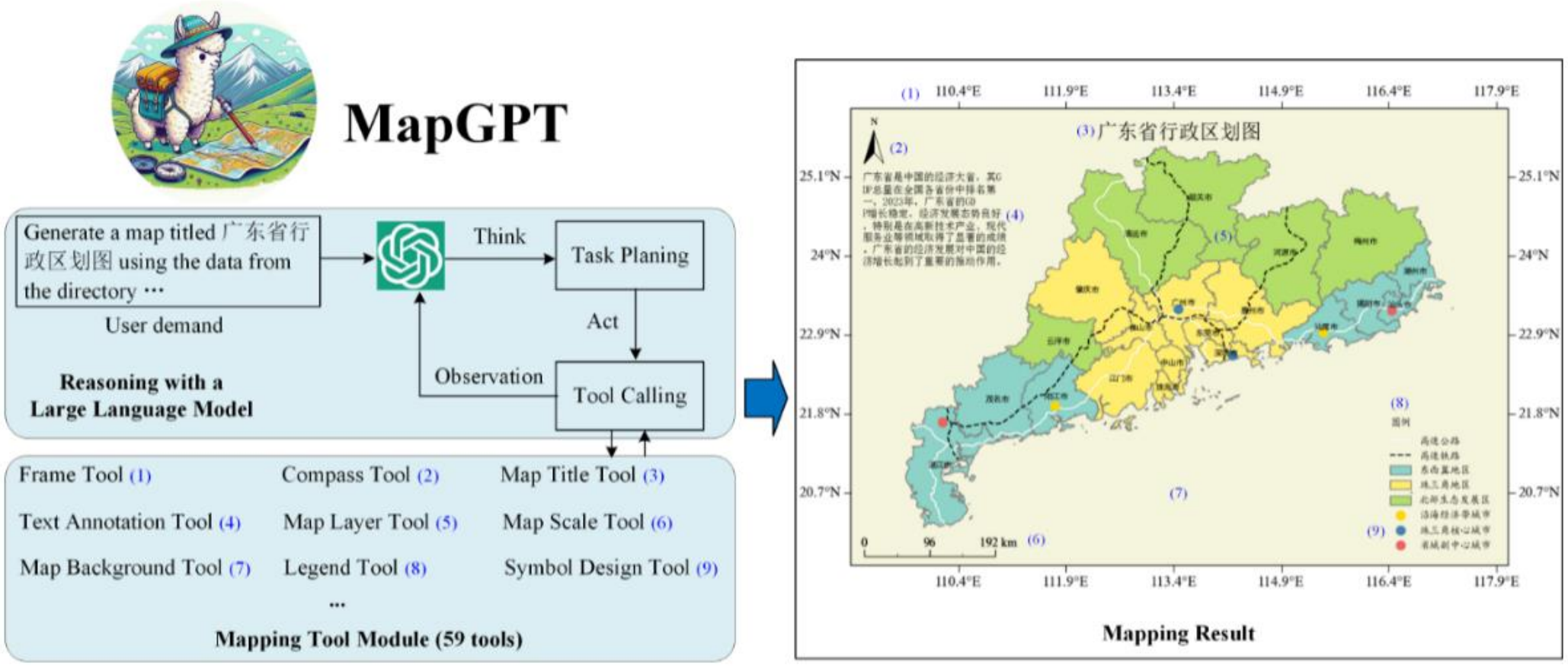

Fig. 3-8 Basic framework of MapGPT

MapGPT is based on the LangChain framework, using OpenAI's GPT4 (0613 version) as the agent of the framework. It defines multiple mapping tools to achieve fine-grained adjustment and drawing of various map elements. Generally, large language models (LLMs) accept text as input and output responses based on that text. Therefore, to enable a large language model to have mapping capabilities, it needs to be equipped with professional mapping tools. Additionally, an environment needs to be established to connect the language model with the mapping tool module, allowing it to "learn to use" the mapping tools. In this paper, the LangChain framework is used to connect the large language model with professional mapping tools. LangChain is a framework designed for developing applications for large language models with the main goal of seamlessly integrating large language models and other data sources and tools, enabling interaction. In this paper, we designed the prompt shown in Figure 3-9 to guide the large language model to identify and call the appropriate mapping tools to complete mapping tasks.

You are a map expert and you are proficient in generating maps using vector or raster data. Your task is to answer the question or solve the problem step by step using the tools provided.

You can only respond with a single complete "Thought, Action, Action Input, Observation" format OR a single "Final Answer" format.
**Complete format**:
Thought: (reflect on your progress and decide what to do next (based on observation if exist), do not skip)
Action: (the action name, should be one of [{tool_names}]. decide the action based on previous Thought and Observation)
Action Input: (the input string to the action, decide the input based on previous Thought and Observation)
Observation: (the result of the action)
(this process can repeat and you can only process one subtask at a time)
**OR**
Thought: (Review original question and check my total process)
Final Answer: (Outputs the final answer to the original input question based on observations and lists all data paths used and generated)

Answer the question below using the following tools: {tool_strings}
Your final answer should contain all information necessary to answer the question and subquestions.

**IMPORTANT**: Your first step is to learn and understand the following rules and examples, and plan your steps accordingly:
The general process of making a map is: first initialize the map, add map layers, add other map components as needed, and finally generate the map. When making a map, the first step must be to initialize the map, and the last step must be to generate the map which is use map_save tool. These two steps are indispensable.

Do not skip these steps.

Begin!
Previous conversation history:{chat_history}
Question: {input}
Thought: {agent_scratchpad}

Fig. 3-9 Frame prompt design

MapGPT defines corresponding mapping tools for multiple map elements to achieve fine-grained control over different map elements, meeting users' needs for detailed mapping. The tools mainly include six aspects: map initialization, designing map symbols using the DALLE-3 model, adding map layers, modifying map element parameters, adding map elements, and saving output maps.

(1) Map Initialization: Build the map framework using the tools in this section based on the specified geographic spatial data provided by the user. Specifically, the tools in the map initialization module are mainly used to define the map extent and corresponding coordinate system based on the user's given geographic spatial data, thereby setting the map background color according to user's requirements.

(2) Designing Map Symbols using the DALLE-3 Model: Map symbol design is a challenging task. Designing reasonable map symbols can effectively express corresponding geographic information. To address the challenges of map symbol design, MapGPT introduces the DALLE-3 model, which can accept text input and then generate images and symbols matching the textual descriptions. To enable DALLE-3 to better generate map symbols representing geographic features, MapGPT designs the following prompt: "Please help me design a map symbol that represents {keywords}. Try to keep it simple and understandable, using only one-color tone and reflecting the style of a simple drawing. There should be as few elements as possible. Try to present only the symbol I need." Here, keywords are the corresponding content input by the large language model based on user's requirements. Additionally, because map symbol design is a subjective task, MapGPT has designed an interactive strategy: in a single tool call, the model simultaneously generates 3 symbols, and users can choose one of them to represent the corresponding geographic feature.

(3) Adding Map Layers: This part of the tools is mainly used to control the addition of map layers, including point, line, and polygon feature layers. The model can automatically identify the corresponding geographic features and load the appropriate map symbols to represent them based on requirements.

(4) Modifying Map Element Parameters: For different map elements, multiple tools are designed to adjust their detailed expressions. For example, for the map compass element, multiple tools such as modify_compass_location, modify_compass_width, modify_compass_color, and modify_compass_style are designed to adjust its expression. Based on these tools, this framework can achieve fine-grained control over map elements.

(5) Adding Map Elements: After modifying the parameters of the corresponding map elements, the add map elements tool need to be drawn on the map. Common map elements include compasses, scales, frames, legends, titles, and annotations.

(6) Saving Output Maps: Save the output map.

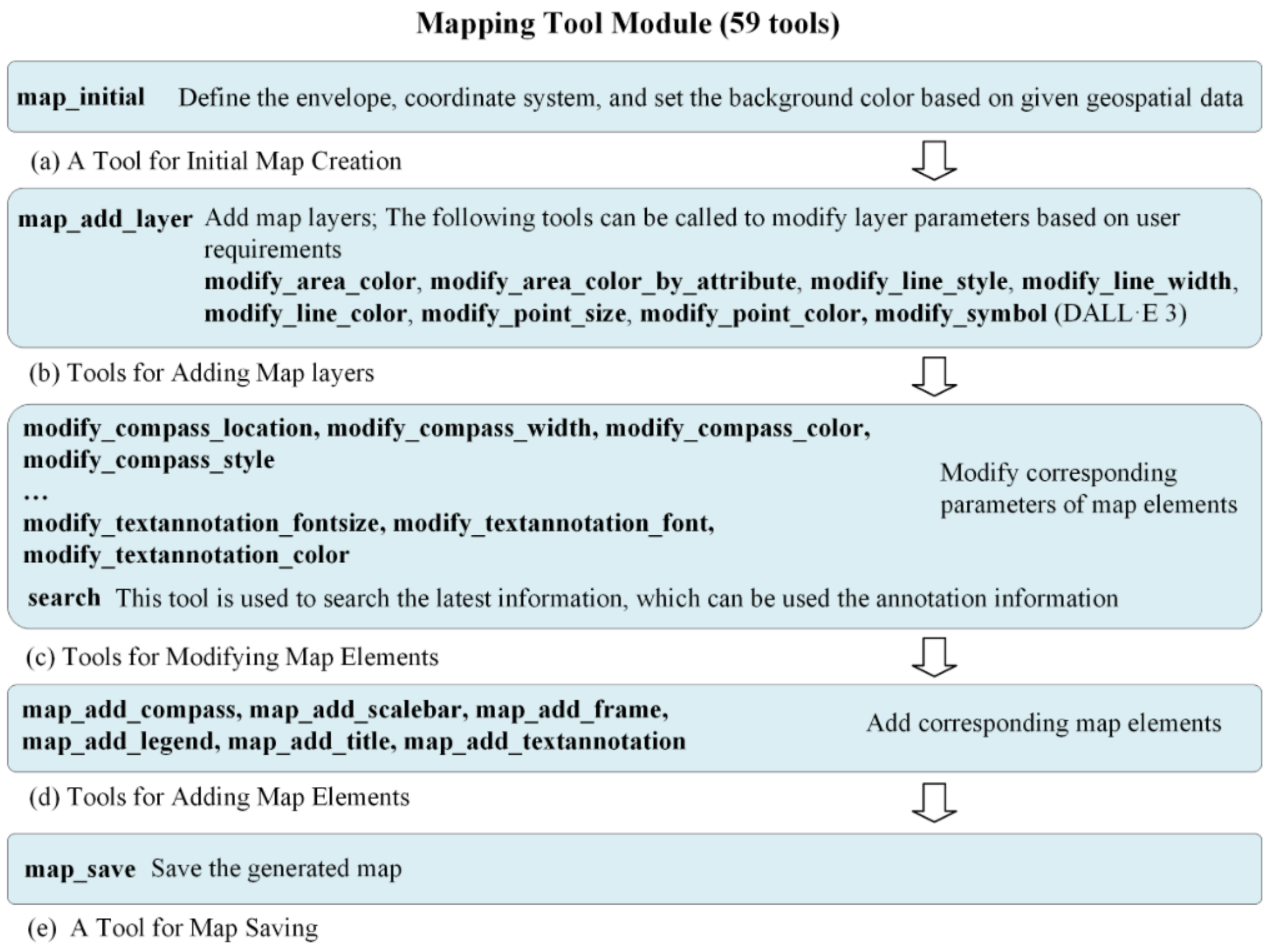

Fig. 3-10 MapGPT Mapping tool module

## 3.3 Geospatial Intelligent Computing

Geospatial intelligent computing is rapidly becoming a major theme in the research and development of geography, geographic information science (GIScience). Many disciplines involving complex patterns and processes that can be found in geographic spaces (i.e., the Earth's surface and near-surface). Geospatial intelligent computing presents a new set of challenges, yet it revitalizes traditional fields of geographic science in fundamentally different directions. It benefits from a perfect confluence of trends: the availability of numerous new data sources from remote sensing, social media, and sensor networks, access to virtually unlimited computational resources, and the integration of powerful new methods in data analytics and machine learning.

### 3.3.1 Deep Learning

In foundation models of spatial data intelligence, deep learning algorithms and neural network structures are the most critical components. Deep learning algorithms refer to methods for learning data features by constructing multi-layer neural networks. Common deep learning algorithms include Convolutional Neural Networks (CNNs), Recurrent Neural Networks (RNNs), and Variational Autoencoders (VAEs). Neural network structure refers to the way nodes and connections are organized within a neural network, including hierarchy, neuron types, activation functions, etc. Different neural network structures are suited to different tasks and data types. In addition to deep learning algorithms, optimization method is another crucial part of foundation model deep learning technology. Optimization methods involve adjusting model parameters to achieve optimal performance on the training dataset. Common optimization methods include Stochastic Gradient Descent (SGD), Adam, and RMSprop. These optimization methods have distinct characteristics and applicability. Also, the appropriate method should be selected based on the specific context. The deep learning algorithms in large spatial data models are also related to the generalization ability of the model, which is one of the key evaluation metrics for model quality. To enhance the model's generalization ability, methods such as data augmentation, regularization, and ensemble learning can be employed within deep learning algorithms. These methods can improve the stability and generalization ability of the model, thereby enhancing its overall performance.

Deep learning algorithms enhance the spatial intelligent computing performance of spatial data intelligence models in the following ways:

(1) Feature Extraction and Representation Learning: Spatial data typically have high dimensionality and complex features. Deep learning can learn high-level feature representations through neural networks, effectively capturing the essential characteristics of the data. Thus, it improves the model's expressive power and generalization ability.

(2) Spatial Data Classification and Recognition: Deep learning can be applied to classification and recognition tasks of spatial data, such as land cover classification in remote sensing images and urban building identification. By training deep learning models, automatic identification and classification of different categories within spatial data can be achieved.

(3) Spatial Data Analysis and Prediction: Deep learning can be utilized for analysis and prediction tasks of spatial data, such as spatiotemporal forecasting of meteorological data and traffic flow prediction. By learning the spatiotemporal relationships within the data, future spatial data can be accurately predicted and analyzed.

(4) Map Generation and Simulation: Deep learning can be used for map generation and simulation tasks, such as generating realistic map images through Generative Adversarial Networks (GANs) or simulating and predicting maps using Recurrent Neural Networks (RNNs).

(5) Spatial Data Association and Reasoning: Deep learning facilitates the association and reasoning between spatial data, such as modeling spatial network structures using Graph Neural Networks (GNNs) to learn and infer relationships between spatial data.

Currently, widely applied deep learning algorithms in large spatial data intelligence models include:

(1) Convolutional Neural Network (CNN): CNNs are deep learning models specifically designed for processing image data. They extract image features and perform classification or regression tasks through components, such as convolutional layers, pooling layers, and fully connected layers. The most fundamental module of CNNs is the convolution operation, which involves using convolutional kernels (filters) to filter the input image, capturing local features like edges and textures. After the convolution operation, activation functions are typically used to apply nonlinear transformations to the feature maps, enhancing the model's nonlinear expressive capability. Pooling operations are used to reduce the dimensionality of the feature maps, decreasing the number of parameters and improving computational efficiency. After multiple convolution and pooling operations, the resulting feature maps are flattened into a one-dimensional vector and processed through fully connected layers for classification or regression tasks. CNNs are primarily applied in large spatial data intelligence models in the following areas:

① Remote Sensing Image Classification and Recognition: CNNs can be applied to classify and recognize remote sensing images, such as identifying different land cover types (e.g., water bodies, forests, buildings) or monitoring surface cover changes. By training CNN models, automated analysis and recognition of remote sensing images can be achieved.

② Geospatial Object Detection and Segmentation: CNNs can be used for detecting and segmenting geospatial objects, such as buildings, roads, and vehicles in remote sensing images. This is crucial for urban planning, traffic management, and other related fields.

③ Map Image Generation and Enhancement: CNNs can be applied to generate and enhance map images. For example, they can generate realistic map images using Generative Adversarial Networks (GANs) or enhancing map image quality and clarity using CNNs.

④ Spatial Data Association and Reasoning: CNNs can process graph data in spatial datasets, such as social networks or transportation networks. By learning the network structure and node features, CNNs can achieve associations and reasoning between spatial data.

(2) Recurrent Neural Network (RNN): RNNs are specialized neural network models designed for processing sequential data. They have a memory function, allowing them to retain information from previous inputs and apply it to the current computation. RNNs process sequential data through a recurrent structure, where each time step receives the current input and the hidden state from the previous time step. Then, outputs the current time step's hidden state and prediction result. The core idea of RNNs is to handle sequential data with a cyclic structure that provides memory capabilities. The basic principles of RNNs can be divided into three parts: the input layer, the hidden layer, and the output layer. The input layer receives the sequence data. The hidden layer is the core part of the RNN, containing a recurrent structure that receives the hidden state from the previous time step and computes the current time step's hidden state using the current input and the previous hidden state. The output layer calculates the output based on the current time step's hidden state, which can be a prediction value or a classification result. Traditional RNNs face issues of vanishing and exploding gradients, making it difficult to handle long sequences of data. To address these issues, more complex variants of RNNs, such as Long Short-Term Memory networks (LSTMs) and Gated Recurrent Units (GRUs), were developed. These advanced structures can more effectively process long sequences of data.

RNN/LSTM/GRU are primarily applied in spatial data intelligence models in several key areas:

① Spatio-temporal data prediction: RNNs can be utilized for predicting spatio-temporal data such as meteorological data and traffic flow data. By training RNN models, relationships between spatio-temporal data can be learned, enabling predictions of future spatio-temporal data.

② Spatio-temporal sequence analysis: RNNs can analyze sequence features of spatio-temporal data, such as studying spatio-temporal correlations between different locations, exploring the periodicity and trends of spatio-temporal data.

③ Geographical environment simulation: RNNs can simulate and generate spatio-temporal data of geographical environments. For instance, by learning the spatio-temporal characteristics of meteorological data, realistic meteorological data can be generated, or the spatio-temporal variations of urban traffic flow can be simulated.

④ Anomaly detection and warning: RNNs can be employed to detect and warn about anomalies in spatio-temporal data, such as monitoring abnormal traffic flow or predicting the occurrence of natural disasters like floods or earthquakes.

⑤ Geographical event prediction: RNNs can predict the occurrence and impact of geographical events by analyzing spatio-temporal data. For example, through spatio-temporal data analysis, predictions can be made regarding urban development trends, changes in land use, and other geographical phenomena.

(3) Graph Neural Network (GNN): GNNs are specialized neural network models designed to handle graph-structured data efficiently for learning and inference tasks. Graph data is typically represented as a network structure composed of nodes and edges, where each node represents an entity and each edge represents a relationship between nodes. GNNs analyze and predict graph data by learning the connections between nodes and the features associated with each node. In GNNs, each node is characterized by a feature vector representing its attribute information. In addition to node features, edges in the graph can also carry feature information. Based on this framework, GNNs learn relationships between nodes and feature representations through information propagation. Apart from node-level feature representations, GNNs can also learn graph-level feature representations. The core component of GNNs is the graph convolutional layer, which facilitates information propagation and feature updates between nodes. By stacking multiple layers of graph convolutional operations, GNNs progressively learn feature representations of nodes and edges in the graph, enabling effective learning and inference tasks on graph data. Through parameter optimization via backpropagation algorithms, GNNs autonomously learn optimal feature representations for nodes and edges, thereby enabling efficient processing and analysis of graph data.

Graph Neural Networks (GNNs) are applied in spatial data intelligence models across several key areas:

① Spatial relationship modeling: GNNs are used to model geographical relationships in spatial data, such as traffic networks between cities or distances between geographic locations. By learning these geographical relationships, GNNs model the connectivity and influence relationships between spatial data points.

② Geographical environment analysis: GNNs analyze complex relationships in geographical environments. For example, by learning correlations between different locations in meteorological data, GNNs enable spatial analysis and prediction of weather patterns.

③ Geographical inference: GNNs infer hidden relationships in geographical information. By learning traffic flow between cities and population movement data, GNNs can infer urban development trends and future planning directions.

④ Geographical event prediction: GNNs predict the occurrence and impact of geographical events. For instance, by learning from historical data of natural disasters like earthquakes and floods, GNNs forecast the likelihood and scope of future disasters.

⑤ Spatial data visualization: GNNs facilitate the visualization of spatial data. For example, by learning relationships between geographic locations, GNNs can visualize map data to aid users in understanding and analyzing spatial data effectively.

(4) Generative Adversarial Network (GAN): GANs are a type of deep learning model composed of a generator and a discriminator, designed to learn to generate realistic data samples through adversarial training. The generator is responsible for producing realistic data samples, while the discriminator's role is to distinguish between the samples generated by the generator and real samples. Through adversarial training, both components continuously optimize their performance, ultimately enabling the generator to produce highly realistic data samples. The generator receives random noise as input and generates realistic data samples, while the discriminator evaluates these samples and provides a probability indicating the likelihood that a given sample is real. The training process of GANs is adversarial with the generator and discriminator continually improving through this adversarial training method.

Generative Adversarial Networks (GANs) are applied in spatial data intelligence models across several key areas:

① Map image generation: GANs can generate realistic map images, such as images of city streets or forests. By training the generator, GANs create visually accurate images with specific map features, which is useful for map visualization and analysis.

② Geographical environment simulation: GANs simulate changes in geographical environments. For example, by learning from meteorological and geographic location data, GANs generate realistic weather scene images or simulate changes in urban traffic flow.

③ Enhancement of geographical information: GANs enhance the visualization of geographical information. By learning from map data, GANs can enhance map images to improve their quality and clarity.

④ Geographical event prediction: GANs predict the occurrence and impact of geographical events. For instance, by generating images of different geographical environments, the discriminator assesses the realism of these images to predict the likelihood and impact of geographical events.

⑤ Anomaly detection and warning: GANs detect anomalies in geographical data. For example, they can monitor abnormal regions in map images or alert about the occurrence of natural disasters like floods or earthquakes.

### 3.3.2 Spatial Optimization and Planning

Spatial optimization and planning in large-scale spatial data intelligence models are crucial for achieving intelligent spatial computing performance, focusing on utilizing spatial data and intelligent algorithms to optimize and plan the utilization and layout of spatial resources for optimal spatial configuration. This topic typically includes several key aspects: ① Spatial Optimization: Utilizing intelligent algorithms to analyze and optimize spatial data to obtain the best spatial layout solutions, applicable in urban planning, transportation planning, and resource allocation to enhance spatial resource utilization efficiency and quality. ② Spatial Planning: Developing rational spatial development and utilization plans within specific spatial boundaries based on planning objectives and constraints, applicable to urban development, land use planning, and natural resource conservation through spatial data analysis and planning to achieve sustainable use and development of spatial resources. ③ Intelligent Algorithms: Commonly used in spatial optimization and planning, intelligent algorithms such as Genetic Algorithms, Ant Colony Optimization, and Particle Swarm Optimization simulate biological evolution and group behavior to find optimal or near-optimal spatial layout solutions, addressing complex spatial optimization and planning challenges. ④ Spatial Data Analysis: The foundation of spatial optimization and planning includes the collection, storage, processing, and analysis of spatial data like geographic information, remote sensing data, and sensor data to derive spatial features, patterns, and trends, providing basis and support for spatial optimization and planning. ⑤ Application Areas: Widely applied in urban planning, transportation planning, environmental protection, and resource management, spatial optimization and planning contribute to achieving sustainable urban development, rational resource utilization, and environmental improvement. The main algorithms involved include Genetic Algorithm (GA), Ant Colony Optimization (ACO), Particle Swarm Optimization (PSO), among others.

(1) Deep Learning-Based Urban Community Spatial Planning

Effective urban community spatial planning plays a crucial role in the sustainable development of cities. Spatial data intelligence models apply AI-based urban planning algorithms to generate spatial plans for urban communities. To overcome the complexities of diverse and irregular urban geography, a graph is constructed to describe the topological structure of any form of city, framing urban planning as a sequential decision-making problem on this graph. Addressing the challenge of vast solution spaces, reinforcement learning algorithms based on graph neural networks are introduced in large-scale models. Experiments involving synthetic and real-world communities demonstrate that computational models outperform human-designed plans based on objective metrics and can generate spatial plans responsive to various scenarios and needs. In the collaborative workflow of urban planning AI, designers benefit from large-scale models to enhance productivity, generating more effective spatial plans in less time.

The large-scale model converts all geographic elements into three geometric types: polygons, polylines, and points. It represents the entire community as a graph. Nodes correspond to these geometric shapes and edges denote spatial adjacency relationships between these shapes; two nodes are connected if their underlying geometric shapes touch each other. Each node stores its geographic information as node features, including the type of geometric shape, coordinates, width, height, length, and area. This approach transforms spatial planning into a decision-making problem on a dynamic graph, evolving based on actions taken by the AI agent. During generative planning, the large-scale model follows a deep reinforcement learning framework where AI agents interact with the spatial planning environment to learn land use and road layout (Fig. 3-11). The Sequential Markov Decision Process (MDP) (Fig. 3-11 e, f) comprises key components: ① the current spatial plan and adjacency graph with rich node features and other information, such as statistics on different land use types; ② actions indicating the placement of current land use or construction of new road segments, derived from selected edges or nodes in the adjacency graph; ③ rewards for all intermediate steps are zero except for the final step of each phase, where spatial efficiency of land use and roads is evaluated; ④ transitions describe changes in the layout given the selected positions, occurring in both original geographic space (new land use and roads on the map) and transformed graph space (new topologies and attributes in the graph).

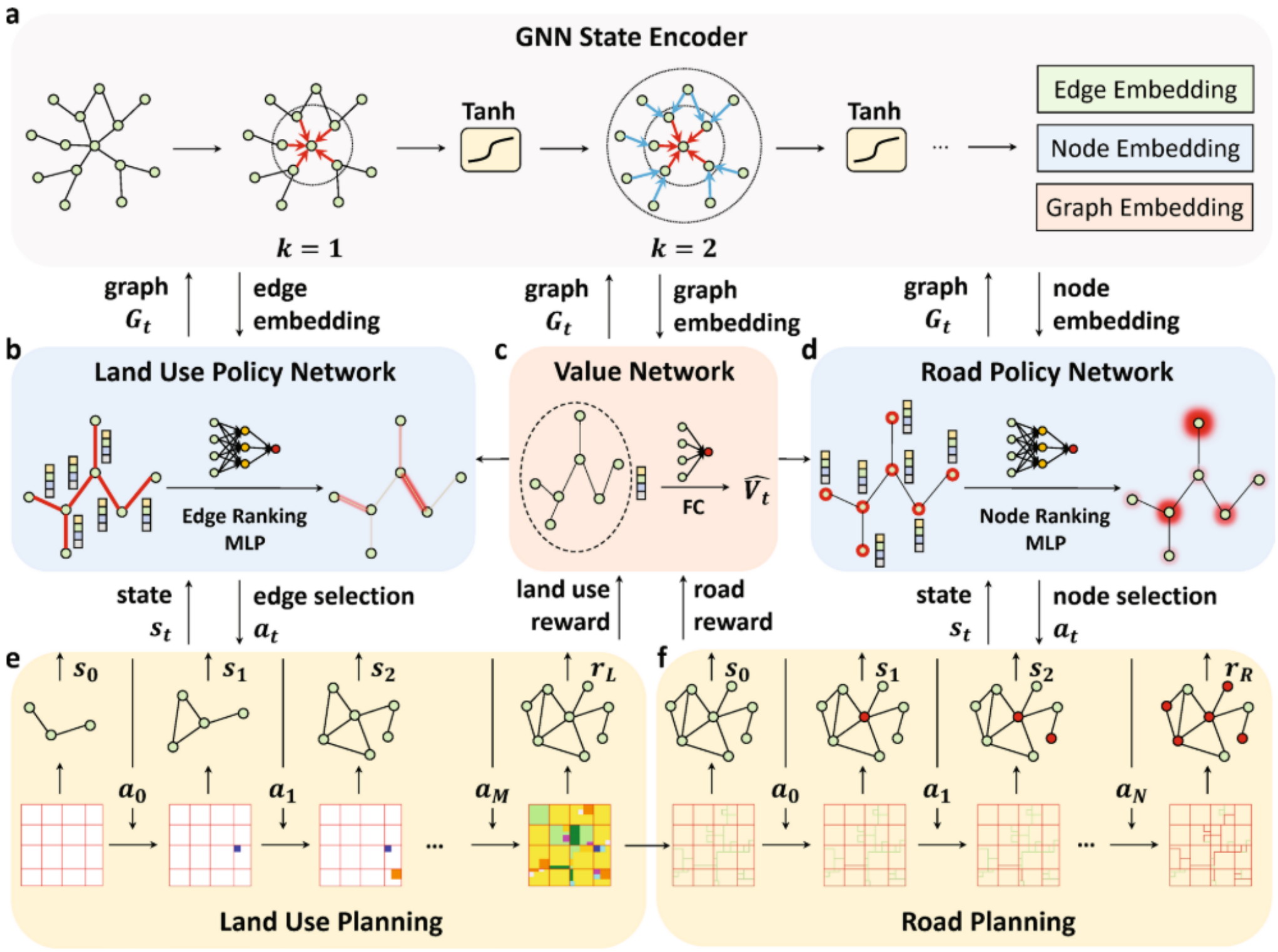

Fig. 3-11 Deep learning urban community spatial planning algorithm framework

In each step, the agent encodes the graph using GNN to represent its state. Through multiple messages passing and non-linear activation layers, the GNN state encoder generates effective representations of edges, nodes, and the entire graph (Fig. 3-11a), which are utilized by the value and policy networks (Fig. 3-11b-d). Specifically, since choosing a location for land use is akin to selecting an edge on the graph, the land use policy network employs edge embeddings and uses a multi-layer perceptron MLP for edge ranking, as shown in Fig. 3-8b. The score obtained for each edge indicates the sampling probability for that edge, which is returned to the environment and becomes the probability of placing land use at the specified edge. Similarly, in road planning, the road policy network uses node embeddings and employs an MLP node ranking (Fig. 3-11d) to score each node, outputting the probability of selecting a parcel boundary and constructing it into a road segment. Finally, the value network utilizes graph embeddings to summarize the entire community and predicts planning rewards through fully connected layers (Fig. 3-11c). During training, to master spatial planning skills, millions of spatial plans are completed using this model, exploring a vast solution space and using this real-time training data to update the parameters of the neural network.

(2) Predictive Simulation of Urban Resident Mobility and Transportation Patterns

Understanding how humans move and choose transportation modes in large-scale transportation networks is crucial for predicting urban congestion and managing traffic efficiently. The Spatial Data Intelligence Grand Model utilizes an intelligent algorithm named DeepTransport, built on extensive heterogeneous data such as GPS records and traffic network data, to simulate and predict human mobility and transportation patterns within cities. DeepTransport's key components are based on deep learning architectures aimed at extracting insights into human mobility and transportation patterns from big and diverse datasets. Given any period, specific city locations, or observations of human movements, the algorithm can autonomously simulate or predict future movements and transportation modes in large-scale transportation networks. Results and validations demonstrate its efficiency and superior performance in predicting and simulating human transportation modes beyond previously perceived capabilities.

The algorithm architecture, as shown in Fig. 3-12, consists of four main components: a database server, a preprocessing module, a deep learning module, and a visualization and evaluation module. The database server manages and stores data sources, providing indexing, retrieval, editing, and visualization services. The preprocessing module cleans data and maps human movements onto the traffic network, generating large-scale human GPS trajectories annotated with transportation mode labels. The deep learning module is crucial to DeepTransport, featuring four LSTM layers: one encoding layer for input sequence separation, one decoding layer for output sequence separation, and two shared hidden layers with identical parameters used for training. Finally, the visualization and evaluation module visualize results and assesses the overall system performance.

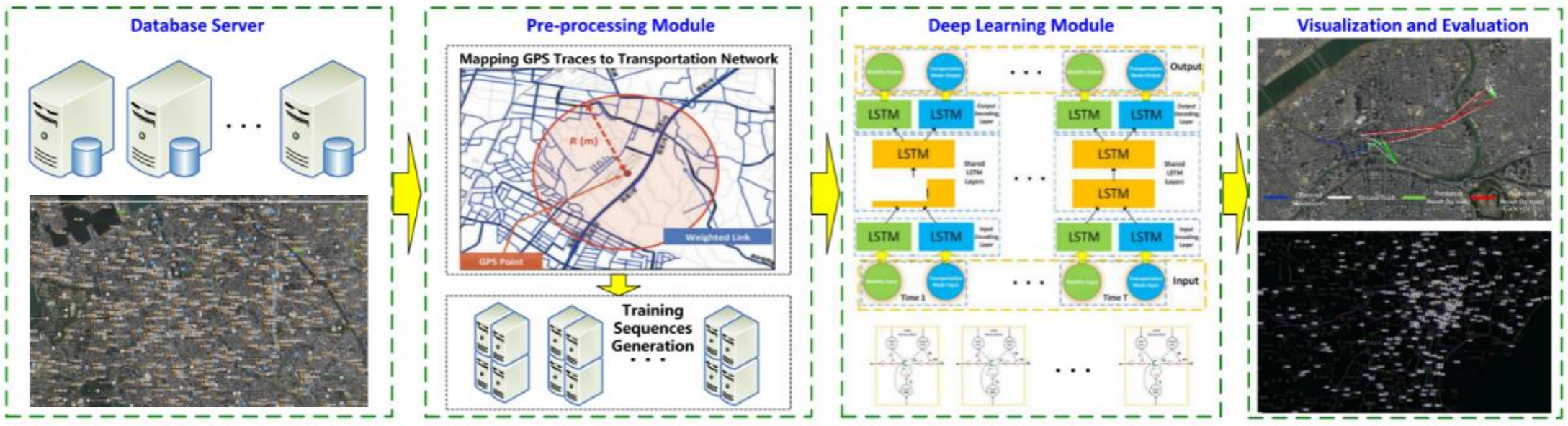

Fig. 3-12 Framework for forecasting and simulating resident mobility and traffic patterns

(3) Deployment Planning of Fire Facilities Based on Network Search and Spatial Optimization

For decades, the efficiency of public investment and services has been a focal point of interest among geographical researchers. In the private sector, inefficiency often leads to price increases, loss of competitiveness, and operational setbacks. Conversely, inefficiencies in public service provision may not immediately precipitate changes. Spatial data intelligence employing large-scale models integrates network search, GIS spatial analysis, and spatial optimization methods to assess the spatial efficiency of fire services at the urban scale. The foundation model initiates a network search process to delineate the current deployment patterns of fire stations in major urban areas, comparing search results with existing databases. Utilizing spatial optimization, the model estimates deployment levels necessary to achieve ideal coverage standards, subsequently evaluating this ideal against the current system as an approach to gauge spatial efficiency. GIS simulates demand locations throughout the document, conducting location-based spatial analysis, visualizing fire station data, and mapping model simulation outcomes.

Fig. 3-13 illustrates the modular design of the network search tool primarily designed for extensive web crawling to identify the presence and locations of fire stations. Given the vastness of the internet, it is imperative to pinpoint segments likely to contain addresses of fire stations, aiming to constrain the search scope and prevent aimless crawling that would squander resources and time. Thus, the initial step

involves determining where web crawlers should commence accessing web pages and from which specific sites.

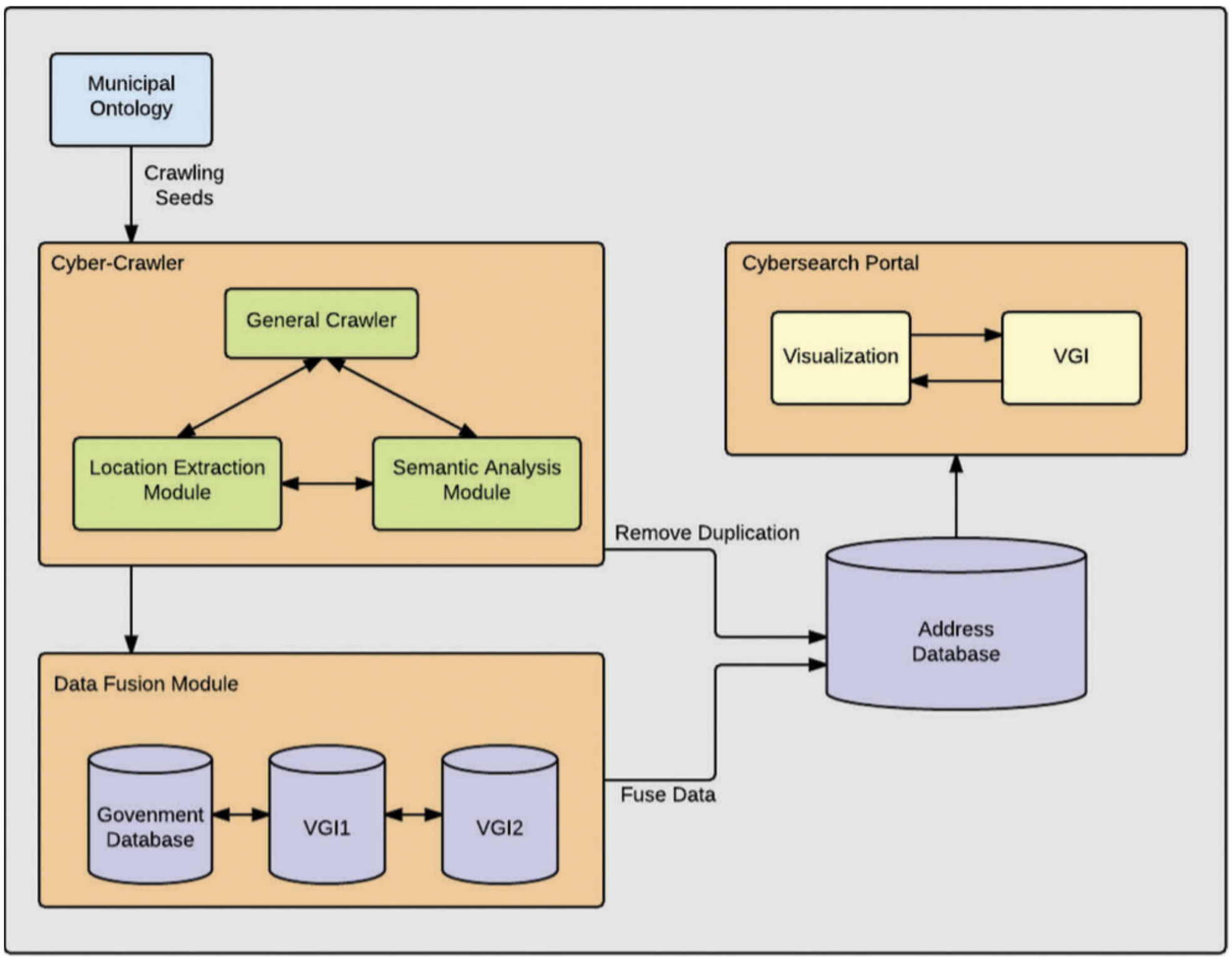

Fig. 3-13 Public service facility network search tool

The structure of LSCP followed by the large-scale model fire facility deployment planning is as follows:

$$Minimize\ Z = \sum_{j \in J} x_j$$

Limited to:

$$\sum_{j \in N_i} x_j \geq 1 \text{ for each } i \in I$$

$$x_j \in \{0,1\} \text{ for each } j \in J$$

In the above formula: i signifies the index representing a specific demand, where the entirety of demands is encompassed within the set I. Similarly, j denotes the index representing a potential site location, with the complete set of all possible sites defined as J. The notation $d_{ij}$ signifies the distance or travel time between a given demand i and a potential site j. The variable s stands for the maximum service distance or time standard applicable in this scenario. $N_i = \{j \mid d_{ij} \leq s\}$, a collection of sites j that can provide coverage for requirement i.

When site j is selected for site layout, $x_j = 1$, and otherwise $x_j = 0$.

### 3.3.3 High-Performance Processing of Big Data

Geospatial big data covers various geographic scopes and rich information content. The data volume is typically substantial and originated from a variety of sources, including satellite remote sensing data, geographic information system (GIS) data, sensor data, etc. with complex and diverse data types and formats. Additionally, geographic data exhibit clear spatial and temporal correlations with relationships based on spatial locations and time. Some geographic data require real-time collection and processing to support immediate decision-making and applications. Furthermore, geospatial data from different sources and formats need to be processed for data fusion and integration. Consequently, geospatial big data is an important target for spatial data intelligence and spatial intelligence computing with foundation models.

(1) Big Data Stream Processing

With the development of GIS technology, the data sources for GIS systems have changed dramatically. In the past, data mainly came from traditional map digitization and measurements collected through devices such as plane workbenches and total stations. The common data format was static vector maps, which lacked update accuracy and versatility. New surveying methods widely employ photogrammetry to collect raw data. The main data sources now include images, videos, radar, and GPS data generated by satellites, aircraft, drones, and survey vehicles. Advanced equipment such as panoramic cameras, street view cameras, observation satellites, and LiDAR systems can acquire comprehensive images and spatial information. Some of these devices support streaming services, allowing data to be dynamically transmitted to users. Today, traditional methods of static data storage, static mapping, and periodic data updates have diminished in importance. This shift has also led to significant changes in how data is stored, processed, analyzed, and utilized.

The Spatial Data Intelligence Foundation model can generate, process, and use real-time data through streaming. Due to the change in data types and the increase in processed data volume, the GIS system structure has been constantly evolving to adapt to this revolution. Currently, there are several streaming practices for foundation models. With distributed computing as the architecture, Spark Stream as the framework for streaming data and integration of message-oriented middleware such as Kafka, which combines message reception, processing, efficient data storage, and real-time data. Time-space analysis as a spatio-temporal integrated software platform to meet the needs of LiveGIS. There have been many successful solutions applied to e-commerce, social media, logistics, and transportation industries. For example, the latest SuperMap GIS platform integrates this system solution with advanced GIS functions, enabling streaming data to take advantage of GIS spatial analysis and visualization capabilities. The platform greatly enriches the capabilities and uses of traditional GIS systems.

Back-end processing capabilities and the flexibility of mobile applications can provide a reliable platform for IoT and applications. Smart devices handle their spatio-temporal data (Figure 3-14). This setup can not only scale with the growth of the business but also migrate quickly between different environments. In summary, it has become the core foundation for the development and operation of smart cities.

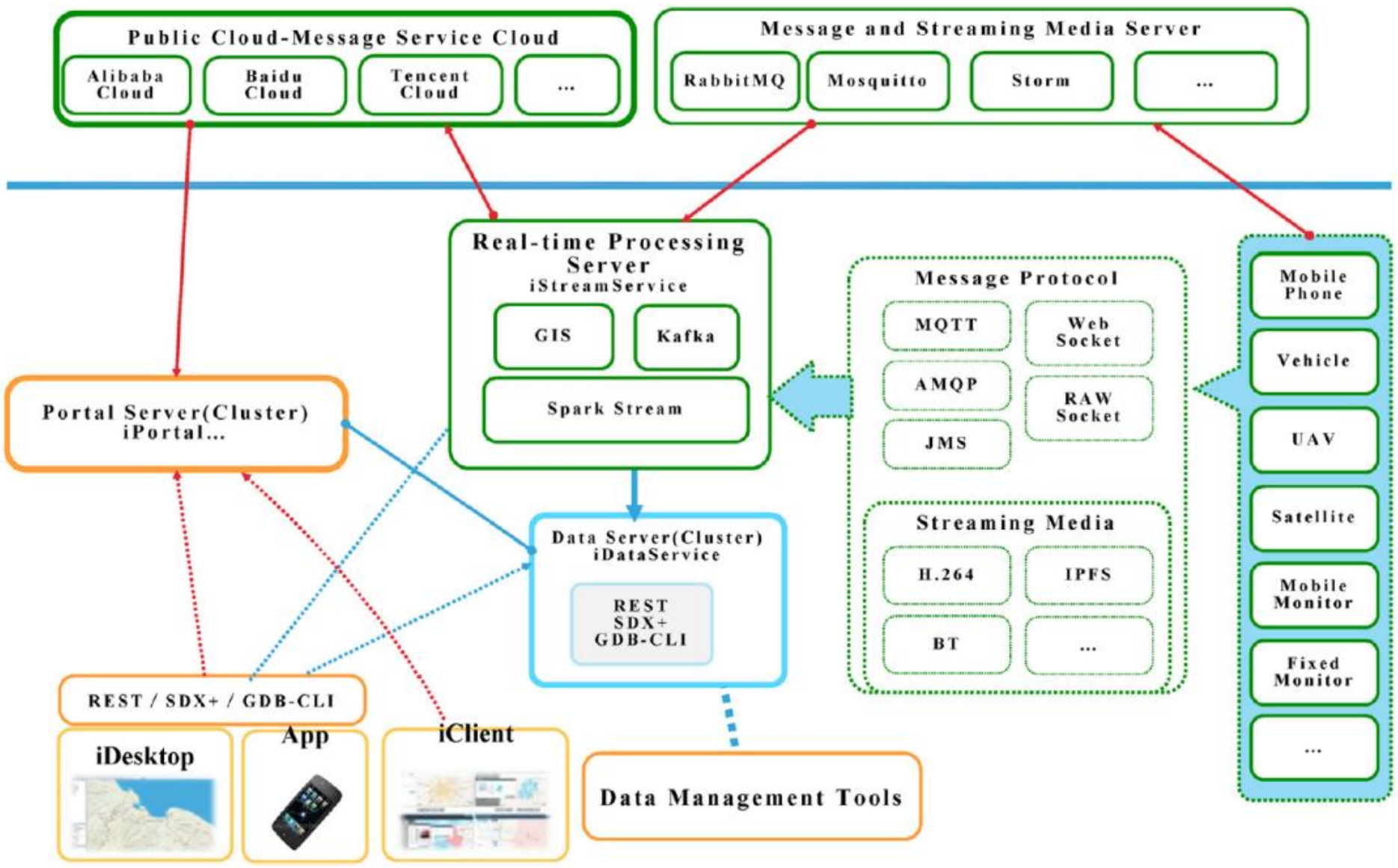

Fig. 3-14 The flow and structure of streaming data processing

(2) A Scalable Framework for Social Media Big Data Analysis

In recent years, social media (e.g., X and Facebook) have grown dramatically in popularity and have become ubiquitous for discourse, content sharing, and social networking. With the widespread adoption of mobile devices and location-based services, social media often allow users to share the whereabouts of daily activities (e.g., check-ins and taking photos), enhancing its role as an agent for understanding human behavior and the complex social dynamics within geographic spaces. Unlike traditional spatio-temporal data, this new form of data is dynamic, massive, and often streamed in unstructured media (e.g., text and photographs), which poses fundamental representational, modeling, and computational challenges for traditional spatio-temporal analyses and geographic information science. The Spatial Data Intelligence Foundation model builds a scalable computational framework to efficiently and systematically analyze spatio-temporal data using massive amounts of location-based social media data. Within this framework, the concept of spatio-temporal trajectories (or paths) is applied to represent the activity profiles of social media users. Based on the aggregation of spatio-temporal trajectories, the Big

Model designs a hierarchical spatio-temporal data algorithm, namely a spatio-temporal data cube model, to represent the collective dynamics of social media users across aggregation boundaries at multiple spatio-temporal scales. The framework is implemented based on the public data streams released by social media X. To demonstrate the advantages and performance of the framework, an interactive stream mapping interface (including single-source and multi-source stream mapping) is developed to allow real-time and interactive visual exploration of movement dynamics in massive location-based social media data at multiple scales.

Figure 3-15 shows the system architecture of the framework and the data flow through the different components. The first step is to retrieve data from X. While millions of social media users are generating a large amount of social media content. As hosts of these data, social media services usually restrict direct or complete access to these contents. In particular, X provides multiple levels of interfaces to access its feed corpus. The X Streaming API enables anyone to retrieve a 1% sample of all data in near real-time by specifying a set of filters (e.g., geographic boundaries of interest). A tweet crawler algorithm is developed based on the X streaming API to collect the posted tweets. The returned tweets are organized into a set of tuples $(u, s, t, m)$. In the second step, text mining methods are applied to unstructured text messages (m) by monitoring a keyword dictionary associated with symptoms of Influenza-Like Illness (ILI), such as "influenza", "cough", "sneeze", and "fever" to diagnose the probability of user X being infected with ILI. It should be noted that, depending on the application scenario, other data mining methods can be incorporated into this step to extract relevant information of interest from each tweet.

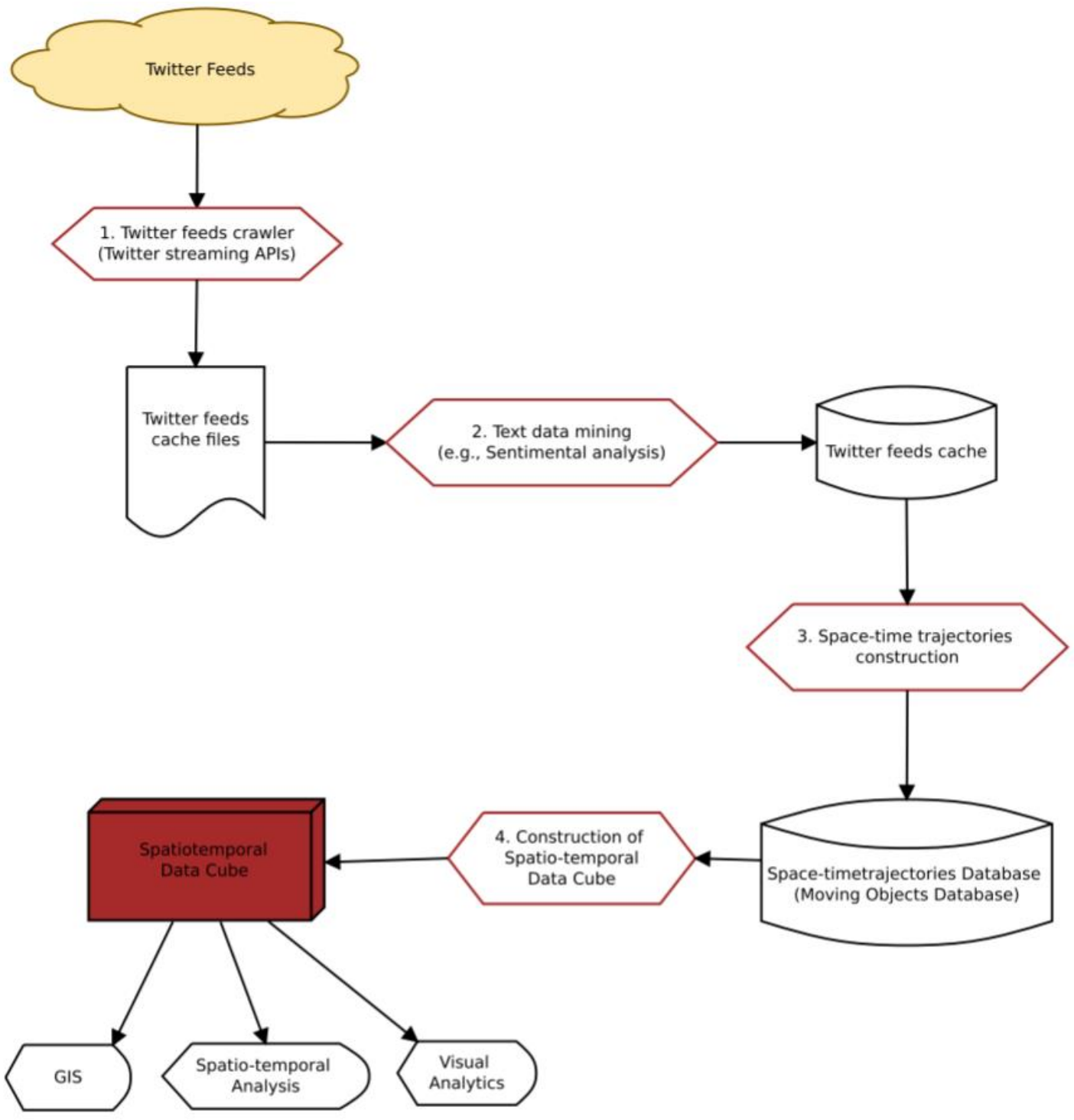

Fig. 3-15 The algorithm framework

(3) Convergence of Big Data and Machine Learning

As a new fuel for geospatial research, the Spatial Data Intelligence Foundation model leverages the latest breakthroughs in machine learning and advanced computing to enable scalable processing and intelligent analysis of geospatial big data. The Big Model of Spatial Data Intelligence sits at the intersection of Artificial Intelligence, Geospatial Big Data, and High-Performance Computing (HPC), providing a promising solution technology for data or computation-intensive geospatial problems. Figure 3-16 shows the Foundation model of Spatial Data Intelligence as a conceptual three-pillar view of GeoAI. As an interdisciplinary extension of artificial intelligence, the goal of the GeoAI Big Model is to give machines the intelligence to reason and analyze spatially as humans do. GeoAI Foundation models have evolved along with AI, which has two main categories of approaches: one is knowledge-driven, known as top-down approaches, and the other is data-driven, known as bottom-up approaches. There is no doubt that the data-driven approach led by machine learning has become the dominant AI today. The reason for this is its superior learning ability to make predictions from large amounts of data without the need to explicitly program analytics rules. Deep learning, the latest breakthrough in machine learning, has changed the data analysis paradigm in two ways.

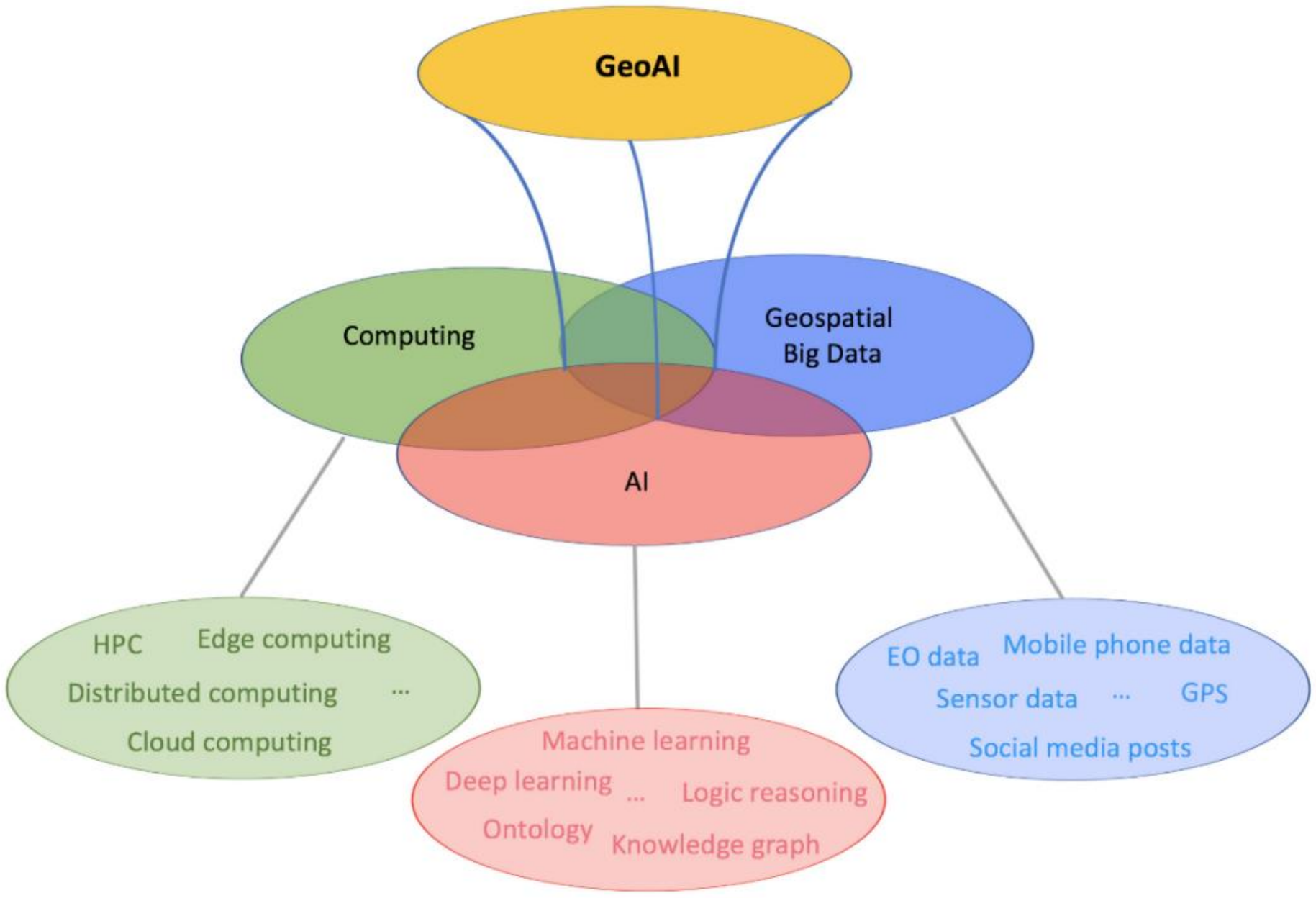

Fig. 3-16 Conceptual three-pillar view of the GeoAI foundation model

Machine learning has also powered more traditional, top-down, ontology-based approaches to GeoAI Foundation models. These approaches address spatial cognition problems by leveraging ontologies and logical reasoning, such as semantic similarity measures. Unlike data-driven approaches, ontology approaches rely on a knowledge base that provides semantic definitions of real-world entities and their relationships in the format of <subject, predicate, object> triples. The knowledge discovery process follows predefined reasoning rules and constraints and uses deductive reasoning to ensure that each newly derived fact can be formally verified to have a clear and traceable reasoning path. Though this approach is highly interpretable, it has two drawbacks: (1) Ontology engineering, the process of building a knowledge base, relies heavily or even entirely on expert knowledge and manual work. While it is possible to establish a deep structure to describe the complex relationships between entities, human-centered approaches are difficult to scale to make the knowledge base comprehensive enough to ensure performance; (2) while ontologies attempt to capture the complexity of human logic, it needs to be implemented in a machine-understandable way, so some simplification and abstraction are inevitable. This adds another layer of performance challenges in making accurate predictions and decisions.

Both methodological threads of GeoAI have broad applications in the geospatial domain. The remote sensing community widely uses Convolutional Neural Networks (CNN) for scene classification (both natural and urban), change detection, and other image analysis tasks. Deep learning has been employed to support mapping tasks such as integrated, intelligent mapping, and map element inspection. Machine learning is increasingly used for semantic and sentiment analysis of social media data and other natural language text documents. In spatial information retrieval, knowledge graphs have become key components and backbone technologies for intelligent question answering, hidden link prediction, and semantic search [11]. Multidimensional geospatial data, such as LiDAR and scientific data from numerical simulation models, can also benefit from processing capabilities like 3D CNN for 3D object detection and event classification. Time series data transmitted from Internet of Things (IoT) sensors can leverage Recurrent Neural Networks (RNN) for real-time prediction and analysis. The diversity of geospatial data and the prevalence of location-based services make GIScience a natural home for these applications and the thriving field of artificial intelligence.

### 3.3.4 Geographical Knowledge Graph

Geographical knowledge graph (GKG) is a form of knowledge representation based on geospatial information, which structurally represents and organizes geographic entities and their attributes, relationships, events and other information. It can help people better understand and use geographic information. Also, it supports the research and application of geographic information system, geographic data mining, geographic intelligence and other fields. The spatial data intelligent foundation model is highly complementary to the geographic knowledge graph, especially in the data processing requirements of processing natural language. The advantage of geographic knowledge graph is that it is a structured way to store and express knowledge, storing many facts in the form of triples. At the same time, GKG can also evolve with the increase of new knowledge. By constructing the knowledge graph of professional knowledge in the expert field, we can add, delete, modify, and check the fact knowledge in the professional field.

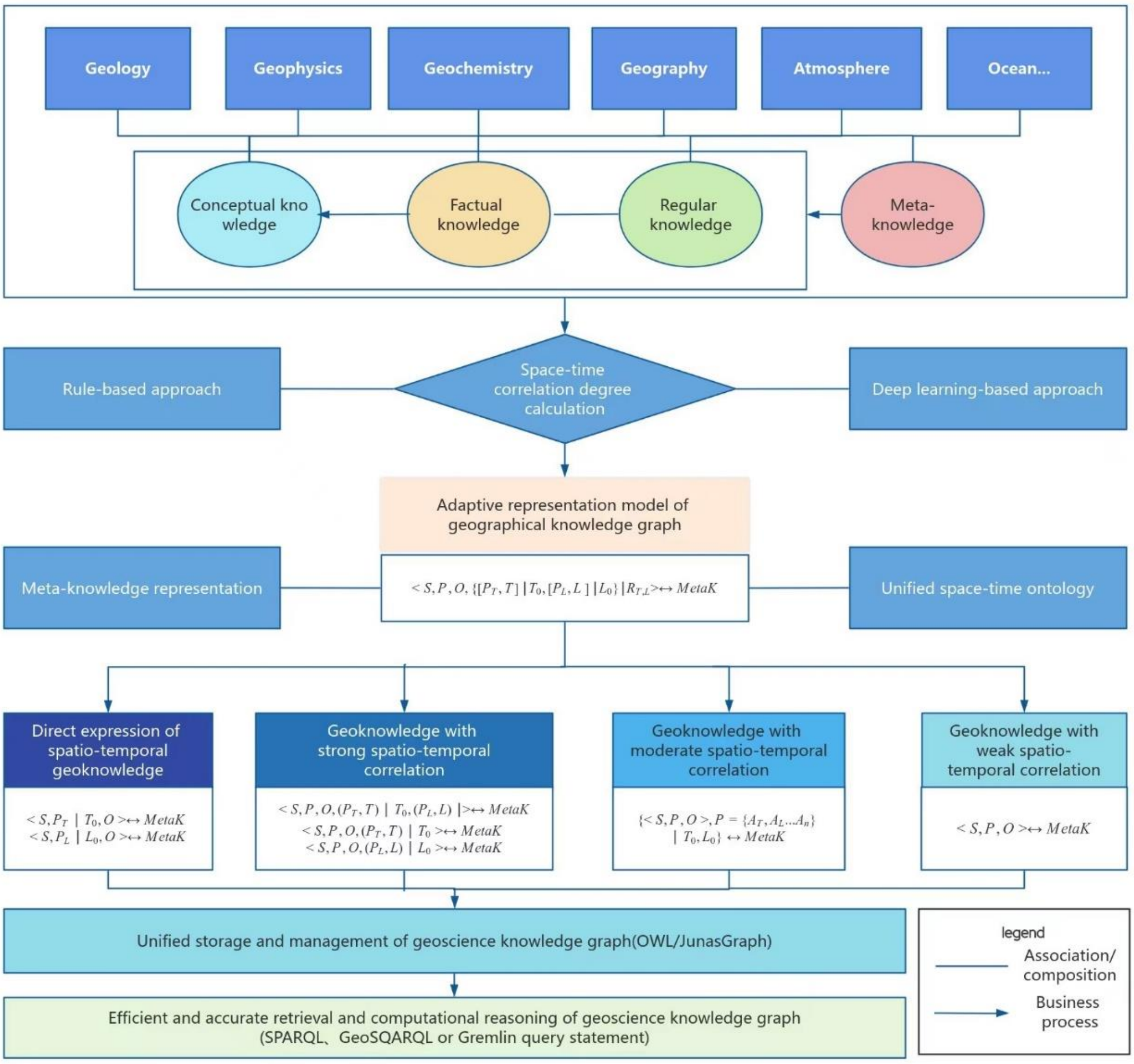

Fig. 3-17 Application process of adaptive expression model of GKG

(1) Adaptive Representation Model of GKG

By organizing all kinds of geoscience knowledge into a semantic network that can be understood and calculated by computers, GKG can realize unified cognition, accurate correlation, computational reasoning and intelligent service of geoscience knowledge, which is the most effective way of geoscience knowledge organization and service at present. It has become the foundation of modern geoscience research based on big data and artificial intelligence. Also, it is becoming the frontier and hot spot of geoscience research. Geoscience knowledge contains much subject domain knowledge, has complex spatial-temporal features and relations, and presents the characteristics of multi-scale, multi-granularity, and multi-dimension. Therefore, it is the basis and prerequisite for the construction and application of GKG to establish the representation model of GKG, which is in line with the characteristics of geographical knowledge and takes into account the complex spatial-temporal features and relations for different disciplines and types of geographical knowledge. The application flow of the adaptive representation model of GKG is shown in Figure 3-17. Firstly, the spatio-temporal correlation degree of the multi-subject and multi-type geoscience knowledge to be expressed is calculated. The spatio-temporal correlation degree can be calculated by the rule-based method or deep learning model mentioned above. According to the spatio-temporal correlation degree, which includes geographical knowledge that directly expresses spatio-temporal information, as well as the geographical knowledge that is strongly, moderately, and weakly correlated with spatio-temporal features, the more compact and accurate representation model matching the spatio-temporal correlation degree is automatically selected based on the adaptive representation model of GKG. These expression models have not only the expression of common topic content tuples and meta-knowledge tuples and the support of unified space-time ontology, but also the expression of spatio-temporal information according to the spatio-temporal correlation degree. Unified description Language and graph database, such as Web Ontology Language (OWL) and JanusGraph graph database, can be used for unified storage and management of GKG. The adaptive representation model can flexibly express geoscience knowledge into triples or quadruples and quintuples closely related to spatio-temporal information according to the different degree of spatio-temporal correlation. Therefore, query languages such as SPARQL (SPARQL Protocol and RDF QueryLanguage), GeoSPARQL or Gremlin can be used to achieve more efficient and accurate retrieval and computational reasoning of geoscience knowledge.

(2) Spatial Explicit Reinforcement Learning Model for Automatic Summarization of GKG

Network scale knowledge graphs, such as the global Linked data cloud, consisting of billions of individual statements about millions of entities, have in recent years sparked interest in the knowledge graph summary technique, which computes representative subgraphs for a given set of nodes. Moreover, many of the most densely connected entities in the knowledge graph are places and regions, which are often represented by thousands of afferent and efferent relationships with other places, actors, events, and objects. In this paper, we propose a new summarization approach that incorporates spatial explicit components into reinforcement learning frameworks to help summarize GKG, which is a topic that has not been considered in related work. Our model considers the internal graph structure as well as external information to obtain a more comprehensive and holistic view of the summary task. By collecting standard data sets and evaluating our proposed models, we prove that spatial explicit models produce better results than non-spatial models, thereby proving that space is indeed special in terms of generalization.

For the reinforcement learning algorithm practice based on GKG in the spatial data intelligent foundation model, Wikipedia abstract is first used to guide the process of GKG summarization using reinforcement learning. The method does not mainly rely on internal information, such as the node group in the grouping and aggregation method and the number of bits required to describe the graph in the bit compression method. However, using Wikipedia abstract to derive the complementary advantages of intrinsic information from graph structure and external knowledge. By framing tasks as sequential decision-making processes, it can be optimized using reinforcement learning. Secondly, the richness of geospatial semantics in geographic knowledge maps is considered. Also, this information is incorporated into the summary process to better capture the relevance of geographic entities and provide better results. The foundation model does this by following an established approach to geographic information, which is to model distance decay from an information theoretic perspective. Third, create a dataset DBP 369 that includes 369 site summaries from Wikipedia and a subgraph from DBpedia for the GKG summary task and make it publicly available. The lack of standard data sets has been one of the obstacles to the development of research in the field of GKG summarization and geographic information retrieval. Fourth, the foundation model establishes different baselines for the GKG summarization task of the DBP369 dataset. The validation results show that the summary graph considering spatial context components better resembles the Wikipedia summary. It is necessary to consider the GKG summarization problem in the spatial data intelligent foundation model, mainly because network-scale knowledge graphs, such as associated data, store tens of millions of locations, often with thousands of related statements (subject-predicate-object triples).

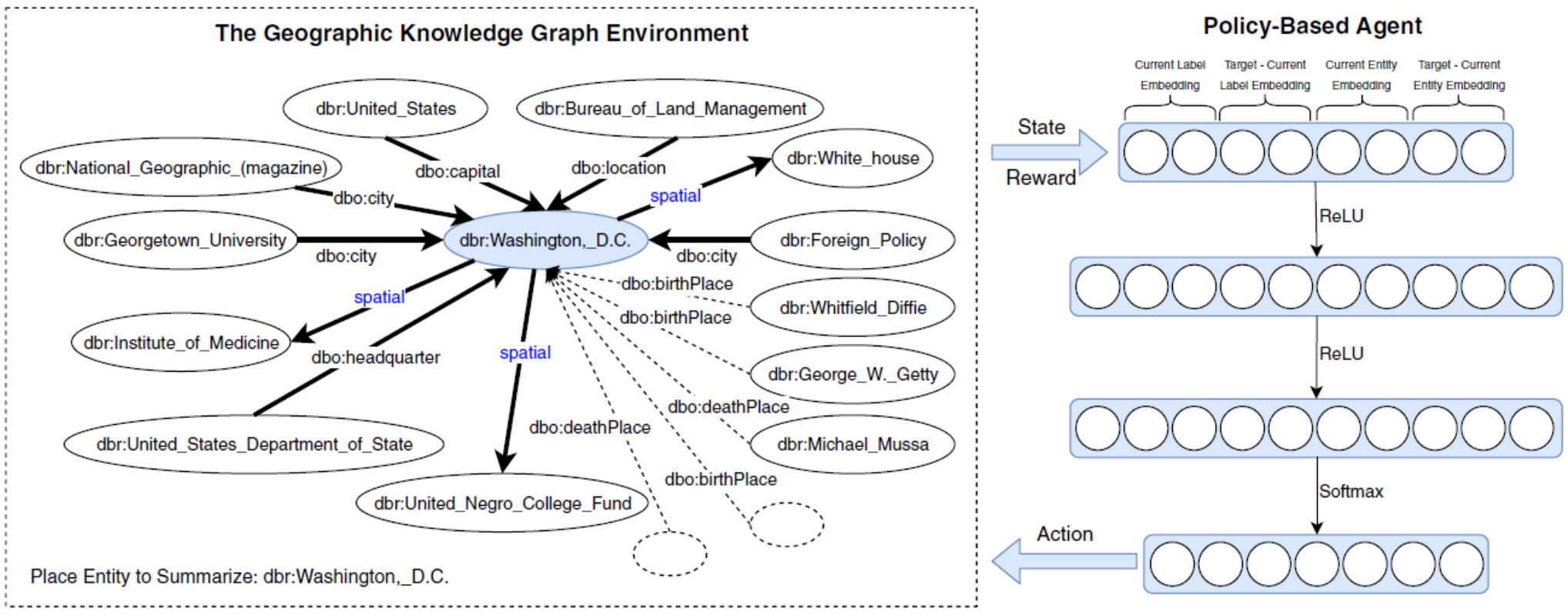

Fig. 3-18 GKG environments and policy-based agents interact in reinforcement learning models

(3) Knowledge Embedding with Geospatial Distance Restriction for GKG Completion

GKG uses the semantics of geographic entities and geographic relations to connect the triples of geographic relations into a large-scale semantic network. However, in the case of sparse distribution of geographically related information on the Web, it is difficult for information extraction system to detect enough geographic information in massive Web resources to establish a relatively complete GKG reference. Due to the absence of geographic entities or geographic relationships in the GKG fact triples, this incompleteness severely affects the performance of GKG applications. The spatial data intelligent foundation model designs a GKG completion knowledge embedding optimization method based on geographic spatial distance restriction, which encodes the semantic information and the geographic spatial distance constraints of geographic entities and geographic relations into a continuous low-dimensional vector space. Then the missing facts of GKG can be supplemented by vector operations. Specifically, the geospatial distance restriction is realized as the weights of the objective functions of current translation knowledge embedding models. These optimized models output optimized representations of geographic entities and geographic relationships to complete the GKG. A real GKG example is used to verify the effectiveness of the proposed method. Compared with the results of the original model, the proposed method has an average improvement of 6.41% in Hits@10 (Filter) for geographic entity prediction and 31.92% in Hits@1 (Filter) for geographic relationship prediction. In addition, the ability of the method to predict the location of unknown entities is also validated. The results show that the restriction of geographical spatial distance reduces the mean error distance of the prediction between 54.43% and 57.24%. All the results support that the geospatial distance restriction hidden in the GKG help to refine the embedding representation of geographic entities and geographic relationships, which plays a crucial role in improving the quality of GKG completion.

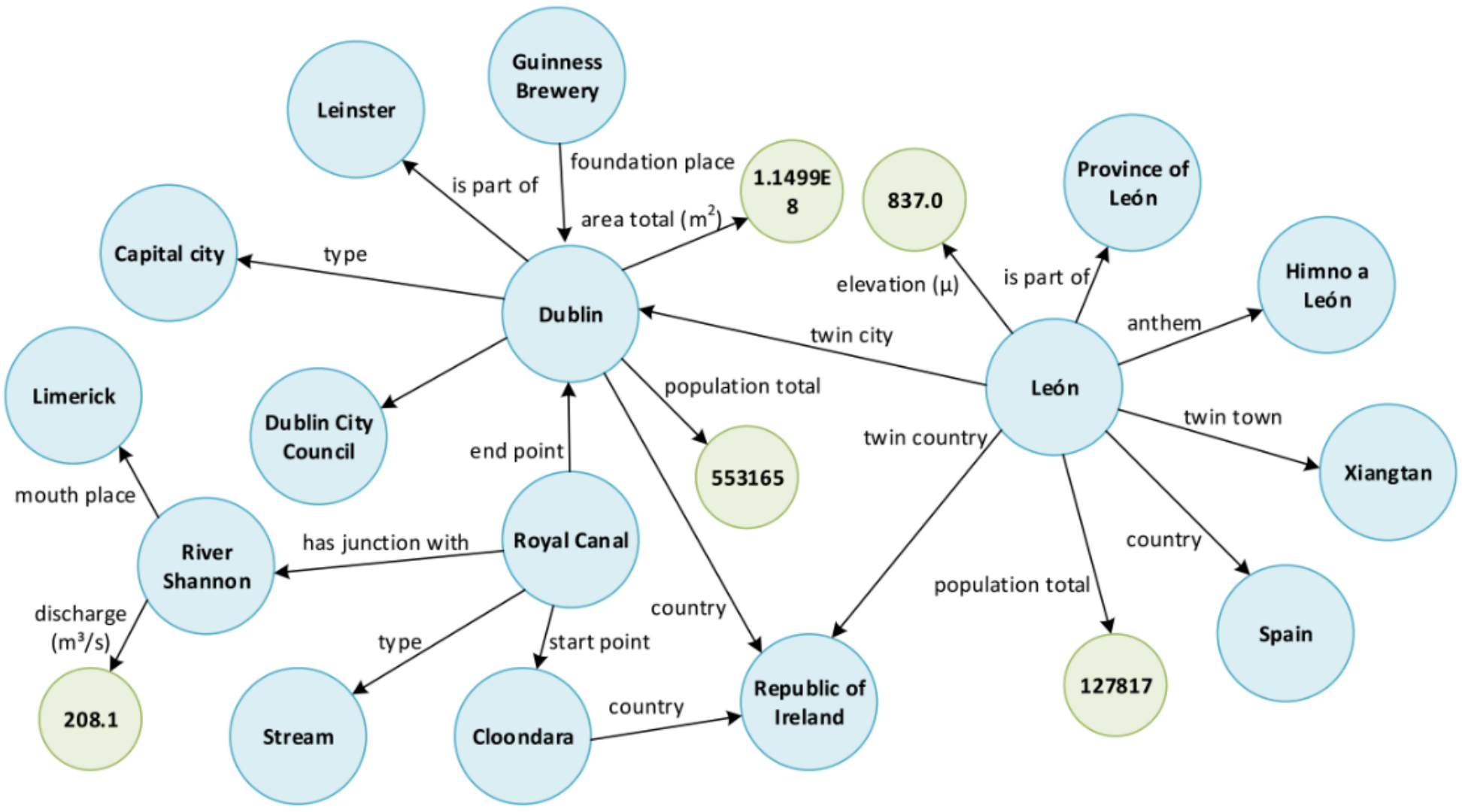

Fig. 3-19 Example of a GKG

## 3.4 Geographical Intelligent Multi-Scenario Simulation

### 3.4.1 Intelligent Land Use Simulation Based on Spatial Data

Multi-scenario land use simulation is one of the key applications of spatial data intelligent foundation models. Its primary purpose is to simulate and predict land use changes under various scenarios, helping decision-makers to formulate reasonable land use planning and management policies. The process begins by determining the simulation scenarios, which include different development strategies, policy measures, or changes in natural conditions that affect land use and land cover changes. Input data typically includes current land use and land cover data, land-use planning data, population data, economic data, etc. Based on the input natural language data and demand analysis, the spatial data intelligent foundation model selects the appropriate model for the given scenarios and objectives. Commonly used models include cellular automata (CA), Markov chain models, and genetic algorithm (GA) models. After setting parameters such as the initial state, transition rules, and influencing factors, the simulation model is run to simulate and predict land use changes under different scenarios. The simulation results can reveal trends and the spatial distribution of land-use changes. Finally, the spatial data intelligent foundation model analyzes and evaluates these results. It compares land use changes across different scenarios and then assessing their impact on land use, providing valuable insights for decision-making.

### 3.4.2 Intelligent Traffic Simulation Based on Spatial Data

Multi-scenario intelligent traffic simulation, as a pivotal application of spatial data intelligent foundation models, aims to simulate and assess traffic phenomena like traffic flow and congestion across various traffic scenarios, aiding transportation planning and management decisions. Initially, the foundation model identifies simulation scenes including traffic network structures, traffic management measures, and traffic demand conditions, encompassing instances such as road construction plans, traffic control strategies, traffic accidents, or emergencies. Data necessary for simulation comprises road network data, traffic flow data, vehicle trajectory data, traffic rule data, etc. The foundation model then intelligently selects the appropriate traffic models for simulation based on input data, simulation complexity, and specific requirements. Commonly employed models include microscopic traffic simulation models, macroscopic simulation models, hybrid simulation models, etc. Subsequently, after configuring parameters for traffic flow, control, and demand across different scenarios, traffic flow and congestion are simulated and predicted to reflect real-world conditions. Finally, the spatial data intelligent foundation model analyzes and evaluates simulation results, comparing traffic conditions and impacts under various scenarios and assessing their effects on the transportation system.

The multi-scenario traffic simulation models and algorithms widely used in spatial data intelligent foundation models include: (1) Microscopic Simulation Models: These models, such as VISSIM and SUMA, simulate the trajectory of each vehicle based on vehicle behavior and traffic rules. They can model interactions between vehicles and capture the dynamic evolution of traffic congestion. (2) Macroscopic Simulation Models: Macroscopic simulation models, including TranSims and MatSim, divide the traffic network into a series of traffic zones and simulate the overall traffic flow for each zone. These models are suitable for large-scale traffic system simulations and can quickly evaluate the effectiveness of transportation planning schemes. (3) Traffic Demand Models: Traffic demand models, including the four-step model, behavioral modeling models, and the taxi repositioning framework based on multi-agent reinforcement learning, are used to estimate traffic demand under different scenarios, including traffic flow and mode choice. These models can analyze factors influencing travel behavior and mode choice, providing data support for transportation planning. (4) Traffic Control Models: Traffic control models, including signal optimization models and traffic control models, are used to evaluate the impact of different traffic control strategies on the transportation system. These models can simulate various traffic control schemes to assess the effectiveness of congestion alleviation and the operational efficiency of the transportation system. (5) Machine Learning and Deep Learning Methods: Machine learning and deep learning methods can be used to optimize parameters in traffic estimation models, predict traffic flow, and analyze traffic congestion. For example, recurrent neural network (RNN), long short-term memory network (LSTM), and reinforcement learning algorithms can be utilized to predict traffic flow and optimize traffic control strategies.

Recently, researchers have applied deep learning techniques like convolutional neural networks (CNNs) to model spatio-temporal data, achieving superior results compared to traditional methods. However, the grid map representation used by these CNN-based models is not well-suited for road-network-based data. To overcome this limitation, a deep spatio-temporal residual neural network (DSTR-RNet) has been developed specifically for road-network-based data modeling. This model introduces locally connected neural network layers (LCNR) to accurately represent the topology of road networks and incorporates residual learning to capture spatio-temporal dependencies. The DSTR-RNet was evaluated using traffic flow data from the Didi cab service, and the results indicated that it preserves the spatial precision and topology of the road network while also enhancing prediction accuracy.

The DSTR-RNet, built on the ResLCNR unit (Figure 3-20), is designed to jointly model spatial and temporal dependencies. It consists of three sub-models, each dedicated to capturing spatio-temporal features from different patterns: recent, daily, and weekly. These features are then combined into a final feature map, which is activated using a *tanh* function to predict values. Each sub-model shares the same structure: (1) an LCNR layer that processes historical road network data and outputs a feature map with elements corresponding to road network segments, and (2) a deep residual LCNR structure composed of multiple ResLCNR units that models the spatio-temporal dependencies within the feature map. By integrating spatial and temporal features, the DSTR-RNet effectively captures correlations across both dimensions. The final feature maps, labeled STFMw, STFMd, and STFMr, are merged using a parameter-based method, defined by the following equation:

$$STFM = STFM_w \circ W_w + STFM_d \circ W_d + STFM_r \circ W_r$$

$$x_t = \tanh(STFM)$$

where $W_w$, $W_d$, and $W_r$ are three parameter vectors with shapes identical to those of the three feature maps. STFM is the final spatio-temporal feature map. Then, a *tanh* function activates the STFM to form the prediction values, $x_t$.

$$loss = MSE = \frac{1}{N}\sum_{i=1}^{N}(y_i - y_i')^2$$

DSTR-RNet computes the loss by comparing ground truth values with predictions, using mean-square error (MSE) as the loss function. In this context, $y_i$ represents the ground truth, $y_i'$ denotes the predicted value, and NNN is the total number of predictions. The input data are split into three subsets: a training set, a validation set, and a test set. The model processes the training set in batches, calculating the loss after each forward propagation. This loss is then minimized through back-propagation using the Adam optimizer, which adjusts the model parameters accordingly. This process continues until the training parameters are optimized and the loss function is minimized.

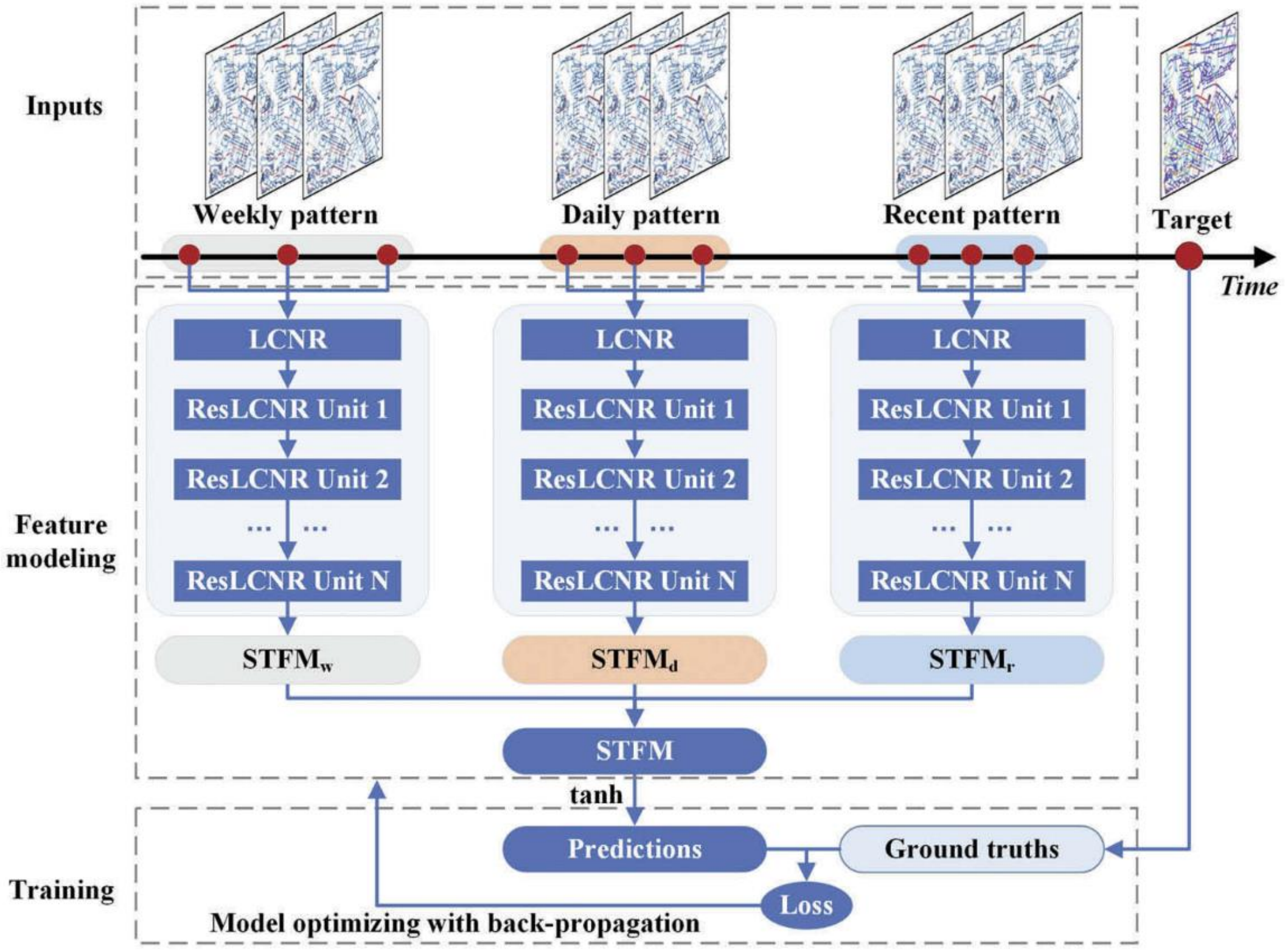

Fig. 3-20 The overall framework of DSTR-RNet

### 3.4.3 Intelligent Public Service Facilities Location Optimization Based on Spatial Data

Intelligent Public Service Facility Site Selection Simulation is a crucial application within spatial data intelligent foundation models. Its purpose is to simulate and evaluate the impact of various site selection schemes on the coverage and service quality of public service facilities, thereby supporting the optimization of their layout and planning. The foundation model collects and prepares the necessary data for simulation, including the current distribution of public service facilities, population distribution, transportation networks, and land use data. These data inputs enable the model to assess the demand for public service facilities, considering population needs, service area requirements, and service quality demands. Data mining and statistical analysis methods are employed to evaluate and forecast these demands. By integrating geographical data with the demand for public service facilities, the foundation model predicts the impact of different site selection schemes using planning algorithms and machine learning methods. Parameters such as site selection rules, weights of influencing factors, and constraints are set by scenarios and objectives. The model then runs based on different site selection schemes, reflecting their impact on the coverage and service quality of public service facilities. The foundation model analyzes and evaluates the simulation results, comparing the merits of different site selection schemes using metrics like coverage area, service quality, and cost-effectiveness. Based on these results, plans for optimizing the site selection of public service facilities are formulated, either by adjusting site selection schemes or improving the layout of service facilities.

The intelligent public service facility site selection employs a variety of algorithms and methods, including:

(1) Programming Algorithm-based Site Selection Models: Models like linear programming and integer programming determine optimal site selection by setting rules and constraints. These algorithms consider factors such as population distribution, transportation networks, and land use to maximize the coverage and service quality of public service facilities.

(2) Optimization Algorithm-Based Site Selection Models: Optimization algorithms, including GA, ant colony optimization (ACO), and simulated annealing (SA) algorithms, determine the best site selection by optimizing the objective function of the site selection scheme while considering multiple goals and constraints.

(3) Machine Learning-Based Site Selection Models: Machine learning algorithms, such as decision trees, random forest (RF), and neural network (NN), predict the outcomes of different site selection schemes by learning from historical data. These algorithms generate predictive models based on data characteristics and requirements, aiding decision-makers in making informed site selection decisions.

(4) Spatial Analysis-Based Site Selection Models: Spatial analysis methods like spatial interpolation and spatial association analysis evaluate and optimize site selection schemes by considering the characteristics of geographic spatial data, helping to identify suitable locations for constructing public service facilities.

(5) Deep Learning-Based Site Selection Models: Deep learning algorithms, including convolutional neural network (CNN) and recurrent neural network (RNN), extract features and predict optimal site selection schemes by learning from large datasets. These algorithms model and solve complex site selection problems, enhancing the accuracy and efficiency of site selection simulations.

By leveraging the advantages of deep learning and reinforcement learning, spatial data intelligent foundation models can incorporate powerful deep reinforcement learning algorithms. This enables them to handle large-scale multimodal data more efficiently, thereby solving multi-scenario intelligent site selection problems for public service facilities. Designing a deep reinforcement learning model for site selection involves several components: state representation, action selection, and reward feedback. The state representation includes information such as geographical location, population density, transportation networks, and the scale and capacity of public service facilities. The action space comprises candidate site selection schemes, including constructing new facilities or expanding existing ones. The reward function evaluates each action based on indicators like coverage area, service quality, and cost-effectiveness. Common deep reinforcement learning models include Deep Q-Network (DQN), Double Deep Q-Network (DDQN), and Deep Deterministic Policy Gradient (DDPG). These models are trained using historical data to learn optimal site selection strategies. During training, they optimize their parameters through environmental interactions to maximize cumulative rewards. Once trained, these models can simulate real site selection problems, choosing actions based on the current state and updating strategies based on reward feedback. Simulation results are evaluated and optimized by comparing different strategies. The performance of site selection algorithms can be enhanced by adjusting the reward function, increasing state-space dimensionality, and improving the model structures. Overall, deep reinforcement learning-based algorithms for public service facility site selection have the following advantages: 1) They can handle complex site selection problems with multiple objectives and constraints; 2) They can learn from historical data and automatically adjust strategies based on environmental changes; 3) They are flexible and can adapt to different scenarios and objectives, demonstrating strong generalization capabilities.

3.4.4 Intelligent Natural Disaster Simulation Based on Spatial Data

Spatial data intelligent natural disaster simulation involves leveraging spatial data and intelligent algorithms to simulate and predict natural disasters, such as floods, earthquakes, and storms. This comprehensive process can evaluate the impact of these disasters on both humans and the environment, guiding the formulation and implementation of effective response measures. The simulation model gathers and collects essential spatial data, including topographic, meteorological, climate, hydrological, and water resources data, which serve as inputs for simulation. Depending on the type of natural disaster, the model selects suitable simulation methods and algorithms and sets the parameters, including topographic, meteorological, climate, hydrological, and water resource parameters. These parameters are adjustable to accommodate varying real-world conditions and requirements. The model then runs the disaster process simulation, mimicking the occurrence and evolution of the natural disaster. The simulation results can indicate the impact range, severity, and duration of the disaster under different conditions. Following the simulation, a rigorous risk assessment, employing methodologies like probabilistic analysis and risk assessment techniques, is conducted to gauge the potential consequences of the natural disaster on both human populations and the environment. Based on the risk assessment results, appropriate natural disaster response measures are formulated, including the construction of early warning systems, disaster prevention planning, and emergency response preparation.

For coastal flood inundation and shelter modeling in Shanghai, a simplified 2D hydraulic model called FloodMap-Inertial is employed to create coastal flood inundation maps. This model uses a computationally efficient inertial algorithm to solve the 2D shallow water equations in a raster-based environment, utilizing the Forward Courant-Friedrichs-Lewy Condition to determine the time step. The model has undergone calibration and validation in several coastal cities, including Shanghai and New York. Coastal inundation modeling requires boundary conditions and floodplain topography. Dynamic boundary conditions (spatial and temporal grids) are created for 100-, 200-, 500-, and 1,000-year flood return periods under current conditions by interpolating station-based water levels and scaling Typhoon Winnie's stage hydrographs. For future flood scenarios (2030 and 2050), these boundary conditions are adjusted for localized sea level rise (SLR) projections, including subsidence, under the RCP 8.5 scenario. A ‘bare earth’ digital elevation model (DEM) constructed from 0.5-meter topographic contours, with a grid cell resolution of 50 meters, is used for Shanghai. Given the uncertainty in improving flood defenses, it is assumed that the current seawalls and floodwalls in Shanghai will remain unchanged over the next few decades. Dike reliability functions are utilized to identify potential failure points along the coast and the Huangpu River. Any segments with potential breaches are removed, and the heights of the remaining flood defenses are overlaid onto the original DEM for each scenario. Additionally, an empirically derived floodplain roughness coefficient (Manning’s n = 0.06) is applied in the simulations to account for the influence of urban features on flow routing.

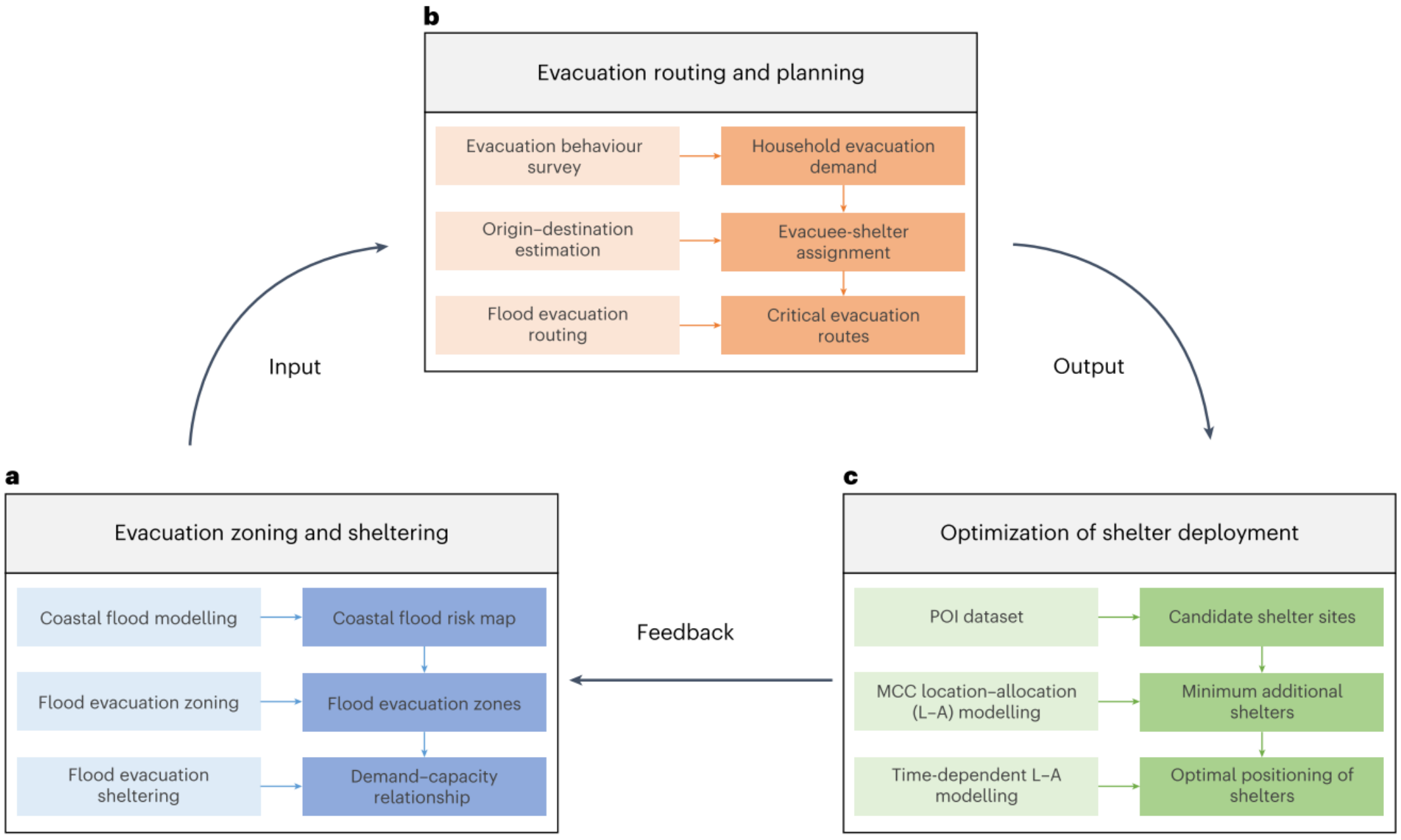

Fig. 3-21 Theoretical framework of rainstorm-flood strategic evacuation planning for effective population transfer in coastal megacities

# 4 The Application of Spatial Data Intelligence to Foundation Models

Spatial data intelligent foundation models exhibit superior performance, which is capable to integrate multiple spatial data sources. They can be trained on large-scale unlabeled data through self-supervised learning, thereby reducing dependency on labeled data and enhancing model efficiency. By leveraging advanced technologies such as deep learning, these intelligent models can accurately predict various phenomena and changes on the Earth's surface, showcasing their robust data processing capabilities. Furthermore, they are capable of effectively fusing diverse spatial data sources, presenting complex spatial data in an intuitive and visual manner, which facilitates a clearer understanding of the data.

Based on the characteristics and advantages of spatial data intelligent foundation models, these models have been widely utilized across various fields, including urban planning and construction, traffic management and optimization, environmental monitoring and protection, disaster risk assessment and response, smart agriculture and precision farming, resource management and conservation, military and national defense security, among others. Here are some of the main application scenarios.

## 4.1 Geospatial Foundation Models and Spatiotemporal Knowledge Graphs

### 4.1.1 JARVIS and Geo-JARVIS: A New Paradigm in GeoAI Based on LLM Agents

In the context of the rapid development of artificial intelligence (AI) technologies, natural language processing (NLP) techniques have also matured. As a result, JARVIS—a collaborative system that connects language models (LLMs) and AI models—has emerged. This system achieves more efficient and accurate natural language processing applications by tightly integrating language models with AI models.

The JARVIS collaborative system, which integrates LLM and AI models, holds significant value among various application scenarios. For instance, in the field of intelligent customer service, this system can provide efficient and accurate customer support for e-commerce platforms, banks, telecommunications companies, and other enterprises. In the realm of intelligent writing, this system can automatically generate text content such as news reports, scientific papers, and advertising copy. Furthermore, it can be employed in intelligent recommendation and intelligent search scenarios, thereby facilitating the expeditious identification of pertinent information by users.

The JARVIS collaborative system, which integrates LLM and AI models, operates as follows: First, the language model preprocesses natural language text to facilitate subsequent analysis and processing by the AI model. Then, the AI model utilizes the preprocessed text data to conduct various types of analysis and processing, such as sentiment analysis, topic classification, and entity recognition. Finally, based on the analysis results, the AI model can automatically generate corresponding text responses or enable other types of intelligent applications.

In comparison to traditional machine learning models, the JARVIS collaborative system, which integrates LLM and AI models, offers several advantages. Primarily, the system is better able to comprehend and leverage the meaning of natural language text, circumventing many limitations associated with traditional machine learning models when processing natural language. Secondly, the introduction of the incorporation of AI models enables the system to automate the learning process and adapt to new knowledge and language phenomena, eliminating the tedious process of manual parameter tuning and model adjustment that is required by traditional machine learning models. Furthermore, the JARVIS collaborative system can significantly enhance the accuracy and efficiency of natural language processing applications, thereby providing users with superior intelligent service experience.

The implementation of the JARVIS collaborative system, which integrates LLM and AI models, necessitates the mastery of advanced deep learning and NLP technologies in addition to extensive knowledge and experience in various business scenarios and applications. The implementation of this system involves the following steps: Firstly, the language model must be constructed and trained to achieve accurate understanding and processing of natural language text. Secondly, the AI model must be designed and trained to enable various types of natural language processing applications. Finally, the language model and the AI model must be tightly integrated to form an efficient collaborative system, facilitating its use in various practical application scenarios.

As a result of the ongoing advancement of technology, the JARVIS collaborative system, which integrates LLM and AI models, will find applications in an increasing number of fields, thereby driving the ongoing progress of natural language processing technology. Earth science research and geographic knowledge discovery are highly complex and multidimensional tasks that require the handling of large amounts of data and the extraction of statistical information and knowledge to address complex problems.

LLMs encode extensive human language, bringing powerful task understanding and reasoning capabilities and making automated earth science research and geographic knowledge discovery through LLMs possibility. Thus, a new architecture called Geo-JARVIS is proposed. The objective of Geo-JARVIS is to introduce a new tool and paradigm in the form of an AIAgent, which will enable the automatic acquisition of geographic data, the automatic processing of geographic data, and the automatic discovery of geographic knowledge. Geo-JARVIS is an intelligent agent that is designed to be human-like and to exhibit the following characteristics: understandability, memorability, plannability, and evolvability. It is comprised of four fundamental spaces that provide the necessary support for Geo-JARVIS: the instruction space, the task space, the model space, and the data space. Additionally, it includes a behavior space that supports the following functions: task decomposition, task modeling, task planning, task calibration, and task integration.

4.1.2 Prithvi: Geospatial AI Foundational Model

Prithvi, which is based on IBM's watsonx, is trained using NASA's Harmonized Landsat Sentinel-2 (HLS) satellite data and fine-tuned with flood and fire scar data. The objective of this initiative is to transform satellite data into high-resolution maps that display changes in floods, fires, and other geographic scenarios. This process aims to reveal environmental developments and prevent potential disasters. The Prithvi model comprises four main modules: Prithvi-100M (the base model), Prithvi-100M-sen1floods11 (the flood mapping model), Prithvi-100M-multi-temporal-crop-classification (the crop and land identification model), and Prithvi-100M-burn-scar (the fire scar identification model). This model will become the largest geospatial foundational model on Hugging Face and the first open-source AI foundational model built in collaboration between IBM and NASA. At present, four single-function demonstrations are available on Hugging Face: multi-temporal image completion, flood detection, fire scar detection, and multi-temporal land feature classification. It is currently unable to support the use of multiple models or data overlays. In order to utilize the model, it is necessary to provide geotiff images from HLS, which should include six bands. The following spectral bands are required: blue, green, red, narrow near-infrared (NIR), shortwave infrared (SWIR), and SWIR 2.

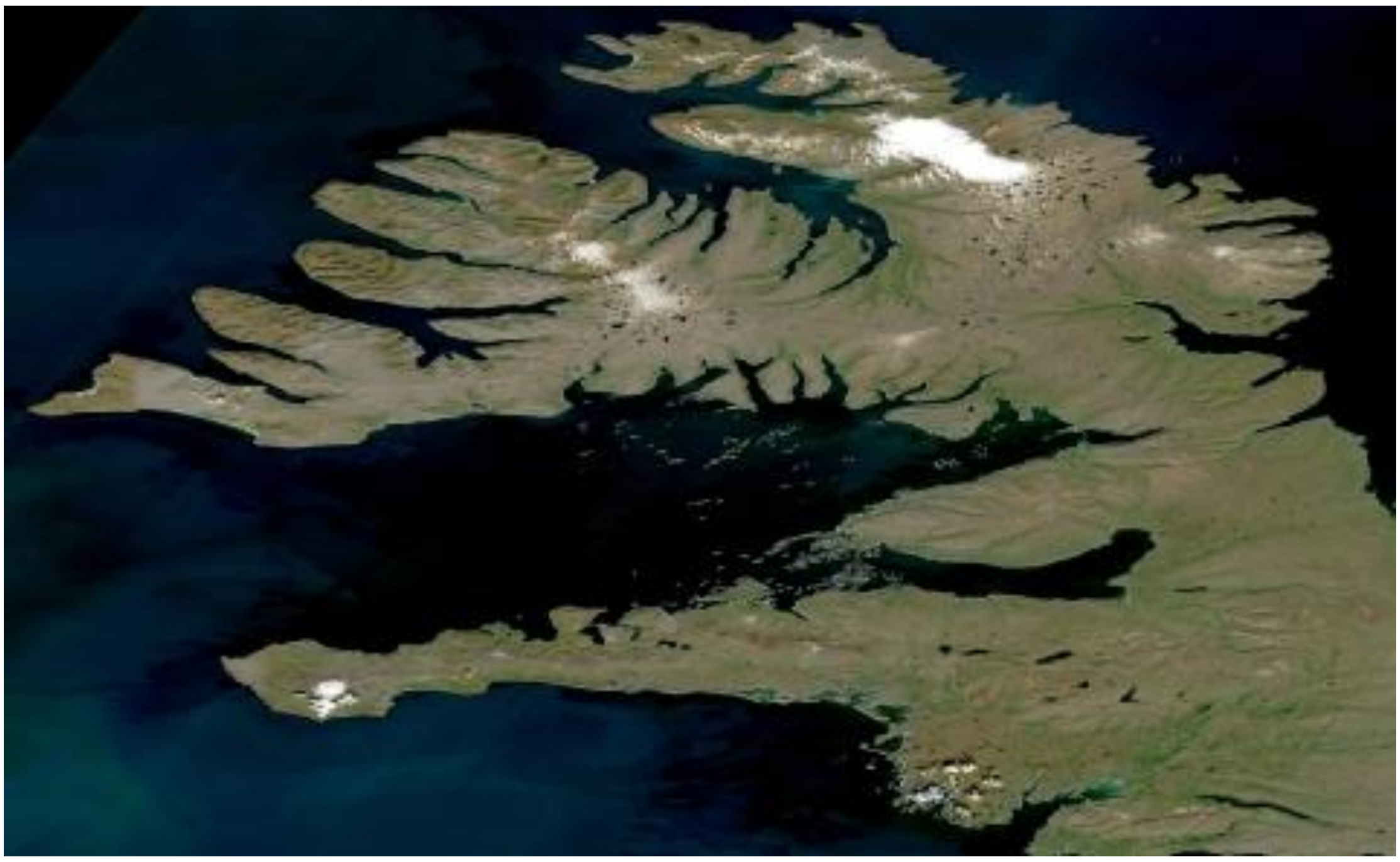

Fig. 4-1 True-color HLS images of northwestern Iceland

The model employs a ViT architecture and a Masked AutoEncoder (MAE) learning strategy to develop a self-supervised encoder, featuring an MSE loss function. The training data comprises continuous HLS imagery. The model incorporates spatial attention across multiple patches and temporal attention within each patch, enabling it to consider spatial relationships between different regions as well as the temporal evolution of the same region. This function enables the model to reconstruct images based on three temporal phases of the same region. First, a set of three HLS images is provided. Also, the model randomly masks a certain proportion of the areas. Then, it reconstructs the images based on the unmasked portions. The figure below shows the random masking and reconstruction results. The reconstruction results are generally consistent with the original images. Although the clarity does not yet match that of the original images, some blurriness can still be observed. Additionally, it is officially stated that the model can also accept remote sensing data in video format, allowing the model to infer the next changes in the scene over time by processing the temporal dimension in the video. This can be applied to scenarios such as flood propagation, fire burning, land cover classification, etc .

The Prithvi-100m model was initially pre-trained using three time series. During fine-tuning, the model can work with any number of time series, which can be applied to simulate flood propagation trends influenced by multiple time series. The model detects the R, G, and B bands from the input imagery. For the sample data, since the image size is relatively small and contains few ground features, the water body segmentation effect is acceptable. However, the edges are still unclear, and the continuity is poor. The model is designed to convert remote sensing data into high-resolution maps displaying flood changes. In the future, with proper packaging, it can achieve simple and user-friendly interaction and visualization. Users can select a region, a task, and a date range, and the model will highlight the flood propagation. Furthermore, users have the option of overlaying other datasets, such as crops, buildings, and road traffic, to ascertain the locations of submerged crops, buildings, or roads. This visual information can be employed for the purpose of planning decisions and risk prevention in similar disaster scenarios with the objective of mitigating the impact of floods.

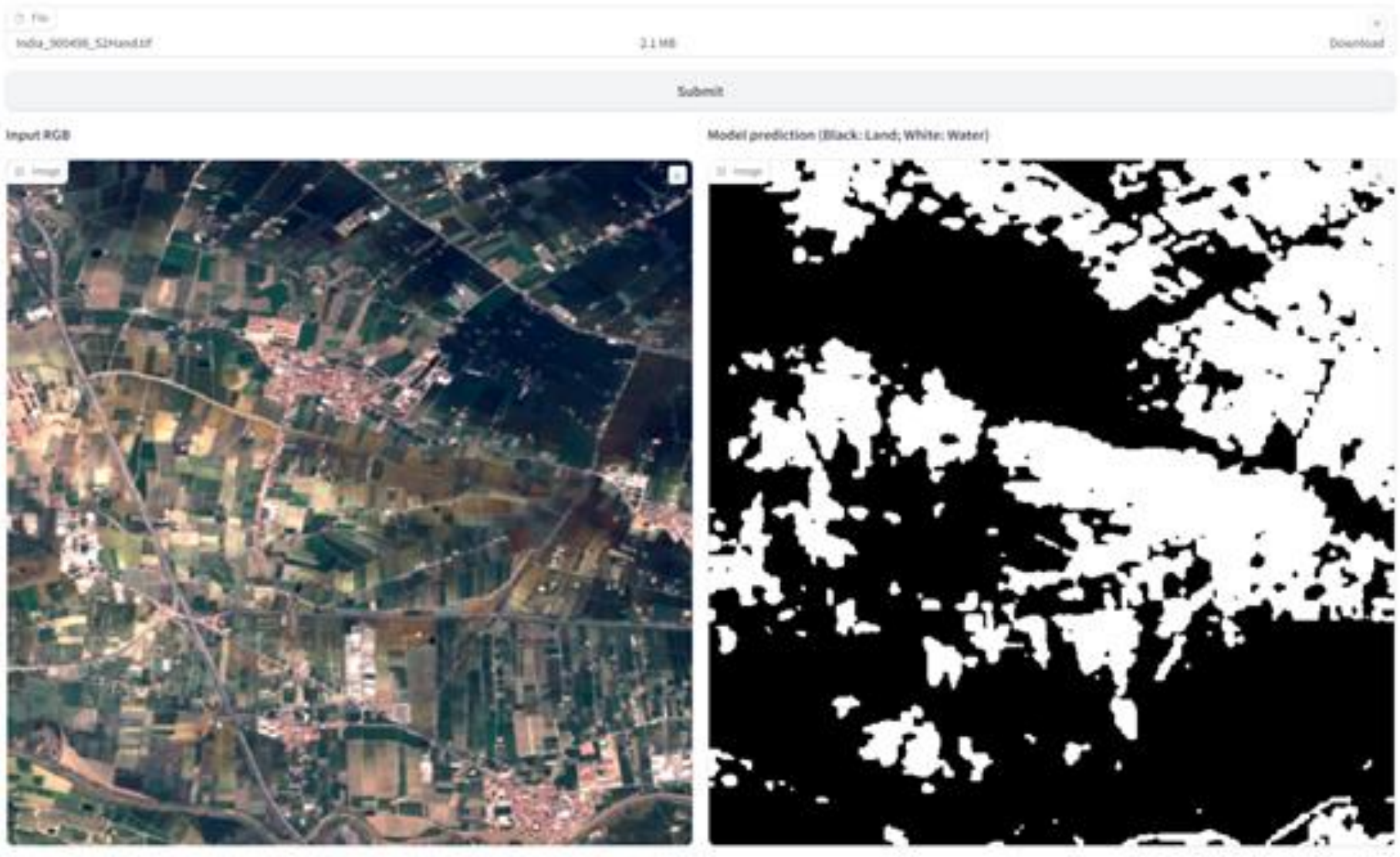

Fig. 4-2 Flood mapping identification (Dark pixels for land and light pixels for water)

The model, when fine-tuned on fire burn scar data, also performs well in detecting fire scars. This is similar to the model's performance in flood detection. For fire scar detection, the model extracts the SWIR, Narrow, NIR, and red bands from the input imagery.

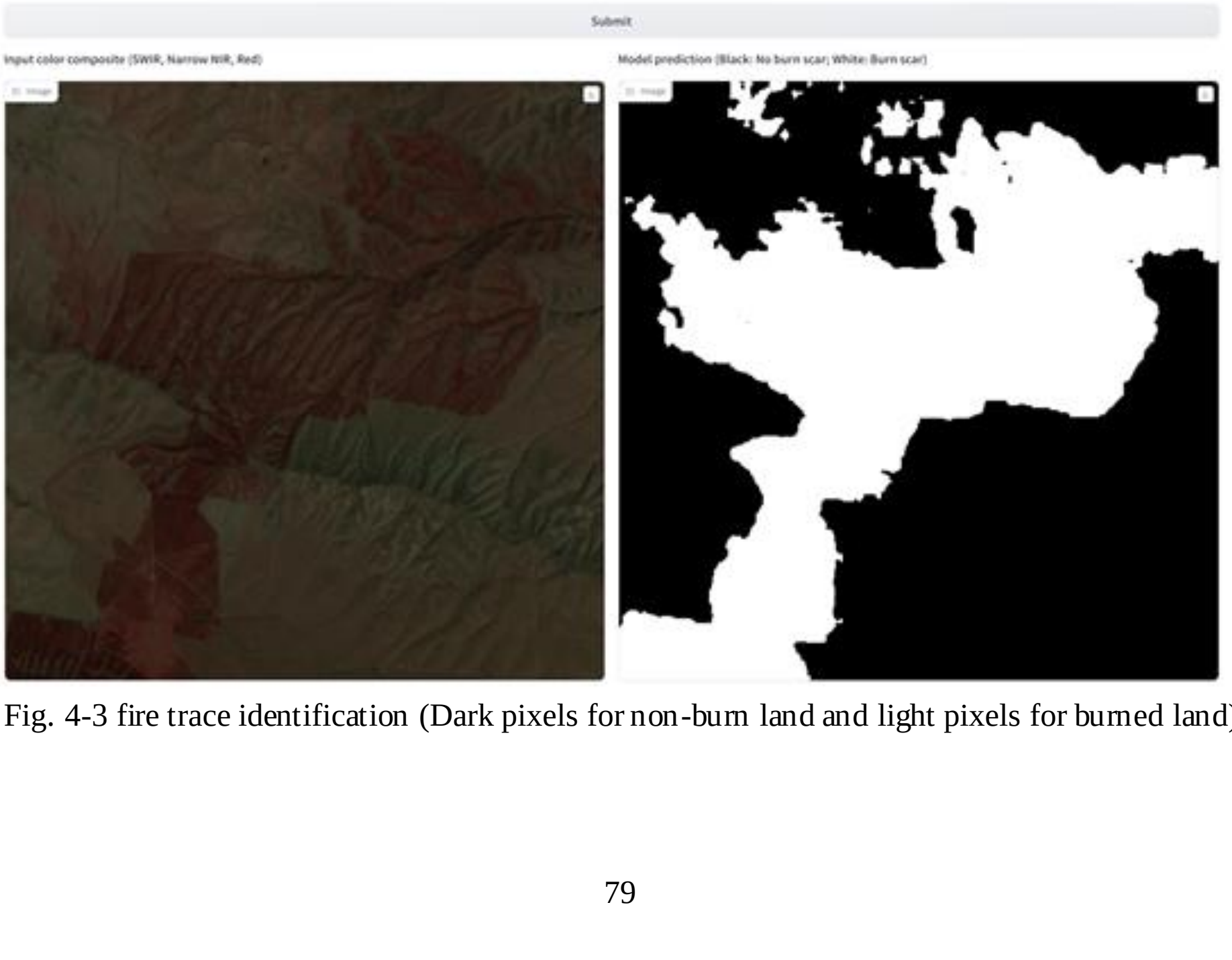

Fig. 4-3 fire trace identification (Dark pixels for non-burn land and light pixels for burned land)

### 4.1.3 The Adaptive Representation of a Geo-Knowledge Graph Considering Complex Spatiotemporal Relationships

The Geoscience Knowledge Graph (GKG) is a tool that can be used to organize various geoscience knowledge into a machine-understandable and computable semantic network, making it an effective means for organizing geoscience knowledge and providing knowledge-related services. As a result, it has received significant attention and has become a frontier in Earth sciences. Geoscience knowledge originates from multiple disciplines and features complex multi-scale, multi-granularity, and multi-dimensional spatiotemporal characteristics and relationships. Therefore, establishing a GKG representation model that aligns with the characteristics of geoscience knowledge is fundamental and essential for GKG construction and application. The existing knowledge graph representation models using fixed tuples are limited in adequately representing complex spatiotemporal features and relationships. To address this issue, this paper firstly resents a systematic analysis of the classification, spatiotemporal characteristics, and relationships of geoscience knowledge. Based on this analysis, a GKG adaptive representation model considering complex spatiotemporal features and relationships is proposed. Under the constraints of a unified spatiotemporal ontology, this is according to their spatiotemporal correlations. This model effectively represents geoscience knowledge, thereby avoiding the isolation of spatiotemporal feature representation and improving the precision and efficiency of geoscience knowledge retrieval. Furthermore, it can achieve alignment, transformation, computation, and reasoning of spatiotemporal information through spatiotemporal ontology.

Existing research employs fixed triples to represent spatiotemporal information as general semantic information or additional information. The fixed representation method may result in a disconnect between the spatiotemporal characteristics and relationships of geoscience knowledge, which could affect the efficiency and accuracy of geoscience knowledge retrieval and even lead to errors. Furthermore, this method presents challenges in tracking and analyzing the state and evolution of geoscience knowledge under different spatiotemporal conditions. The absence of a unified spatiotemporal ontology further complicates the process of spatiotemporal computation and reasoning, as different geoscience knowledge systems employ disparate spatiotemporal references and patterns to express spatiotemporal information. Consequently, the outcomes of computations and reasoning may occasionally be erroneous. To address these issues, this paper proposes an adaptive representation model for a Geo-Knowledge Graph (GKG) based on spatiotemporal associations. This model employs an automated process for selecting the most appropriate tuples to represent geoscience knowledge based on spatiotemporal associations. The fundamental concept is to initially represent the content of geoscience knowledge through basic tuples and subsequently determine whether to separately represent temporal or spatial information based on the relevance of the knowledge's spatiotemporal characteristics. In order to ensure accurate spatiotemporal calculation and reasoning across knowledge, it is essential that the representation of spatiotemporal information refers to the unified spatiotemporal ontology. This is achieved by establishing consistent spatiotemporal reference and expression. Meta-knowledge can be employed to record the generation and updating process of geoscience knowledge, thereby enabling the analysis of geoscience knowledge evolution and traceability.

Fig. 4-4 illustrates the application process of the GKG adaptive representation model. The initial step is to calculate the spatiotemporal associations of the geoscience knowledge to be represented. This can be accomplished through the application of either rule-based methods or deep learning methods. The second step entails the automatic selection of the most appropriate representation model from the adaptive representation model of the GKG, which is based on the strength of the spatiotemporal association of the knowledge and its category (i.e., weak, moderate, strong, and super-strong spatiotemporal associations). These representation models encompass not only common features, such as the representation of thematic content tuples and meta-knowledge tuples, which are supported by a unified spatiotemporal ontology, but also the capacity to represent personalized spatiotemporal information according to the spatiotemporal associations.

In the third step, a unified formal language (e.g., Web Ontology Language, OWL) and a graph database (e.g., JanusGraph) can be employed for unified storage and management of the GKG. The adaptive representation model permits the flexible representation of geoscience knowledge as triples, or as quads and quintuples, based on the closely related spatiotemporal information of the spatiotemporal associations. Consequently, the use of query languages such as SPARQL (SPARQL Protocol and RDF Query Language), GeoSPARQL, or Gremlin can facilitate more efficient and accurate retrieval, computation, and reasoning of geoscience knowledge.

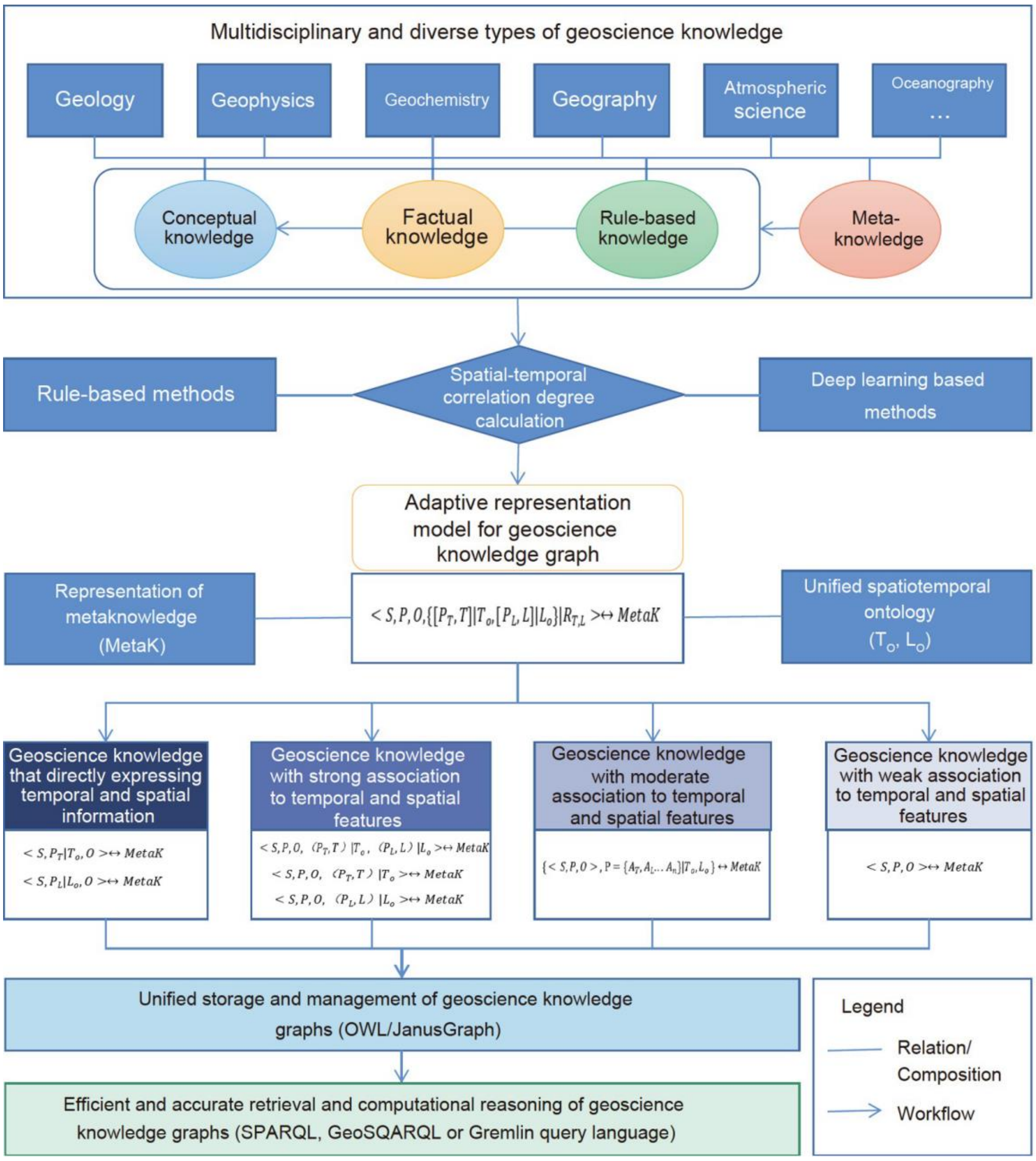

Fig. 4-4 The application process of the geographic knowledge graph adaptive representation model

The GKG is a semantic network that organizes various geoscience knowledge into a machine-understandable and computable format. This enables unified understanding, precise association, computational reasoning, and intelligent services for geoscience knowledge. It has become the most effective method for organizing geoscience knowledge and providing knowledge services. Consequently, geoscience research based on big data and artificial intelligence has become the foundation of modern geoscience e research and is a prominent and contentious topic in the field.

Geoscience knowledge is derived from a multitude of disciplines and exhibits intricate spatiotemporal characteristics and relationships, encompassing multi-scale, multi-granularity, and multi-dimensional features. To represent the diverse forms of geoscience knowledge derived from different disciplines, it is crucial to develop a GKG representation model that aligns with the intrinsic characteristics of geoscience knowledge and takes into account the complex spatiotemporal features and relationships. This serves as the foundation and prerequisite for the establishment and application of GKG.

This paper first analyzes the status and limitations of existing knowledge graph representation models. It then systematically explains the categorical patterns of geoscience knowledge and their spatiotemporal characteristics and relationships. Based on this analysis, it proposes an adaptive representation model for GKG that can fully express spatiotemporal features and relationships. The core of this model is to represent different types of geoscience knowledge as different tuples according to their spatiotemporal associations. Furthermore, it is constrained by a unified spatiotemporal ontology. This model not only enables efficient representation and storage of geoscience knowledge but also supports efficient and accurate retrieval and utilization of geoscience knowledge. Additionally, by establishing links with the spatiotemporal ontology, it can help facilitate the unified alignment, transformation, computation, and reasoning of spatiotemporal information.

The research on the GKG adaptive representation model is still in its preliminary stages. Some preliminary experiments have been conducted in the context of certain studies, such as the construction of geological time-scale knowledge maps in DDE.

4.1.4 The Geo-Science Knowledge Graph (GKG): Development, Construction, and Challenges

For a considerable period, a variety of symbols have served as the primary vehicles for the generation, accumulation, and dissemination of human knowledge. These symbols exist in machines in multiple forms, including text, images, videos, audio, and graphics among other multimodal data. However, it is not straightforward for machines to comprehend and utilize the diverse systems of human knowledge. Consequently, the necessity arises for the development of a novel cognitive mechanism to address this predicament.

In the field of geoscience, knowledge can be obtained from a variety of sources, including standard specifications, professional books, scientific articles, terminology dictionaries, social media, and observation stations. Nevertheless, contemporary methods of representing, organizing, and applying geoscience knowledge exhibit notable divergences from both human and machine perspectives. In the context of human-machine collaborative thinking, the content of geoscience knowledge should be consistent across different sources. Consequently, the Geo-Science Knowledge Graph (GKG) is in urgent need of a unified human-machine collaborative mechanism.

In this context, a cognitive mechanism is proposed to understand geoscience knowledge from both human and machine perspectives. Humans perceive the Earth system through the five senses and express their cognition through language. Over time, knowledge can be extracted from these languages, gradually forming a system of geoscience knowledge. With the proliferation of Earth big data, machines can acquire geoscience knowledge from multimodal data, constructing various knowledge bases. In comparison to human knowledge systems, geoscience knowledge bases contain a wealth of knowledge but lack common sense. To bridge this gap, human-machine interaction must be introduced to connect these two branches. This mechanism can systematically express geoscience knowledge, addressing the issues of traceability, interpretability, and computability of geoscience knowledge in a computer environment. A rule-based knowledge representation method based on the "condition-result" model is proposed. This model incorporates the straightforward structure of knowledge graphs and addresses the challenge of representing intricate rule-based knowledge, paving the way for further knowledge reasoning. It is comprised of two fundamental components: the condition part and the result part. The condition part encompasses a set of nodes, in which disparate nodes are aggregated through logical computation to form constraint rules.

Completion of the GKG is critical to maintaining its integrity. The completion process involves evolving the knowledge graph by adding new triples, including the three subtasks of link, entity, and relation prediction. This can be achieved through various methods such as embedding-based models, relational path reasoning, path discovery reasoning, meta-relation learning, and rule-based reasoning. Each method has its specific advantages and disadvantages and can be used based on the needs of practical applications.

There are two key issues that need to be carefully considered in the GKG completion process: (1) As mentioned earlier, the earth system is a constantly evolving open system where elements and relationships are constantly changing. Therefore, the completion model should adopt an open scenario and be able to handle dynamic knowledge skillfully. (2) Since the earth is a complex entity with many interrelated elements, the completion of GKG must consider multilevel knowledge reasoning.

Geoscience knowledge is complex and vast. The data in question can be obtained from earth science experts through manual input as well as from Big Earth data through the use of artificial intelligence (AI) methods. However, relying on either method alone is not sufficient to build a comprehensive geoscience knowledge system. On the one hand, it is very challenging to capture and formalize all expert knowledge, which is often highly uncertain or ambiguous. On the other hand, while big earth data contains substantial geoscience knowledge, automatically constructed knowledge systems are not complete and exclude some essential expert experience.

Therefore, by combining geoscience experts with computer systems can leverage the advantages of experts in different fields, utilizing continuously optimized AI methods promote the long-term sustainable development of geoscience knowledge engineering systems. Based on this, it is proposed that GKG requires a human-machine collaboration mechanism, which can be realized by at least three modes.

The initial mode of operation is designed to facilitate access to more complex and highly accurate professional geoscience knowledge. This can be achieved through crowdsourcing from geoscientists or by implementing a two-step strategy: initially acquiring knowledge automatically, followed by expert verification. The second mode can be achieved by a natural language question-and-answer system. As a human-machine interaction system, users can assess the accuracy, completeness, and systematic nature of GKG based on responses to a series of questions. The third mode entails the utilization of specific geoscience models to validate related knowledge, thereby assessing the practical applicability of GKG within geoscience models from a problem-solving perspective.

The research on GKG is still in its early stages and requires more in-depth studies for further development. In particular, the following challenges urgently need to be addressed:

(1) Representation of geoscience knowledge across different disciplines: Geoscience knowledge represents human understanding of the elements of the Earth system. Unlike other types of knowledge, it covers a broader range and has significant spatiotemporal characteristics. In the previous section, we proposed a representation framework to address this distinction. However, such a framework is generic to all fields of geoscience and needs further refinement in different sub-disciplines. Future research should aim to develop the proposed framework by incorporating more domain-specific knowledge to meet practical application requirements.

(2) Collaborative management of geoscience knowledge based on collective intelligence: Expert knowledge is a vital source of geoscience knowledge. Traditional methods tend to manually input this knowledge into databases. However, barriers between different disciplines significantly increase the difficulty. To enhance collaboration among geoscientists and improve the interpretability of geoscience knowledge, it is necessary to establish a collaborative mechanism and corresponding management system based on collective intelligence. This system should leverage the strengths of geoscientists, enhance collective intelligence collaboration capabilities, and continuously acquire geoscience knowledge through iteration and optimization.

(3) Quality assessment of GKG: There is no universally accepted method for evaluating the quality of GKG. Therefore, it is necessary to establish a quality assessment system based on professional Earth data and geoscience knowledge content. To achieve this, research can be conducted in the following aspects: defining evaluation dimensions based on existing assessment methods; determining indicators for each dimension and corresponding evaluation methods; proposing a quality assessment system combining qualitative and quantitative indicators.

(4) Geoscience knowledge reasoning in problem solving: Knowledge reasoning can uncover implicit semantic relationships between different elements, which is crucial for solving geoscience problems. The complexity of geoscience knowledge systems mainly lies in the multiple associative relationships commonly present among geosciences. Modern geoscience methods usually handle unary or binary relationships by simplifying these multiple associative relationships, leading to significant information loss. Therefore, future research should be guided by geoscience problem solving, exploring the combination of inductive and deductive reasoning methods with deep learning and knowledge representation models. Additionally, addressing semantic reasoning involving interaction, evolution, and hierarchical structures is necessary to improve simulation and prediction capabilities of GKG.

## 4.2 Multi-Modal Foundation Model

### 4.2.1 Pangu: A Multi-Modal Weather Forecast Foundation Model

Weather forecasting is one of the most important scenarios in the field of scientific computing. The prediction of future weather changes, especially extreme weather events such as rainstorms, typhoons, droughts, and cold waves, is crucial. Traditional numerical forecasting uses mathematical and physical equations to model the state of the atmosphere and employs computer simulation methods to solve these equations to obtain future weather conditions. Over the past thirty years, this approach has achieved remarkable success. However, as the growth of computational power slows, physical models become increasingly complex and numerical meteorological forecasting methods have gradually encountered bottlenecks. On the one hand, traditional numerical forecasting consumes a vast amount of computational power. For example, a 10-day numerical forecast with a resolution of 0.25°×0.25° requires several hours of simulation on a supercomputer with over 3,000 nodes. On the other hand, complex parameterized physical models are inherently incomplete. The parameterization of physical processes inevitably introduces systematic errors into numerical forecasting.

Since the 1920s, especially in the past three decades, with the rapid development of computing power, numerical weather prediction has achieved great success in daily weather forecasts, extreme disaster warnings, climate change predictions, and other fields. However, with the slowdown in computing power growth and the gradual complexity of physical models, the bottleneck of traditional numerical prediction has become increasingly prominent. Researchers are beginning to explore new weather forecasting paradigms such as using deep learning methods to predict future weather. In the fields where numerical methods are most widely used, such as medium and long-term forecasting, the accuracy of existing AI forecasting methods is still significantly lower than numerical forecasting methods and is restricted by problems such as lacking interpretability and inaccurate extreme weather predictions.

AI weather forecasting first achieved great success in short-term forecasting. This is due to the huge advantage of AI forecasting in prediction speed: numerical forecasting methods cannot give minute-level weather forecasts, while the ability of AI methods to fit radar echo data exceeds that of extrapolation methods such as the optical flow method. When AI forecasting methods are applied to medium and long-term weather forecasts (one of the most successful areas for the application of numerical weather forecasting), although AI methods can greatly improve the prediction speed, the resolution and accuracy of AI forecasting methods are significantly behind those of numerical weather forecasting methods. In March 2022, NVIDIA launched the FourCastNet model. This model, for the first time, increased the horizontal resolution of forecasts to a level comparable to that of numerical forecasts, which is 0.25°×0.25°. However, its forecast accuracy still lags far behind that of numerical forecasting methods. For example, the root mean square error (RMSE) of FourCastNet's 5-day potential forecast is 484.5. Even if 100 models are used for integrated forecasting, its root mean square error is still as high as 462.5, which is much higher than the 333.7 reported by the European Meteorological Center's operational IFS. Before the Pangu meteorological model was proposed, AI weather forecasting was mainly used as a rapid alternative model to numerical forecasting and could not directly replace traditional numerical forecasting methods. Some meteorologists even pointed out that it would take some time for AI forecasting methods to surpass traditional numerical methods.

Researchers from Huawei Cloud have proposed a new high-resolution global AI weather forecast system: the Pangu Meteorological Model. The Pangu Meteorological Model is the first AI method whose accuracy exceeds traditional numerical forecasting methods. The 1-hour to 7-day forecast accuracy is higher than traditional numerical methods (European Meteorological Center's operational IFS). It can provide second-level global weather forecasts, including position Potential, humidity, wind speed, temperature, sea level pressure, etc. The Pangu meteorological model has a horizontal spatial resolution of 0.25°×0.25°, a temporal resolution of 1 hour and covers 13 vertical layers, allowing it to accurately predict fine-grained meteorological characteristics. As a basic model, the Pangu meteorological foundation model can also be directly applied to multiple downstream scenarios. For example, in the tropical storm prediction task, the prediction accuracy of the Pangu Meteorological Model significantly exceeded the results of the European Meteorological Center's high-precision forecast (ECMWF HRES Forecast).

To use the current foundation models in the CV field for meteorological data analysis, this paper proposes a 3D Earth-Specific Transformer (3DEST) to deal with complex and non-uniform 3D meteorological data. It also uses a hierarchical time-domain aggregation strategy to reduce the number of forecast iterations, thereby reducing iteration errors. Figure 4-5 is a schematic diagram of the 3D Earth-Specific Transformer proposed in this paper. The main idea is to use a 3D variant of a visual transformer to process complex and non-uniform meteorological elements. Due to the large resolution of meteorological data, compared with the common vision transformer method, the researchers reduced the encoder and decoder of the network to 2 levels (8 blocks). They, they adopted the sliding window attention mechanism of the Swin transformer to reduce the computational workload of the network. It should be noted that even with these methods. the overall FLOPs of the current network still exceed 3000G. In the future, with sufficient computing power, larger networks can be used to further improve forecast accuracy.

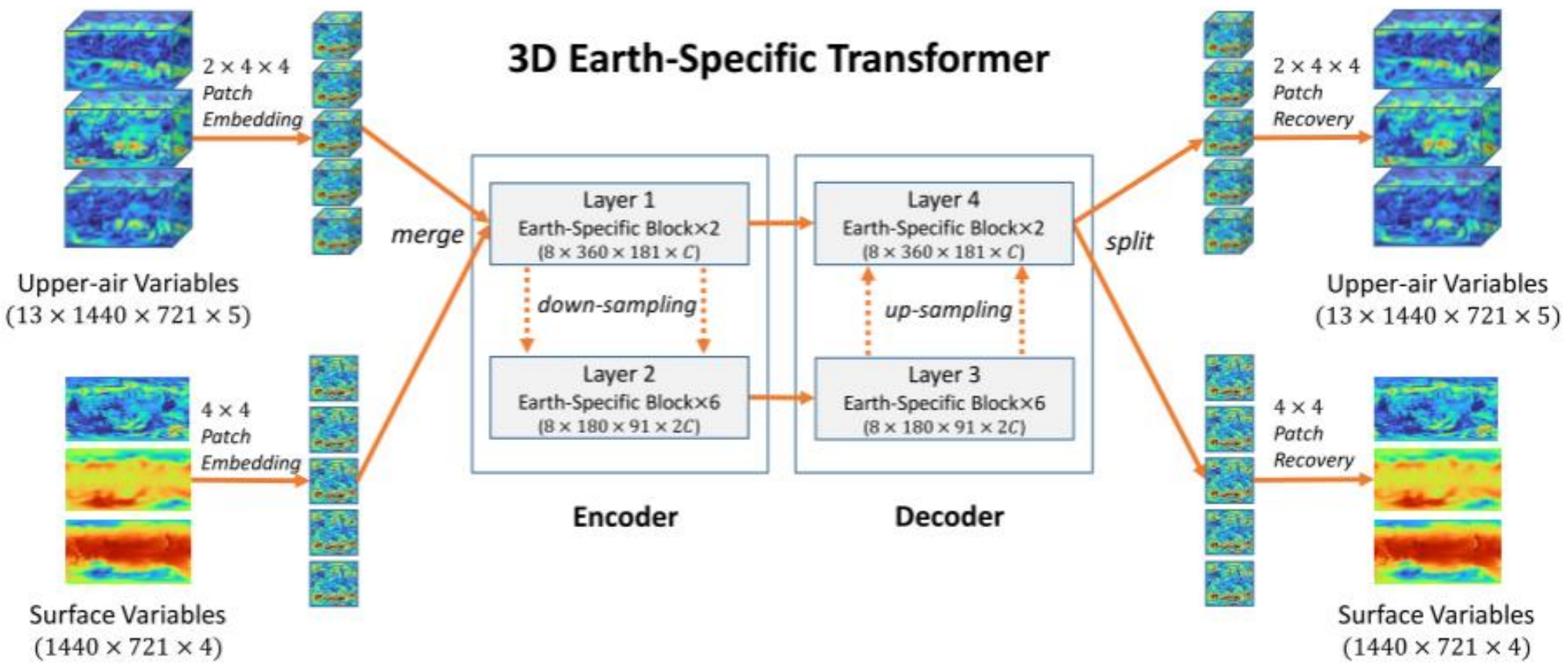

Fig. 4-5 The structure of the 3D Earth-Specific Transformer

The Pangu Meteorological Big Model has, for the first time, surpassed traditional numerical methods in medium and long-term weather forecasting. Both training and testing were conducted on the ERA5 dataset, which includes 43 years (1979-2021) of global real-time meteorological data. Specifically, data from 1979 to 2017 were used as the training set, data from 2019 served as the validation set, and data from 2018, 2020, and 2021 were used as the test set. The data used by the Pangu Big Model includes 13 different pressure levels at vertical heights. At each level, it considered as five elements: temperature, humidity, potential, and wind speed in both longitude and latitude directions. Additionally, it includes four meteorological elements at the Earth's surface: 2-meter temperature, and wind speed in both longitude and latitude directions, and sea-level pressure. Figure 1 illustrates some results of the Pangu Meteorological Big Model. The model outperforms existing numerical forecasting methods, such as the European Centre for Medium-Range Weather Forecasts' operational IFS comprehensively. For example, the root means square error (RMSE) of the Z500 five-day forecast provided by the Pangu Meteorological Foundation model is 296.7, which is significantly lower than the best previous numerical forecasting method (operational IFS: 333.7) and AI method (FourCastNet: 462.5). Moreover, the Pangu Meteorological Big Model can complete a 24-hour global weather forecast in just 1.4 seconds on a single V100 GPU, achieving a speedup of over 10,000 times compared to traditional numerical forecasting methods.

### 4.2.2 SkySense Multi-Modal Remote Sensing Foundation Model

Recently, foundation models have demonstrated exceptional performance across various tasks, leading to increased interest in developing a versatile Remote Sensing Feature Model (RSFM) for Earth Observation (EO). A key requirement for an effective RSFM is its ability to handle multi-modal temporal Remote Sensing Imagery (RSI). Earth observation relies heavily on multi-modal time series data, such as optical and Synthetic Aperture Radar (SAR) imagery, each providing unique benefits and complementing each other. For example, optical images offer detailed spectral bands and texture but can be affected by weather conditions.

An ideal RSFM should be adaptable to various modes (single-modal and multi-modal) and spatial granularities (pixel-level, object-level, and image-level) for different EO tasks. Additionally, since remote sensing data is inherently tied to spatio-temporal coordinates, the model should effectively leverage geographical context to enhance RSI interpretation.

While previous research has shown the potential of universal models for EO, they have largely focused on single modalities without incorporating time or geographic contexts, which limits their versatility. To address this, we introduce SkySense, a comprehensive model trained on a dataset of 21.5 million multi-modal RSI time series. SkySense features a decomposed multi-modal spatiotemporal encoder, which processes optical and SAR data over time. This encoder is trained using multi-granularity contrastive learning to capture representations across different modalities and spatial scales. Additionally, SkySense incorporates geographic context prototype learning to enhance RSI representation by learning region-aware prototypes based on the multi-modal spatiotemporal characteristics of the data.

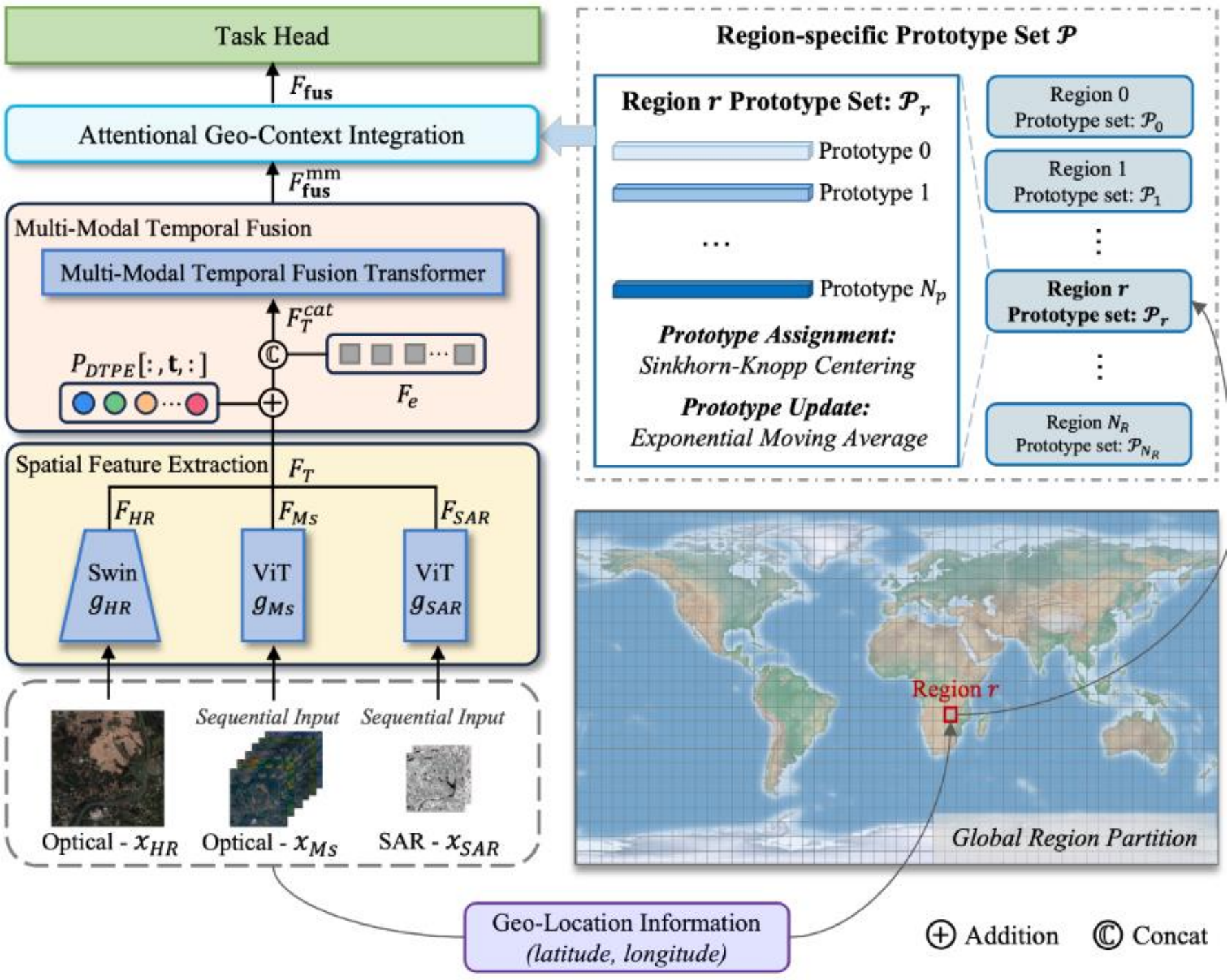

Fig. 4-6 SkySense Model architecture

In Fig. 4-7, the map visualizes the learned prototypes by calculating the pre-training features for each pixel and matching them to the most similar prototype. When compared to the ESRI LandCover Map, the results are promising, especially in distinguishing different regions. Notably, the central section of Figure 4-7 highlights GCP's superiority in fine-grained segmentation, revealing details such as cultivated land within towns that the LandCover Map does not capture. The visualization maintains the same spatial resolution as the ESRI LandCover Map.

SkySense stands out as the largest multi-modal Remote Sensing Feature Model (RSFM) available. It offers flexibility with its modular design, allowing for both combined and standalone use across various tasks. In extensive evaluations involving 16 datasets and 7 different tasks—including single-modal to multi-modal, static to temporal, and classification to localization—SkySense demonstrates exceptional generalization. It outperforms 18 recent RSFMs across all test scenarios, showing average improvements of 2.76%, 3.67%, and 3.61% over the latest models such as GFM, SatLas, and Scale-MAE, respectively.

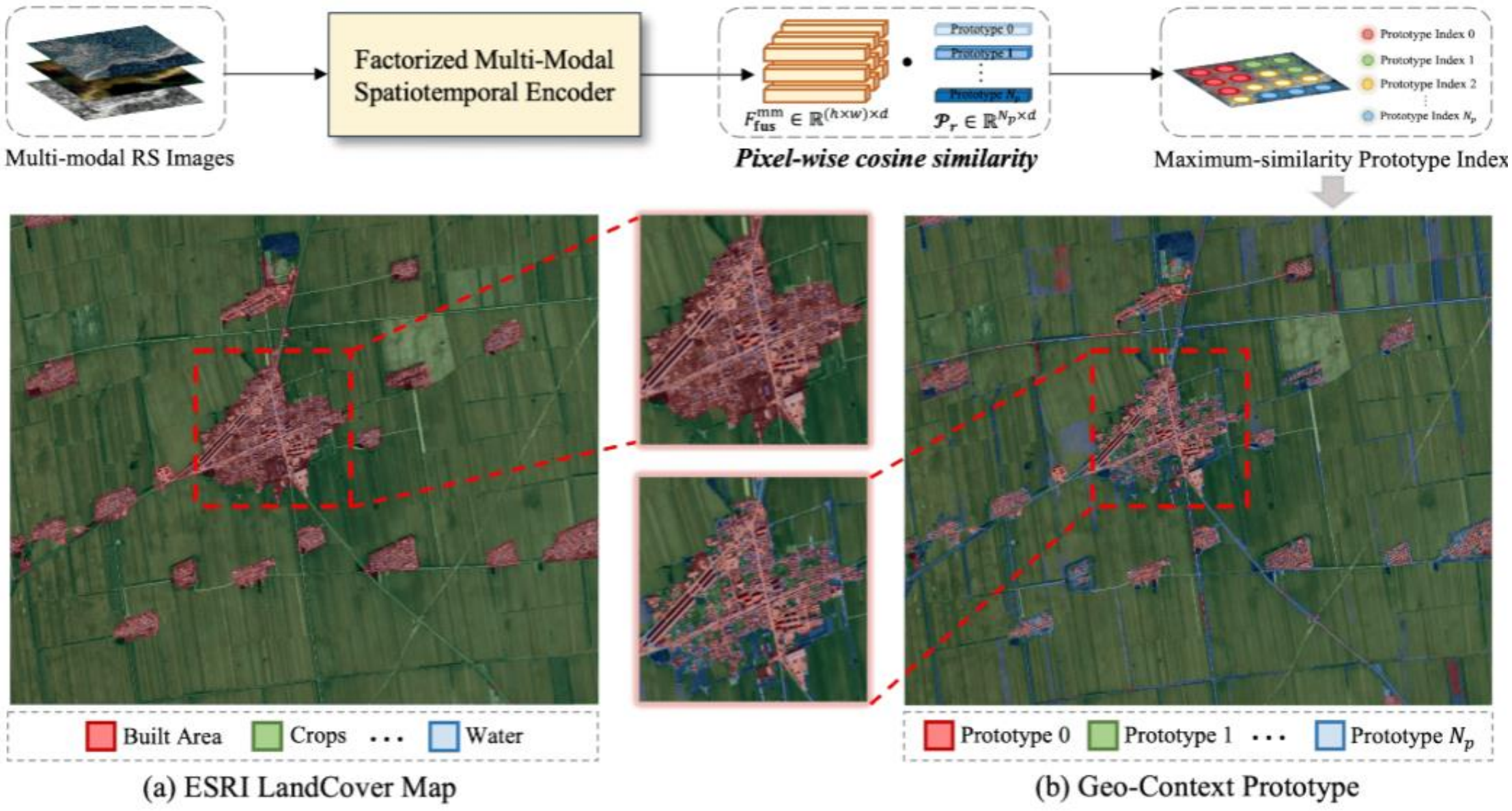

Fig. 4-7 Comparison between (a) ESRI LandCover Map and (b) Geo-Context Prototype

4.2.3 RingMo: Generative Pre-trained Foundation Model for Cross-Modal Remote Sensing

Deep learning, as a representative of artificial intelligence technology, has been applied to various remote sensing image interpretation tasks. Remote sensing data is characterized by large spatial coverage and complex scene content. A standard remote sensing image can often contain billions of pixels, covering tens of thousands of square kilometers, exhibiting significant differences from natural scene data. Most existing deep neural network models are initialized with weights pre-trained on natural scene images. Meanwhile, their performance and generalizability on remote sensing data interpretation tasks need further improvement.

The Aerospace Information Research Institute, Chinese Academy of Sciences (AIRCAS) has led the development of the first generative pre-training model for cross-modal remote sensing data, named "RingMo" (Remote Sensing Foundation Model). It is a general-purpose, multi-model, and multi-take model, aiming to build a universal, convenient, and high-performance solution for various applications in the remote sensing field. The model has the following features.

Driven by remote sensing characteristics and unlike existing remote sensing pre-training methods that typically follow supervised or contrastive learning paradigms, the "RingMo" model leverages a masked autoencoder structure. It is a generative self-supervised pre-training model designed for complex scenarios with enhanced general representation capabilities in remote sensing (Figure 4-8). For example, addressing issues such as differing imaging mechanisms and target characteristics from various platforms, large observation areas with relatively small targets in remote sensing images and significant variations in target sizes and uneven distribution. the "RingMo" model adopts a self-supervised learning approach guided by target characteristics. By incorporating constraints on geometry, electromagnetism, and target structure, the model automatically extracts general features of remote sensing objects, demonstrating strong generalization capabilities for new tasks. Notably, the "RingMo" model employs the latest popular Transformer-based backbone networks, such as ViT and Swin Transformer, effectively modeling both local and global feature dependencies in remote sensing data.

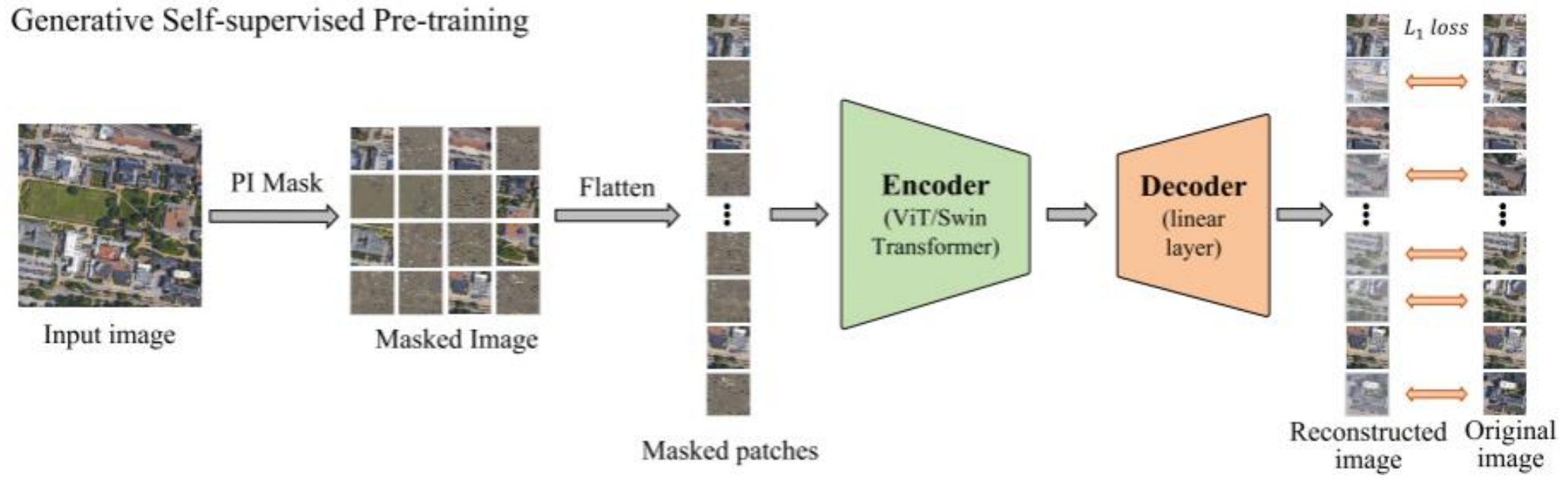

Fig. 4-8 Remote sensing-generated self-supervised pre-training algorithm

To possess cross-modal remote sensing datasets, the existing remote sensing sample libraries rely on manual annotation by professionals, which is highly labor-intensive and time-consuming, making it difficult to meet the large-scale, high-diversity, and rapidly expandable data requirements for foundation model training. To enhance the feature representation capability of remote sensing pre-training models, the training dataset for the "RingMo" model includes over 2 million remote sensing images with resolutions ranging from 0.1m to 30m. These images are sourced from platforms such as the China Remote Sensing Satellite Ground Station and aerial remote sensing aircraft, as well as sensors like the Gaofen (GF) series satellites, Jilin satellites, and QuickBird satellites (Fig. 4-9). Additionally, the dataset contains over 100 million target instances with arbitrary angle distributions, covering more than 150 typical cities and towns worldwide, as well as common scenes like airports and ports. The sample data used has distinctive remote sensing features. At the meantime, the entire dataset does not require annotation, significantly reducing the cost of training data annotation.

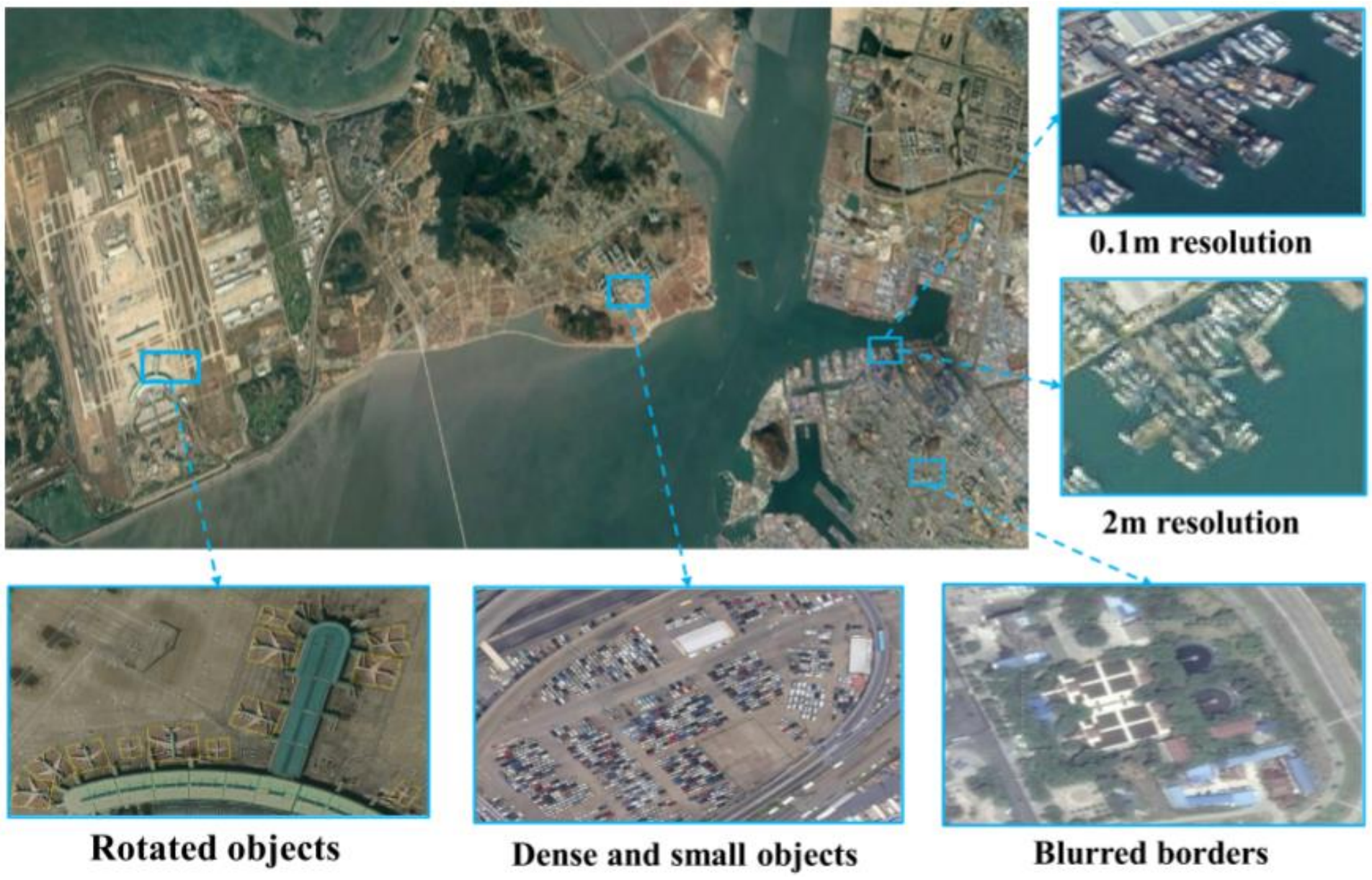

Fig. 4-9 Cross-modal remote sensing dataset

To possess the capability for application task generalization, due to the varying challenges of different application tasks, the data and targets used also differ. Existing interpretation methods require designing specialized network structures for different downstream tasks and fine-tuning with a large amount of labeled data. Consequently, the resulting remote sensing models often lack general applicability and robust task generalization capabilities, being suitable only for specific application tasks. The "RingMo" model possesses the ability to understand and restore remote sensing data, enabling the common semantic space representation of cross-modal remote sensing data (Figure 4-10). For different downstream tasks, only the prediction head network needs to be modified, allowing for flexible and rapid transfer to different fields' downstream tasks. Simple fine-tuning enables the model to adapt to multi-target fine-grained classification, small target detection and recognition, and complex object extraction tasks.

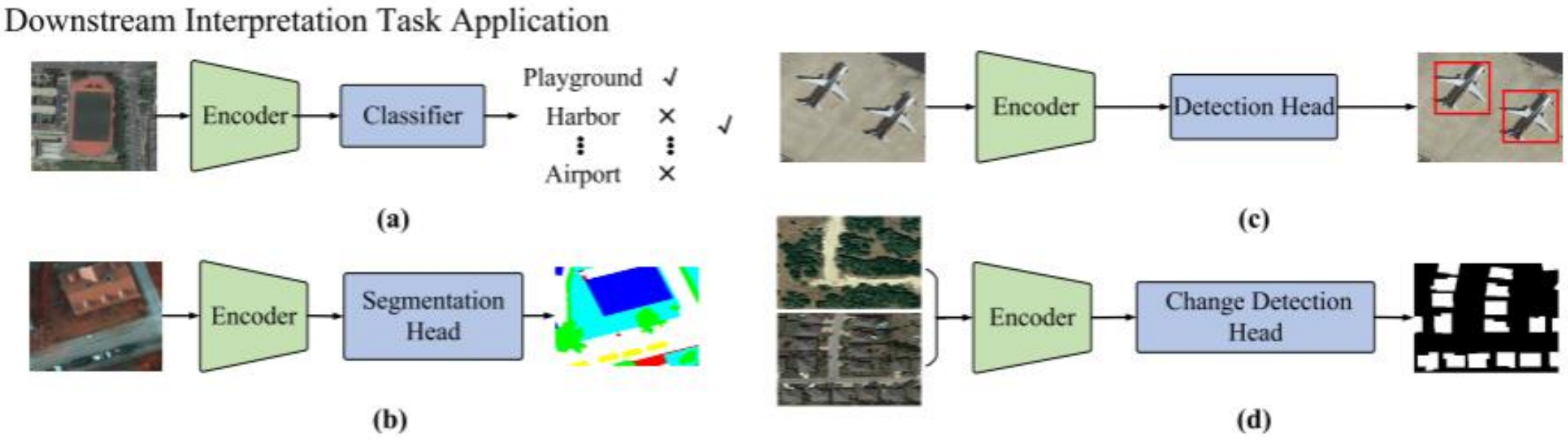

Fig. 4-10 Application task generalization

To achieve domestic adaptation and independent innovation, the Aerospace Information Research Institute has collaborated extensively with Huawei. The Beijing Ascend AI Ecosystem Innovation Center provides technical support, leveraging the computational power of the Chengdu Intelligent Computing Center, a benchmark project under the "Eastern Data Western Computing" initiative. The model and training methods have been adapted to domestic platforms based on the Ascend base and the MindSpore AI framework. Additionally, performance optimizations have been made for self-supervised large-scale data training. This effort provides robust support for researchers across various industries to conduct remote sensing pre-training and develop downstream tasks on domestic software and hardware platforms, thereby promoting application and implementation in business scenarios.

Currently, the relevant outcomes of the "RingMo" model have been published in the prestigious remote sensing journal IEEE Transactions on Geoscience and Remote Sensing. At the same time, the model has also been trialed in various fields such as national defense and security, real-life 3D imaging, demonstrating significant improvements in target detection and recognition as well as feature classification, compared to general visual models. Future plans include expanding their application to more industries such as land resources, housing and transportation, water conservation, and environmental protection, providing a comprehensive solution for integrated aerial, space, and ground applications.

## 4.3 Foundation Model for Remote Sensing Intelligent Computation

### 4.3.1 SpectralGPT: Spectral Remote Sensing Basic Foundation model

Spectral imaging excels in capturing extensive spectral data, offering highly precise analysis and identification of objects and scenes that surpasses the capabilities of traditional RGB data. This capability makes multi/hyperspectral (MS/HS) remote sensing (RS) data invaluable for numerous Earth Observation (EO) applications, including land use/land cover mapping, ecosystem monitoring, weather forecasting, energy development, biodiversity conservation, and geological exploration. The surge in available and accessible spectral data from remote sensing satellites like Landsat-8/9, Sentinel-2, and GF-1/2/6 has opened up new opportunities for discoveries and advancements in EO fields. However, this growth has also introduced two significant challenges that need urgent and effective solutions.

The Masked Autoencoder (MAE) is a straightforward auto-encoding technique that reconstructs the original signal. Unlike traditional auto-encoders, MAE features an asymmetric design where the encoder processes only part of the observed signal without mask tokens, while the lightweight decoder reconstructs the complete signal from latent representations and mask tokens. Drawing inspiration from spatiotemporal agnostic sampling in video data, we model multiband spectral images as 3D tensor data. We employ a 3D cube masking strategy to facilitate efficient processing of spectral tensor data, capturing spatial and spectral representations effectively with a 90% masking ratio and extracting more comprehensive knowledge from the input data.

The SpectralGPT model comprises three main components: 3D masking for spectral data processing, an encoder for learning spectral visual representations, and a decoder for reconstructing multiple objects. Our approach stands out due to its progressive training strategy, where the model is trained on various types of spectral data. This enhances the SpectralGPT base model's flexibility, robustness, and generalization capabilities. Fig. 4-11 illustrates the workflow of SpectralGPT across different downstream tasks.

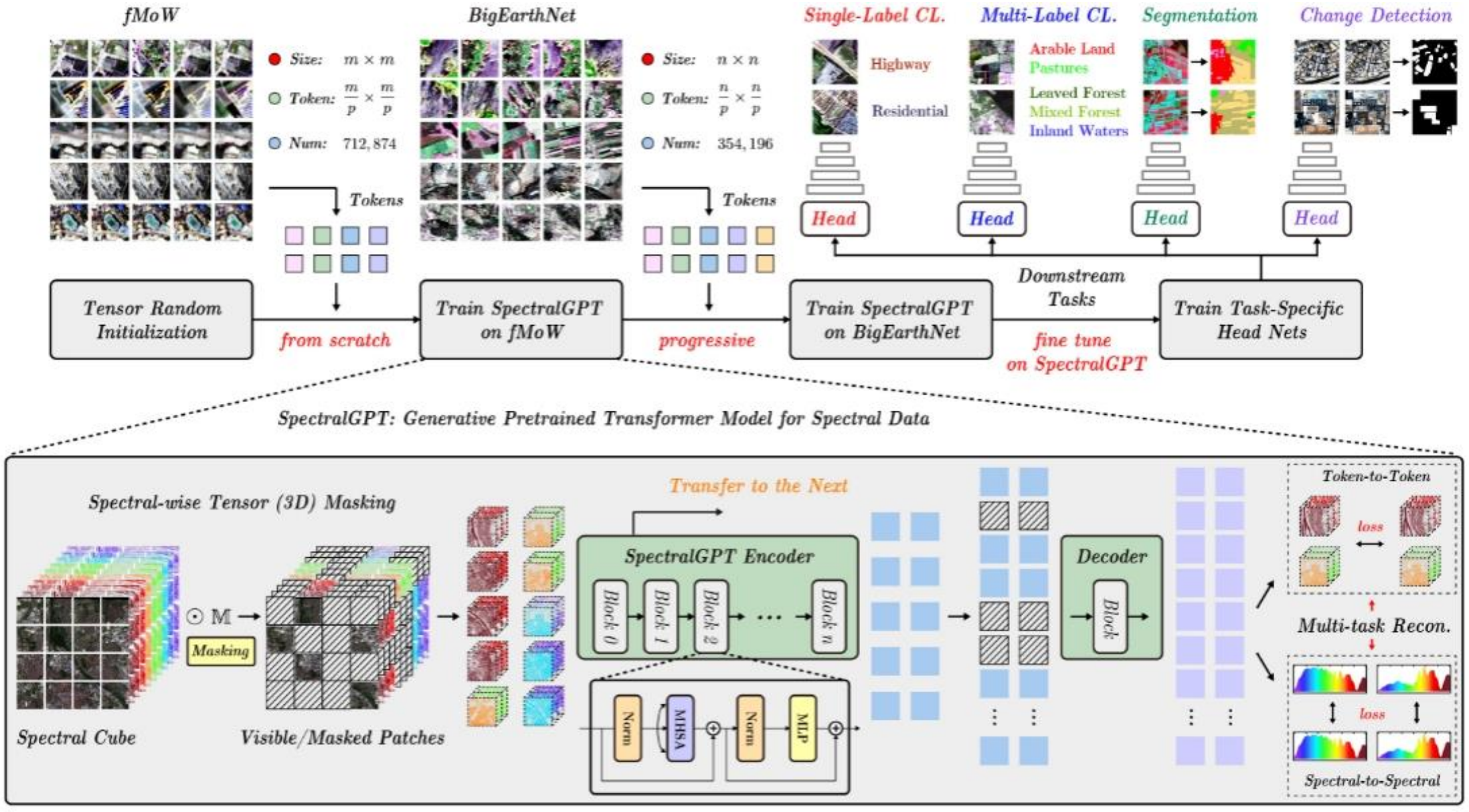

Fig. 4-11 An illustrative workflow of the underlying model and adaptation to downstream tasks of SpectralGPT

The SpectralGPT model underwent a thorough performance evaluation by comparing it with several state-of-the-art (SOTA) base models, including ResNet50, SeCo, ViT, and SatMAE. Its effectiveness was assessed across four downstream Earth Observation (EO) tasks: single-label scene classification, multi-label scene classification, semantic segmentation, and change detection. Quantitative metrics were used to evaluate the model's pre-trained performance, including recognition accuracy, macro and micro mean accuracy (mAP) for single-label scene classification, macro-mAP and micro-mAP for multi-label scene classification, overall accuracy (OA) and mean intersection over union (mIoU) for semantic segmentation, and accuracy, recall, and F1 score for change detection.

Additionally, extensive ablation studies were conducted to investigate various factors such as masking ratio, decoder depth, model size, patch size, and training duration. These studies utilized the computational capabilities of four NVIDIA GeForce RTX 4090 GPUs to fine-tune the pre-trained base model, providing a comprehensive understanding of SpectralGPT's performance and adaptability within the remote sensing domain.

Future research aims to broaden the dataset used for training by incorporating a greater variety of remote sensing data, including different modes, resolutions, time series, and image sizes. This expansion will enhance the robustness of the remote sensing model. Additionally, efforts will be made to extend SpectralGPT's functionality to cover a wider range of downstream tasks, transforming it into a versatile AI model with improved generalization capabilities for various Earth Observation and geoscience applications.

#### 4.3.2 Multi-Modal Artificial Intelligence Model Enables Earth Observation

Advancements in Earth observation (EO) technologies have greatly enhanced our ability to measure and monitor various aspects of the planet, including its surface, subsurface, atmosphere, and water quality, as well as the health of humans, plants, and animals. Remote sensing (RS), a prominent non-contact EO method, allows us to gather valuable information about geophysical properties and their environmental contexts from space. This wealth of remote sensing data introduces the concept of multimodality, where diverse data types—such as images, text, audio, social media, and video—are integrated to describe the same object from multiple perspectives.

With the increasing volume and variety of remote sensing data from spaceborne, airborne, and ground-based platforms, there is a pressing need to employ artificial intelligence techniques to enhance the processing and analysis of this multimodal data. To address this, a high-precision remote sensing interpretation system has been developed. This system, depicted in Fig. 4-12, operates through a circular process: acquiring multimodal remote sensing data from observation platforms, developing foundational multimodal AI models, applying these models to real-world applications, and then feeding back insights for payload and platform refinement. The system's effectiveness relies on integrating vast amounts of multimodal data, leveraging high-performance computing, and incorporating advanced remote sensing models.

While achieving the first two elements—data fusion and computing power—is largely feasible, a major challenge remains: the need for specialized multimodal AI models that effectively connect remote sensing data with computational resources. These base models are crucial for extracting detailed information from remote sensing data and represent a shift towards an era focused on advanced models that combine statistical, physical, and big data techniques. Recent developments have seen a surge in pre-training methods for remote sensing base models, particularly using spectral data. The introduction of SpectralGPT5 marks a significant milestone as the first base model designed specifically for spectral remote sensing data. Trained on over one million multimodal RS images across various sizes, resolutions, time series, and regions, SpectralGPT5 is currently the largest spectral base model with over 600 million parameters. Its capabilities have shown substantial promise in advancing multimodal remote sensing applications in geosciences, particularly in tasks such as single-label scene classification, multi-label scene classification, semantic segmentation, and change detection.

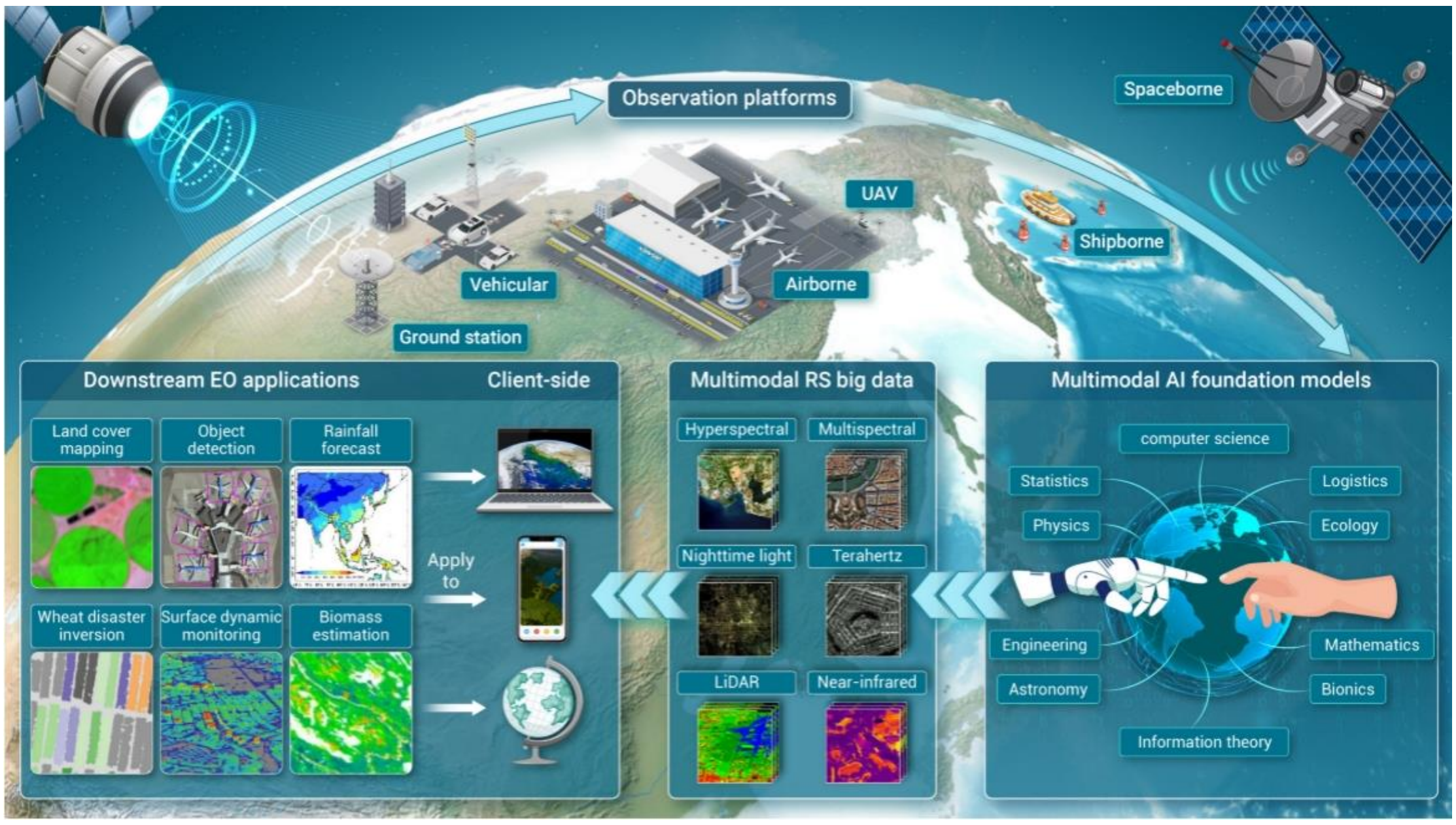

Fig. 4-12 EO remote sensing intelligent interpretation system supported by the basic model of remote sensing big data multi-modal AI

Multimodal AI Foundation Models are poised to revolutionize the analysis of RS big data, harnessing the full potential of diverse RS datasets for various EO missions. These advanced models integrate multiple data types and patterns, offering a robust framework for tackling the complexities of EO applications. By synthesizing information from different modalities, they significantly improve our ability to understand and analyze the Earth's surface and environment. This transition to multimodal base models represents a significant advancement in optimizing RS big data, ushering in a new era of enhanced capabilities and applications in the field.

4.3.3 Shangtang: Integrated Remote Sensing Intelligent Interpretation Foundation model

Nowadays, the application of satellite remote sensing technology greatly reduces the difficulty of obtaining surface information. Also, the application of artificial intelligence technology significantly shortens the time for analyzing massive remote sensing image data. SenseRemote and SenseEarth intelligent remote sensing image interpretation foundation models launched by Shangtang Technology intelligently complete road, building and other information extraction, land classification, aircraft and ship target detection, regional change monitoring, etc. Through deep learning technology, they provide reliable and objective data support for natural resource planning, ecological protection, business decision-making, emergency disaster reduction, etc.

The platform is based on Shangtang AI remote sensing foundation model. It has the foundation of general vision foundation model and high generalization ability. It can interpret different object types, image types, time and spectral segments, and generate image patch effects comparable to manual labeling. SenseEarth 3.0 platform releases 25 semantic segmentation models in Shangtang remote sensing foundation model. These models greatly reduce the running time and save users' time cost. The platform covers 5 types of target monitoring, 4 types of change detection and 2 types of super-resolution algorithms. The average accuracy of Shangtang AI remote sensing foundation model exceeds 80% on the million-level map validation set, which can directly meet the application needs of various business scenarios. Shangtang remote sensing business has served more than 20,000 industry users, covering natural resources, agriculture, finance, environmental protection, photovoltaic, and other fields. Especially in the field of natural resources, the ability advantage of general change detection of Shangtang AI has been widely used in natural resources law enforcement supervision in more than 14 provinces and cities, which has improved the work efficiency of users by 3~5 times. In addition, in the fields of non-agricultural and non-grain monitoring, food security monitoring, photovoltaic roof survey, green finance, airport activity, bare ground dust, etc., Shangtang AI remote sensing foundation model has also been applied on a large scale to provide high-quality interpretation services for users in various industries, helping to reduce costs and improve efficiency. Since SenseEarth 3.0 platform adopts DaaS (Data-as-a-Service) Innovative service mode, users do not need to upload remote sensing data and can directly obtain a set of remote sensing images + structured data, lowering the threshold of intelligent remote sensing applications. In addition, the platform also has GIS data rendering and analysis capabilities, providing free structured data viewing and analysis services. So far, SenseEarth3.0 platform has launched more than 120 "intelligent remote sensing" related products, covering 87146 square kilometers. Shangtang AI remote sensing foundation model is a powerful, accessible, prevalent remote sensing intelligent interpretation platform, providing efficient remote sensing interpretation solutions for various industries.

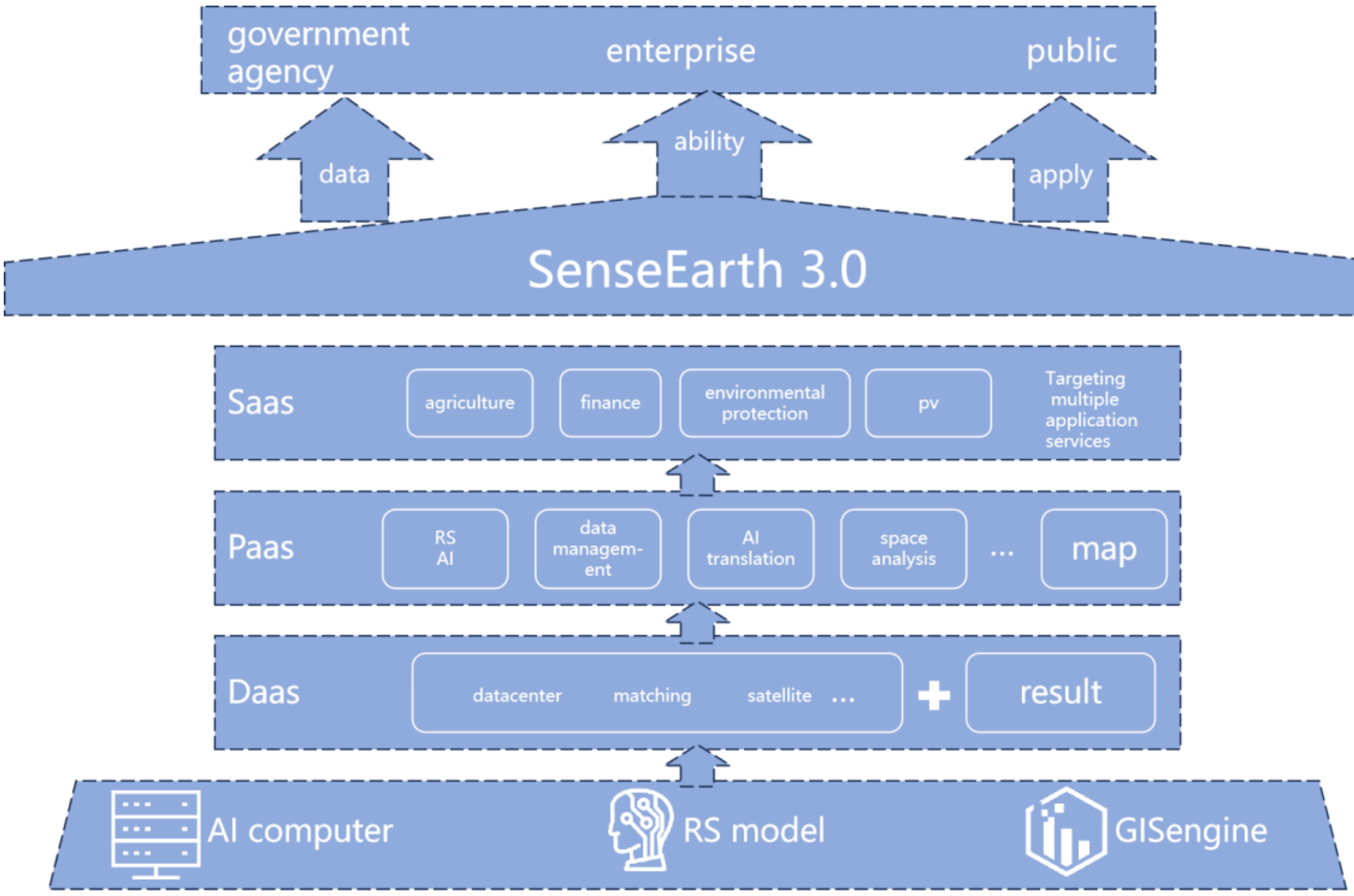

Fig. 4-13 From traditional remote sensing application model to intelligent remote sensing application innovation service model

In the interface of SenseEarth, users can slide and circle operations, wait a few seconds, and get the interpretation results of selected image areas in real time, which is simple and efficient. Roads, vehicles, or various land use categories, including farmland, forest, grassland, shrub, water, impermeable layer, wasteland, snow, wetland, etc., are visually visible. Moreover, satellite image data (Beijing and Shanghai) on SenseEarth platform are updated at high frequency every month. Users can compare the changes of remote sensing images of target areas in different periods of time by month. SenseEarth will automatically present the changes in map spots for visual display.

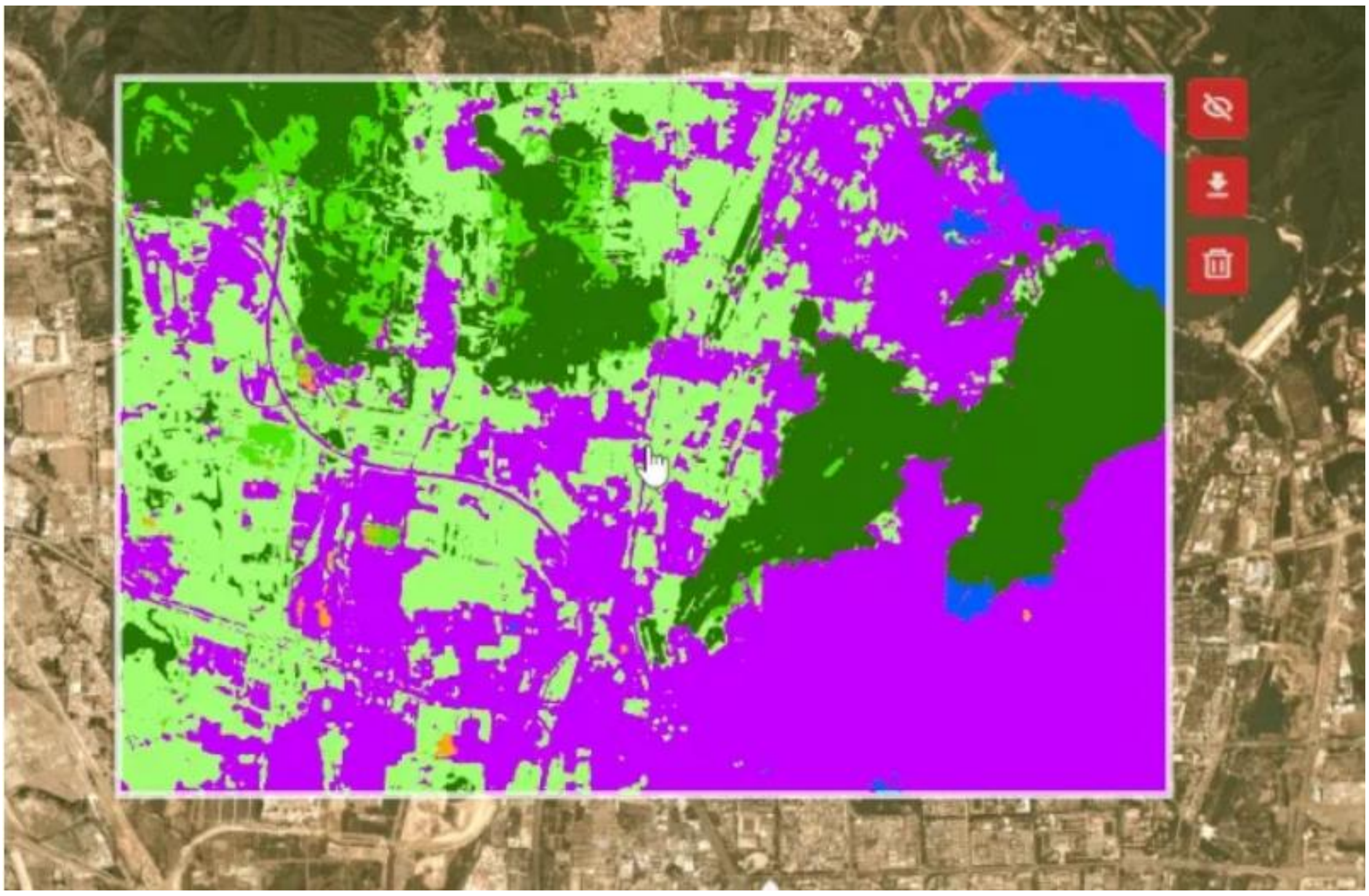

Fig. 4-14 SenseEarth intelligent remote sensing image interpretation platform land classification

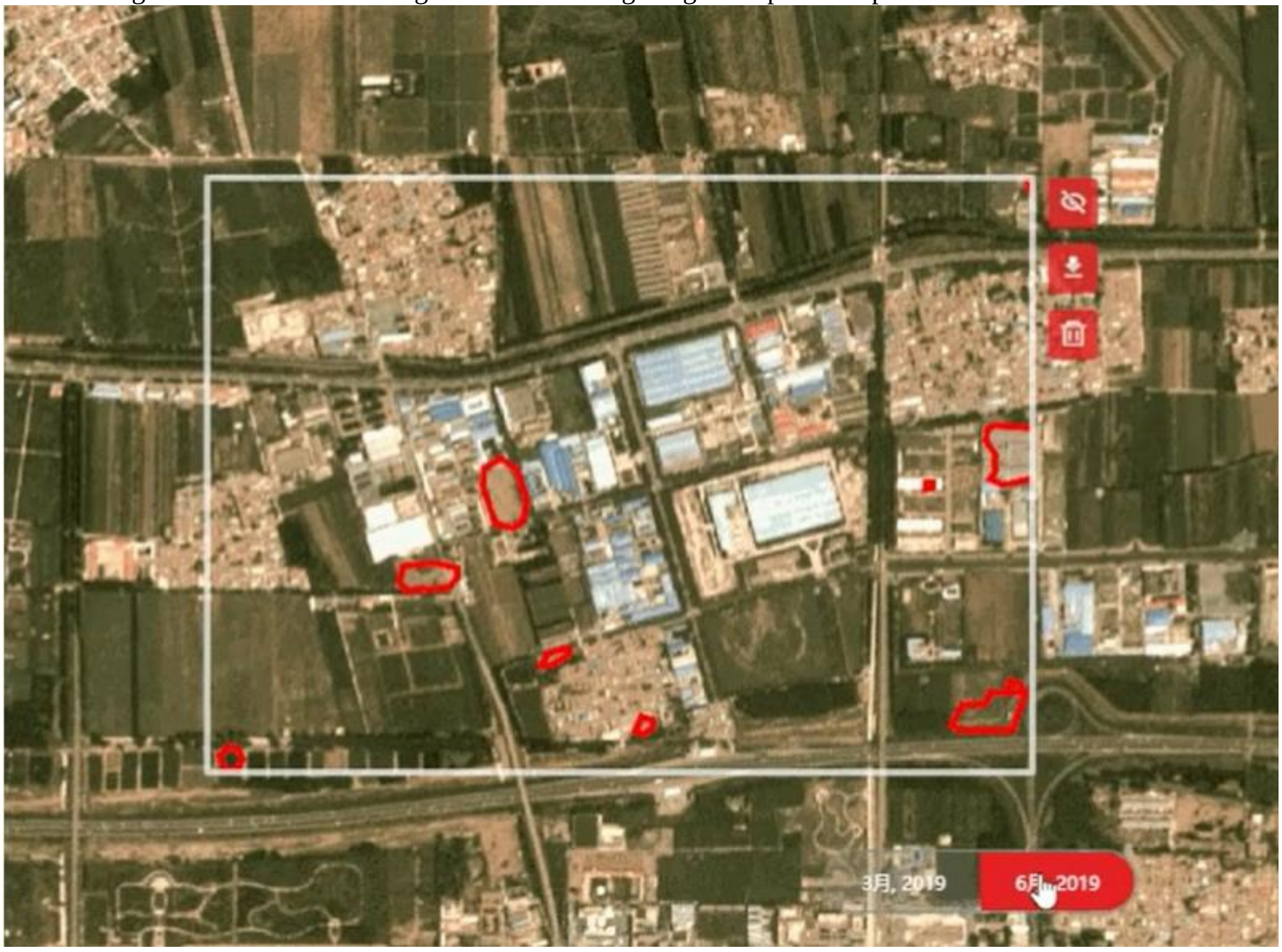

Fig. 4-15 SenseEarth intelligent remote sensing image interpretation platform changes monitoring

As an efficient, real-time, and easy-to-use intelligent remote sensing image interpretation platform, SenseEarth can be regarded as the terminal application of SenseRemote remote sensing image intelligent interpretation algorithm of Shangtang Technology, which fully reflects its high precision and high efficiency. Under multi-scene interpretation, SenseRemote's accuracy is better than 90%; SenseEarth can interpret 5000p × 5000p images in only 20 seconds, making up for the problems of low revisit rate and low timeliness of interpretation technology of traditional satellite images, creating powerful data analysis and insight capability without manual intervention. With the continuous updating of the platform, SenseEarth will add more city and regional data in the future, support multiple resolutions and higher frequency data update cycles, and integrate functions such as building extraction, aircraft detection, and water extraction into interpretation. In addition, it can also support users to upload image data for interpretation, fully reaching personalized needs. Based on SenseEarth's powerful interpretation capabilities and excellent interactive experience, road information extraction will be more convenient and efficient, thereby accelerating the production of auxiliary electronic maps and urban traffic planning; The accuracy of land use classification will be higher. By generating various thematic maps such as cultivated land, forest land, human activity area, and water body, high-quality basic geographical information will be provided to assist land resources survey and geographical survey; the results of land change will be more precise to follow the impact of human activities or natural changes on the surface more macroscopically and comprehensively. This will provide reliable basis for urban construction management and ecological environment monitoring. In addition, Shangtang Technology also demonstrated the "Intelligent Remote Sensing City Integrated Solution" based on SenseRemote, which assists the comprehensive construction of smart cities through the interpretation and analysis capabilities of remote sensing images supported by AI. It provides dynamic monitoring data for customers in units of "days" and grasps the dynamics and changes of cities. At the same time, the solution can optimize traditional business processes and improve urban planning, construction, and management to an intelligent stage marked by rapid investigation and monitoring, scientific diagnosis and analysis, and efficient decision-making and management. With the four advantages of "high frequency, high definition, high precision, and high efficiency", it brings more imagination to the development of smart cities in the future. Artificial intelligence enables remote sensing image interpretation, improving the insight and efficiency of geographic information and space science and technology industry. It promotes fine management and high-quality development, benefiting people’s livelihood and society.

#### 4.3.4 Cangling ImageBot: Integrated Intelligent Interpretation and Foundation Model Application

The Cangling AI team, from Aerospace Information Research Institute, Chinese Academy of Science (AIRCAS), has developed the first general integrated foundation model "Cangling ImageBot" for remote sensing intelligent interpretation and applications. The team has overcome key technologies based on deep learning for remote sensing image classification and segmentation, object detection and recognition, and change detection. They have developed a full-process, full-system intelligent remote sensing analysis platform, which enables automatic, rapid, and precise production from raw images to thematic information.

For global-wide object detection, thematic mapping, and remote sensing classification, Cangling AI team have established a large-scale remote sensing knowledge database with millions of samples, including thematic targets such as energy, minerals, environmental protection, infrastructure, as well as features from land-use and land-cover classes. Additionally, they have created a nationwide comprehensive remote sensing sample library that full coverage of China, full land cover classes with multi-source and multi-temporal images.

The team has established an integrated intelligent interpretation and application system that connects all processes and links from raw data processing, sample production, model design, and model training to product production and result release. This system can realize fully automated integrated application services from raw satellite images to thematic products.

Cangling ImageBot, based on a massive sample database, independently developed the remote sensing inference framework CanglingInferEngine (CLInferEngine) to reduce the coupling of the remote sensing inference framework and to facilitate deployment and arrangement in information engineering scenarios. This framework strips away training functions and other redundant data, focusing solely on task prediction, thus enhancing the professionalism and accuracy of remote sensing applications.

The Cangling AI inference engine, designed to improve information engineering quality, incorporates a comprehensive set of optimized functional modules and management mechanisms for remote sensing engineering. These include a distributed data automatic distribution mechanism, a distributed hardware storage architecture, an optimized memory and video memory mapping mechanism, automatic large-image cropping and stitching, test-time enhancement mechanisms, automatic matching of geographic coordinate information, lossless data compression, automatic coloring, and post-processing vectorization workflows. These modules enable a complete and automated end-to-end remote sensing inference process without human intervention.

At the remote sensing task level, the remote sensing task is composed of four relatively independent foundation models: object detection (CanglingDetection, CLDet), semantic segmentation (CanglingSegmentation, CLSeg), change detection (CanglingChangeDetection, CLCD), and remote sensing inversion (CanglingInversion, CLInv).

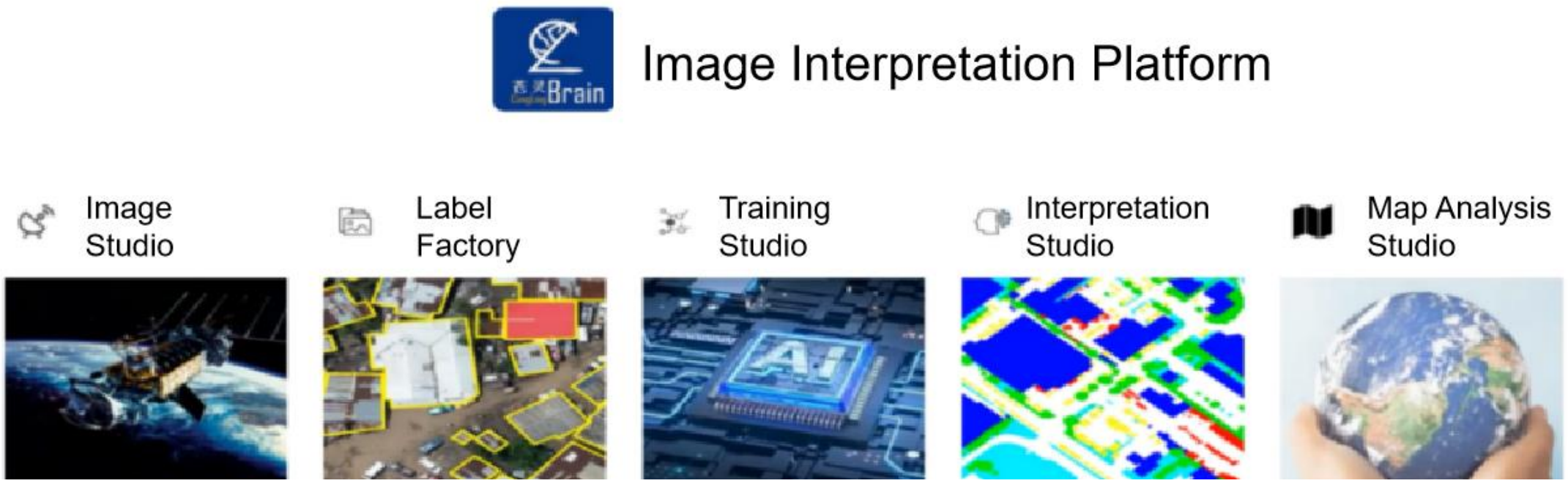

Fig. 4-16 Cangling Imagebot integrated intelligent interpretation and application platform

(1) Object Detection Model (CanglingDetection, CLDet)

Using multisource remote sensing images and text data containing geographical information descriptions, a multimodal dataset was constructed. By combining remote sensing visual feature processing with geographic text matching, a remote sensing object detection foundation model was developed for regional object localization. Through iterative training on a nationwide scale, it supports the recognition of objects in high-resolution satellite remote sensing images (1 m and 2 m) with three and four bands across multiple time periods. Recognizable targets include tailings, wind turbines, thermal power plants, steel plants, cement plants, and sewage treatment plants. The integration of remote sensing visual features and geographic information greatly enhances the interpretability of regional target recognition, providing support for applications in national energy facility development activities, disaster emergency monitoring, and more.

(2) Semantic Segmentation Model (CanglingSegmentation, CLSeg)

Based on large-scale multisource remote sensing images and textual attribute knowledge descriptions, a remote sensing classification interpretation foundation model was developed by integrating spatio-temporal-spectral features of remote sensing data and text knowledge prompts. This model supports common resolutions (0.5 m, 2 m, 10 m, 15 m, 30 m) of optical, SAR, and hyperspectral multisource satellite remote sensing data, enabling land use/land cover classification of all types of land-use/land-cover features, as well as single-element feature classification and extraction of various specified objects of interest such as arable land, forests, residential areas, roads, and water bodies. It has completed the fine classification and thematic product production of nationwide 2m domestic high-resolution images over five periods, the generation and application of one period of 0.8m fine thematic products, 40 years of Landsat series classification products, and five periods of Sentinel data products.

(3) Change Detection Model (CanglingChangeDetection, CLDet)

A multimodal dataset was constructed based on large-scale, long-term, multisource remote sensing images and text descriptions. By integrating vision foundation models, text foundation models, and remote sensing data features, a remote sensing change detection foundation model was developed. Through iterative training on a nationwide scale, the “Cangling Sentinel” was developed. It supports satellite remote sensing data from 0.1 m to 30 m, including optical, SAR, multispectral, and hyperspectral data. The output information includes change type and change status text descriptions, change map spots, and position vectors. Cangling Sentinel has been widely used in various complex scenarios for remote sensing change detection tasks, serving applications such as national ecological red line monitoring and disaster risk element extraction.

(4) Remote Sensing Inversion Model (Cangling Inversion, CLInv)

A multimodal dataset was constructed based on large-scale, long-term, multisource remote sensing images and extensive ground-truth data provided by weather stations, hydrological stations, and agricultural experiment stations. By integrating vision foundation models, text foundation models, and remote sensing data features, and through iterative training nationwide, the remote sensing quantitative inversion foundation model CLInv was developed. CLInv supports multisource, multi-resolution data input from optical, radar, hyperspectral satellite data, and UAV data, outputting various surface and biological parameters such as vegetation index, soil moisture, surface temperature, leaf area index, crop type, crop growth, and estimated yield. CLInv has been widely applied to various quantitative inversion tasks such as growth detection and pest and disease warning, providing support for agricultural detection and forestry resource management.

The Cangling ImageBot integrated intelligent interpretation and application foundation model covers remote sensing image classification and feature extraction, object detection, change detection, and quantitative inversion. It enables fully automated integrated application services from raw satellite images to thematic products. Over 50 types of thematic products have been produced and applied globally, providing technical or product services to more than 60 organizations, including the Ministry of Emergency Management, Ministry of Ecology and Environment, Ministry of Natural Resources, and Ministry of Agriculture and Rural Affairs. It is the first remote sensing big data intelligent recognition system reported by CCTV News.

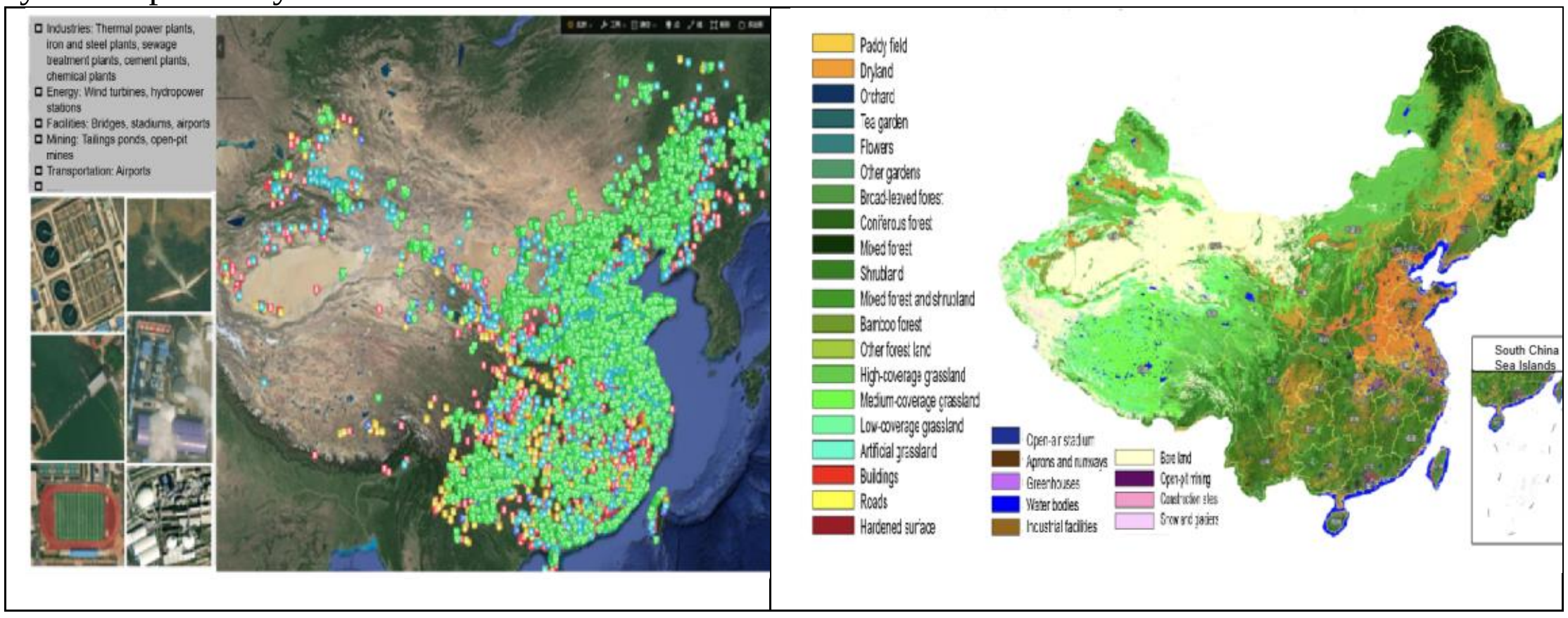

Fig. 4-17 National high-precision target and classification product diagram produced by Cangling Imagebot

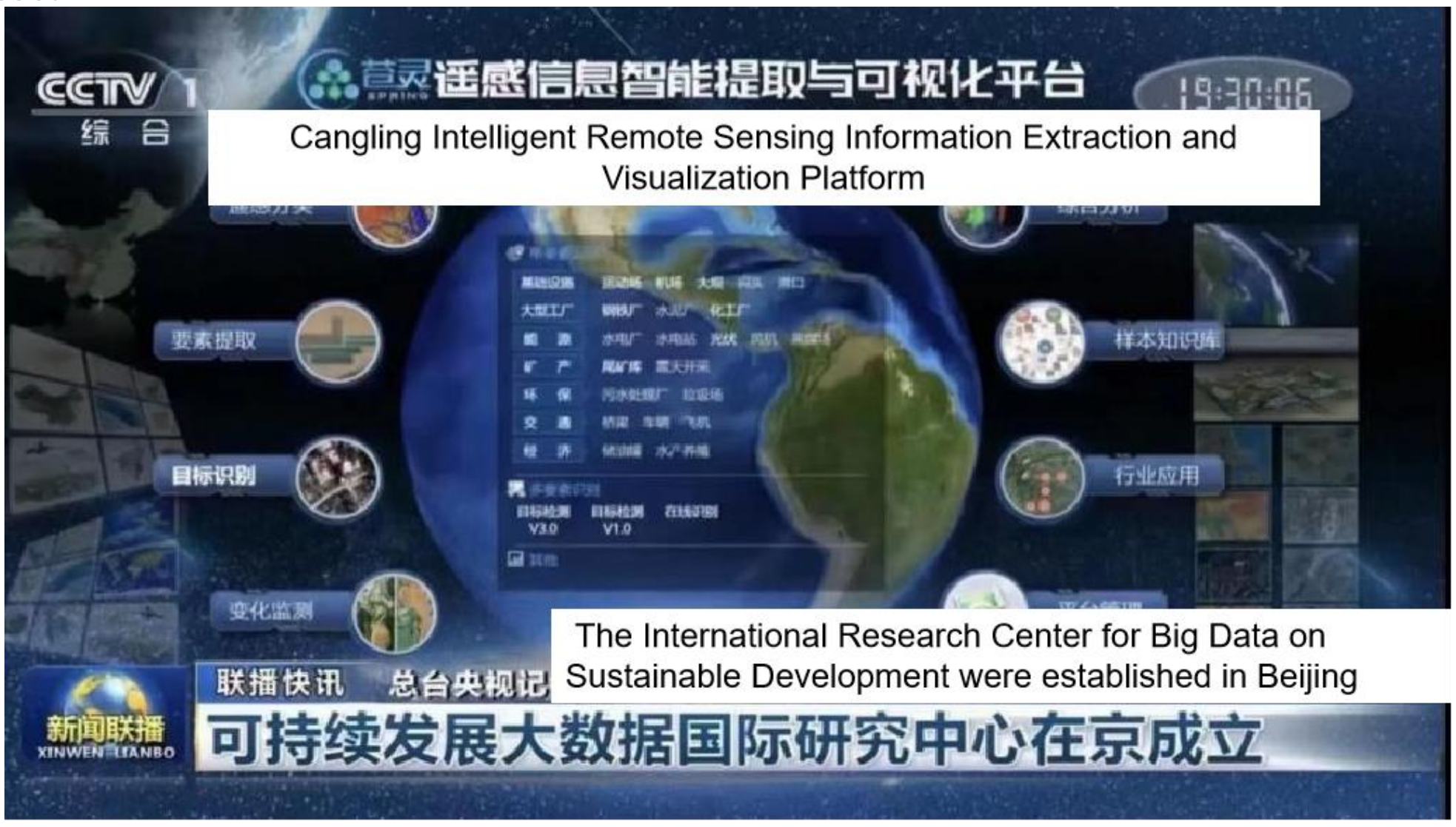

Fig. 4-18 Cangling ImageBot System reported by CCTV News

## 4.4 Intelligent Foundation Model of Urban Transportation and Public Facilities Services

### 4.4.1 TrafficGPT: A Foundation Model of Urban Traffic Management

As chat technologies become more prevalent, large-scale language models have showcased remarkable abilities in common sense, reasoning, and planning, offering valuable insights for urban traffic management and control. Despite these advancements, such models struggle with numerical data and simulations, limiting their effectiveness in tackling traffic issues. Specialized traffic models exist but are often tailored for specific tasks with restricted input-output interactions. Integrating these models with large language models can enhance their capacity to address complex traffic challenges and deliver better recommendations. To address this, the TrafficGPT model was proposed, combining ChatGPT with traffic base models (Fig. 4-19). This integration brings several key improvements: it enables ChatGPT to analyze and interpret traffic data for urban transportation system management, breaks down complex tasks into manageable parts using transportation infrastructure models, supports traffic control decisions through natural language interactions, and incorporates feedback for refining results. By merging large language models with traffic expertise, TrafficGPT not only enhances traffic management but also introduces a novel way of leveraging AI capabilities in this domain.

While the latest advances in AI and NLP have created new possibilities, large language models like ChatGPT, despite their impressive reasoning and planning skills, still struggle with the intricacies of traffic management. To overcome these limitations, TrafficGPT integrates ChatGPT with traffic base models, allowing it to handle complex operations and offer more insightful suggestions. TrafficGPT utilizes multimodal data sources, including video, detector, and simulation data, with an intermediate database manager layer facilitating access. The LLM at the outer layer manages user needs and task execution through the Traffic Foundation Models (TFM). This integration aims to transform traffic management by harnessing AI's potential to address the complex challenges of traffic data analysis and decision-making.

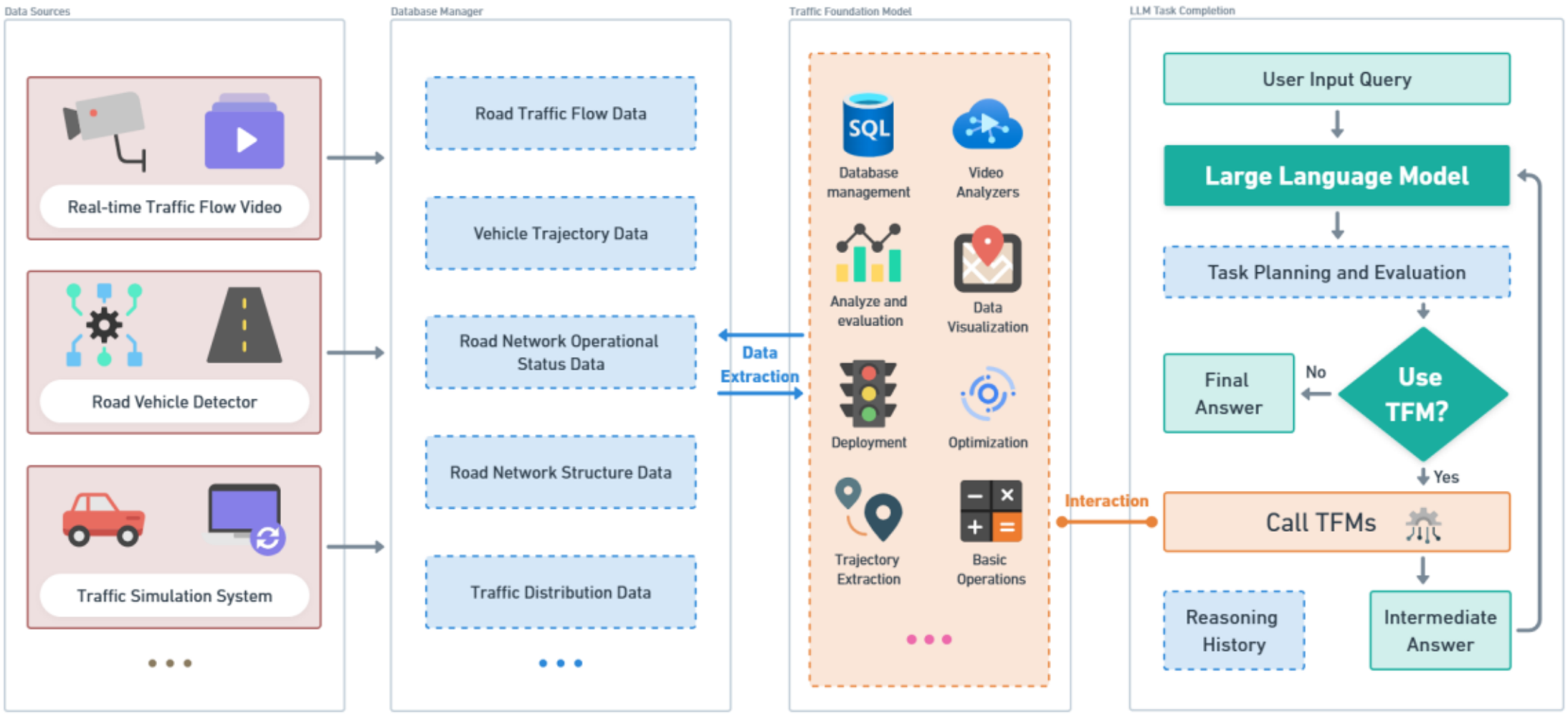

Fig. 4-19 TrafficGPT Framework

Integrating Large Language Models (LLMs) into complex traffic management tasks, such as analyzing and processing diverse traffic datasets, remains an underexplored area in current research. This is primarily due to the challenges LLMs face in handling numerical data effectively. The limited application of LLMs in addressing complex traffic problems and aiding decision-making underscores a significant research gap. Traffic management tasks often involve various subtasks that require LLMs to process and analyze numerical traffic data. Efficient processing, analysis, and visualization of this data are crucial for supporting traffic management decisions.

Although Traffic-Based Models (TFMs) are typically designed for specific tasks with single input-output interactions, the presence of numerous mature TFMs and the potential to integrate multiple TFMs provide a promising foundation for utilizing LLMs in solving complex traffic issues. In this context, TrafficGPT allows users to start tasks through natural language inputs. These inputs serve as prompts that are managed through a defined framework for the LLM agent, outlining the agent's working mechanism, key considerations, and available tools. Historical dialogue contexts can also be integrated to enhance multi-turn interactions.

The prompt management step includes user requests, system instructions, available tools, inference history, and dialogue history. By combining these elements, the agent is provided with the necessary context to deconstruct and execute tasks effectively. The LLM's cognitive abilities allow it to understand natural language prompts, perform deductive reasoning, and merge task requests with available tools and inference history. The selected TFM is then called upon to perform various tasks such as database retrieval, analysis, visualization, and system optimization, generating the required results.

During tool execution, the LLM agent retrieves output via an API interface and integrates it into an intermediate response for further processing. For tasks requiring multimodal outputs, structured content is provided in Markdown format, and additional files like images and data are shared as file paths. The agent assesses task completion by comparing user requests with intermediate responses and uses the LLM's capabilities to generate a final response. This response is then delivered to the user, and conversation records are stored to provide context for future interactions.

This comprehensive framework aims to revolutionize traffic data analysis by integrating large-scale language models with intelligent transportation systems. The following sections will delve into the key elements of this framework, offering detailed insights into their significance and components.

#### 4.4.2 Prediction of Shared Bicycle Demand Based on Irregular Convolutional Neural Networks

In recent decades, shared bicycles have gained significant attention as a component of urban transportation. As an eco-friendly mode of transport for short urban trips, bike-sharing services help reduce carbon emissions and enhance the "last mile" connection to public transit systems. During the COVID-19 pandemic, shared bicycles proved to be a more resilient transportation option, alleviating concerns about crowded public transport. Due to the role of bike-sharing services in urban mobility, accurately forecasting demand is essential for effective daily rebalancing operations. Numerous studies have sought to develop models to estimate city-wide bicycle demand using both traditional methods and machine learning techniques.

In recent years, deep learning methods have become popular for predicting short-term traffic demand, with a major focus on modeling the spatiotemporal dependencies of travel patterns. Convolutional neural networks (CNN) and recurrent neural networks (RNN) are often combined to effectively capture spatial and temporal information. Typically, CNNs use conventional convolutional filters to extract spatial features from input data, while RNNs leverage temporal dynamics from previous elements in a sequence to predict future demand. To enhance the capture of spatiotemporal data, several hybrid deep learning models integrating CNN and RNN architectures have been developed, achieving strong results in various traffic forecasting applications.

However, CNNs have limitations in modeling the spatiotemporal characteristics of shared bicycle demand. While CNNs perform well in image object detection due to the high correlation between adjacent pixels of the same object, bicycle usage in neighboring urban areas can vary greatly because of differences in travel behavior and the built environment. Conversely, areas that are geographically distant may show similar bicycle usage patterns due to similar temporal rhythms. Conventional CNN architectures, with their regular-shaped filters, struggle to capture these similarities in distant urban areas, which could be beneficial for accurately predicting bike-sharing demand. By integrating these distant yet similar patterns into the predictive model, it is possible to enhance the accuracy and reliability of forecasts. To address this gap, this study proposes an irregular convolutional long short-term memory model (IrConv + LSTM) designed to improve short-term demand predictions for urban bike-sharing systems.

To fill the research gap, this paper introduces an irregular convolutional long short-term memory model (IrConv + LSTM) to improve the short-term demand prediction of urban bicycle sharing systems. This model uses an irregular convolutional framework to capture the relationship between bicycle usage in distant urban areas. Given a predicted region, the model can perform irregular convolution operations on its semantic neighbors, which refer to places displaying similar bicycle usage patterns over time. Then, model use Pearson correlation coefficient (IrConv + LSTM: P) and dynamic time warping (IrConv + LSTM: D) as similarity measures to identify semantic neighbors in the predicted region. The two variants of the proposed model (IrConv + LSTM: P and IrConv + LSTM: D) and several benchmark models were evaluated and compared on bike sharing systems in five cities, including a pileless bike sharing system in Singapore and four stations respectively located in Washington D.C., Chicago, New York, and London.

Fig. 4-20 illustrates the overall structure of the model developed in this study. The model comprises three distinct modules, each sharing the same architecture. These modules process different sets of historical bicycle-sharing demand data as input. Instead of utilizing the entire history of observations for training, this model selectively identifies key time periods that vary in recency relative to the target prediction period, feeding these selected data sets into the three modules. This selective approach helps to minimize the impact of redundant information from historical data and significantly reduces the computational complexity involved in training the model. Research has shown that this method outperforms models trained with a full set of historical observations.

As depicted in Fig. 4-20, each module incorporates a three-layer irregular convolutional structure designed to capture the unique characteristics of urban bicycle demand. The output of these irregular convolutions is flattened into a vector sequence, which serves as input to the LSTM model, enabling it to extract temporal features from the data. The outputs from the three modules are then combined in a feature fusion layer. This fused output is processed through a nonlinear activation function, which generates the final predictions. These predictions are then compared with the actual observed values, and the resulting differences are used in the loss estimation and backpropagation processes to refine and update the model parameters.

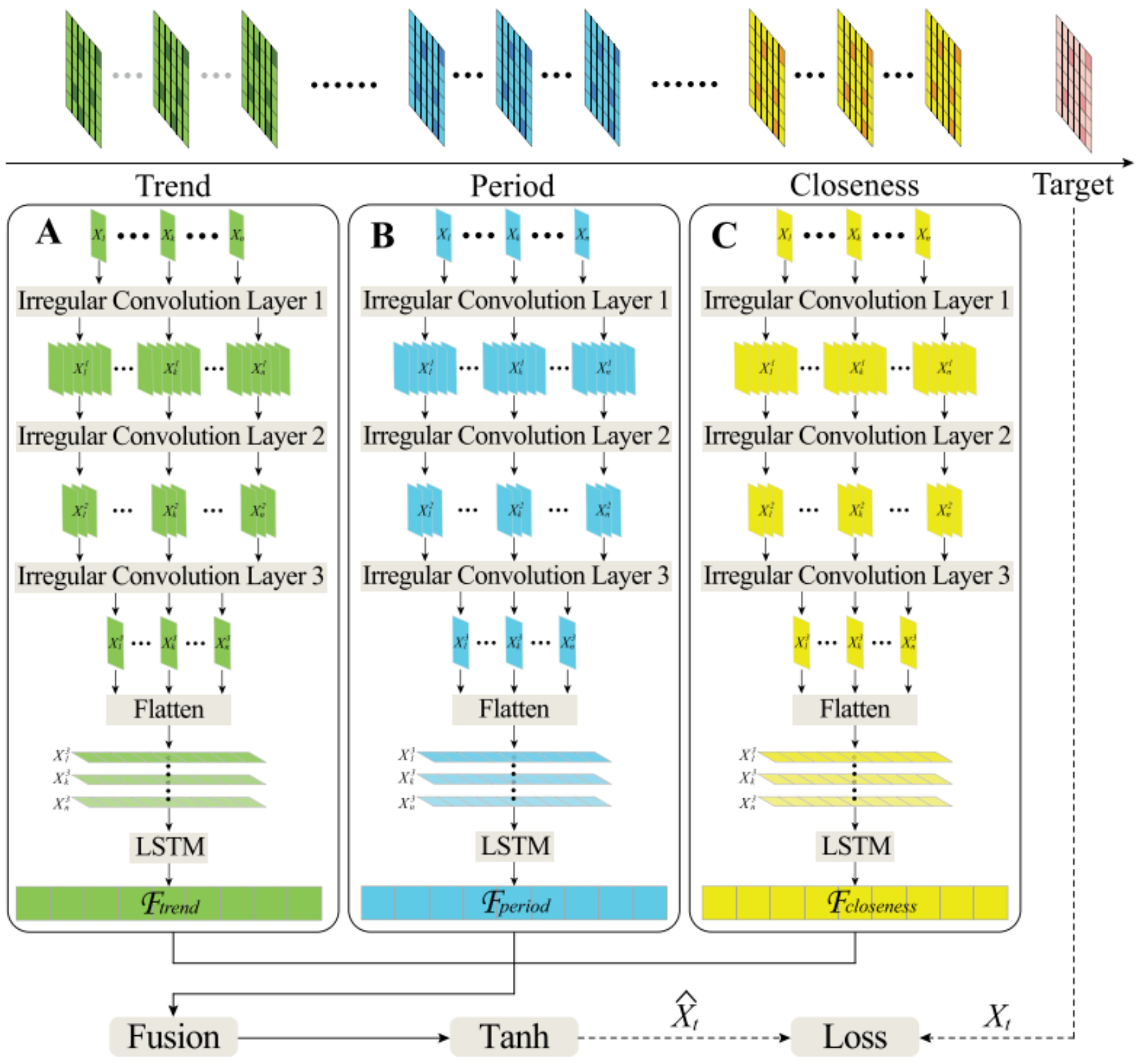

Fig. 4-20 The overall structure of the model

This study introduces an Irregular Convolutional Long Short-Term Memory Model (IrConv + LSTM) to enhance the accuracy of short-term demand forecasting for shared bicycles. By employing an irregular convolutional design, this model modifies conventional CNNs to uncover hidden relationships between "semantic neighbors." The model was tested against several benchmark models across five study locations, which included a dockless bike-sharing system in Singapore and four station-based systems in Chicago, Washington D.C., New York, and London. The results showed that the IrConv + LSTM model outperformed the benchmark models in all five cities, demonstrating strong predictive capabilities across areas with varying levels of bicycle use and during peak periods. The findings suggest that looking beyond traditional spatial neighbors can significantly improve predictions for short-term travel demand in urban bike-sharing systems.

4.4.3 Application of Foundation Model for Deconstructing the Contribution of Urban Facilities

The automobile era has severely reduced the quality of urban life through expensive travel and significant environmental impacts. A new urban planning paradigm must become the core of the roadmap for the coming years, in which, residents can reach their basic living needs by bicycle or on foot within minutes. The deconstructed foundation model of urban facility distribution presents new insights into the interaction between facility distribution and population to maximize accessibility on the existing road network. Survey results from six cities show that by integrating spatial data intelligent models and reallocating facilities, combining multi-source data, and conducting holistic analysis of the collected data. Hence, travel costs can be reduced by half. In an optimal scenario, the average travel distance can be modeled as a function of the number of facilities and population density. As an application of this finding, the number of facilities needed to achieve the desired average travel distance for a given urban population distribution can be estimated.

Greenhouse gas emissions are generated by heating and cooling networks in buildings and widespread gasoline transportation, leading some cities to become unbreathable at a time when climate impacts on urban life are very evident. When transportation becomes the largest source of CO2 emissions, new approaches to urban space utilization are needed, requiring a better understanding of the spatial distribution of facilities and population. The advent of the information age and the revolution in online maps enable the study of human interactions with their built and natural environments on a global scale. Pioneering work in multi-city studies reveals proportional relationships between population, facility distribution, and socioeconomic activities at a macro scale. For instance, cities with larger populations have higher per capita consumption efficiency. Also, the occupational diversity of the population can be modeled as social networks embedded in space. However, systematically understanding the interactions between urban morphology, facility distribution, and accessibility at multiple scales remains a challenging task. By introducing location-based social network datasets, the demand for different types of cultural resources has been identified. urban areas lacking venues have been pinpointed. While efforts have been made to solve the optimal configuration for specific cities, there is still a lack of systematic understanding of optimal facility configuration from an urban science perspective.

To address this issue, a multi-city study was proposed to measure the accessibility of urban neighborhoods to different types of facilities through road networks and to investigate the role of population distribution (Xu et al., 2020). While on a broad scale, travel costs can be replaced by the Euclidean distance from residents to facilities, road networks and geographical constraints play a crucial role in the movement of people within cities. It is well known that road network attributes affect residents' daily travel, urban morphology, and accessibility. As a supplement to most studies on commuters' travel costs, this work also analyzed the road network distance from individuals to the nearest various amenities, dividing the space into high-resolution blocks of 1 square kilometer each. For each city and facility type, existing facilities were optimally reallocated. The results were compared with empirical distributions. It was observed that the accessibility of some blocks increased in the reallocation, while the accessibility of others decreased. This means that to maximize the use of existing facilities and achieve more equitable accessibility, some neighborhoods will benefit while others will have facilities removed accordingly. At a diversified urban level, the gap between empirical facility distribution and optimal planning provides new insights into evaluating the quality of urban facility planning.

Studies on the empirical distribution of facilities in multiple cities across different regions of the world collected population data with a spatial resolution of 30 arc seconds (approximately 1 $km^2$ near the equator) for each city block and gathered road networks. Facilities were collected from service applications. The boundaries of each city were delineated along metropolitan areas, including urban and rural regions. In cities, the distance people travel within the road network is constrained by infrastructure and landscape. To quantify the accessibility of populations to facilities, routing distance became the representative measure of accessibility from residences to each amenity. The study results confirmed that the optimal strategy based on Euclidean distance is similar in cost to the actual distribution of facilities but is far less effective than the optimization strategy based on routing distance. Accessibility refers to the level of service facilities provided to residents. In network science, accessibility is defined as the ease of reaching points of interest within a given cost budget. One of the core issues in urban facility planning is how to reasonably allocate urban facilities to maximize overall accessibility. Reallocating facilities by minimizing the total routing distance from the population to the nearest facility can effectively address this issue.

Due to differences in morphology, economy, and population distribution among cities, the interaction between population and facility distribution presents a challenge for future urban planning. The accessibility of facilities is constrained by their availability, road networks, and transportation modes. While managing daily commuting and transit-oriented development, the distribution planning of facilities in different cities should focus on achieving a paradigm shift towards walkable cities. The empirical conditions within cities do not follow a continuous approximation of the power law of population density because facilities are not evenly planned. The number of facilities is significant compared to the number of population blocks. The study found that centralized cities require fewer facilities than polycentric cities to achieve the same level of accessibility. The application of this framework is the optimal way to reallocate resources for providing emergency services.

The optimal planning of facilities in the deconstructed foundation model of urban facility distribution assumes that all residents have equal demand for resources with accessibility measured by place of residence. In reality, socioeconomic segregation in cities leads to heterogeneous demand for resources. Cities with different social systems and levels of economic development exhibit different demands for various types of facilities, necessitating the consideration of economic factors. On the other hand, due to people's time-varying mobility behaviors, their needs are naturally dynamic, changing over time and space. All these factors result in complex interactions between facility allocation and residential areas, which can become an important avenue for future research. Another important avenue is to consider the limited capacity of facilities in optimal planning.

#### 4.4.4 Deep Reinforcement Learning for the Location of Urban Facilities

Urban space computing is a methodological approach for investigating the characteristics, patterns, and complexities of urban areas. This approach involves evaluating urban spatial features through tools like geographic information systems (GIS), statistical methods, and spatial data analysis techniques. One critical component of urban space computing is urban space optimization, which focuses on achieving the best possible spatial layouts by applying various constraints to minimize costs or maximize specific objectives based on actual urban conditions. By providing scientifically-based solutions, spatial optimization plays a crucial role in enhancing urban planning, improving transportation systems, and promoting sustainable development.

The discrete facility location problem is a well-known NP-hard issue in operations research and represents one of the key challenges in spatial optimization. The Maximal Covering Location Problem (MCLP), initially introduced by Church, is a fundamental problem in this domain with broad applications in logistics, urban planning, and related areas. The goal of the MCLP is to choose a set number of facilities from a pool of potential sites to maximize the coverage of demand points. This involves determining the optimal locations for facilities so that the maximum number of demand points falls within their service areas, taking into account the reachability and service range of each facility.

Research on the MCLP has led to the development of several solution methods, including exact algorithms, approximation algorithms, and heuristics. Exact algorithms are effective for finding optimal solutions in smaller-scale problems but become impractical for larger instances due to their high computational demands. Approximation algorithms, on the other hand, provide solutions that are close to optimal, with a known approximation ratio, $\alpha$, indicating the difference between the suboptimal and optimal solutions. Heuristic methods are faster and more scalable but do not guarantee finding the optimal solution.

MCLPs are commonly formulated as mixed-integer linear programming (MILP) problems, allowing solver-based methods to be applied effectively. Solvers like Gurobi, Cplex, OR-tools, SCIP, and COPT are popular choices for tackling these problems, with SCIP offering an open-source alternative. These solvers employ a range of specialized algorithms and heuristics to efficiently address MCLPs within certain problem sizes. However, due to the NP-hard nature of MCLPs, finding exact solutions remains a significant challenge.

In recent years, deep learning models have demonstrated their ability to extract meaningful features from complex data. Building on this, a novel algorithm has been created to solve the MCLP using deep reinforcement learning, as depicted in Fig. 4-21. This algorithm incorporates attention mechanisms to capture the relationships between demand and facility points, allowing the deep learning model to effectively solve the MCLP. When compared to genetic algorithms, this deep reinforcement learning approach achieves greater solution accuracy while maintaining computational efficiency. To evaluate the robustness of this algorithm, it was tested on both synthetic and real-world datasets. The experimental results highlight the algorithm's effectiveness in solving MCLPs, showcasing its valuable contributions to urban spatial optimization.

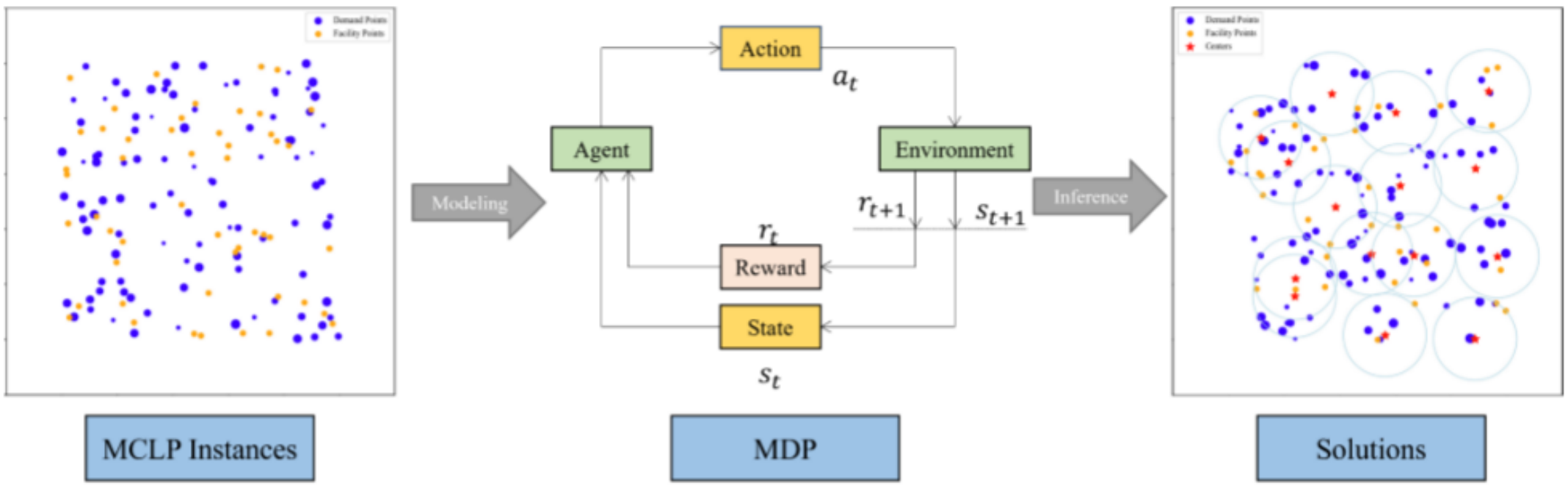

Fig. 4-21 The workflow

The main challenge in the Maximal Covering Location Problem (MCLP) is devising an optimal strategy to select $p$ facility locations to maximize the coverage of demand points. To address this, a constructive approach is used to generate solutions by modeling the problem as a Markov Decision Process (MDP). Deep reinforcement learning algorithms are then employed to train models that assist decision-makers in choosing facility locations step-by-step until a final solution is achieved. The deep learning framework utilizes an Encoder-Decoder architecture with multi-head attention layers in both the encoder and decoder.

In essence, MCLP is a crucial spatial optimization problem with wide-ranging applications, including the placement of public facilities like parks and hospitals, as well as emergency facility allocation. This problem is vital for urban planning and sustainable development. However, due to its NP-hard nature, finding the optimal solution for MCLP remains a complex challenge. This study presents a novel deep learning-based algorithm designed to tackle this issue. The approach incorporates attention mechanisms to identify intricate relationships between demand points and facility locations. Using deep reinforcement learning, the model learns an effective strategy for selecting facility locations to maximize coverage. Once trained, the model provides fast and efficient solutions for MCLP across various problem scales. Experimental results using both synthetic and real-world data demonstrate the algorithm's effectiveness. Compared to the Gurobi solver, this method offers quicker solution times and achieves smaller deviations from optimal solutions than genetic algorithms.

For future research, several avenues could be explored. Integrating additional constraints, such as facility capacity and budget limits, could expand the algorithm's applicability. Investigating the scalability of the approach to handle larger instances of MCLP would also be valuable. Additionally, conducting comprehensive case studies across different real-world scenarios would further illustrate the algorithm's robustness and effectiveness.

### 4.5 Application of Resource and Environment Foundation model

4.5.1 Distribution and Dynamics of Global Soil Inorganic Carbon

Soil inorganic carbon (SIC) has traditionally been considered a stable carbon reservoir with a turnover time spanning thousands of years. However, this perspective is changing as new evidence emerges showing that SIC dynamics are accelerating, leading to significant disturbances within just a few decades. This includes an increasing trend in alkalinity in major rivers globally and the accumulation of new bicarbonate ions from soil in groundwater. These changes in SIC are affecting soil properties, including its ability to buffer acidity, the availability of nutrients, plant productivity, and the stabilization of organic carbon. This highlights the critical role SIC plays not only in carbon sequestration but also in maintaining soil health, ecosystem services, and overall ecosystem functionality.

SIC in soils is made up of lithogenic, biogenic, and pedogenic carbonates. Pedogenic carbonate forms through the dissolution of solid minerals into cations, which then reprecipitate with dissolved inorganic carbon (DIC) as carbonate minerals within the soil. This process is influenced by soil hydrology and microenvironmental conditions that govern the equilibrium of the carbonate system. Water movement plays a crucial role by either reprecipitating SIC into deeper soil layers or removing DIC through drainage, thereby influencing carbon dynamics in both freshwater and oceanic systems. SIC serves as a link between organic and inorganic carbon processes within the carbon cycle, connecting land, water, and atmospheric systems across various timescales, from rapid carbonate reactions to Earth's geological history. Despite its importance, SIC is often overlooked in carbon budgeting, leaving many aspects of its size, distribution, influencing factors, and future uncertain. Addressing these gaps and clarifying the role of SIC in the global carbon cycle is increasingly urgent, especially given the rapid pace of carbonate reactions and the massive global SIC stock, which ranges from 695 to 940 billion tons of carbon (GtC) in the top 1 meter of soil and exceeds 1000 GtC in the top 2 meters. Even minor changes in this stock could have significant effects on atmospheric CO2 levels and global climate change.

To better understand SIC, we have compiled a global database that includes 223,593 measurements from 55,077 soil profiles, drawing on a wide array of field measurements, national inventories, coordinated field studies, and standardized global soil databases. This comprehensive dataset covers all 12 soil orders recognized by the United States Department of Agriculture, and includes samples from nearly every continent, climate zone, and biome around the world (see Fig. 4-22). The data reveal a highly variable SIC content, ranging from 0 to over 100 g(C) per kilogram of soil in the top 2 meters (Fig. 4-22 A and C), with 42% of samples containing no detectable SIC (0 g(C) per kilogram of soil) (Fig. 4-22). For soils that do contain SIC, the average content generally increases with soil depth (top 2 meters, Fig. 4-22 B) and is higher in soils with stronger alkalinity (pH above 9 compared to pH between 7 and 9). In contrast, acidic soils (pH below 5) are typically depleted in SIC (Fig. 4-22 D). Nonetheless, SIC content can vary significantly even among soils with similar pH levels.

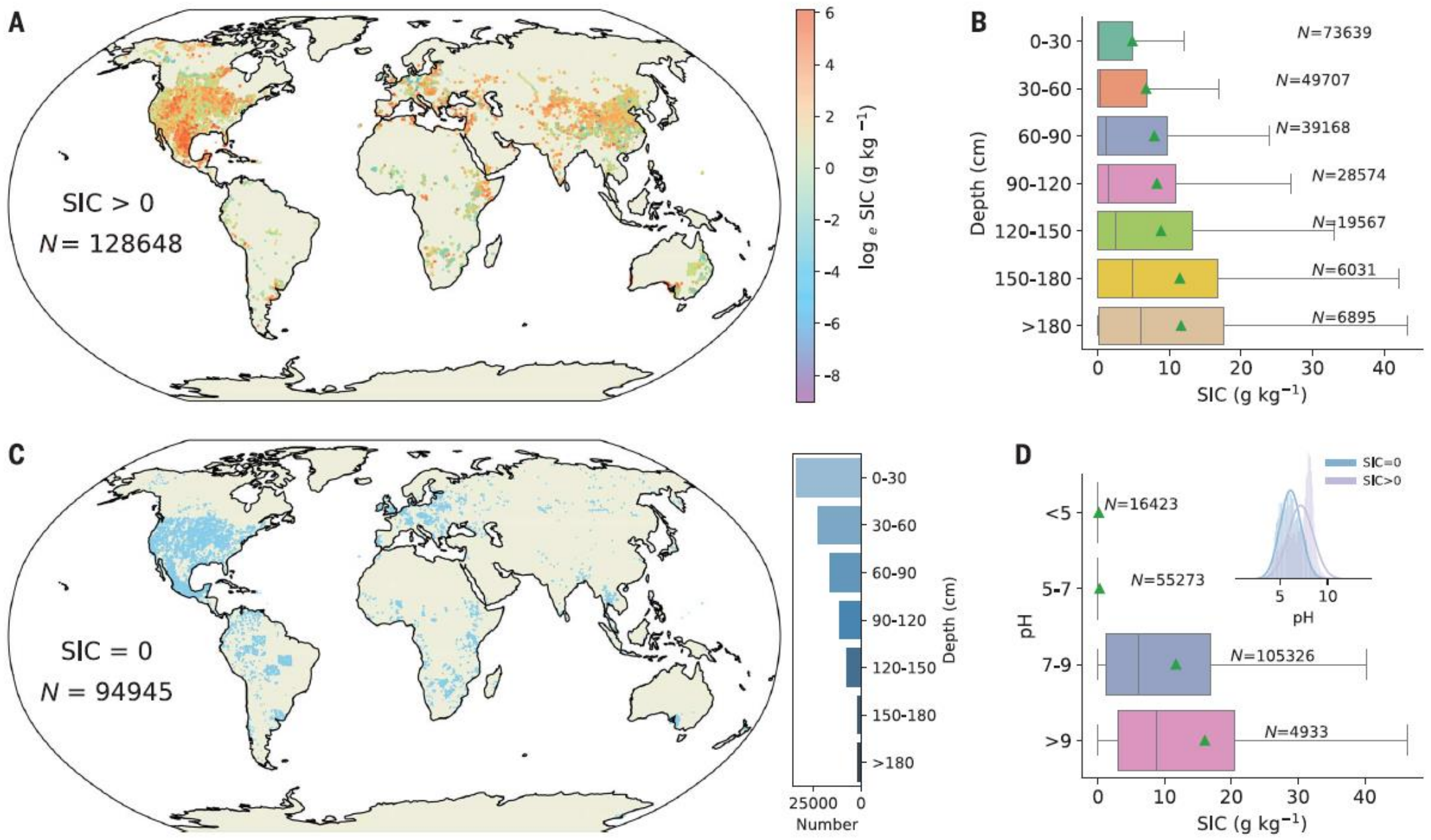

Fig. 4-22 Distribution of raw observations of SIC content

Machine learning models link measured soil inorganic carbon (SIC) content to spatially detailed data on factors such as climate, topography, geology, vegetation, soil characteristics, and human activities. These models use this information to infer the global distribution of SIC by combining insights into the sources, formation, transport, and persistence of SIC with recent advances in observational techniques, theoretical frameworks, and computational methods (see materials and methods for more details). To avoid bias towards zero in predictions, a two-step modeling approach is employed: first, a classification model is developed to predict whether a soil sample (with particle size $\leqslant 2$ mm) contains no SIC (i.e., SIC = 0). For samples where SIC is present, a regression model then estimates the actual SIC content.

Leveraging established data-driven relationships, the classification and regression models achieve high performance (classification: AUC = 0.99, F score = 0.95; regression: $R^2 = 0.79$, root mean square error = 6.17 g $kg^{-1}$, using 10-fold cross-validation) (Fig. 4-23. D and E). These models provide a spatially detailed global estimation of SIC at a 30 - arc second resolution (roughly 1 square kilometer at the equator) (Fig. 4-23) to a depth of 2 m. Additionally, they offer quantitative insights into the factors influencing SIC storage. This methodology captures more real-world variability and heterogeneity in SIC compared to previous methods that relied on land or soil units, owing to the comprehensive integration of multiple environmental covariates and the extensive SIC measurement database.

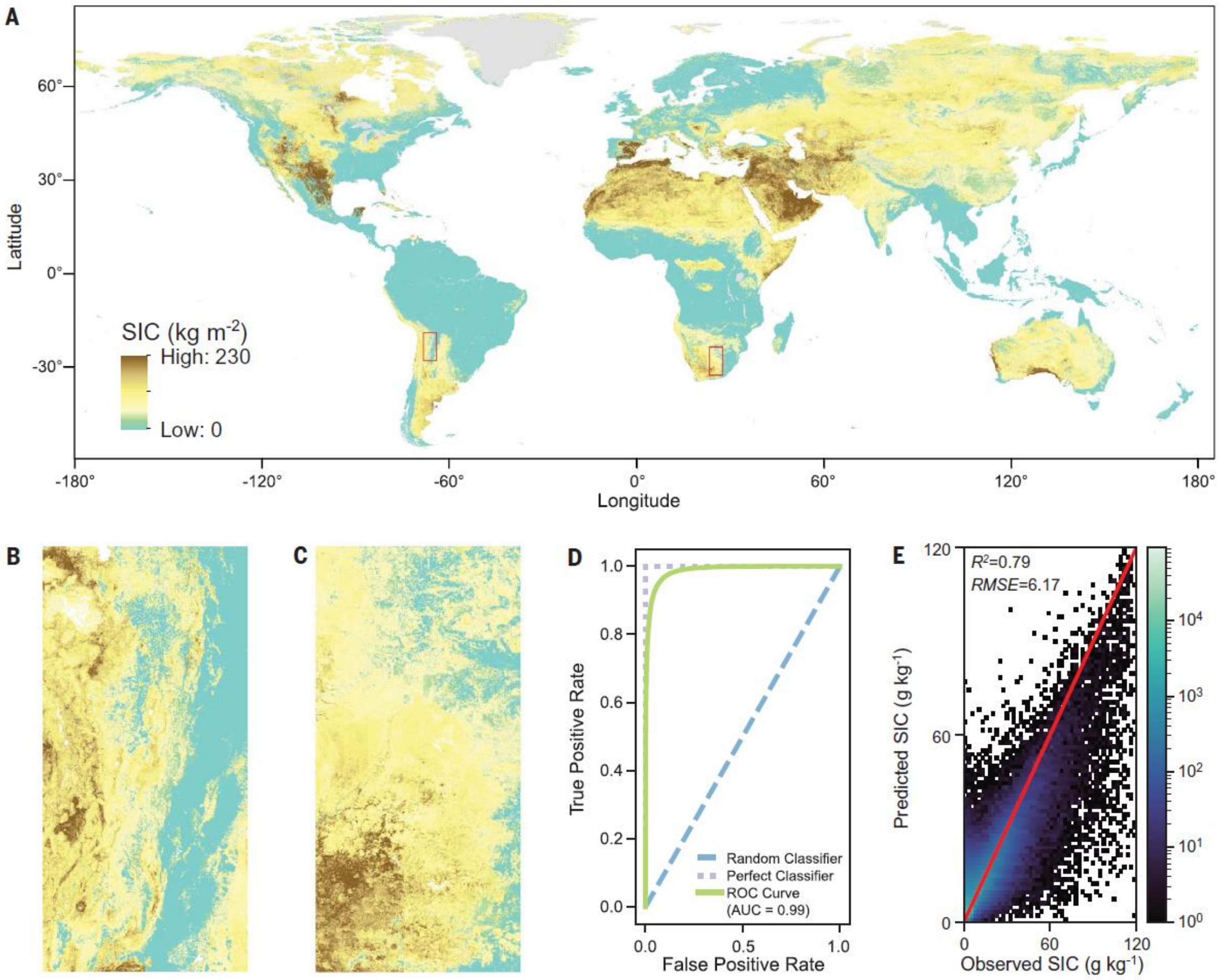

Fig. 4-23 The global map of SIC of the top 2 m of soil at the 1- km$^2$ spatial resolution

The amount and distribution of soil inorganic carbon (SIC) are shaped by complex interactions between soil parent materials, soil characteristics, biological factors, climate, topography, and human influences (Fig. 4-24). Among these, soil pH is the most influential factor for predicting SIC presence, accounting for 29% of the variability in the model. Other important factors include the annual temperature range (4.9%), temperature seasonality (3.0%), cation exchange capacity (2.6%), precipitation during the coldest quarter (2.3%), and soil silt content (2.3%). These percentages are derived from Shapley values, which assess the average marginal effect of each predictor using cooperative game theory. Soil pH serves as a comprehensive indicator that mirrors the complex interplay between soil properties and environmental factors like climate and water balance. While soil pH affects the dissolution and loss of carbonate minerals, SIC also plays a key role in buffering and stabilizing soil pH.

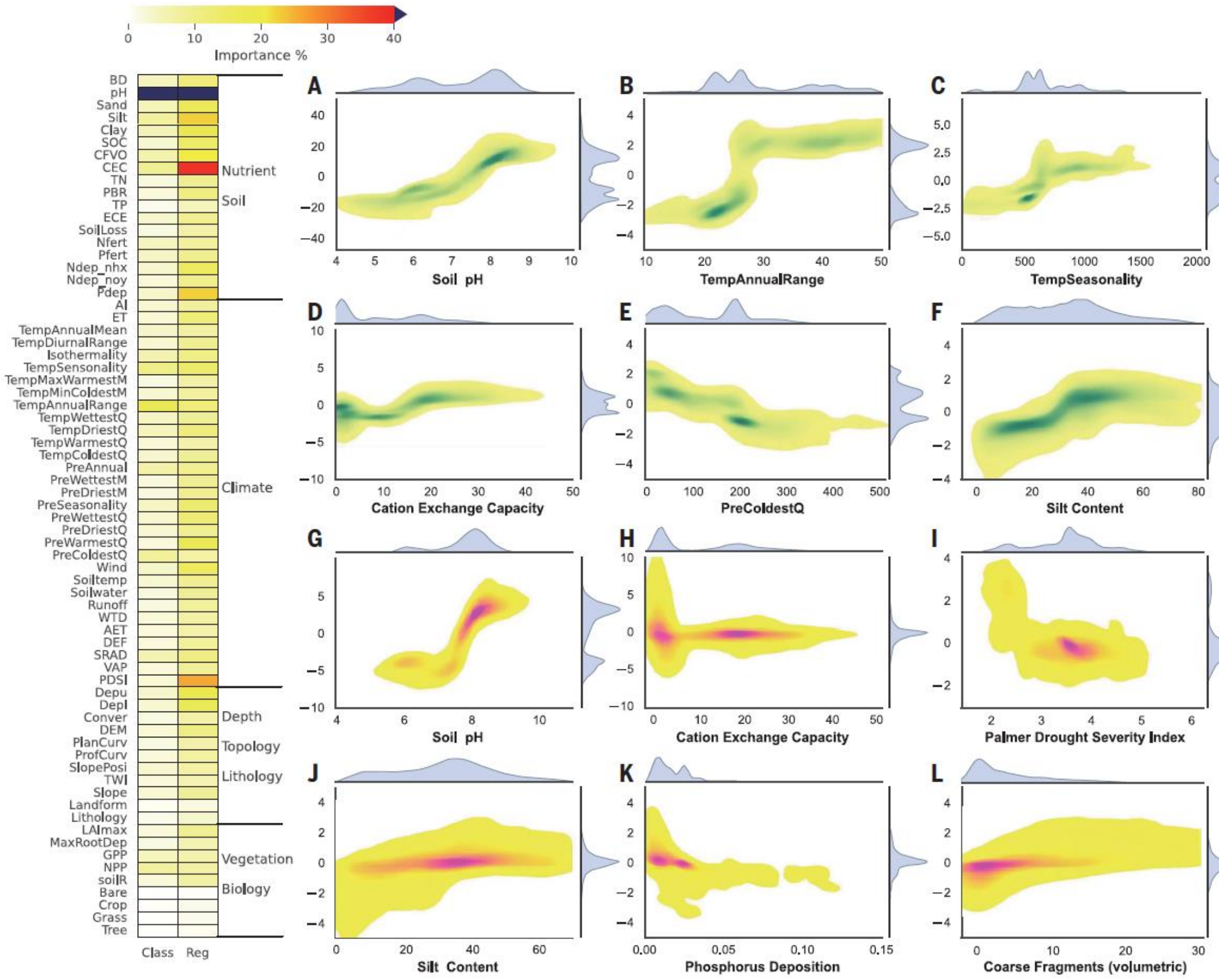

Fig. 4-24 Predictors of SIC

Soil acidification has accelerated the loss of SIC across the globe. Sensitivity analysis indicates that a uniform reduction of 0.1 to 0.5 units in soil pH (within the top 0.3 meters of soil) worldwide could potentially release an additional 9 to 55 GtC of SIC (Fig. 4-25). Regionally, the United States is the most sensitive to SIC losses due to acidification, followed by Australia, Argentina, Russia, and Mexico (Fig. 4-25). In reality, the extent of acidification varies from one region to another. Among the various natural processes and human activities that drive changes in soil pH, climate change and nitrogen additions are identified as two of the most significant contributors.

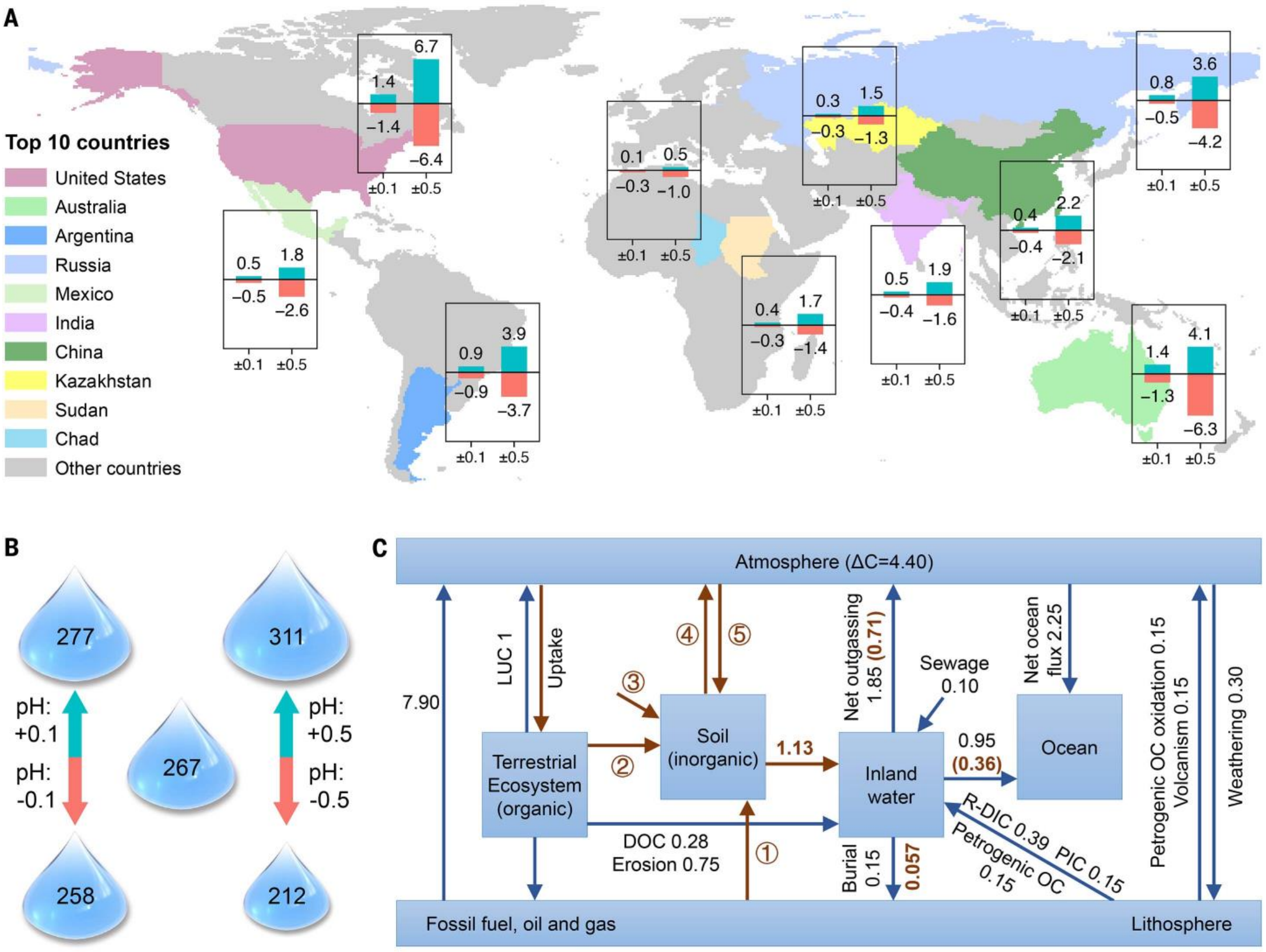

Fig. 4-25 SIC-relevant global budgets

The response of soil inorganic carbon (SIC) to climate change, precipitation patterns, increasing atmospheric $CO_2$, and land use modifications is distinct from how biological systems react to these same drivers. This difference has the potential to reshape our understanding of terrestrial carbon dynamics and carbon-climate feedbacks. Additionally, the success of carbon sequestration methods—such as enhanced rock weathering, afforestation, and soil organic carbon stabilization—depends significantly on SIC, which affects soil and plant health through elements like nutrient supply, soil structure, interactions between organic matter and minerals, and water retention. The linkages between SIC and the atmosphere, biosphere, hydrosphere, and lithosphere highlight SIC's complex role in the global carbon cycle and its significant, often overlooked, influence. While there are challenges in accurately estimating SIC-related carbon fluxes over periods ranging from decades to centuries, the prevailing assumption that SIC has remained static since preindustrial times, as implied by the IPCC and Global Carbon Project reports, requires reassessment. A more comprehensive understanding of SIC's function in the carbon cycle is crucial.

The global map of SIC content is a valuable tool for advancing our understanding of the biogeochemical cycle of inorganic carbon, monitoring its changes, identifying areas at high risk of SIC loss, determining key influencing factors, and evaluating human impacts. It also supports efforts at local, national, and international levels to remediate and sequester carbon. For instance, the effectiveness of pH control for SIC preservation varies by region, and spatial information on SIC content can help minimize disturbances from agricultural practices, such as nitrogen fertilization or irrigation, that could impact SIC levels.

### 4.5.2 Spatiotemporal Analysis of Hotspots of Global River Changes

Rivers are among the most dynamic ecosystems and components of the water cycle on Earth's surface. They hold significant importance for the socio-economic development of human societies, the sustainability of watershed ecological environments, and the stability of regional climates. Against the backdrop of global changes, e.g. global warming, glacial and permafrost melting, and flood disasters, and intensified human activities, e.g. reservoir construction and aquaculture, impacting hydrological systems, the hydrological regimes of rivers have undergone large-scale and significant changes. Monitoring these changes on a global scale and understanding the driving factors behind them is highly challenging.

This study, based on the latest SWOT satellite river database (SWORD) and the Global Surface Water (GSW) dataset, comprehensively investigated the changes in river water areas globally, covering a total length of 2,097,799 km and a total area of 769,390 km², during the early 21st century (2000-2018) compared to 1984-1999. The research compiled a dataset of newly constructed reservoirs worldwide since 2000. By establishing a massive set of manually interpreted samples and employing machine learning methods, it distinguished for the first time three types of global river water area changes: dam-driven river expansion (Type-R), river morphology evolution (Type-M), and hydrological signal-dominated type (Type-H). Focusing on the hydrological signal-dominated type, the study reported the spatial patterns and hotspots of global river water area expansion/contraction and analyzed the main influencing factors of river water area changes using long-term meteorological data, nighttime light data, and published literature.

The results indicate that approximately one-fifth of the world's rivers have experienced significant changes in river geomorphology, such as river migration and braided river oscillations) (Type-M). About 25% of these morphological changes occurred around the high mountainous regions of Asia (Yarlung Tsangpo River, Indus River, The Ganges River, Irrawaddy River, The Amu Darya River) and the middle and upper reaches of the Amazon River in South America, where the proportion of river changes reaches 40-80%. These river morphology changes occur under specific hydrological and geological conditions, such as meandering and multi-channel rivers. They are related to geological activity, runoff intensity, slope, bank erosion intensity, and sedimentation rates, reflecting the instability characteristics of rivers. Besides the rivers' inherent characteristics, climate change and human activities may have enhanced river instability. For instance, in the Yarlung Tsangpo, Ganges, and Indus basins, seasonal runoff changes due to glacial meltwater and water regulation projects have highlighted river instability.

The expansion of river water areas due to reservoir construction (Type-R) is particularly significant: on the scale of six-level basins, new reservoirs have led to an overall increase of 30.5% in river water areas, most notably in developing countries and regions of Asia, South America, and Central and West Africa. Brazil, China, and India are the top three countries where newly constructed reservoirs have had the largest impact on river water area, contributing 21.7%, 18.5%, and 10.5% respectively. Compared to other types (Type-H) of river water area expansion signals, the expansion effect due to dam construction cannot be ignored, as it accounts for 31.9% of the global river water area expansion.

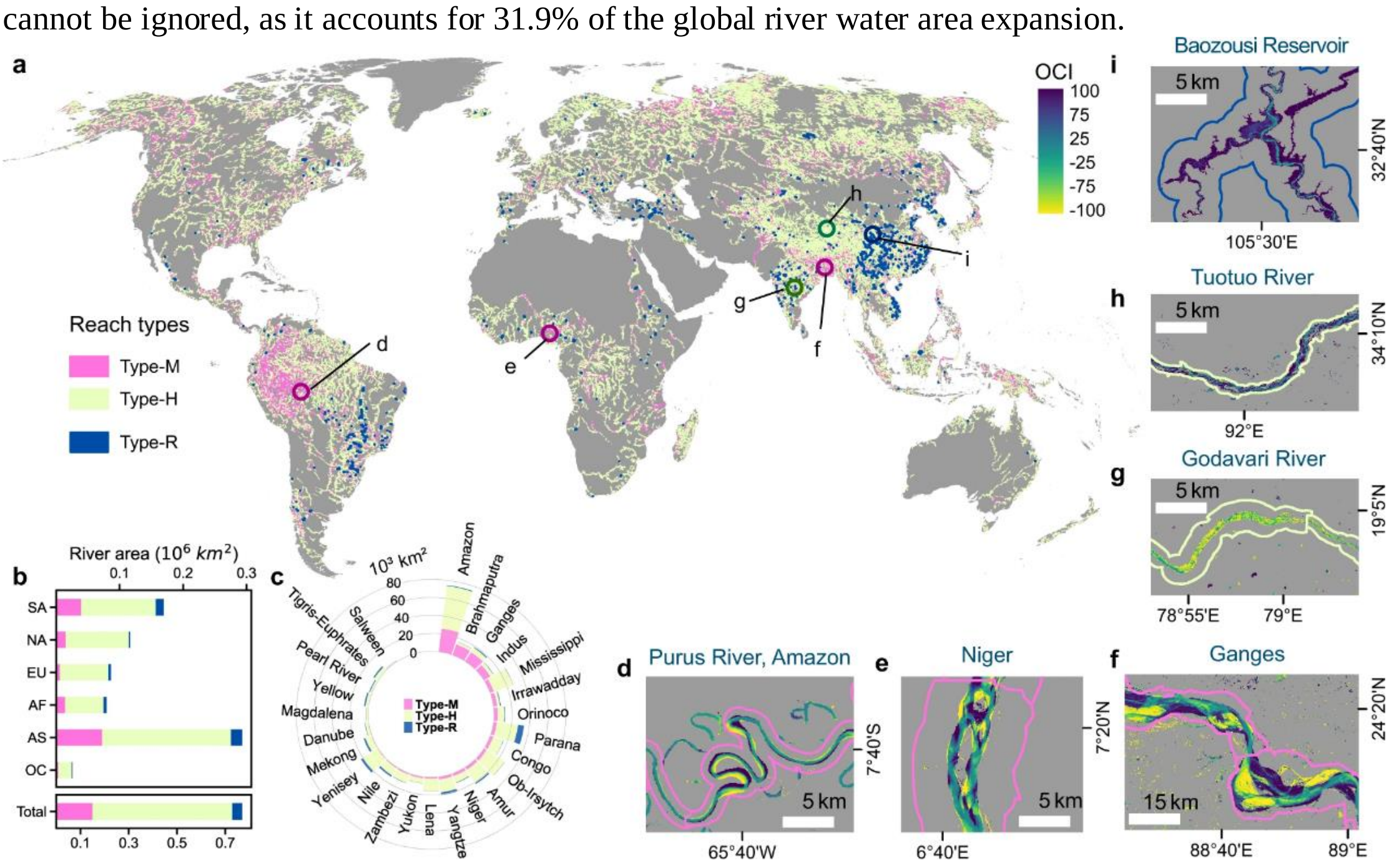

Fig. 4-26 Distribution of change types of different rivers globally

(Type-M: River Morphology Evolution; Type-H: Hydrological Signal-Dominated Type; Type-R: Dam-Driven River Expansion). (a) Global distribution of different types of river changes. (b) Area statistics of different types of river changes across the six continents. (c) Statistics of different types of changes in 25 major river basins worldwide. (d-f) Examples of water body frequency change patterns (OCI, Occurrence Change Intensity) for the three types of changes. Red, green, and blue lines represent the maximum statistical range of river water areas for Type-M, Type-H, and Type-R, respectively.

Excluding river morphology evolution (Type-M) and dam-driven river expansion (Type-R), the study focuses on the changes in river water areas dominated by hydrological signals (Type-H). The results (as shown in Fig. 4-27) indicate that on a global scale, the percentage of areas with significant (moderate) increases in river water areas is 9.0% (8.6%), higher than the significant (moderate) decreases of 4.8% (7.4%). By quantifying the net change in river area for each river basin unit, this study reveals the characteristics of the eight most significant hotspot areas for increases and decreases (indicated as positive and negative hotspots) and their relationships with major climate factors (precipitation, temperature, and evapotranspiration). The positive hotspots are all located in Asia, including Eastern Siberia, the Tibetan Plateau, North-Central Siberia, and East-Central Asia, mainly due to the more sensitive response of high-latitude or high-altitude areas to climate change. In contrast, the negative hotspots are distributed in the Central Great Plains of North America, East-Central South America, Western Siberia, and Northern India, primarily dominated by arid or semi-arid climates. The study also explores the reasons for the relative expansion of river water areas in the Yellow River Basin in China since the 21st century, which may be related to the unified water regulation and a series of water-saving measures that have restored the Yellow River's flow since the 21st century.

In terms of the proportion of river water area expansion and contraction, more than half (70.2%) of the world's rivers remain relatively stable with the highest proportions found in North America (82.1%), followed by Europe (79.5%) and South America (70.5%). Rivers in developed regions, such as Northwestern Europe (e.g., Finland, Sweden) and North America (e.g., Canada, the United States), are relatively more stable compared to those in developing regions such as Asia (e.g., Myanmar, China) and South America (e.g., Bolivia, Peru). This stability is somewhat correlated with nighttime light intensity, indirectly reflecting a possible relationship with the level of socio-economic development. On one hand, populated areas are generally located away from areas with high river changes (e.g., river sources, floodplains). On the other hand, early development of river embankment projects in developed regions has stabilized the extent of river water areas.

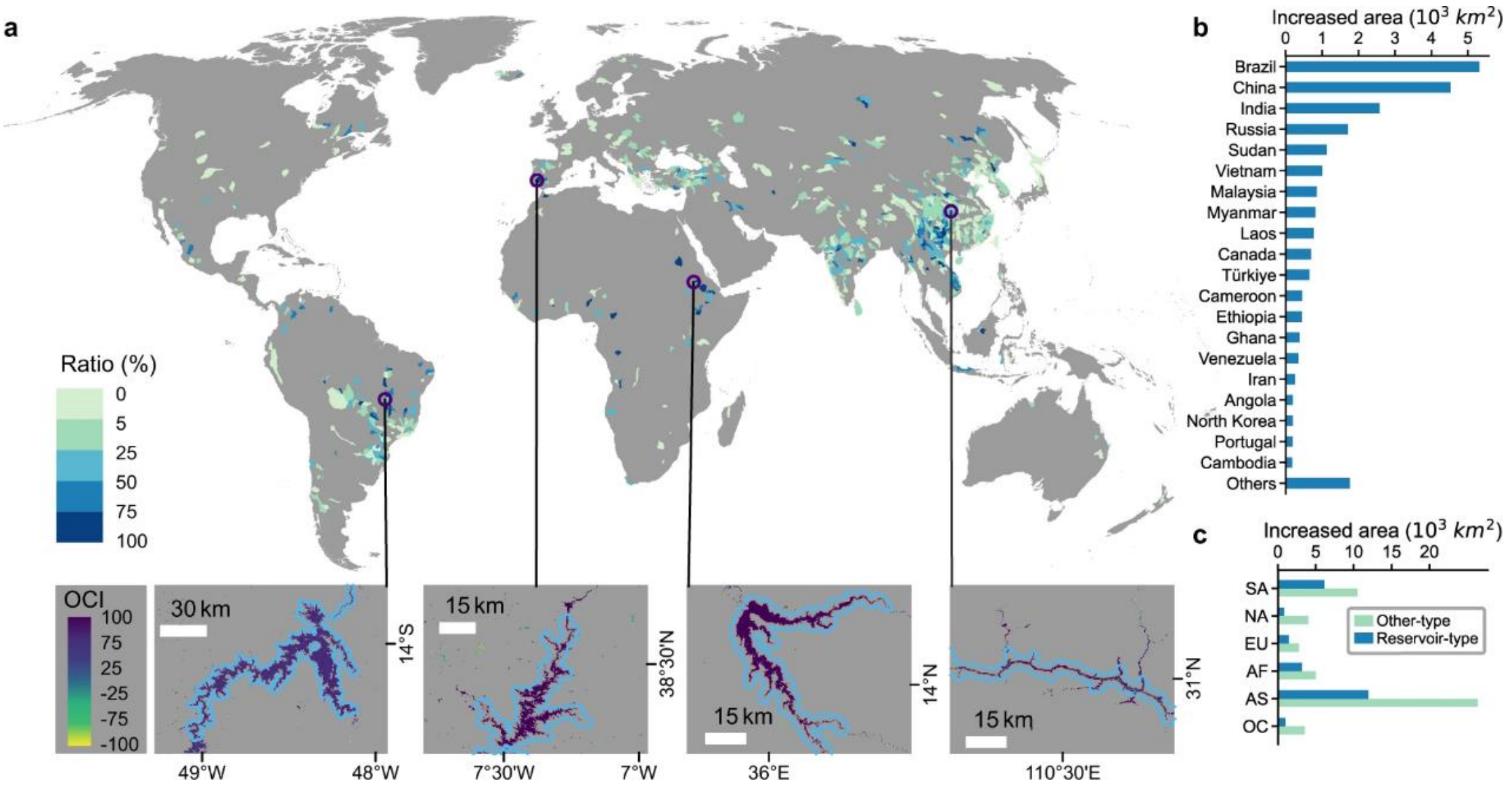

Fig. 4-27 Dry and wet hydrological signal characteristics of global rivers

The upper figure shows the characteristics of river water area changes from 2000-2018 relative to 1984-1999 (PI-PD-PGS: Proportion of river water area increase-decrease-relative stability). The bottom figure highlights the major hotspot regions for water area expansion and contraction.

Overall, this study, based on long-term satellite observations, reveals the characteristics of global river water area changes in the early 21st century and the dominant driving mechanisms. The findings provide scientific evidence for formulating future priority river protection and restoration plans under the United Nations 2030 Sustainable Development Agenda. The study also calls for international action to strengthen long-term monitoring and protection of river water ecosystems.

### 4.5.3 Monitoring Urban Building Damage Based on Satellite Remote Sensing

Major natural geological disasters and armed conflicts can cause extensive damage to urban buildings, leading to casualties and property loss, severely impacting the normal functioning and social stability of cities. Accurately assessing the extent of urban building damage is a critical foundation for post-disaster relief, emergency management, and urban reconstruction. Traditionally, knowledge of urban building damage has primarily relied on eyewitness reports and news coverage. However, these information sources are limited and subjective. They lack the timeliness required for large-scale and real-time assessments.

Satellite remote sensing technology offers an objective and efficient means for assessing urban building damage. Through remote sensing imagery, it is possible to quickly obtain extensive urban building information and identify damaged structures. However, there are several challenges in assessing urban building damage. A major issue is the severe imbalance in sample data. Damaged buildings from natural disasters and armed conflicts constitute only a small fraction of the total urban buildings, resulting in far fewer positive samples than negative ones. This imbalance affects the performance and accuracy of traditional machine learning classification algorithms.

Acquiring high-resolution remote sensing images also presents difficulties. After disasters, it is challenging to promptly obtain high-resolution images, while low-resolution images fail to accurately identify building details, thereby impacting the accuracy of damage assessments. Even with high-resolution remote sensing images, detecting damaged buildings can still result in high rates of false positives and misclassifications. Factors such as shadows and vegetation can be mistakenly identified as damaged buildings, distorting the assessment results and complicating the evaluation of urban building damage.

Satellite remote sensing technology combined with foundation model methods provides an efficient and accurate solution for urban damage assessment. By leveraging deep learning techniques such as convolutional neural networks, foundation models can automatically extract features of damaged buildings from remote sensing imagery, achieving high-precision detection of damaged structures. Foundation models can integrate multiple data sources, including social media data and historical imagery, to comprehensively depict the extent of damage in disaster-affected areas, thereby enhancing assessment accuracy. Additionally, foundation models can effectively address sample imbalance issues through techniques like transfer learning and data augmentation, improving assessment performance in scenarios with limited samples.

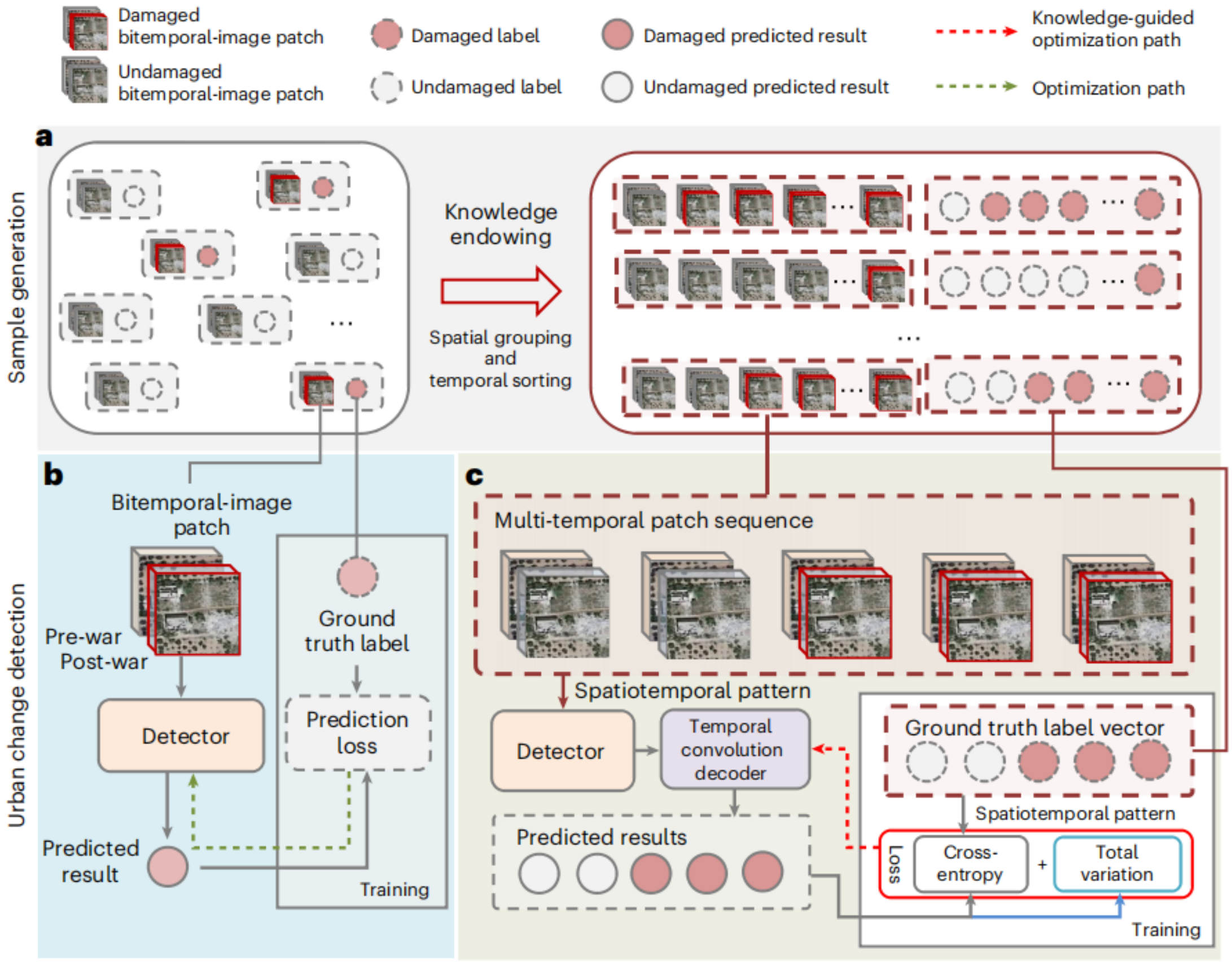

Fig. 4-28 Comparison of existing change detection methods and TKDS

Medium-resolution remote sensing images, with advantages such as global coverage, high revisit frequency, and free access, can be widely used in disaster emergency situations. However, the resolution limitation of medium-resolution remote sensing images means that each building is represented by only a few pixels, making it challenging to effectively identify building shapes, textures, and shadows. Additionally, due to lighting variations, the same building may appear in different colors in remote sensing images taken at different times. This temporal shift can significantly increase the false positive rate, where undamaged buildings are mistakenly identified as damaged. Severe sample imbalance in complex urban environments means that even a small false positive rate can lead to a high number of prediction errors.

Existing building damage detection methods focus on differences between satellite images before and after damage at a specific point in time, neglecting the temporal pattern of damage. However, damaged buildings cannot be rebuilt during a disaster or armed conflict, meaning that building damage has a distinct temporal pattern. Inspired by this, Zhang Liqiang's team proposed a temporal-knowledge-guided detection scheme (TKDS), drawing on ideas from the field of natural language processing. In change detection, different machine learning models can be embedded as detectors in TKDS. To significantly enhance urban damage identification results, they developed a Pixel-based Transformer model (PtNet) as the detector for TKDS.

Next, they used 0.5m and 10m resolution remote sensing images to detect building damage in six cities during the Syrian civil war (2011-2018) and 10m resolution Sentinel-2 data to detect building damage in four Ukrainian cities during the 2022-2023 conflict. The results showed that the F1 score of TKDS-PtNet was twice as high as that of ResNet (2.5 times higher for 10m resolution remote sensing images). Before conducting damage assessment, they thoroughly verified the transferability, interpretability, and reliability of TKDS-PtNet's detection results.

TKDS-PtNet provides a near-real-time urban damage monitoring method, which can generate high-quality damage information from medium and high-resolution remote sensing images in environments where ground data is sparse or difficult to access. TKDS-PtNet is also suitable for detecting infrastructure damage caused by natural geological disasters, aiding in the estimation and assessment of disaster losses.

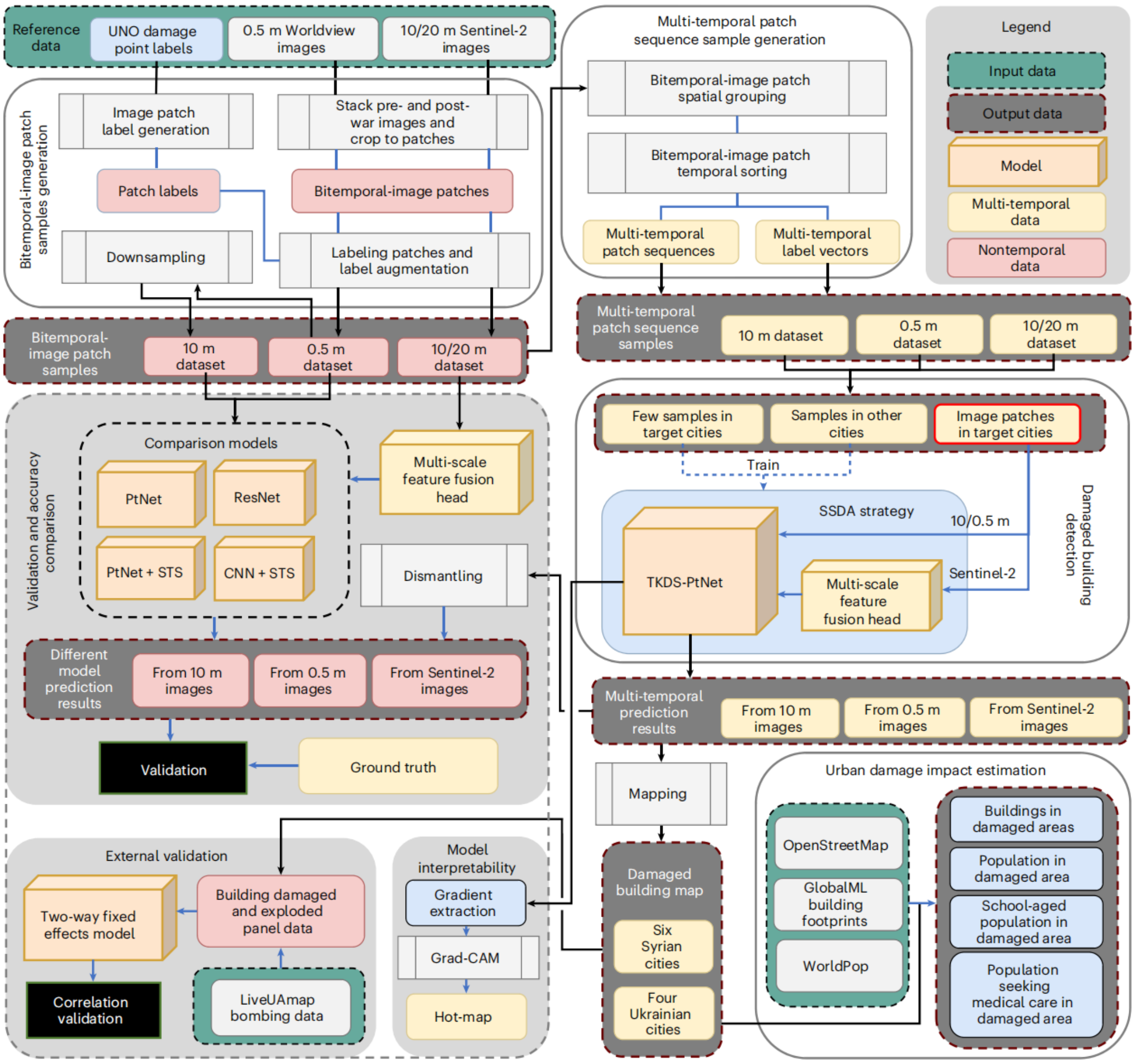

Fig. 4-29 Urban damage monitoring workflow

With the continuous development of foundation model technology and remote sensing image acquisition methods, the integration of foundation model technology and remote sensing imagery is promising for urban disaster monitoring and assessment. Foundation models can efficiently process massive amounts of remote sensing data, extracting key information to provide real-time and accurate support for disaster monitoring. Remote sensing imagery, on the other hand, can overcome temporal and spatial limitations, capturing comprehensive information about disaster areas and aiding foundation models in precise analysis. The synergy between the two will elevate urban disaster monitoring and assessment to new heights.

The deep integration of foundation model technology and remote sensing imagery will bring revolutionary changes to urban disaster monitoring and assessment systems. Intelligent monitoring systems can achieve real-time monitoring and precise early warnings, shortening response times. In-depth analysis of remote sensing imagery data provides scientifically accurate disaster assessment results, supporting post-disaster reconstruction. Additionally, information sharing and collaborative decision-making can enhance overall emergency management efficiency. This integration will propel disaster prevention and control efforts into a new stage of intelligence and efficiency.

# 5. Summary and Prospect

## 5.1 Summary

Based on multiple technical means such as advanced communication technology, artificial intelligence method, big data analysis, and advanced computer technology, a comprehensive model that can conduct comprehensive and in-depth analysis and processing of massive and heterogeneous spatial data is constructed. This model can not only efficiently integrate all kinds of spatial data resources and realize the integration and cross-application of multi-source data, but also intelligently extract the potential value and law of spatial data. It provides accurate spatial information services and decision support for various industries. The intelligent foundation model of spatial data covers the main development directions of data perception, data management, data analysis, and data security. It realizes the all-round intelligent processing and application of spatial data through the comprehensive perception, fine management, in-depth analysis, and security guarantee of data. The model not only focuses on data acquisition and perception but also pays attention to data storage and management, processing and in-depth analysis, and data privacy and security. This ensures the integrity, accuracy and reliability of spatial data.

With the continuous development of science and technology and the explosive growth of data, the demand for accurate spatial information services and decision support in various industries is also increasing. In this context, the emergence of spatial data intelligent foundation model fills the technical gap in the field of spatial data analysis and provides a new way and method to solve various complex problems. This report introduces the importance and research value of spatial data intelligent foundation model in detail from four parts: the background of spatial data intelligent foundation model, thematic foundation model, key technology, and application. In the background of spatial data intelligent foundation model, the definition, development history, research status, development trend, and challenges of spatial data intelligent foundation model are discussed in detail. In the section of spatial data intelligence thematic foundation model, the application of spatial data intelligence large-scale model in different thematic fields is discussed in detail. These thematic foundation models cover many fields such as cities, space remote sensing, geography, and transportation, providing important support and help for the development of various industries. For example, in the field of urban planning, intelligent foundation models of spatial data can analyze urban population distribution, traffic flow, and other information to provide scientific basis for urban development and planning. In the field of space remote sensing, it can use remote sensing data to monitor and analyze the surface and provide support for resource management and environmental protection. However, there are some common problems between different thematic foundation models, such as data quality and model accuracy, which need to be further studied and solved. In the key technologies of spatial data intelligent foundation model, the core technologies supporting spatial data intelligent foundation model are introduced. The development of spatio-temporal big data platform, spatial analysis and visualization, geospatial intelligent computing, spatial intelligent geography multi-scenario simulation, and other technologies provides technical support for the application of spatial data intelligent foundation model. It also provides an opportunity for future technological innovation. The continuous evolution and improvement of these key technologies will further promote the development and application of spatial data intelligent foundation models. In the application part of spatial data intelligent foundation model, the specific application cases in different fields are introduced in detail. Application cases such as dynamic multi-dimensional spatio-temporal deep learning smart city foundation model, multi-modal spatial data smart foundation model, remote sensing smart computing foundation model, geographic smart transportation foundation model and resource, and environment foundation model demonstrate the wide application and rich practice of spatial data smart foundation model in urban planning, resource management, environmental protection, transportation, and other fields. These application cases provide precise decision support and intelligent solutions for various industries and promote the development and progress of related fields.

The emergence and development of spatial data intelligent foundation model fill the technical gap in the field of spatial data analysis and provide new ideas and methods to solve various complex problems. However, to achieve the sustainable development and application of foundation models of spatial data intelligence, many challenges still need to be further studied and addressed, including:

(1) Scaling Law of Foundation models: In the process of building and applying intelligent foundation models of spatial data, we are faced with the challenge of scaling law. The law of scale means that the distribution and characteristics of data show different regularities and changing trends at different spatial scales. How to build effective models at different scales and make them have good generalization ability and adaptability is an important challenge. It is necessary to research and develop model construction and optimization methods for data at different scales. This approach will enable intelligent data analysis and processing at different scales.

(2) Effectiveness of Foundation models: With the increasing and complex spatial data, building and maintaining large-scale intelligent foundation models of spatial data faces the challenge of effectiveness. The validity includes the computational efficiency, resource utilization, and prediction accuracy of the model. How to improve the effectiveness of foundation models so that they can process massive spatial data quickly and accurately is an important challenge. It is necessary to research and develop efficient model construction and optimization algorithms to improve the calculation and prediction performance of the model.

(3) Generative Intelligence for Foundation models: Generative intelligence plays an important role in intelligent foundation models of spatial data, but it also faces some challenges. Generative intelligence can be affected by data bias, time bias, and other factors, resulting in inaccurate or unstable results. How to improve the stability and reliability of generative intelligent models so that they can generate high-quality spatial data is an important challenge. It is necessary to research and develop model training and optimization methods for generative intelligence to improve the accuracy and controllability of the generated results.

### 5.2 Future Prospects

(1) Intelligent Interaction Method of Multimodal Foundation model

With the increasing diversity and complexity of spatial data, the fusion and interaction of multimodal information will become a key challenge. Future research will focus on developing cross-modal information fusion and interaction techniques to achieve effective alignment and interaction between different modal data. Through deep learning and other technologies, models can better understand the meaning of spatial data and make intelligent decisions and outputs according to specific scenarios, providing more rich and effective support for spatial data applications.

(2) Security Theory and Practice of Foundation models

With the wide application of foundation models in various fields, people pay more and more attention to the security and privacy protection of models. In the future, the theory and practice of foundation model security will become the key link in the development of foundation model of spatial data intelligence. The research will focus on developing data security and privacy protection technologies to effectively protect sensitive data and personal privacy involved in the model through data encryption, secure computing, and other technologies. At the same time, the robustness and security evaluation of the model will also be a research focus to ensure that the model can maintain robustness and reliability in the face of various attacks and threats.

(3) Research on Interpretability of Foundation models of Spatial Intelligence Based on Neuroscience

With the wide application of spatial intelligent foundation model in various fields, people pay more and more attention to the interpretability and security of the model. In the future, the intersection of neuroscience and artificial intelligence will become a key link in the development of this field. Research will focus on exploring the inner workings of models and developing interpretability techniques to help people more clearly understand the basis and logic of model decisions. At the same time, the robustness and reliability evaluation of the model will also be the focus of research to ensure the robustness and reliability of the model in various environments and scenarios. This effort will promote the integration of neuroscience and artificial intelligence, laying the foundation for building smarter and more reliable foundation models of spatial intelligence.

(4) A Foundation model of Spatial Intelligence Based on Brain-like Computing

In the future development, the combination of brain-like computing and spatial data intelligence foundation models will open completely new prospects. As a computational method that mimics the structure and function of human brain (Zhong, 2022), brain-like computing can inject new vitality and possibilities into the development of spatial data intelligence foundation models. First, through the characteristics of brain-like computing, the model can realize dynamic adjustment of its structure and parameters to better adapt to different spatial data environments and task requirements. It can also improve the flexibility and adaptability of the model. Secondly, the parallel processing and distributed learning mechanism of brain-like computing are in line with the processing requirements of spatial data, which can provide better computing support for the intelligent foundation model of spatial data and realize efficient parallel computing and distributed processing. In addition, the learning method based on brain-like computing is more in line with the cognitive mode of human beings, which helps to improve the model's understanding and processing ability of spatial data, so as to realize the deep understanding and intelligent inference of spatial data. In conclusion, the combination of brain-like computing and spatial data intelligence model will bring more comprehensive and in-depth development to the field of spatial data intelligence and provide new solutions to solve complex problems in spatial data processing and analysis.

## Acknowledgement

This report is written by Dr. Wang Shaohua and his team from the Aerospace Information Research Institute, Chinese Academy of Sciences. The authors would like to thank a number of research papers from the fields of geographic information science and computer science in China and abroad, including Peking University, Renmin University of China, Wuhan University, Jilin University, Institute of Geographic Sciences and Natural Resources Research, Chinese Academy of Sciences, China University of Geosciences (Beijing), Beijing Normal University, Guangzhou University, Lanzhou Jiaotong University, Jimei University, and Aerospace Information Research Institute, Chinese Academy of Sciences. It was written by a team of outstanding scholars and graduate students. Also involved in the preparation of this report were: Liang Haojian, Huang Zhuonan, Li Xiao, Su Cheng, Liu Chang, Xu Dachuan, Zhou Junyuan, Wei Wenyu, Zhang Jiawei, Pan Yongyi, Zhong Yang, Gu Haoyang, Li Pengze, Wang Kaimin, Guo Heyi, Li Jingbo, Gao Xiaotong, Yang Yujie, Hu Ruichen, Zhang Yu, Zhang Fan, Mei Qiang, Zhou Liang, Sun Yeran, Qi Xiangkun, Wang Peng, Xu Min, Yang Xinwei, Zhang Chunxiao, Chen Yingbiao, Yue Jianwei, Bai Yongqing, Liu Caixia, Zeng Jiangyuan, Liu Xiliang, Zhang Hengcai, Cao Hanghe, etc. I would like to express my heartfelt thanks to the staff members mentioned above.

*Note: The preparation of this paper started in December 2023, the first draft was completed in March 2024, the revised draft was published in April 2024, and the first final version was completed in May 2024.*